\documentclass[11pt]{book}

\usepackage{a4wide,chicago,amsmath,amsthm,epic,eepic,ecltree,lingmacros,psfig}

\newcommand{\Reals}{\mathrm{I} \! \mathrm{R}}
\newcommand{\Rn}{{\mathrm{I} \! \mathrm{R}}^n}
\newcommand{\Nats}{\mathrm{I} \! \mathrm{N}}

\newcommand{\Xcal}[1]{\mbox{${\cal #1}$}}
\newcommand{\Xsf}[1]{\mbox{\sf #1}}

\newcommand{\denI}[1]{\mbox{$ [ \! [  {#1} ] \! ] ^ {\cal I} $}}
\newcommand{\denIa}[1]{\mbox{$ [ \! [ {#1} ] \! ] ^ {\cal I} _ 
{\alpha} $}}
\newcommand{\denIV}[1]{\mbox{$ [ \! [  {#1} ] \! ] ^ {\cal I}_{\sf V} $}}
\newcommand{\denA}[1]{\mbox{$ [ \! [  {#1} ] \! ] ^ {\cal A} $}}
\newcommand{\denAi}[2]{\mbox{$ [ \! [  {#1}  ] \! ] ^ {{\cal A}_{#2}} $}}
\newcommand{\denAV}[1]{\mbox{$ [ \! [  {#1} ] \! ] ^ {\cal A}_{\sf V} $}}
\newcommand{\denX}[2]{\mbox{$ [ \! [  {#1} ] \! ] ^ {\cal {#2}} $}}
\newcommand{\Po}{\mbox{$\mathcal{P} \:$}}
\newcommand{\Pf}{\mbox{${\cal P}_F \:$}}
\newcommand{\Rl}{\mbox{${\cal R(L)} \:$}}
\newcommand{\La}{\mbox{${\cal L}\:$}}
\newcommand{\A}{\mbox{${\cal A}\:$}}
\newcommand{\I}{\mbox{${\cal I}\:$}}
\newcommand{\R}{\mbox{${\cal R}\:$}}
\newcommand{\ASS}{\mbox{${\sf ASS}\:$}}
\newcommand{\gr}{\mbox{$\stackrel{r}{\longrightarrow}\:$}}
\newcommand{\cs}{\mbox{$\stackrel{c}{\longrightarrow}\:$}}

\newtheorem{de}{Definition}[chapter]
\newtheorem{po}{Proposition}[chapter]
\newcommand{\then}{\item[$\Longrightarrow$]}

\allowdisplaybreaks

\begin{document}

   
\thispagestyle{empty}
\begin{center}
\mbox{\psfig{file=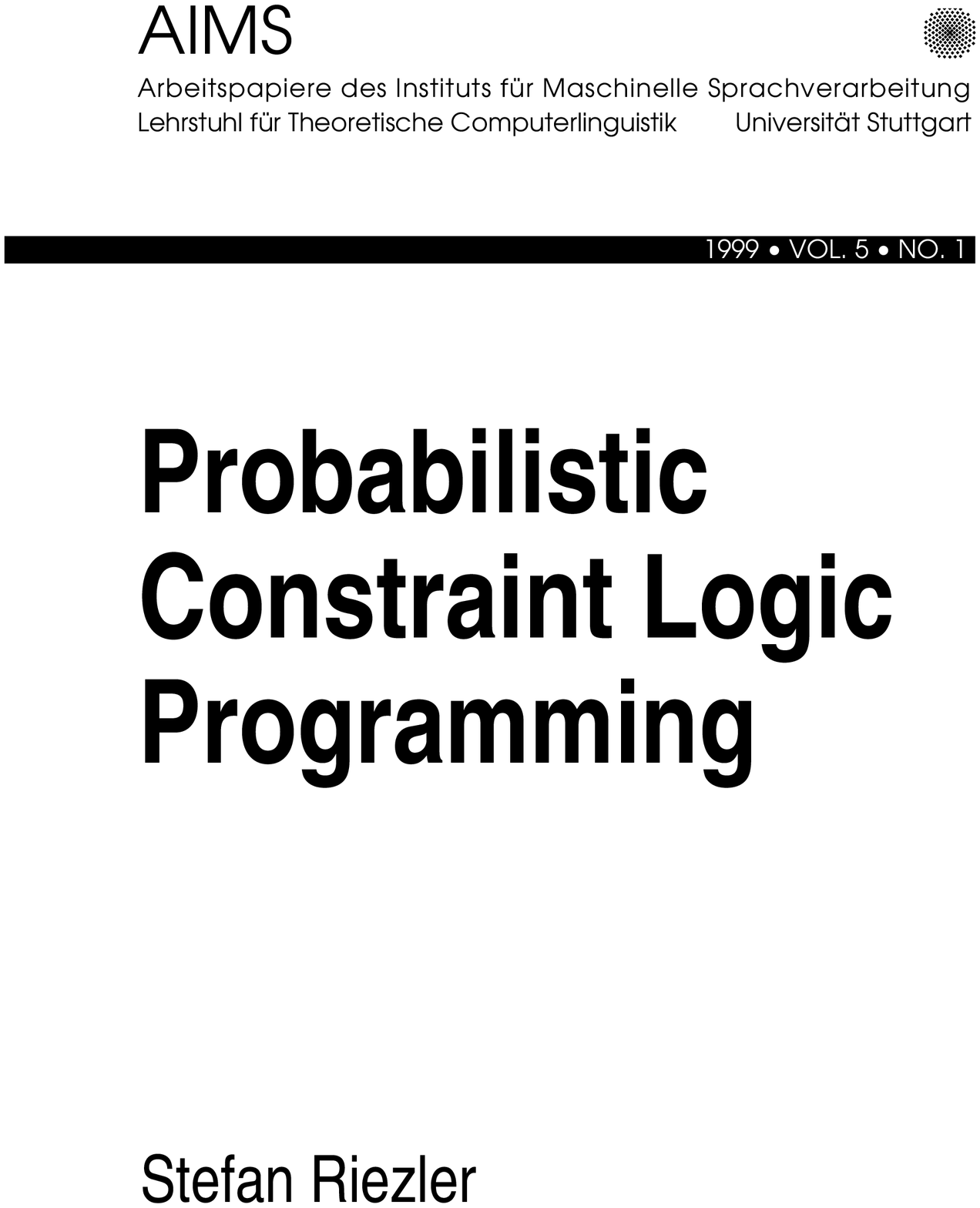,width=15cm,height=20cm}}
\end{center}






\newpage
\thispagestyle{empty}

{\small
{
\setlength{\parindent}{0ex}

\textbf{Previous Issues of AIMS:}

Vol.1 (1) 1994: \textit{Wigner Distribution in Speech Research}.
Master thesis by Wolfgang Wokurek (in German), papers by Grzegorz
Dogil, Wolfgang Wokurek, and Krysztof Marasek (in English), and a
bibliography.

Vol.2 (1) 1995: \textit{Sprachentst\"orung}. Doctoral Dissertation.
University of Vienna, 1994 by Wolfgang Wokurek (in German with abstract
in English). Full title: Sprachentst\"orung unter Verwendung eines
Lautklassendetektors (Speech enhancement using a sound class
detector).

Vol.2 (2) 1995: \textit{Word Stress}. Master thesis by Stefan Rapp (in
German) and papers mostly by Grzegorz Dogil, Michael Jessen, and
Gabriele Scharf (in English).

Vol.2 (3) 1995: \textit{Language and Speech Pathology}. Master theses
by Gabriele Scharf and by J\"org Mayer (in German) and papers by
Hermann Ackermann, Ingo Hertrich, J\"urgen Konczak, and J\"org Mayer
(mostly in German).

Vol.3 (1) 1997: \textit{Tense versus Lax Obstruents in German}.
Revised and expanded version of Ph.D. Dissertation, Cornell
University, 1996 by Michael Jessen (in English). Full title: Phonetics
and phonology of the tense and lax obstruents in German.

Vol.3 (2) 1997: \textit{Electroglottographic Description of Voice Quality}.
Habilitationsschrift, University of Stuttgart, 1997 by Krysztof
Marasek (in English).

Vol.3 (3) 1997: \textit{Aphasie und Kernbereiche der Grammatiktheorie (Aphasia
  and core domains in the theory of grammar).} Doctoral Dissertation,
University of Stuttgart, 1997 by Annegret Bender (in German with
abstract in English).

Vol.3 (4) 1997: \textit{Intonation and Bedeutung (Intonation and
  meaning).} Doctoral Dissertation, University of Stuttgart, 1997 by
J\"org Mayer (in German with abstract in English).

Vol.3 (5) 1997: \textit{Koartikulation und glottale Transparenz
  (Coarticulation and glottal transparency).} Doctoral Dissertation,
University of Bielefeld, 1997 by Kerstin Vollmer (in German with
abstract in English).

Vol.3 (6) 1997: \textit{Der TFS-Repr\"asentationsformalismus und seine
  Anwendung in der maschinellen Sprachverarbeitung (The TFS
  Representation Formalism and its Application to Natural Language
  Processing).} Doctoral Dissertation, University of Stuttgart, 1997
by Martin C. Emele (in German).

Vol.4 (1) 1998: \textit{Automatisierte Erstellung von Korpora f\"ur die
  Prosodieforschung (Automated generation of corpora for prosody
  research).} Doctoral Dissertation, University of Stuttgart, 1998 by
Stefan Rapp (in German with abstract in English). 

Vol.4 (2) 1998: \textit{Theoriebasierte Modellierung der deutschen
  Intonation f\"ur die Sprachsynthese (Theory-based modelling of
  German intonation for speech synthesis).} Doctoral Dissertation,
University of Stuttgart, 1998 by Gregor M\"ohler (in German with
abstract in English).

Vol.4 (3) 1998: \textit{Inducing Lexicons with the EM Algorithm.}
  Papers by Mats Rooth, Stefan Riezler, Detlef Prescher, Sabine
  Schulte im Walde, Glenn Carroll, and Franz Beil. Chair for
  Theoretical Computational Linguistics, Institut f\"ur Maschinelle
  Sprachverarbeitung, Universit\"at Stuttgart. 
}
}


\begin{titlepage}

\centering

\vspace*{25ex}

\textbf{\Huge Probabilistic Constraint Logic Programming}

\vspace{3ex}

\textbf{\Large Formal Foundations of Quantitative and Statistical
  Inference in Constraint-Based Natural Language Processing}  

\vspace{10ex}

\textbf{\Large Stefan Riezler}

\end{titlepage}


\newpage
\thispagestyle{empty}
{
\setlength{\parindent}{0ex}

\vspace*{35ex}

CIP-Kurztitelaufnahme der Deutschen Bibliothek

\vspace{1ex}

\fbox{
\begin{minipage}{14cm}
\textbf{Stefan Riezler:} \\
Probabilistic Constraint Logic Programming. Formal Foundations of
Quantitative and Statistical Inference in Constraint-Based Natural
Language Processing / Stefan Riezler - Stuttgart, 1999.\\
AIMS - Arbeitspapiere des Instituts f\"ur Maschinelle
Sprachverarbeitung, Vol. 5, No. 1, 1999, Stuttgart, Germany. \\
ISSN 1434-0763
\end{minipage}
}

\vfill

Copyright $\copyright \;$  by the author

\vspace{3ex}

Lehrstuhl f\"ur Theoretische Computerlinguistik \\
Institut f\"ur Maschinelle Sprachverarbeitung \\
Universit\"at Stuttgart \\
Azenbergstra{\ss}e 12 \\
70174 Stuttgart 

\vspace{3ex}

www: \texttt{http://www.ims.uni-stuttgart.de/projekte/gramotron/} \\
email: \texttt{gramotron@ims.uni-stuttgart.de}

}


\newpage
\thispagestyle{empty}

\begin{center}

\vspace*{25ex}

{\Large Probabilistic Constraint Logic Programming}

{\large Formal Foundations of Quantitative and Statistical Inference in Constraint-Based Natural Language Processing}  
\vspace{5ex}

von

\vspace{5ex}

Stefan Riezler

\vspace{10ex}

Philosophische Dissertation \\
angenommen von der Neuphilologischen Fakult\"at \\
der Universit\"at T\"ubingen \\
am 17. Dezember 1998

\vfill

Stuttgart \\
1999

\newpage 
\thispagestyle{empty}

\centering

\vspace*{25ex}

Gedruckt mit Genehmigung der Neuphilologischen Fakult\"at \\
der Universtit\"at T\"ubingen 

\vspace{10ex}

Hauptberichterstatter: Prof. Dr. Erhard Hinrichs, Universit\"at T\"ubingen\\
Mitberichterstatter: PhD Steven Abney, AT\&T Labs, Florham Park, NJ \\
Mitberichterstatter: Prof. Dr. Uwe M\"onnich, Universit\"at T\"ubingen\\
Mitberichterstatter: Prof. Dr. Mats Rooth, Universit\"at Stuttgart\\
Dekan: Prof. Dr. Hartmut Engler, Universit\"at T\"ubingen

\end{center}


\newpage

\begin{center}
\textbf{Acknowledgements}
\end{center}

Since the time when I graduated on the metatheoretical foundations
of linguistics I have wanted to learn more about the solid mathematical
basis I assumed to underly computational linguistics. Even if I did
not find a definite answer to my naive questions, miraculously, my
search for these foundations has led to this thesis. The miracle
clearly is due to the people who accompanied me on this way.

First of all, this thesis is dedicated to Sabine, whom I want to thank
for making these years in T\"ubingen the best ones of my life.

Next, I want to thank my supervisors. Steve Abney initiated me with incredible patience into the area of probabilistic modeling and statistical inference. Being an autodidact in mathematics himself, he was the only one to know how to spoon-feed me these topics. He always had time and, more importantly, also always an answer to my many questions.
Erhard Hinrichs made my research possible by taking me on the Graduiertenkolleg in T\"ubingen and accepting me as doctoral student in spite of the naive thesis proposal I handed in then. Furthermore, I would like to thank him for letting me step into these theoretical spheres and yet taking me down to earth when necessary.
Uwe M\"onnich was so kind to take on the time-consuming job of a third
supervisor and was very helpful in making the formulation of the final
draft clearer. 
Special thanks go to my new boss, Mats Rooth, who took
on the job of the fourth supervisor. I would like to thank him for
giving me all the time and support I needed to make up a final version
of the thesis. Moreover, I am very glad that he took a theoretician
like me on his team and gave me the possibility to move on the
practical side of statistical modelling and experimenting in my new job.

Furthermore, I am grateful to my new colleagues at the Institut f\"ur
maschinelle Sprachverarbeitung in Stuttgart whom I got to know during
a two-year reading group mostly on papers relevant to my thesis. They are
Glenn Carroll, Marc Light, Detlef Prescher, and Helmut Schmid.
Similar thanks go to my
colleagues at the Seminar f\"ur Sprachwissenschaft and the
Graduiertenkolleg in T\"ubingen, Thilo G\"otz, Graham Katz, Paul
King, Frank Morawietz, and Andreas Wagner. Thank you all for all these
endless discussions and for reading all 
these drafts of my papers and thesis chapters. Special thanks to
Detlef Prescher for proof-reading my thesis several times, and above
all, for his never-ending patience which makes him the
very best colleague one can wish to work with.

Exceptional thanks go to Mark Johnson who made it possible for me to
discuss my ideas with the computational linguists and applied
mathematicians at Brown University, and moreover, who initiated a group working
on estimation of probabilistic constraint-based grammars and did most
of the work himself.

Moreover, I would like to thank all the people who accompanied me on
my way to probabilistic modeling of natural language, i.e., Karel Oliva and Hans Uszkoreit, who helped me to get into the black art of computational linguistics in Saarbr\"ucken; Hanspeter Ortner and Lorelies Ortner, who supported my break with classical linguistics as my supervisors and teachers at the German linguistics department in Innsbruck; and finally my parents, who supported my extended studies in Innsbruck and Vienna.

Last, I would like to thank the Deutsche Forschungsgemeinschaft for supporting my work with a three-year scholarship at the Graduiertenkolleg Integriertes Linguistikstudium in T\"ubingen.


\newpage

\begin{center}
\textbf{Abstract}
\end{center}

Structural ambiguity in linguistic analyses is a severe problem for
natural language processing. For grammars describing a nontrivial
fragment of natural language, every input of reasonable length may
receive a large number of analyses, many of which are implausible or
spurious. This problem is even harder for highly complex
constraint-based grammars. Whereas the mathematical foundation of such
grammars as instances of constraint logic programming is clear enough,
there is so far no mathematically well-defined method for extending
constraint logic programs by using weights to introduce
graded distictions between analyses. Previous approaches to ambiguity
resolution for context-dependent natural language processing models
either are tailored to specific applications and based on uncertain
mathematical grounds, or they are sufficiently well-defined and
expressive but infeasible in practice. 

In this thesis, we present two approaches to a rigorous mathematical
and algorithmic foundation of quantitative and statistical
inference in constraint-based natural language processing.
The first approach, called quantitative constraint logic programming, is
conceptualized in a clear logical framework, and presents a sound and
complete system of quantitative inference for definite clauses
annotated with subjective weights. This approach combines a rigorous
formal semantics for quantitative inference based on subjective
weights with efficient weight-based pruning for constraint-based systems.
The second approach, called probabilistic constraint logic
programming, introduces a log-linear probability distribution on the
proof trees of a constraint logic program and an algorithm for
statistical inference of the parameters and properties of such
probability models from incomplete, i.e., unparsed data. The
possibility of defining arbitrary properties of proof trees as
properties of the log-linear probability model and efficiently
estimating appropriate parameter values for them permits the
probabilistic modeling of arbitrary context-dependencies in constraint
logic programs. The usefulness of these ideas is evaluated empirically
in a small-scale experiment on finding the correct parses of a
constraint-based grammar. In addition, we address the problem of
computational intractability of the calculation of expectations in the
inference task and present various techniques to approximately solve this
task. Moreover, we present an approximate heuristic technique for
searching for the most probable analysis in probabilistic constraint
logic programs. 

\clearpage
\thispagestyle{empty}
~\clearpage


\newpage

\tableofcontents


\chapter{Introduction}
\pagenumbering{arabic}

This thesis presents a novel mathematical treatment of the problem of
structural ambiguity in constraint-based natural language processing
(NLP). This problem will be attacked from two different angles. On the
one side we will present a novel formalism for quantitative
constraint-based inference with subjective weights. On the other side
we will approach this problem by novel methods for statistical
inference and probabilistic modeling for constraint-based NLP.

In this chapter we introduce the general problem of structural
ambiguity and a general solution to this problem, namely weighted grammars.
Furthermore, we will specify the notion of constraint-based NLP and
sketch the general idea of the two different approaches to ambiguity
resolution for weighted constraint-based grammars which
constitute the main contribution of this thesis.

\section{Overview}
\markboth{Chapter 1. Introduction}{1.1 Overview}

Following this introduction, Chap. \ref{Foundations} discusses the formal
framework in which the informal notion of constraint-based NLP will
be dealt with in the course of this thesis. To this end, we
discuss the formal basics of Constraint Logic Programming (CLP), which
is used here to provide an operational treatment of various declarative
constraint-based grammars. This is done by an embedding of 
the logical description languages of such grammars into a CLP scheme, yielding
Constraint Logic Grammars (CLGs).

Chap. \ref{Q} presents a quantitative extension of CLP which
allows us to assign subjective numerical weights to the structural
components of a constraint logic program. We present a sound and
complete system for quantitative inference with such subjective
weights based on concepts of fuzzy set algebra.
Furthermore, the general concepts of quantitative
CLP will be exemplified with a simple quantitative CLG and we will
show how the search technique of alpha-beta pruning can be adapted to
efficiently finding the best parse in quantitative CLGs. 

A completely different approach to weighted CLP is presented
in Chap. \ref{P}. Here, instead of concentrating on a formal
specification of the handling of subjective weights, the aim is to use
methods of probabilistic modeling and statistical inference to
automatically induce weights from empirical data. 
We introduce a powerful log-linear probability model for CLP
and present a novel technique for statistical inference of the
parameters and properties of such models from incomplete training
data. We show monotonicity and convergence of the algorithm to the
desired maximum likelihood estimates and discuss various methods for
approximate computation for the inference task. We
present an instantiation of probabilistic CLP to a simple
probabilistic CLG and show how the structure of the 
probabilistic model can be used to guide the search for the most
probable analysis. Furthermore, the main concepts of this statistical
approach are evaluated empircally in a small experiment on finding the
correct parses of a constraint-based grammar.

Chaps. \ref{Q} and \ref{P}, presenting the two different approaches to
weighted CLP and CLGs, are conceptualized completely independent of
each other. Whereas Chap. \ref{Q} is based upon the general
concepts of Chap. \ref{Foundations}, namely classical CLP with CLGs as
a special instance, the work of Chap. \ref{P} is entirely
self-contained and even more general. That is, the presented methods of
probabilistic modeling, statistical inference and approximate
computation can easily be abstracted away from the CLP application to
more general data structures.

Chap. \ref{Conclusion} presents a summary of the work of this thesis,
and compares the advantages and shortcomings of the two presented approaches
relative to each other and relative to other approaches. Furthermore,
directions of future work are sketched.

The rest of this chapter presents a motivation of the why and how of
the work of this thesis.

\section{A Practical  Problem: Structural Ambiguity}
\markright{1.2 A Practical Problem: Structural Ambiguity}

Structural ambiguity is a practical problem for every grammar
describing a nontrivial fragment of natural language. That is, for
such grammars every input of reasonable length may receive a large
number of different analyses, many of which are not in accord with
human perceptions. The problem to be addressed is how to differentiate
between these analysis and how to efficiently find the correct
analysis out of the set of all possible ones.

A simple example illustrating the ubiquity and severity of the problem
of structural ambiguity has been presented by \citeN{Church:82}.
Consider the following sentence with two PPs. It has the following two
analyses in terms of PP-attachment:
\eenumsentence{\item Put the block [in the box on the table].
\item Put [the block in the box] on the table.}
If we have three PPs, the number of analyses is five. 
\eenumsentence{\item Put the block [[in the box on the table] in the kitchen].
\item Put the block [in the box [on the table in the kitchen]].
\item Put [[the block in the box] on the table] in the kitchen.
\item Put [the block [in the box on the table]] in the kitchen.
\item Put [the block in the box] [on the table in the kitchen].}
Continuing this list further,  a number of more than thousand analyses
is achieved quickly with only eight PPs. The pattern behind this
list can be explained as a combinatorial growth of ambiguity in the
number of PPs. This growth pattern follows the combinatorial principle
of the the Catalan numbers, where
$Cat(n)$ describes the number of ways to parenthesize a sentence of
length $n$, or equivalently the set of binary trees that can be
constructed over $n$ terminal elements\footnote{The Catalan numbers
are generated by the following formula:
$Cat(n) = \left( \begin{array}{c} 2n \\ n \end{array} \right) 
- \left( \begin{array}{c} 2n \\ n-1 \end{array} \right)$.}.
Clearly, this pattern can be found also in other linguistic
combinations such as conjuncts, nominal modifications, or relative
clauses.

Whereas ambiguities of this kind are only problematic if the number of
linguistic elements to be combined is large, there is another source
of ambiguity depending simply on the number of analyses the grammar
can produce at all. Let us consider the standard linguistic example
sentence \emph{John saw Mary} and the two analyses given below.

\eenumsentence{\item $ [\textrm{John}_N \; [ \textrm{saw}_V\;
  \textrm{Mary}_N \; ]_{VP}\; ]_S$.
\item $[[\textrm{John}_N \; \textrm{saw}_N \;]_{NP} \; \textrm{Mary}_N  \;]_{NP}$.}
Even if the first analysis is perfectly plausible and might be
considered as the unique analysis of this sentence, the second
analysis has to be accepted if the grammar also licenses other nominal
modifications such as
\enumsentence{$[[\textrm{school}_N \; \textrm{committee}_N \; ]_{NP}\;
  \textrm{meeting}_N \; ]_{NP}$.}
Following \citeN{Abney:96}, the second analysis furthermore can be
given a perfectly plausible interpretation as the reference to a
person named Mary who is associated with a kind of saw called John
saw. 
Clearly, such spurious ambiguities may be characterized as resulting
from rare usages of words and constructions, but they will appear in
every grammar which covers a reasonable fragment of natural
language and thus produces a large number of analyses. Furthermore,
in most cases such spurious ambiguities cannot be given a plausible
interpretation, but just have to be accepted as a side-effect of high
coverage.

Together combinatorial and spurious ambiguity can confront NLP
systems with severe problems. Clearly, there is a need to distinguish more
plausible analyses of an input form less plausible or even totally
spurious ones. A practical and general approach to this problem is the use of
weighted grammars for resolving structural ambiguities.

\section{A Practical Solution: Weighted Grammars}
\markright{1.3 A Practical Solution: Weighted Grammars}

We will approach the problem of structural ambiguity by using weighted
grammars for ambiguity resolution. Weighted grammars can be
characterized very generally as follows. They assign numerical values,
called weights, to the structure-building components of a grammar and
calculate the weight of an analysis from the weights of the structural
components that make it up. The simple but effective assumption is to
connect the plausibility of an analysis with its weight. That is, a
ranking of analyses is defined by the weighted grammar, and more
plausible analyses are differentiated from less plausible analyses in
terms of their weights. The most plausible or correct analysis then is
chosen from among the in-principle possible analyses by assuming the
analysis with the greatest weight to be the correct one.
Furthermore, when we are interested only in the highest weighted parse,
the weight calculation scheme can be used to guide the search for the
highest weighted parse efficiently instead of simply listing all possible
parses and choosing the highest weighted one.

There are three basic problems to be solved for every weighted grammar to
be a useful device in real-world NLP applicatons. These problems can
be described by the following questions.

\begin{enumerate}
\item How can the values of the weights be obtained?
\item How should the weights be applied to the components of the
grammar and how should the weight of an analysis be calculated from
the weights of the components? 
\item How can the structure of the weight calculation scheme be used
to guide the search for the highest weighted analysis efficiently?
\end{enumerate}

Clearly, the answers to these questions depend on each other and on
the non-weighted framework to be extended. In the
following we will sketch the basic ideas of two different approaches
to answer these questions consistently for a framework of
constraint-based systems.

\section{Towards a Mathematical Foundation of Weighted
Constraint-Based Grammars}
\markright{1.4 Weighted Constraint-Based Grammars}

The NLP systems of choice in this thesis are constraint-based
grammars. The term constraint-based is a collective name for highly expressive
frameworks for declarative description of natural language in
terms of logical description languages. 
Throughout this thesis, the informal concept of constraint-based
grammars will be replaced by the formal concept of constraint logic
grammars. That is, constraint-based grammars are formalized
here by an embedding of the logical description languages of such
grammars into a CLP scheme, yielding CLGs as special applications of CLP. 
The advantages of this approach are on the one hand the
(Turing-)power of the underlying logic, which is conceived as a
welcome property to overcome the inadequacy of regular and
context-free grammars for the description of natural language.
On the other hand this approach permits an operational treatment of,
e.g., the parsing problem for arbitrary constraint-based grammars in a
consistent and unique way. Since CLGs can be seen as special
applications of CLP, the mathematical work of this thesis will be
based upon CLP in general, and CLGs will serve as running example
illustrating the applicability of the general work to NLP. 
The reference to the general framework of CLP will generalize the
results of this thesis in a welcome manner. 

However, most CLP applications require some form of graded
distinctions which are not provided by a classical CLP
scheme. A very important example for
this demand for gradedness is the task of structural ambiguity
resolution in CLGs. A crucial assumption in this thesis is the claim
that a framework of weighted CLP is the solution of choice for the
ambiguity resolution problem for CLGs. In the following chapters we
will present a rigorous mathematical formulation of two different
approaches to weighted CLP and weighted CLGs.

The first approach we will present is motivated by the aim to give the
grammar designer and implementer maximal freedom in choosing
appropriate values for the weights of the weighted grammar. That is,
the values of the weights are only restricted to be some quantities
lying in a certain interval of real numbers. Such weights can be
restricted to meet the axioms of probability theory, but there is no
need to do so. Besides subjective probabilities, such quantities could
be subjective preference values, or values obtained from experiments
on preferences in human language processing, or values describing
human judgements on degrees of grammticality, or others. In order to
stress the generality of this approach to weighted CLP and weighted
CLGs, we will henceforth refer to it as \emph{quantitative CLP} and
\emph{quantitative CLGs}, respectively. The main task of this approach is to
specify the questions of how to establish a proper weight calculation scheme
for given values and of how to use such a scheme for efficient
disambiguation. Since it is the grammar designer and implementer who
has to specify the grammar and the weights, it makes sense to tie these
two tasks together as closely as possible. That means, in the same way
as the inference system of classical CLP is coupled with a clear
formal semantics, one would like to relate a quantitative inference
system to a quantitative formal semantics, instead of adding an
extralogical calculation scheme to the well-defined logic of CLP. Thus
the task to be addressed is to provide a precise, but yet
simple formal semantics for quantitative inference in CLP. To this
end, we present a formal semantics for quantitative CLP based upon the
simple and intuitive concepts of fuzzy set algebra. This semantics and
the corresponding sound and complete quantitative inference system
furthermore are designed in a way which enables the search technique
of alpha-beta pruning to be used quite directly for efficient
disambiguation. Quantitative CLP then provides an efficient,
well-defined quantitative deduction system, which can be adapted for
specific applications by embedding specific constraint languages into CLP
and attaching appropriate weights to them.

A completely different approach to weighted CLP and weighted CLGs is
presented by our models of \emph{probabilistic CLP} and
\emph{probabilistic CLGs}. The aim of this approach is to specify
a probability distribution over the set of proof trees of CLP or the
parses of CLGs, and to provide statistical methods to
infer the values of the parameters of such probabilitic models from
empirical data. For a given sample of training data and a parametric
probability model, both the parameters of the
probabilistic model and the properties of the model associated with
these parameters can be induced automatically by methods of
statistical inference. We present a highly expressive log-linear
probability model for CLP, and a novel algorithm to infer the
parameters and properties of log-linear models from incomplete
data. We show monotonicity and convergence of the new algorithm and
discuss methods for efficient approximate computation of the formulae
involved in the algorithm. This algorithm is applicable to log-linear
models in general, and especially provides the means for automatic and
reusable training of arbitrary probabilistic constraint-based grammars
from unparsed data. The usefulness of these concepts is shown
empirically in a small-scale experiment on finding preferences in
parse-data from a constraint-based grammar.
Furthermore, we discuss the possibilities of using the structure of
the probabilistic model to guide the search for the most probable
proof tree or analysis, and present a heuristic search algorithm for
this task. Clearly, in this setting a model-theoretic semantics for
probabilistic inference is superfluous since the values of the
probabilistic parameters are obtained by automatic statistical methods
which are not manipulable by the user. Rather, we are interested in a
stochastic semantics for CLP inference which is determined by the log-linear
probability model together with the statistical methods for parameter
estimation and property selection from given input data.

\section{Bibliographical Note}
\markright{1.4 Bibliographical Note}

Various parts of this thesis are based upon previously published work of
the author. Chap. \ref{Q} is an extended version of
\citeN{Riezler:96}. Chap. \ref{P} is based upon work presented in
\citeN{Riezler:97}, \citeN{RiezlerKon:98}, \citeN{Riezler:98}, and \citeN{Johnson:99}.

\chapter{Foundations: Basic Concepts of CLP and CLGs}
\label{Foundations}

In this chapter we report the central formal concepts of the CLP
scheme of \citeN{HuS:88}. In preparation for the following work we
give some proofs missing in the original paper and present the CLP
scheme in a slightly modified fashion. Furthermore, in order to
prepare the running example of the next chapters, we report the main
concepts of a feature-based constraint language for HPSG and show how
to embed this constraint language into the CLP scheme, yielding
feature-based CLGs.

\section{Introduction and Overview}
\markboth{Chapter 2. Foundations}{2.1 Introduction and Overview}

{\em Constraint logic programming} is a powerful extension of
conventional logic programming \cite{Lloyd:87}, and involves the 
incorporation of constraint languages and constraint solving methods
into logic programming languages. 
The name CLP was first introduced by \citeN{JuL:86} for a general
framework of a logic programming language that is parametrized with
respect to constraint language and a domain of computation, and yields
soundness and completeness results for an operational semantics
relying on a constraint solver for the employed constraint language.
For example, conventional logic programming or Prolog is obtained from
CLP by employing equations between first order terms as
constraint language and by interpreting these equations in the
Herbrand universe. In this case the operational semantics of SLD-resolution can be
seen to rely on a constraint solver which solves term equations in the
Herbrand universe by term unification. Recent extensions,
refinements, and various applications of CLP are discussed in
\citeN{Jaffar:94}. In the following we will rely on the
general CLP scheme of \citeN{HuS:88}, which has been shown to be a
useful tool for our intended application of linguistic knowledge
representation (see \citeN{Doerre:93}, \citeN{Goetz:95},
\citeN{GoetzMeurers:95}).

The term {\em constraint logic grammars} expresses the
connection between CLP and constraint-based grammars. That is, CLGs are
understood as grammars formulated by means of a suitable logical
language which can be used as a constraint language in the CLP scheme of
\citeN{HuS:88}. The idea behind this connection is to provide an
operational treatment of purely declaratively specified grammars. This
needs further explanation:
Constraint-based grammars enable a clear model-theoretic
characterization of linguistic objects by specifying grammars as sets of
descriptions from a suitable logical description language, called the
constraint language. The descriptions, called constraints, are stated
as axioms required to be true of every
object in the domain to be described, i.e., they constrain the
admissible models of the grammar.
The parsing problem (and similarly the generation
problem) can be defined as follows: Given a set of
axioms (encoding the grammar) and some constraint $\phi$
(encoding the string/logical form we want to parse/generate from), we ask if
there is some model of our axioms which satisfies $\phi$. Following
\citeANP{Goetz:98} (to appear), we will call
this the prediction problem.

A well-known subclass of these grammars widely used in computational
 linguistics are grammars based upon feature description languages such
 as simple PATR grammars \cite{Shieber:86}  or more expressive 
grammars such as LFG \cite{Bresnan:82} or HPSG \cite{PuS:94}.
Formalizations of the more or less
informal notions of these grammars in terms of first-order
languages were firstly presented by 
\citeN{Smolka:88} for PATR and
by \citeN{Johnson:88} and \citeN{King:89}, \citeN{King:94} for LFG and
HPSG, respectively. 

However, such model-theoretic approaches do not necessarily provide an
operational interpretation of their declarative
specifications. This may lead to problems with an operational
treatment of model-theoretically well-defined problems such as parsing or
generation. CLP provides one possible approach to an operational
treatment of various such frameworks by embedding arbitrary logical
languages into constraint logic programs. 
Definite clause specifications over such constraint languages then
define grammars as constraint
logic programs, i.e., as sets of axiomatic interpreted definite clauses.
The prediction problem is in this setting as follows: Given a program ${\cal
  P}$ (encoding a grammar) and a definite goal $G$
(encoding the string/logical form we want to parse/generate from), we ask if we
can infer an answer $\varphi$ of $G$ (which is a satisfiable
constraint encoding an analysis) proving  the implication 
$\varphi \rightarrow G$ to be a logical consequence of ${\cal
  P}$.

For feature-based grammars an embedding of a logical
language close to that of \citeN{Smolka:88} into the CLP scheme of
\citeN{HuS:88} is done in the formalism CUF
\cite{Doerre:91,Doerre:93}.
This approach quite directly offers the operational properties of the
CLP scheme, but unfortunately gives up the connection to the
model-theoretic specifications of the underlying feature-based grammars. 
A different approach is given by \citeN{Goetz:95},
\citeN{GoetzMeurers:95}, who defines an
explicit translation from a logical language close to that of
\citeN{King:94} into constraint logic programs. This translation
procedure preserves the prediction problem by generating a constraint logic program
${\cal P(G)}$ from a feature-based grammar ${\cal G}$ in an explicit
way.
Other approaches to an operational semantics for the prediction
problem of feature-based languages have been presented, e.g., by
\citeN{Carpenter:92}, \citeN{Ait-Kaci:93} or \citeANP{Goetz:98} (to
appear). These approaches are tailored especially for specific
feature-based languages and clearly suit the particular frameworks better
than an embedding of the specific languages into a CLP scheme. 
However, under the CLP approach, arbitrary constraint-based grammars
can receive an unique operational semantics by an embedding into definite clause
specifications\footnote{For example, an embedding of a the logical
  language for tree-description grammars of \citeN{Rogers:94} into the
  CLP scheme of \citeN{HuS:88} is given in \citeN{Morawietz:97}.}.

We see the main advantage of the CLP approach in the possibility to rely on
the well-understood paradigm of logic programming. This allows the
resulting programs to run on existing architectures and to use
well-known optimization techniques worked out in this area. The
possibility to embed arbitrary constraint languages into the CLP
scheme and the broad applicability of CLP itself should generalize the
work of the following chapters in a welcome manner.

This chapter is organized as follows. In Sect. \ref{CLP} we will
report the main concepts of constraint logic programming following
the CLP scheme of \citeN{HuS:88}.
As the work in the next chapters will build upon this 
scheme, we will reformulate the main definitions and propositions
of \citeN{HuS:88} in a form convenient for the following discussions,
and give some missing proofs which will be helpful to make this work
parallel to the work of the next chapters.

In order to provide a concrete instantiation of this CLP scheme to
constraint logic grammars, we will report in Sect. \ref{CLG} a
feature-based constraint language and show how this language can be
embedded into the CLP scheme to yield feature-based CLGs.

\section{Constraint Logic Programming}
\markright{2.2 Constraint Logic Programming}
\label{CLP}

The scheme presented by \citeN{HuS:88} generalizes
conventional logic programming \cite{Lloyd:87} and also the constraint
logic programming scheme of \citeN{JuL:86} to a scheme of definite clause
specifications over arbitrary constraint languages. 
Relying on terminology well-known for conventional logic programming,
H\"ohfeld and Smolka's generalization of the key result of
conventional logic programming can be stated as follows:
First, for every definite clause specification \Xcal{P} in the extension of
an arbitrary constraint language \La, every interpretation of \La can
be extended to a minimal model of \Xcal{P}. Second, the SLD-resolution method for
conventional logic programming can be generalized to a sound and
complete  operational semantics for definite clause specifications,
which are not restricted to Horn theories.
In contrast to \citeN{JuL:86}, in this scheme constraint languages are
not required to be sublanguages of first order predicate logic and do not have to be
interpreted in a single fixed domain. Instead, a constraint is
satisfiable if there is at least one interpretation in which it has a
solution. This makes this scheme usable for a wider range of
applications. Furthermore, such interpretations do not have to be solution
compact\footnote{That is, it is not necessary that every element of an
interpretation must be obtainable as the unique solution of a possibly
infinite set of constraints. See \citeN{JuL:86}.}. This was necessary in \citeN{JuL:86} to provide a sound and
complete treatment of negation as failure. \citeN{HuS:88} do not
include negation as failure but rather let the embedded constraint
language provide for logical negation. 

\subsection{Constraint Languages}

A very general characterization of the concept of constraint language
can be given as follows.

\begin{de}[\La] A constraint language \La consists of
\begin{itemize}
\item an \La-signature, specifying the non-logical elements of
  the alphabet of the language,
\item a decidable infinite set \Xsf{VAR} whose elements are called variables,
\item a decidable set \Xsf{CON} of \La-constraints which are
  pieces of syntax built from the \La-signature, the variables in
\Xsf{VAR}, and the logical elements of the alphabet of the language,
\item a computable function \Xsf{V} assigning to every constraint
  $\phi \in \Xsf{CON}$ a finite set $\Xsf{V}(\phi)$ of variables, the
  variables constrained by $\phi$,
\item a nonempty set of \La-interpretations \Xsf{INT},
where each \La-interpretation $\I \in \Xsf{INT}$ is defined
w.r.t. a nonempty set \Xcal{D}, the domain of \I, and a set \ASS
of variable assignments $\alpha:\Xsf{VAR} \rightarrow \Xcal{D}$,
\item a function \denI{\cdot} mapping
  every constraint $\phi \in \Xsf{CON}$ to a set \denI{\phi} of
  variable assignments, the solutions of $\phi$ in \I.
\item Furthermore, a constraint $\phi$ constrains only the variables
  in $\Xsf{V}(\phi)$, i.e., if $\alpha \in \denI{\phi}$ and $\beta$ is
  a variable assignment that agrees with $\alpha$ on $\Xsf{V}(\phi)$,
  then $\beta \in \denI{\phi}$.
\end{itemize}
\end{de}
In order to state certain closure conditions on constraint languages,
further definitions are necessary. The following definitions are made
with respect to some given constraint language. 

\begin{de} \label{renaming} \mbox{}
\begin{itemize}
 \item A {\bf renaming} is a bijection $\Xsf{VAR} \rightarrow
\Xsf{VAR}$ that is the identity except for finitely many exceptions.
\item A constraint $\phi'$ is a {\bf $\rho$-variant} of a
constraint $\phi$ under a renaming $\rho$ iff
$\phi' = \phi \rho$, i.e., $\phi'$ is the constraint
obtained from $\phi$ by simultaneously replacing each occurence of a
variable $X$ in $\phi$ by $\rho(X)$ for all variables $X$ in
$\Xsf{V}(\phi)$, 
and so $\denI{\phi} = [ \! [  { \phi' } ] \! ] ^ {\cal I}_{\alpha \circ
  \rho} :=
\{ \alpha \circ \rho |$ $\alpha \in \denI{\phi'} \}$,
i.e., the function compositions of the solutions of $\phi'$ and a
renaming $\rho$ yield the solutions of $\phi$, for all
interpretations \I.
\item A constraint $\phi'$ is a {\bf variant} of a constraint $\phi$ if
there exists a renaming $\rho$ s.t. $\phi'$ is a $\rho$-variant of
$\phi$.
\end{itemize}
\end{de}
The following closure conditions on constraint languages will be
convenient in the further discussion.

\begin{de} \label{closure}
A constraint language is 
\begin{itemize}
\item {\bf closed under renaming} iff every
constraint has a $\rho$-variant for every renaming $\rho$,
\item {\bf closed under intersection} iff for every two constraints
  $\phi$ and $\phi'$ there exists a constraint $\psi$
  s.t. $\denI{\phi} \cap \denI{\phi'} = \denI{\psi}$ for every
  interpretation \I,
\item {\bf decidable} iff the satisfiability of its constraints is
  decidable. A constraint $\phi$ is satisfiable iff there exists at
  least one interpretation in which $\phi$ has a solution.
\end{itemize}
\end{de}

\subsection{Relationally Extended Constraint Languages}

To obtain constraint logic programs, a given constraint language
\Xcal{L} has to be extended to a constraint language \Xcal{R(L)}
providing for the necessary relational atoms and propositional
connectives.

\begin{de}[\Rl] A constraint language \Rl extending a
  constraint \\
language \La is defined as follows:
\begin{itemize}
\item The signature of \Rl is an extension of the signature of \La
  with a decidable set \Xcal{R} of relation symbols and an arity
  function $\Xsf{Ar}: \Xcal{R} \rightarrow \Nats$.
\item The variables of \Rl are the variables of \La.
\item The set of \Xcal{R(L)}-constraints is the smallest set s.t.
\begin{enumerate}
\item $\phi$ is an \Xcal{R(L)}-constraint if $\phi$ is an
\Xcal{L}-constraint,
\item $r(\vec{x})$ is an \Xcal{R(L)}-constraint, called an {\bf
atom},  if $r \in \Xcal{R}$ is a relation symbol with arity n and
$\vec{x}$ is an n-tuple of pairwise distinct variables,
\item $\emptyset$, $F \:\&\:  G$, $F \rightarrow G$ are 
\Xcal{R(L)}-constraints, if F and G are \Xcal{R(L)}-constraints,
\item $\phi$ $\&$ $B_1$ $\&$ $\ldots$ $\&$ $B_n \rightarrow A$  is 
an \Xcal{R(L)}-constraint, called a {\bf definite clause}, if A, $B_1,
\ldots, B_n$ are atoms and $\phi$  is an \Xcal{L}-constraint. We may
write a definite clause also as
$A \leftarrow \phi$ $\&$ $B_1$ $\&$ $\ldots$ $\&$ $B_n$.
\end{enumerate}
\item The variables constrained by an \Rl-constraint are defined as follows:
  If $\phi$ is an \La-constraint, then $\Xsf{V}(\phi)$ is defined as
  in \La;
$\Xsf{V}(r(x_1, \ldots, x_n)):= \{ x_1, \ldots, x_n \}$; 
$\Xsf{V}(\emptyset):= \emptyset$;
$\Xsf{V}(F \;\&\; G):= \Xsf{V}(F) \cup \Xsf{V}(G)$;
$\Xsf{V}(F \rightarrow G) := \Xsf{V}(F) \cup \Xsf{V}(G)$.
\item For each \La-interpretation \I,
  an \Xcal{R(L)}-interpretation \Xcal{A} is an extension of an
\Xcal{L}-inter-pretation \Xcal{I} with
relations $r^{\cal A}$ on the domain $\cal D$  of \Xcal{A} with
appropriate arity for every $r \in \Xcal{R}$, and
the domain of \Xcal{A} is the domain of \Xcal{I}.
\item For each \Rl-interpretation \A, for each \La-interpretation \I, \denA{\cdot} is a
  function mapping every \Rl-constraint to a set of
  variable assignments s.t.
\begin{enumerate}
\item $\denA{\phi} = \denI{\phi}$ if $\phi$ is an 
\Xcal{L}-constraint,
\item $\denA{r(\vec{x})} = \{ \alpha \in \Xsf{ASS}|$ 
$\alpha(\vec{x}) \in r^{\cal A} \}$,
\item $\denA{\emptyset} = \Xsf{ASS}$,
\item $\denA{F \;\&\; G} = \denA{F} \cap \denA{G}$,
\item $\denA{F \rightarrow G} = (\Xsf{ASS} \setminus \denA{F}) \cup \denA{G}$.
\end{enumerate}
\end{itemize}
\end{de}
Note that we slightly abuse the notation $\alpha(\vec{x})$ to
abbreviate the notation $(\alpha(x_1), \alpha(x_2), \ldots,
\alpha(x_n))$ for a n-tuple of objects assigned to a n-tuple
$\vec{x}$ of variables by a variable assignment $\alpha$.

\subsection{Syntax and Declarative Semantics of Definite Clause Specifications}

The concept of a constraint logic program now can be defined as a
definite clause specification over a constraint language.

\begin{de}[Definite clause specification] A definite clause
  specification \Xcal{P} over a constraint language \Xcal{L} is a set of
  definite clauses from a constraint language \Xcal{R(L)} extending \Xcal{L}.
\end{de}
Models of definite clause specifications are determined by
the definite clauses constituting these specifications, i.e., a definite
clause specification has its definite clauses as its axioms.
For reasons of generality, the following two definitions are made with
respect to general sets of \Rl-constraints.

\begin{de}[Model] \label{model}
An \Xcal{R(L)}-interpretation \Xcal{A} is a model of a set $\Psi$ of
\Rl-con-straints iff for every
  $\alpha \in \Xsf{ASS}$, for every $\psi \in \Psi$: $\alpha \in
  \denA{\psi}$.
\end{de}
For convenience we furthermore introduce the concept of logical consequence.

\begin{de}[Logical consequence] \label{lc}
An \Xcal{R(L)}-constraint $\psi$ is a logical 
consequence of set $\Psi$ of \Rl-constraints iff,
  for every \Xcal{R(L)}-interpretation \Xcal{A},
  \Xcal{A} is a model of $\Psi$ implies that \Xcal{A} is a model of
  $\psi$.
\end{de}
A {\bf goal} $G$ is defined as a possibly empty conjunction of
\Xcal{L}-constraints and \Xcal{R(L)}-atoms.

Given a definite clause specification \Xcal{P} and a goal $G$, a {\bf 
\Xcal{P}-answer} of $G$ is defined as a satisfiable \Xcal{L}-constraint $\phi$
such that the implication $\phi \rightarrow G$ is a logical consequence of \Xcal{P}.

In order to show that the semantic properties of conventional
logic programming extend to CLP, \citeN{HuS:88} first define a partial
ordering on the set of \Xcal{R(L)}-interpretations.
\Rl-interpretations extending the same \Xcal{L}-interpretation \I are called base
equivalent, and \I is called the base of these \Rl-interpretations. A partial
ordering on such \Rl-interpretations is defined via a partial ordering
on the set of the denotations of the relation symbols in these
interpretations. We get for all base equivalent \Rl-interpretations
${\cal A}, \;{\cal A'}$:
\begin{itemize}
\item ${\cal A} \subseteq {\cal A'}$ iff
for each n-ary relation symbol $r \in \R$:
$r^{{\cal A}} \subseteq r^{{\cal A'}}$, 
\item $\A = \bigcup X$ iff 
for each n-ary relation symbol $r \in \R$:
$r^{\cal A} = \bigcup \{ r^{\cal A'} | \; {\cal A'} \in X \}$,
\item $\A = \bigcap X$ iff 
for each n-ary relation symbol $r \in \R$:
$r^{\cal A} = \bigcap \{ r^{\cal A'}| \; {\cal A'} \in X \}$.
\end{itemize}
This set of base equivalent \Xcal{R(L)}-interpretations is a complete
lattice under the partial order of set inclusion. That is, for every
set of base-equivalent \Rl-interpretations we have a supremum, given by the
union, and an infimum, given by the intersection of the
interpretations in the set. The top element is the \Rl-interpretation
$\mathcal{A}^\top$ such that for each n-ary relation symbol
$r \in \R: r^{\mathcal{A}^\top} = \mathcal{D}^{\mathsf{Ar}(r)}$,
and the bottom element is $\mathcal{A}^\bot$ s.t for each n-ary
relation symbol $r \in \R: r^{\mathcal{A}^\bot} = \emptyset$.

Proposition \ref{definiteness}, due to \citeN{HuS:88}, generalizes the
fixpoint- or lattice-theoretic semantics of conventional logic
programming to CLP.
It says that for each \La-interpretation \I, a definite clause specification \Po
in \Rl defines unique minimal denotations for the relation symbols of
\R. That is, every \La-interpretation \I can be used to construct a
minimal model for \Po in \Rl. All questions concering
the declarative semantics of CLP can then be dealt with in terms of a
minimal model semantics. Moreover, a minimal model semantics is
crucial for the construction of a sound and complete deduction system
for CLP.

\begin{po}[\citeN{HuS:88}, Theorem 4.4.] \label{definiteness}
Let \Xcal{I} be an \Xcal{L}-interpre-tation and \Xcal{P} be a definite clause
  specification in \Xcal{R(L)}. Then the equations
\begin{quote}
$r^{{\cal A}_0} := \emptyset $, \\
$r^{{\cal A}_{i+1}} := \{ \alpha(\vec{x})|$ there is a clause $(r(\vec{x}) 
\leftarrow G ) \in \Xcal{P}$ and $\alpha \in [ \! [  G  ] \! ] ^ 
{{\cal A}_i} \} $
\end{quote}
(i) define a chain $\Xcal{A}_0 \subseteq \Xcal{A}_1 \subseteq 
\ldots $ of \Xcal{R(L)}-interpretations extending \Xcal{I},  \\
(ii) the union $\Xcal{A} := \bigcup_{i \ge 0} \Xcal{A}_i$ is a model
of \Xcal{P} extending \I, \\
(iii) \A is the minimal model of \Xcal{P} extending \I.
\end{po}

Proposition \ref{CLPLogicalConsequence} connects the concept of a
\Xcal{P}-answer with the minimal model semantics of \Xcal{P} (see
\citeN{HuS:88}, Proposition 4.5.). This proposition justifies the restriction of the
declarative semantics of CLP to a minimal model semantics.
We prove this proposition explicitly with reference to the
concept of logical consequence.

\begin{po} \label{CLPLogicalConsequence}
For each definite clause specification \Xcal{P} in \Rl, for
  each goal $G$, for each \La-constraint $\phi$:
$\phi \rightarrow G$ is a logical consequence of \Xcal{P} iff
each minimal model \Xcal{A} of \Xcal{P} is a model of $\phi \rightarrow G$.
\end{po}

\begin{proof}
{\em If:} For each minimal model \A of \Xcal{P}:
\Xcal{A} is a model of $\phi \rightarrow G$ 
\begin{description}
\then for every model \Xcal{B} of \Xcal{P} base equivalent to some minimal model
\A of \Xcal{P}: \Xcal{B}
is a model of $\phi \rightarrow G$, since $\Xcal{A} \subseteq
\Xcal{B}$ by Proposition \ref{definiteness}
\then $\phi \rightarrow G$ is a logical consequence of \Xcal{P}.
\item[]{\em Only if:}
$\phi \rightarrow G$ is a logical consequence of \Xcal{P}
\then every model of \Xcal{P} is a model of $\phi \rightarrow G$, by
Definition \ref{lc}
\then \Xcal{A} is a model of $\phi \rightarrow G$.
\qed
\end{description}
\renewcommand{\qed}{}
\end{proof}

\subsection{Operational Semantics of Definite Clause Specifications}
The following definitions are made with respect
to some implicit \La, \R, \Xcal{P}, and \Xsf{V}, where \Xsf{V}
denotes the finite set of variables in the query and the \Xsf{V}-solutions of
a constraint $\phi$ in an interpretation \I are defined as
$\denIV{\phi} := \{ \alpha|_\mathsf{V} |\; \alpha \in \denI{\phi} \}$ and
$\alpha|_\mathsf{V}$ is the restriction of $\alpha$ to \Xsf{V}.

\citeN{HuS:88} define the generalization of the SLD-resolution rule by a binary relation
$\stackrel{r}{\longrightarrow}$, called {\bf goal reduction},
on the set of goals. The rule selects the leftmost atom in the goal,
looks for a variant of a program clause with the selected atom as
head, and replaces the selected atom in the goal by the body of the
variant clause. Furthermore, the rule ensures that no accidental variable
sharing is introduced by the variant.

\begin{description}
\item[$A$ $\&$ $G$ $\stackrel{r}{\longrightarrow}$ $F$ 
$\&$ $G$]
if $A \leftarrow F$ is a variant of a clause in \Xcal{P}\\
s.t. $(\Xsf{V} \cup \Xsf{V}(G)) \cap \Xsf{V}(F) \subseteq \Xsf{V}(A)$.
\end{description}
A second rule takes care of {\bf constraint solving} for the
\Xcal{L}-constraints appearing in subsequent goals. The rule takes the
conjunction of the \Xcal{L}-constraints from the reduced goal and the
applied clause and gives, via the black box of a suitable \Xcal{L}-
constraint solver, a satisfiable \Xcal{L}-constraint in solved form if the
conjunction of \Xcal{L}-constraints is satisfiable. If the conjunction
of \La-constraints is not satisfiable, an \La-constraint $\bot$ denoting
failure is returned. The constraint solving rule can then be defined as a total function
$\stackrel{c}{\longrightarrow}$ on the set of goals. 

\begin{description}
\item[$\phi$ $\&$ $\phi'$ $\&$ $G$ $\stackrel{c}{\longrightarrow}$ 
$\phi''$ $\&$ $G$]
if $ [ \! [  \phi \:\&\: \phi' ] \! ] ^{\cal I}
_{{\sf V} \cup {\sf V}(G)} = [ \! [   \phi'' ] \! ] ^{\cal I}
_{{\sf V} \cup {\sf V}(G)}$\\
for all \Xcal{L}-interpretations \Xcal{I} and for all \La-constraints
$\phi, \phi'$ and $\phi''$.
\end{description}

Furthermore, a complexity measure that mirrors the
construction steps of a minimal model in the complexity of goal
reduction is introduced. This measure will be crucial for proving completeness of
goal reduction.
\begin{itemize}
\item The complexity of a variable assignment $\alpha$ for an atom $A$
in the minimal model \Xcal{A} where $\alpha \in \denA{A}$ 
is defined as 
\[
comp(\alpha,A,\Xcal{A})
:= \min \{ i |\;\alpha \in \denAi{A}{i} \}
 ; 
\]
\item The complexity of
$\alpha$ for goal $G$ in \Xcal{A} where $\alpha \in \denA{G}$
is 
\[
comp(\alpha, G,\Xcal{A}) := \{ comp(\alpha, A, \A) |\; A \textrm{ is an
atom in } G \} 
\]
where $ \{ \ldots \}$ is a multiset;
\item The \textsf{V}--complexity of $\alpha$ for $G$ in
\Xcal{A} where $\alpha \in \denAV{G}$
is 
\[
comp_\mathsf{V}(\alpha,G,\Xcal{A}) := \min \{ comp(\beta,G,
\Xcal{A}) | \; \beta \in \denA{G} \textrm{ and } \alpha = \beta
|_\mathsf{V} \}
\]
where $\beta|_\mathsf{V}$ is the restriction of $\beta$ to the variables in
\Xsf{V}, and the minimum is taken with respect to a total ordering on
multisets such that $M \leq M'$ iff $\forall x \in M \setminus M',
\exists x' \in M' \setminus M$ s.t. $x < x'$. 
\end{itemize}

\citeN{HuS:88} prove the following propositions showing that goal
reduction is a sound and complete rule for deducing \Xcal{P}-answers
from general definite clause specifications. We prove the main results
explicitly in Propositions \ref{CLPSoundness} (soundness) and
\ref{CLPCompleteness} (completeness). Note that soundness and
completeness can be proven without reference to constraint solving,

\begin{po}[\citeN{HuS:88}, Proposition 5.1.]\label{sound} 
If $G_1 \stackrel{r}{\longrightarrow} G_2$, then
$\denA{G_2} \subseteq \denA{G_1}$ for every model \Xcal{A} of
\Xcal{P}.
\end{po}

\begin{po}\label{CLPSoundness}
If $G  \stackrel{r}{\longrightarrow} \!\mbox{}^\ast
\phi$, then $\phi \rightarrow G$ is a
logical consequence of \Xcal{P}.
\end{po}

\begin{proof}
$G \stackrel{r}{\longrightarrow}\!\mbox{}^\ast
\phi$
\begin{description}
\then $\denA{\phi} \subseteq \denA{G}$ for every model \Xcal{A} of
\Xcal{P}, by Proposition \ref{sound} and transitivity of $\subseteq$
\then for every model \Xcal{A} of \Xcal{P}: $\denA{\phi\rightarrow G}
= \Xsf{ASS}$, since for every model \Xcal{A} of \Xcal{P}:
$\denA{\phi} \subseteq \denA{G}$
\then for every model \Xcal{A} of \Xcal{P}: \Xcal{A} is a model of
$\phi \rightarrow G$, by Definition \ref{model}
\then $\phi \rightarrow G$ is a logical consequence of \Xcal{P}.
\qed
\end{description}
\renewcommand{\qed}{}
\end{proof}

\begin{po}[\citeN{HuS:88}, Theorem 5.2.] \label{comp}
Let \Xcal{L} be closed under renaming, \Xcal{A} be a minimal model of
\Xcal{P}, $G_1$ be a goal, $A$ be an atom in $G_1$, and
  $\alpha \in \denAV{G_1}$. Then there exists a clause $C$ in \Xcal{P}
  and a goal $G_2$ s.t.
$G_1 \stackrel{r}{\longrightarrow} G_2$ using a variant of $C$
on $A$ is possible,
$\alpha \in \denAV{G_2}$ and
$comp_\mathsf{V}(\alpha,G_2,\Xcal{A}) < comp_\mathsf{V}(\alpha,G_1,\Xcal{A})$.
\end{po}

\begin{po}[\citeN{HuS:88}, Corollary 5.3.]\label{CLPCompleteness}
 Let \Xcal{L} be closed
under renaming, \Xcal{A} be a minimal model of \Xcal{P}, $G$ be a goal
and $\alpha \in \denAV{G}$. Then there exists a \Xcal{P}-answer $\phi$ of
$G$ s.t. $G \stackrel{r}{\longrightarrow}\!\mbox{}^\ast\phi$ and
$\alpha \in \denAV{\phi}$.
\end{po}

\begin{proof}
The result is proven by induction on 
$comp_\mathsf{V}(\alpha,G,\Xcal{A})$.
\begin{description}
\item[{\rm Base:}] Goals with mulitset complexity $\emptyset$ have to be a satisfiable
\Xcal{L}-constraint $\phi$. Then $\phi \stackrel{r}{\longrightarrow}
\!\mbox{}^0 \phi$ and $\phi$ is a \Xcal{P}-answer of itself.
\item[{\rm Hypothesis}:] Suppose the result holds for goals with multiset complexity less than
some multiset $N$.
\item[{\rm Step}:]
$comp_\mathsf{V}(\alpha',G_1,\Xcal{A})=N$  and $\alpha' \in
\denAV{G_1}$ 
\begin{description}
\then there exists a clause $C$ of \Xcal{P} and a goal $G_2$ s.t. $G_1
\stackrel{r}{\longrightarrow} G_2$ and $\alpha' \in \denAV{G_2}$ and
$comp_\mathsf{V}(\alpha',G_2,\Xcal{A}) < comp_\mathsf{V}(\alpha',G_1,\Xcal{A})$, by
Proposition \ref{comp}
\then there exists a \Xcal{P}-answer $\phi$ of $G_2$ s.t. $G_2
\stackrel{r}{\longrightarrow}\!\mbox{}^\ast \phi$ and
$\alpha' \in \denAV{\phi}$,
by the hypothesis
\end{description}
\then there exists a \Xcal{P}-answer $\phi$ of $G_1$ s.t. $G_1 \stackrel{r}{\longrightarrow}\!\mbox{}^\ast \phi$
and
$\alpha' \in \denAV{\phi}$, and by Proposition \ref{CLPSoundness},
$\phi \rightarrow G_1$ is a logical consequence of \Xcal{P}.
\item[] The result follows by arithmetic induction.
\qed
\end{description}
\renewcommand{\qed}{}
\end{proof}

In all following examples, we will use a standard Prolog resolution
procedure for the CLP scheme of \citeN{HuS:88}, i.e., we combine the
left-right selection rule defined in goal reduction with a depth-first
search rule. Furthermore, after each goal reduction step, constraint
solving is applied, and another clause is tried immediately if constraint solving fails.  
Moreover, it will be convenient in the following discussion to view
the search space determined by the derivation rules \gr and \cs as a
search of a tree. A derivation tree is defined as follows.

\begin{de}[Derivation tree] \label{DerivationTree} 
A derivation tree determined by a query $G_1$ and a definite clause
specification \Po has to satisfy the following conditions:
\begin{enumerate}
\item Each node is either a relation node or a constraint node.
\item The successors of every relation node are all constraint nodes
  s.t. for every \gr-resolvent $G'$ obtainable by a clause $C$ from
  goal $G$ in a relation node, there is a successor constraint node
  labeled by $C$ and $G'$.
\item The successors of every constraint node are all relation nodes
  s.t. for the unique \cs-resolvent $G \:\&\: \phi''$ obtainable from
  goal $G \:\&\: \phi \:\&\: \phi'$ in a constraint node, there is a
  successor relation node labeled by $G \:\&\: \phi''$.
\item The root node is a relation node labeled by $G_1$.
\item A success node is a terminal relation node labeled by a
  satisfiable \La-constraint.
\end{enumerate}
\end{de}

Successful derivations correspond to subtrees of derivation trees
which are labeled by terminal success nodes. Such trees can be
defined as proof trees as follows.

\begin{de}[Proof tree] \label{ProofTree}
A proof tree for a query $G_1$ from \Po is a subtree of a derivation
tree determined by $G_1$ and \Po and is defined as follows:
\begin{enumerate}
\item A relation node of the proof tree is a relation node of the
  supertree and takes one of the successors of the relation node of the
  supertree as its successor node.
\item A constraint node of the proof tree is a constraint node of the
  supertree and takes the unique successor of the constraint node of
  the supertree as its successor node.
\item The root node of the proof tree is the root node of the
  supertree.
\item The terminal node of the proof tree is a success node of the
  supertree, labeled by a satisfiable \La-constraint, called answer
  constraint.
\end{enumerate}
\end{de}

Let us illustrate the basic concepts of CLP with an example. A
simple program consisting of clauses \texttt{1} to \texttt{3} is
depicted in Fig. \ref{CLPprogram}. 

\begin{figure}[htbp]
\begin{center}
\begin{tabular}{l}
\texttt{1} $\texttt{q}(X) \leftarrow \texttt{p}(X).$ \\
\texttt{2} $\texttt{p}(X) \leftarrow X=a.$\\
\texttt{3} $\texttt{p}(X) \leftarrow X=b.$
\end{tabular}
\end{center}

\caption{Constraint logic program}
\label{CLPprogram}
\end{figure}

\begin{figure}[htbp]
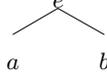

{\footnotesize
\setlength{\GapWidth}{30pt}
\setlength{\GapDepth}{10pt}
\begin{center}
  \begin{bundle}{$e$}
    \chunk{$a$}
    \chunk{$b$}
  \end{bundle}
\end{center}
}
\caption{Type hierarchy}
\label{CLPtypes}
\end{figure}

The \La-constraints are considered to come from a language of hierarchical types, where
the ordering on types is defined via the operation of set inclusion on
their denotations. In our example, we have $\denI{a} \subseteq
\denI{e}, \; \denI{b} \subseteq \denI{e}$ and $\denI{a} \cap \denI{b}
= \emptyset$. This hierarchy is depicted graphically in
Fig. \ref{CLPtypes}.

The construction of a minimal model for the program of
Fig. \ref{CLPprogram} is shown in Fig. \ref{CLPMinimalModel}. The
unique minimal denotations of the relation symbols \texttt{p} and
\texttt{q} are obtained in step 1 and 2 of the minimal model
construction respectively.

\begin{figure}[htbp]
\begin{center}
\begin{tabular}{l}
$\texttt{p}^{\mathcal{A}_0} = \emptyset,
\texttt{q}^{\mathcal{A}_0} = \emptyset,$ \\
$\texttt{p}^{\mathcal{A}_1} = \{ \denI{a}, \denI{b} \},
\texttt{q}^{\mathcal{A}_1} = \emptyset, $\\
$\texttt{p}^{\mathcal{A}_2} = \{ \denI{a}, \denI{b} \},
\texttt{q}^{\mathcal{A}_2} = \{ \denI{a}, \denI{b} \}, $\\
$\vdots $\\
$\texttt{p}^{\mathcal{A}} = \{ \denI{a}, \denI{b} \},
\texttt{q}^{\mathcal{A}} = \{ \denI{a}, \denI{b} \}, \textrm{ where }
\mathcal{A} = \bigcup_{i \geq 0} \mathcal{A}_i$.
\end{tabular}
\end{center}
\caption{Minimal model construction for constraint logic program}
\label{CLPMinimalModel}
\end{figure}

\begin{figure}[htbp]
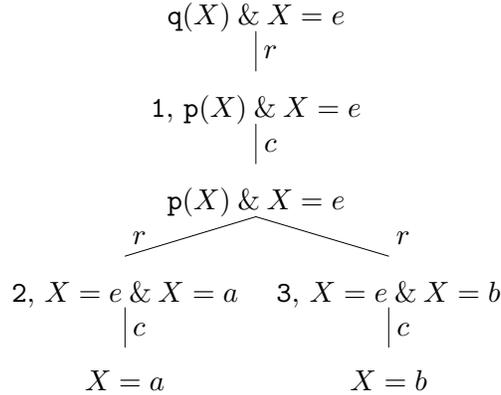

\begin{center}
\setlength{\GapWidth}{70pt}

\begin{bundle}{$\texttt{q}(X) \:\&\: X=e$}
  \chunk[$\quad r$]{
    \begin{bundle}{\texttt{1}, $\texttt{p}(X) \:\&\: X=e$} 
      \chunk[$\quad c$]{
        \begin{bundle}{$\texttt{p}(X) \:\&\: X=e$}
          \chunk[$\quad r$]{
            \begin{bundle}{\texttt{2}, $X=e \:\&\: X=a$}
              \chunk[$\quad c$]{$X=a$}
            \end{bundle}
            }
          \chunk[$\quad r$]{
            \begin{bundle}{\texttt{3}, $X=e \:\&\: X=b$}
              \chunk[$\quad c$]{$X=b$}
            \end{bundle}
            }
        \end{bundle}
        }
    \end{bundle}
    }
\end{bundle}

\end{center}
\caption{Derivation tree for constraint logic program}
\label{CLPDerivationTree}
\end{figure}

A derivation tree for the query $\texttt{q}(X) \:\&\: X=e$ from the
program of Fig. \ref{CLPprogram} is given in
Fig. \ref{CLPDerivationTree}. We depict only the success branches of
the derivation tree, yielding two distinct proof trees for the query, with answer
constraints $X=a$ and $X=b$ respectively.

Soundness of the CLP scheme implies that corresponding to the
derivation of $X=a$ and $X=b$, we know that the implications
$X=a \rightarrow \texttt{q}(X) \:\&\: X=e$ and
$X=b \rightarrow \texttt{q}(X) \:\&\: X=e$ are logical consequences of
the program of Fig. \ref{CLPprogram}. This is easily verified from the
minimal model given in Fig. \ref{CLPMinimalModel}.
Furthermore, completeness of the CLP scheme is easily verified from
the fact that for solutions 
$\alpha \in \denAV{\texttt{q}(X) \:\&\: X=e}$ and
$\alpha' \in \denAV{\texttt{q}(X) \:\&\: X=e}$, we can derive
\Po-answers with $\alpha \in \denAV{X=a}$ and $\alpha' \in
\denAV{X=b}$.

\section{Constraint Logic Grammars}
\markright{2.3 Constraint Logic Grammars}
\label{CLG}

In this section, we will explicate the concept of constraint logic
grammars. To this end, we will restrict our attention to feature-based
CLGs and discuss in particular the main properties of an
HPSG instance of such grammars.

We will show how a feature-based constraint language
can be obtained from a feature-based logical description language, and
how such a constraint language can be embedded into the CLP scheme of
\citeN{HuS:88}, yielding a feature-based CLG. The
language to be discussed is that of \citeN{Goetz:95},
\citeANP{Goetz:98} (to appear),
which is close to that of \citeN{King:89}, \citeN{King:94} (modulo the usage of
variables) and \citeN{Smolka:88}, \citeN{Smolka:92} (modulo appropriateness
conditions).
This language provides a description language \Xcal{FD}
specifying the logical foundations of HPSG grammars and is extendable
to a constraint language \Xcal{FL}, in the sense of \citeN{HuS:88}.
The expressive power of the
language is smaller than or equal to the expressive power of
first-order predicate logic with equality.

\subsection{A Feature-Based Constraint Language}

The language is based on a notion of signature, i.e., the non-logical
elements of the alphabet, declaring the structures the linguist is 
interested in.  A signature specifies a set of feature
symbols, a lattice of sort symbols and
appropriateness conditions restricting the functional 
properties of the feature symbols. All subsequent work should be
understood with respect to an implicit signature $\Sigma$.

\begin{de}[Signature]
A signature is a quadruple $\left< \Xcal{T}, \preceq, \Xcal{F}, 
approp \right>$ s.t.
\begin{itemize}
\item $\left< \Xcal{T}, \preceq \right>$ is  a finite join-semilattice
  of types,
\item $\Xcal{S} =\{ t \in \Xcal{T}|$ if $ t' \preceq t $ then $
 t' = t \}$ is a finite set of minimal types,
\item \Xcal{F} is a finite set of feature symbols,
\item $approp: \Xcal{S} \times \Xcal{F} \rightharpoonup 
\Xcal{T}$ is a partial function from pairs of minimal types and
features to types.
\end{itemize}
\end{de}

The well-formed formulae of the feature-based description language \Xcal{FD}, 
called feature descriptions, are built from the symbols in the signature, 
a countably infinite set of
variables {\sf VAR}, the symbol : assigning features to their values,
and the standard boolean connectives.  
Expressions of this kind can be seen as the formal equivalent of the 
AVM notation used in \citeN{PuS:94}. The set \Xsf{Desc} of feature
descriptions is defined as follows.

\begin{de}[Feature descriptions]  The set \Xsf{Desc} of feature
  descriptions is the smallest set s.t.

\renewcommand{\arraystretch}{1.5}
\begin{tabular}{ll}
$\bullet$ X is a description &  if X $\in$ {\sf VAR}, \\
$\bullet$ t is a description & if t $\in$ \Xcal{T}, \\
$\bullet$ f:D is a description & if f $\in$ \Xcal{F}, $D \in
\Xsf{Desc}$, \\
$\bullet$ $ D_1 \wedge D_2, \; D_1 \vee D_2, \; \neg D_1, \; D_1  
\rightarrow D_2$ are descriptions & if $D_1\in \Xsf{Desc}, \; D_2 \in \Xsf{Desc}.$ 
\end{tabular}
\end{de}

An interpretation of a signature is based on an arbitrary domain of
objects, and assigns to every object exactly one minimal type, and to
every feature symbol a partial function on the domain. The domains and
ranges of these functions are determined by the $approp$
function. This function specifies that for each object $u$ of a minimal
type $s$, there is a connected object $\Xsf{F}(f)(u)$ defined iff
$approp(s,f)$ is defined, and the type $\Xsf{S}(\Xsf{F}(f)(u))$ of
this connected object has to be appropriate.

\begin{de}[Interpretation]
An interpretation is a quadruple  $\Xcal{I} = \left< \Xsf{U, S, F}
\right>$ s.t. 
\begin{itemize}
\item \Xsf {U} is a set of objects, the domain of \Xcal{I},
\item  \Xsf{S: U} $\rightarrow$ \Xcal{S}  is a total function
  from the domain to the set of minimal types,
\item \Xsf{F:} \Xcal{F} $\rightarrow \Xsf{U}^{\Xsf{U}}$ is a
is a total feature interpretation function s.t. 
\begin{enumerate}
\item for each u $\in$ \Xsf {U}, for each f $\in$ \Xcal{F}, if
$approp(\Xsf{S}(u), f)$ is defined and \\
$approp(\Xsf{S}(u),f) = t$, then
$\Xsf{F}(f)(u)$ is defined and $\Xsf{S}(\Xsf{F}(f)(u))
 \preceq t$,
\item for each u $\in$ \Xsf{U}, for each f $\in$ \Xcal{F}, if
  $\Xsf{F}(f)(u)$ is defined, then \\
$approp(\Xsf{S}(u),f)$ is defined
  and $\Xsf{S}(\Xsf{F}(f)(u)) \preceq approp(\Xsf{S}(u),f)$.
\end{enumerate}
\end{itemize}
\end{de}

The denotation of feature descriptions with respect to an interpretation \Xcal{I}
and a variable assignment $\alpha$ is defined to be a subset of
the domain for every feature description. By abstracting away from the
variable assignment, we arrive at a concept of abstract denotation
comprising the denotation of a feature description under every possible
variable assignment.

\begin{de}[Variable assignment]
A variable assignment $\alpha : \Xsf{VAR} \rightarrow \Xsf{U}$ is a
 total function from the set of variables to the domain. Write
 \Xsf{ASS} for the set of variable assignments. 
\end{de}

\begin{de}[Feature description denotation]  \mbox{}

\renewcommand{\arraystretch}{1.5}
\begin{tabular}{ll}
$\bullet$  $\denIa{X} = \{ \alpha(X) \}$ &  if X $\in$ {\sf VAR},\\
$\bullet$ $ \denIa{t} = \{ u \in \Xsf{U} |$ $ \Xsf{S}(u) \preceq  t \}
$  & if $t \in \Xcal{T}$,\\ 
$\bullet$ $ \denIa{f:D} = \{ u \in \Xsf{U} |$ $ \Xsf{F}(f)(u)$ is defined,
$\Xsf{F}(f)(u) \in \denIa{D}\} $ & if f $\in$ \Xcal{F}, $D \in
\Xsf{Desc}$,\\ 
$\bullet$ $\denIa{D_1 \wedge D_2} = \denIa{D_1} \cap  
\denIa{D_2}$ & if $D_1, \; D_2 \in \Xsf{Desc}$,\\
$\bullet$ $\denIa{D_1 \vee D_2} = \denIa{D_1} \cup  
\denIa{D_2}$ & if $D_1, \; D_2 \in \Xsf{Desc}$,\\
$\bullet$ $\denIa{\neg D_1} = \Xsf{U} \setminus \denIa{D_1}$ &
if $D_1 \in \Xsf{Desc}$,\\  
$\bullet $ $\denIa{D_1 \rightarrow D_2} = (\Xsf{U} \setminus  
\denIa{D_1}) \cup \denIa{D_2}$ & if $D_1, \; D_2 \in \Xsf{Desc}.$
\end{tabular}
\end{de}

\begin{de}[Abstract denotation] \mbox{}

$\denI{D} = \bigcup\limits_{\alpha \in \Xsf{ASS}} \denIa{D}$ 
\hspace{1cm} if $D \in \Xsf{Desc}.$ 
\end{de}

To obtain a feature-based constraint language \Xcal{FL} fulfilling the
closure requirements on constraint languages stated by \citeN{HuS:88}, first
we simply have to attach every feature description $D$ in \Xcal{FD}
with a new variable not occuring in the set $\Xsf{V}(D)$ of variables
in $D$. This avoids accidental variable sharing and guarantees renaming closure of
\Xcal{FL}. Furthermore, an explicit definition of conjunction of feature
constraints ensures intersection closure of \Xcal{FL}.

\begin{de}[Feature constraints] \label{FeatureConstraints} \mbox{}

\renewcommand{\arraystretch}{1.5}
\begin{tabular}{ll}
$\bullet$ $X = D$ is a constraint & if $X \in \Xsf{VAR},$ $X \not 
\in \Xsf{V}(D)$, $D \in \Xsf{Desc}$, \\
$\bullet$ $\phi \:\&\: \phi'$ is a constraint &
if $\phi, \; \phi'$ are constraints.
\end{tabular}
\end{de}

The denotation of a constraint is defined by a function mapping every
constraint to a set of variable assignments, called solutions.  The
solutions of a constraint $X = D$ are the variable assignments in \ASS which
constrain the value of the variable $X$ to the objects in the
denotation of $D$. The denotation of a conjunction of constraints is
the intersection of the respective denotations.

\begin{de}[Feature constraint solutions] \mbox{}

\renewcommand{\arraystretch}{1.5}
\begin{tabular}{ll}
$\bullet$ $\denI{X = D} = \{ \alpha \in \Xsf{ASS} |$ $ \alpha(X) \in
\denIa{D} \}$ & if $X \in \Xsf{VAR}$, $X \not\in \Xsf{V}(D)$, $D \in 
\Xsf{Desc}$, \\
$\bullet$ $\denI{\phi \:\&\: \phi'} =
\denI{\phi} \cap \denI{\phi'}$ & if $\phi, \; \phi'$ are constraints.
\end{tabular}
\end{de}

Next we have to consider the problem of deciding satisfiability
of feature descriptions and feature constraints.

\begin{de}[Satisfiability of feature descriptions] 
A feature description $D$ is satisfiable iff  there is an interpretation \Xcal{I}
s.t. $\denI{D} \neq \emptyset$. 
\end{de}

This problem has been shown to be decidable for feature-based description
languages closely related to the above reported one. For the
description language reported above, a decision algorithm is given by
\citeANP{Goetz:98} (to appear), for the variable-free notational
variant of \citeN{King:94} by \citeN{Kepser:94}, for a less expressive
version of the language not employing appropriateness conditions by
\citeN{Smolka:88}, \citeN{Smolka:92}, or
for an even less expressive version employing conjunction as only
boolean operator by \citeN{Ait-Kaci:93}. 

Most of these approaches adapt for satisfiability checking a
constraint solving method similar to that of \citeN{Smolka:88},
\citeN{Smolka:92}. This method is a three-step transformation process from
feature descriptions to a solved form of feature constraints displaying (un)satisfiability.
Following \citeANP{Goetz:98} (to appear), constraint solving for the
feature-based constraint language reported above can be illustrated as follows:
Firstly, every feature description is transformed to disjunctive normal form;
secondly, every feature description in disjunctive
normal form is transformed into a (disjunctively interpreted) set of
(conjunctively interpreted) sets of feature constraints of the simple form
 $X=Y$, $X= \neg Y$, $X=t$ or $X=f:Y$;
thirdly, every such set of sets of simple feature constraints is
transformed into a set of sets of feature constraints in solved form.

For reasons of readability, we will consider the constraint solver for
the feature-based constraint language \Xcal{FL} in the following as a
black box. The interested reader is referred for details and proofs to
\citeANP{Goetz:98} (to appear). In all subsequent examples, we will
depict only the result of constraint solving, re-translated from
simple feature constraints in solved normal form to feature
constraints in a more readable form according to Definition
\ref{FeatureConstraints}. 

The notion of satisfiability defined for feature constraints is as follows.

\begin{de}[Satisfiability of feature constraints] 
A feature constraint $\phi$ is satisfiable iff there exists an interpretation
\Xcal{I} s.t. $\denI{\phi} \neq \emptyset$.
\end{de}

Since every feature constraint  is satisfiable whenever the embedded
feature description is satisfiable, and since  satisfiability of
feature descriptions is decidable, we get immediately the desired decidability
result for the feature-based constraint language \Xcal{FL}. To sum up,
since \Xcal{FL} is closed under renaming and intersection, and due
to the decidability algorithm for \Xcal{FL} constraint solving of
\citeANP{Goetz:98} (to appear), we can state the following proposition.

\begin{po} \Xcal{FL} is a decidable constraint language closed under
  renaming and intersection.
\end{po}

\subsection{Feature-Based Constraint Logic Grammars}

Feature-based grammars can be built in a pure declarative way simply as sets
of axiomatic interpreted feature descriptions from the
feature description language \Xcal{FD}.

\begin{de}[Grammar] 
A feature-based grammar \Xcal{G} is a finite set of feature descriptions
s.t. $\Xcal{G} \subseteq \Xsf{Desc}$.
\end{de}

The feature descriptions comprising a grammar constrain the
admissible models of the grammar in that in every model of a grammar
every feature description must be true of every object.

\begin{de}[Model]
A model of a feature constraint grammar \Xcal{G}
 is an interpretation  $\Xcal{I} =
\left< \Xsf{U, S, F} \right>$ s.t. for every $u \in \Xsf{U}$, for
every $D \in \Xcal{G}$: $u \in \denI{D}$.
\end{de}

The central problem of prediction can then be defined
model-theoretically as a relation between grammars and feature
descriptions encoding the questioned input.

\begin{de}[Prediction]\label{prediction}
A feature description $D$ is predicted by a grammar \Xcal{G} iff there
is a model \Xcal{I} of \Xcal{G} s.t. $\denI{D}\neq\emptyset$.
\end{de}

In contrast to this definition, the linguistic problem of
grammaticality is sometimes considered as a relation between grammars 
and objects. As we will see below, the syntactic coding of Def.
\ref{prediction} enables a
connection of the model-theoretic concept of prediction with the
implementational parsing/generation problem.
The problem of prediction has shown to be undecidable for various
feature-based description languages (see \citeN{Ait-Kaci:93},
\citeN{Smolka:92}, \citeANP{Goetz:98} (to appear)). As shown by
\citeANP{Goetz:98} (to appear), decidable fragments of such languages are
obtainable, e.g., in the form of grammars fulfilling the finite model
property. 

\begin{de}A grammar \Xcal{G} has the finite
  model property iff for all descriptions $D$, 

\Xcal{G} predicts $D$ iff \Xcal{G} has a finite model \I s.t
$\denI{D} \not= \emptyset$.
\end{de}

Note that even if for grammars having the finite model property the
prediction problem is decidable, it is undecidable if a
grammar has the finite model property of not. Thus it has to be kept
in mind that decidability of the prediction problem for linguistically
interesting CLGs is based on an assumption of finiteness of linguistic
structures.

To obtain feature-based CLGs from feature-based grammars, the feature
descriptions from \Xcal{FD} have to be extended to feature constraints
from \Xcal{FL}, which then can be embedded as \Xcal{FL}-constraints into
a suitable definite clause specifcation in \Xcal{R(FL)}. An example
for such an embedding of a feature-based grammar into the CLP scheme of
\citeN{HuS:88} is given below. The resulting feature-based CLG can be
seen as a notational variant of a CUF-grammar
\cite{Doerre:91,Doerre:93}. Alternatively, when replacing the
predicates of this feature-based CLG by a single predicate \texttt{gram}
in all clauses, we arrive at a program which would result from a
direct application of the compilation algorithm of
\citeN{Goetz:95}, \citeN{GoetzMeurers:95} to a feature-based grammar\footnote{Based on a
  differentiation of types in distinct sets according to whether and
  how they appear as antecedents of grammar constraints, this
  compilaton procedure introduces a set of clauses defining the single
  predicate \texttt{gram} for each such set of types. Actually, for the
  example given below, this compilation scheme would also produce a clause
  $\texttt{gram}(X) \leftarrow X=t$ for each minimal type $t$ of the
  grammar signature which is not the antecedent of a grammar
  description. For ease of readability, we will omit clauses
  introduced for (minimal or non-minimal) non-antecedent types in our example.}.
This compilation scheme connects the model-theoretic concept of
prediction with the logic programming 
concept of \Xcal{P}-answer directly. This is done by an
automatic generation of a \Xcal{R(FL)}-program \Xcal{P} for every
\Xcal{FD}-grammar \Xcal{G}, where the program defines an unary
relation \texttt{gram} encoding prediction. This encoding is said to
be correct under the following conditions.

\begin{de} \label{correctTranslation} Let \Xcal{P} be a definite clause specification in
  \Xcal{FL} defining the relation $\mathtt{gram}$, and let \Xcal{G} a
  grammar from \Xcal{FD}. Then \Xcal{P} is a correct translation of
  \Xcal{G} iff 

  \Xcal{G} predicts feature description $D$ iff the goal
  $\mathtt{gram}(X)$ $\&$
  $X = D$ has a \Xcal{P}-answer.
\end{de}

The compilation scheme presented by \citeN{Goetz:95} is sound and for a
large class of grammars complete. A sufficient condition to receive
correct translations in the sense of Def. \ref{correctTranslation} is
again the finite model property. Thus under the assumption that
linguistic structures are finite, CLP can be seen as a useful parsing
scheme for linguistically interesting feature-based CLGs.

Let us illustrate these concepts with an example.
Suppose a simple grammar licensing,
among others, analyses such as 
\[
[Peter\; believes\; [ Clinton_N \; talks_V ]_S \;]_S
\]
or
\[
[Peter\; believes\; [ Clinton_N \; talks_N]_{NP} \;]_S .
\]
We will define now a feature-based grammar presenting a
\Xcal{FD}-encoding of the part of this grammar which is relevant for
the structural ambiguity. It is a modified and extended version of an example
from \citeN{Carpenter:92}.

\begin{figure}[htbp]
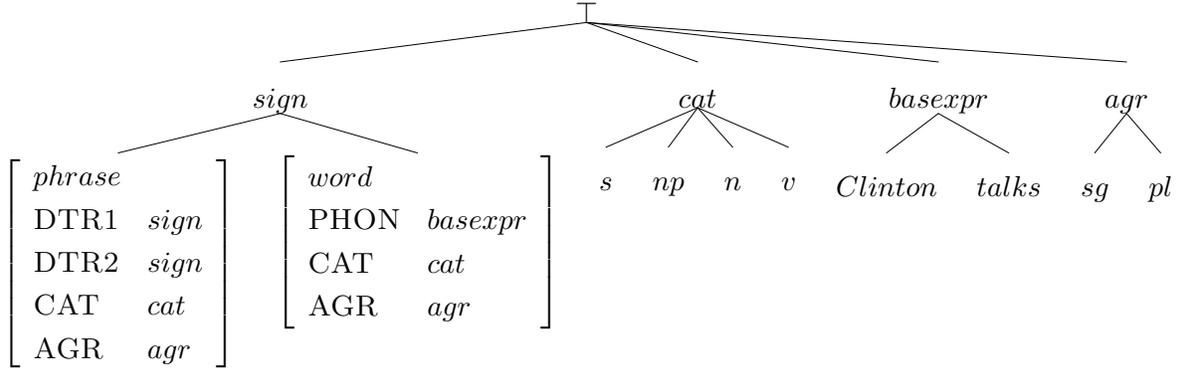

\setlength{\GapWidth}{15pt}
\begin{bundle}{$\top$}
  
  \chunk{
    \begin{bundle}{$sign$}
      \chunk{
        $\left[ \begin{array}{ll}
            phrase &  \\
            \textsc{DTR1} & sign \\
            \textsc{DTR2} & sign \\
            \textsc{CAT} & cat \\
            \textsc{AGR} & agr
          \end{array}
        \right]$}
      
      \chunk{
        $\left[ \begin{array}{ll}
            word  &  \\
            \textsc{PHON} & basexpr\\
            \textsc{CAT} & cat \\
            \textsc{AGR} & agr
          \end{array}
        \right]$}
    \end{bundle}}
  
  \chunk{
    \begin{bundle}{$cat$}
      \chunk{$s$}
      \chunk{$np$}
      \chunk{$n$}
      \chunk{$v$}
    \end{bundle}}
  
  \chunk{
    \begin{bundle}{$basexpr$}
      \chunk{$Clinton$}
      \chunk{$talks$}
    \end{bundle}}
  
  \chunk{
    \begin{bundle}{$agr$}
      \chunk{$sg$}
      \chunk{$pl$}
    \end{bundle}}
  
\end{bundle}
\caption{Signature for feature-based grammar}
\label{signature}
\end{figure}

The signature comes with a type hierarchy with top element
$\top$, feature symbols, and appropriateness conditions, and is depicted in
the graph in Fig. \ref{signature}.
Feature symbols are depicted in \textsc{small caps}
font, type symbols in \textit{lower case italics}, and appropriateness
conditions are expressed in a matrix notation, reading, e.g., $approp (phrase,
\textsc{DTR1}) = sign$. 

The relevant \Xcal{FD}-descriptions are given in
Fig. \ref{grammar}. The first implication encodes the rules $S
\rightarrow N \; V$ and $NP \rightarrow N \; N$. Context-sensitivity
is introduced by the agreement requirement on the first rule. The
second implication encodes the rules $N \rightarrow Clinton$, $V
\rightarrow talks$, and $N \rightarrow talks$.

\begin{figure}[htbp]
\begin{tabular}{ll}
$phrase \rightarrow$ & $(\textsc{CAT}:s \: \wedge\:
\textsc{DTR1:CAT}:n \:\wedge\: 
\textsc{DTR2:CAT}:v \:\wedge\:
\textsc{DTR1:AGR}:Y)$\\
& 
$ \:\wedge\: 
\textsc{DTR2:AGR}:Y)
$\\ 
 &  $\vee\: (\textsc{CAT}:np \:\wedge\: 
\textsc{DTR1:CAT}:n \:\wedge\:\textsc{DTR2:CAT}:n)
$\\ 
$word \rightarrow$ & $(\textsc{CAT}:n \:\wedge\:
\textsc{PHON}:Clinton \:\wedge\: \textsc{AGR}:sg)$\\ 
 & $\vee\: (\textsc{CAT}:v \:\wedge\: \textsc{PHON}:talks \:\wedge\:
 \textsc{AGR}:sg) 
$\\  
 & $\vee\: (\textsc{CAT}:n \:\wedge\: \textsc{PHON}:talks \:\wedge\: \textsc{AGR}:pl)$
\end{tabular} 
\caption{Feature-based grammar}
\label{grammar}
\end{figure}

The CLG obtained from a simplified compilation of the grammar in
Fig. \ref{grammar} to a definite clause specification in \Xcal{FL} is
given in Fig. \ref{CLGprogram}. The embedded \Xcal{FL}-constraints are
depicted graphically in the same way as
\Xcal{FD}-descriptions. \Xcal{R(FL)}-atoms are depicted in
\texttt{typewriter} font.

\begin{figure}[htbp]
\begin{description}
\item[{\tt 1}  $\texttt{phrase}(X) \leftarrow$] $ X = (phrase \:\wedge\:
\textsc{CAT}:s \:\wedge\: \textsc{DTR1:CAT}:n \:\wedge\:
\textsc{DTR2:CAT}:v \:\wedge\: \textsc{DTR1:AGR}:Y \:\wedge\:
\textsc{DTR2:AGR}:Y \:\wedge\: \textsc{DTR1}:Z_1
\:\wedge\: \textsc{DTR2}:Z_2) \:\&\: \texttt{sign}(Z_1) \:\&\: \texttt{sign}(Z_2)$.
\item[{\tt 2} $\texttt{phrase}(X) \leftarrow$] $ X = (phrase \:\wedge\:
\textsc{CAT}:np \:\wedge\:\textsc{DTR1:CAT}:n \:\wedge\: \textsc{DTR2:CAT}:n \:\wedge\: \textsc{DTR1}:Z_1
\:\wedge\: \textsc{DTR2}:Z_2) \:\&\: \texttt{sign}(Z_1) \:\&\: \texttt{sign}(Z_2)$.
\item[{\tt 3}  $ \texttt{word}(X) \leftarrow$] $ X = (word \:\wedge\:
  \textsc{CAT}:n \:\wedge\: \textsc{PHON}:Clinton \:\wedge\:
  \textsc{AGR}:sg)$.
\item[{\tt 4} $\texttt{word}(X) \leftarrow$] $ X = (word \:\wedge\:
  \textsc{CAT}:v \:\wedge\: \textsc{PHON}:talks \:\wedge\:
  \textsc{AGR}:sg)$.
\item[{\tt 5} $\texttt{word}(X) \leftarrow$] $ X = (word \:\wedge\:
  \textsc{CAT}:n \:\wedge\: \textsc{PHON}:talks \:\wedge\:
  \textsc{AGR}:pl)$.
\item[{\tt 6} $\texttt{sign}(X) \leftarrow$] $\texttt{phrase}(X)$.
\item[{\tt 7} $\texttt{sign}(X) \leftarrow$] $\texttt{word}(X)$.
\end{description}
\caption{Feature-based constraint logic grammar}
\label{CLGprogram}
\end{figure}

Given this program and a goal 
\[X = (sign \: \wedge \: \textsc{DTR1: PHON}:Clinton \:\wedge\:
\textsc{DTR2: PHON}:talks) \:\&\: \texttt{sign}(X)
\]
encoding the phrase \emph{Clinton talks}, we can infer
two answers 
\begin{center}
$\begin{array}{c}
X = (phrase \:\wedge\:  \textsc{CAT}:s 
\:\wedge\: \textsc{DTR1}:word 
\:\wedge\: \textsc{DTR1: CAT}:n  \\
\:\wedge\: \textsc{DTR1: PHON}:Clinton 
\:\wedge\: \textsc{DTR1: AGR}:Y
\wedge\: \textsc{DTR1: AGR}:sg \\
 \:\wedge\: \textsc{DTR2}:word 
\:\wedge\: \textsc{DTR2: CAT}:v 
 \:\wedge\: \textsc{DTR2: PHON}:talks \\
\:\wedge\: \textsc{DTR1: AGR}:Y
\:\wedge\: \textsc{DTR1: AGR}:sg) 
\end{array}$
\end{center}
and 

\begin{center}
$\begin{array}{c}
X = (phrase \:\wedge\:  \textsc{CAT}:np 
\:\wedge\: \textsc{DTR1}:word  
\:\wedge\: \textsc{DTR1: CAT}:n  \\
\:\wedge\: \textsc{DTR1: PHON}:Clinton 
\:\wedge\: \textsc{DTR1: AGR}:sg
 \:\wedge\: \textsc{DTR2}:word \\
\:\wedge\: \textsc{DTR2: CAT}:n 
 \:\wedge\: \textsc{DTR2: PHON}:talks
\:\wedge\: \textsc{DTR2: AGR}:pl)
\end{array}$
\end{center}
encoding the parses $[Clinton_N \; talks_V ]_S$ and
$[Clinton_N \; talks_N ]_{NP}$ respectively. The parses are depicted
in Figs. \ref{CLG1st} and \ref{CLG2nd}. Note that goal reduction and
constraint solving are applied in one step. Furthermore, only success
branches are depicted and the the constraint solver is viewed as a black box.

\begin{figure}[htbp]
\begin{center}
\setlength{\GapWidth}{50pt}

\begin{bundle}
  {$\begin{array}{c} 
      X = (sign \: \wedge \: \textsc{DTR1: PHON}:Clinton 
      \:\wedge\: \textsc{DTR2: PHON}:talks) \\
      \:\&\: \texttt{sign}(X)  
    \end{array}$} 
  
  \chunk[$r,c$]
  {\begin{bundle}
      {$\begin{array}{c} 
          {\tt 6,} \; X = (sign \: \wedge \: \textsc{DTR1: PHON}: Clinton 
          \:\wedge\: \textsc{DTR2: PHON}:talks) \\
          \:\&\: \texttt{phrase}(X) 
        \end{array}$} 
      
      \chunk[$r,c$]
      {\begin{bundle}
          {$\begin{array}{c}
              {\tt 1,} \; X = (phrase \:\wedge\:  \textsc{CAT}:s \:\wedge\: 
              \textsc{DTR1}:word  \:\wedge\: \textsc{DTR1: CAT}:n \\ 
              \:\wedge\: \textsc{DTR1: PHON}:Clinton 
              \:\wedge\: \textsc{DTR1: AGR}:Y
              \:\wedge\: \textsc{DTR2}:word\:  
              \wedge\:\textsc{DTR2: CAT}:v \\ 
              \:\wedge\: \textsc{DTR2: PHON}:talks 
              \:\wedge\: \textsc{DTR2: AGR}:Y
              \:\wedge\: \textsc{DTR1}:Z_1 \:\wedge\: \textsc{DTR2}:Z_2) \\
              \:\&\: \texttt{sign}(Z_1)
              \:\&\: \texttt{sign}(Z_2)
            \end{array}$}
          
          \chunk[$r,c$]
          {\begin{bundle}
              {$\begin{array}{c}
                  {\tt 7,} \; X = (phrase \:\wedge\:  \textsc{CAT}:s \:\wedge\: 
                  \textsc{DTR1}:word  \:\wedge\: \textsc{DTR1: CAT}:n \\ 
                  \:\wedge\: \textsc{DTR1: PHON}:Clinton 
                  \:\wedge\: \textsc{DTR1: AGR}:Y
                  \:\wedge\: \textsc{DTR2}:word\:  
                  \wedge\:\textsc{DTR2: CAT}:v \\ 
                  \:\wedge\: \textsc{DTR2: PHON}:talks 
                  \:\wedge\: \textsc{DTR2: AGR}:Y
                  \:\wedge\: \textsc{DTR1}:Z_1 \:\wedge\: \textsc{DTR2}:Z_2) \\
                  \:\&\: \texttt{word}(Z_1) \:\&\: \texttt{sign}(Z_2)
                \end{array}$}

              \chunk[$r,c$]
              {\begin{bundle}
                  {$\begin{array}{c}
                      {\tt 3,} \; X = (phrase \:\wedge\:  \textsc{CAT}:s \:\wedge\: 
                      \textsc{DTR1}:word  \:\wedge\: \textsc{DTR1: CAT}:n \\
                      \:\wedge\: \textsc{DTR1: PHON}:Clinton 
                      \:\wedge\: \textsc{DTR1: AGR}:Y
                      \:\wedge\: \textsc{DTR1: AGR}:sg 
                      \:\wedge\: \textsc{DTR2}:word \\
                      \: \wedge\:\textsc{DTR2: CAT}:v 
                      \:\wedge\: \textsc{DTR2: PHON}:talks 
                      \:\wedge\: \textsc{DTR2: AGR}:Y
                      \:\wedge\: \textsc{DTR2: AGR}:sg \\
                      \:\wedge\: \textsc{DTR2}:Z_2) 
                      \:\&\: \texttt{sign}(Z_2) 
                    \end{array}$
                    }

                  \chunk[$r,c$]
                  {\begin{bundle}
                      {$\begin{array}{c}
                          {\tt 7,} \; X = (phrase \:\wedge\:  \textsc{CAT}:s \:\wedge\: 
                          \textsc{DTR1}:word  \:\wedge\: \textsc{DTR1: CAT}:n \\ 
                          \:\wedge\: \textsc{DTR1: PHON}:Clinton 
                          \:\wedge\: \textsc{DTR1: AGR}:Y
                          \:\wedge\: \textsc{DTR1: AGR}:sg 
                          \:\wedge\: \textsc{DTR2}:word \\
                          \: \wedge\:\textsc{DTR2: CAT}:v 
                          \:\wedge\: \textsc{DTR2: PHON}:talks 
                          \:\wedge\: \textsc{DTR2: AGR}:Y
                          \:\wedge\: \textsc{DTR2: AGR}:sg \\
                          \:\wedge\: \textsc{DTR2}:Z_2) 
                          \:\&\: \texttt{word}(Z_2) 
                        \end{array}$
                        }
                      
                      \chunk[$r,c$]
                      {$\begin{array}{c}
                          {\tt 4,} \; X = (phrase \:\wedge\:  \textsc{CAT}:s \:\wedge\: 
                          \textsc{DTR1}:word  \:\wedge\: \textsc{DTR1: CAT}:n \\ 
                          \:\wedge\: \textsc{DTR1: PHON}:Clinton 
                          \:\wedge\: \textsc{DTR1: AGR}:Y
                          \:\wedge\: \textsc{DTR1: AGR}:sg 
                          \:\wedge\: \textsc{DTR2}:word \\
                          \: \wedge\:\textsc{DTR2: CAT}:v 
                          \:\wedge\: \textsc{DTR2: PHON}:talks 
                          \:\wedge\: \textsc{DTR2: AGR}:Y
                          \:\wedge\: \textsc{DTR2: AGR}:sg )
                        \end{array}$
                        }
                    \end{bundle}
                    }
                \end{bundle}
                }
            \end{bundle}
            }
        \end{bundle}
        }
    \end{bundle}
    }
\end{bundle}

\end{center}
\caption{A derivation of $[Clinton_N \;talks_V]_S$}
\label{CLG1st}
\end{figure}

\begin{figure}[htbp]
  \begin{center}
    \setlength{\GapWidth}{50pt}
    
    \begin{bundle}
      {$\begin{array}{c} 
          X = (sign \: \wedge \: \textsc{DTR1: PHON}:Clinton 
          \:\wedge\: \textsc{DTR2: PHON}:talks) \\
          \:\&\: \texttt{sign}(X)  
        \end{array}$} 
      
      \chunk[$r,c$]
      {\begin{bundle}
          {$\begin{array}{c} 
              {\tt 6,} \; X = (sign \: \wedge \: \textsc{DTR1: 
                PHON}: Clinton 
              \:\wedge\: \textsc{DTR2: PHON}:talks) \\
              \:\&\: \texttt{phrase}(X) 
            \end{array}$} 
          
          \chunk[$r,c$]
          {\begin{bundle}
              {$\begin{array}{c}
                  {\tt 2,} \; X = (phrase \:\wedge\:  \textsc{CAT}:np \:\wedge\: 
                  \textsc{DTR1}:word  \:\wedge\: \textsc{DTR1: CAT}:n \\ 
                  \:\wedge\: \textsc{DTR1: PHON}:Clinton 
                  \:\wedge\: \textsc{DTR2}:word\:  
                  \wedge\:\textsc{DTR2: CAT}:n \\ 
                  \:\wedge\: \textsc{DTR2: PHON}:talks 
                  \:\wedge\: \textsc{DTR1}:Z_1 \:\wedge\: \textsc{DTR2}:Z_2) \\
                  \:\&\: \texttt{sign}(Z_1)
                  \:\&\: \texttt{sign}(Z_2)
                \end{array}$}
              
              \chunk[$r,c$]
              {\begin{bundle}
                  {$\begin{array}{c}
                      {\tt 7,} \; X = (phrase \:\wedge\:  \textsc{CAT}:np \:\wedge\: 
                      \textsc{DTR1}:word  \:\wedge\: \textsc{DTR1: CAT}:n \\ 
                      \:\wedge\: \textsc{DTR1: PHON}:Clinton 
                      \:\wedge\: \textsc{DTR2}:word\:  
                      \wedge\:\textsc{DTR2: CAT}:n \\ 
                      \:\wedge\: \textsc{DTR2: PHON}:talks 
                      \:\wedge\: \textsc{DTR1}:Z_1 \:\wedge\: \textsc{DTR2}:Z_2) \\
                      \:\&\: \texttt{word}(Z_1) \:\&\: \texttt{sign}(Z_2)
                    \end{array}$}

                  \chunk[$r,c$]
                  {\begin{bundle}
                      {$\begin{array}{c}
                          {\tt 3,} \; X = (phrase \:\wedge\:  \textsc{CAT}:np \:\wedge\: 
                          \textsc{DTR1}:word  \:\wedge\: \textsc{DTR1: CAT}:n \\
                          \:\wedge\: \textsc{DTR1: PHON}:Clinton 
                          \:\wedge\: \textsc{DTR1: AGR}:sg 
                          \:\wedge\: \textsc{DTR2}:word \\
                          \: \wedge\:\textsc{DTR2: CAT}:n 
                          \:\wedge\: \textsc{DTR2: PHON}:talks 
                          \:\wedge\: \textsc{DTR2}:Z_2) \\
                          \:\&\: \texttt{sign}(Z_2) 
                        \end{array}$
                        }

                      \chunk[$r,c$]
                      {\begin{bundle}
                          {$\begin{array}{c}
                              {\tt 7,} \; X = (phrase \:\wedge\:  \textsc{CAT}:np \:\wedge\: 
                              \textsc{DTR1}:word  \:\wedge\: \textsc{DTR1: CAT}:n \\ 
                              \:\wedge\: \textsc{DTR1: PHON}:Clinton 
                              \:\wedge\: \textsc{DTR1: AGR}:sg 
                              \:\wedge\: \textsc{DTR2}:word \\
                              \: \wedge\:\textsc{DTR2: CAT}:n 
                              \:\wedge\: \textsc{DTR2: PHON}:talks 
                              \:\wedge\: \textsc{DTR2}:Z_2) \\
                              \:\&\: \texttt{word}(Z_2) 
                            \end{array}$
                            }
                          
                          \chunk[$r,c$]
                          {$\begin{array}{c}
                              {\tt 5,} \; X = (phrase \:\wedge\:  \textsc{CAT}:np \:\wedge\: 
                              \textsc{DTR1}:word  \:\wedge\: \textsc{DTR1: CAT}:n \\ 
                              \:\wedge\: \textsc{DTR1: PHON}:Clinton 
                              \:\wedge\: \textsc{DTR1: AGR}:sg 
                              \:\wedge\: \textsc{DTR2}:word \\
                              \: \wedge\:\textsc{DTR2: CAT}:n
                              \:\wedge\: \textsc{DTR2: PHON}:talks 
                              \:\wedge\: \textsc{DTR2: AGR}:pl)
                            \end{array}$
                            }
                        \end{bundle}
                        }
                    \end{bundle}
                    }
                \end{bundle}
                }
            \end{bundle}
            }
        \end{bundle}
        }
    \end{bundle}

  \end{center}
  \caption{A derivation of $[Clinton_N \; talks_N]_{NP}$}
  \label{CLG2nd}
\end{figure}

\section{Summary}
\markright{2.4 Summary}

In this chapter we discussed the basic formal concepts of the CLP
scheme of \citeN{HuS:88}. These concepts provide a formal
specification of the notions of constraint language and of a
constraint logic program embedding a constraint language. For
convenience, we gave some missing proofs and introduced the notions of
logical consequence, derivation tree and proof tree into the CLP
scheme. These concepts will be useful in latter chapters.

Furthermore, we reported the central formal details of feature-based
CLGs and presented a simple linguistic grammar which will be used as a
running example in the following chapters.

Proof trees or parses in constraint-based NLP can be quite complex
even if simple grammars are used to analyze two-word phrases as in the
example given above. Clearly, for complex grammars and phrases of
reasonable length, structural ambiguity in constraint-based NLP is a
severe problem. The task of the next two chapters is to provide a
rigorous mathematical foundation of ambiguity resolution in
constraint-based NLP.

\chapter{Quantitative CLP: Quantitative
Inference with Subjective Weights and its Formal Semantics}
\label{Q}

In this chapter we present a novel framework for
quantitative inference with subjective weights for CLP. 
We show soundness and completeness of the
quantitative system with respect to a simple and intuitive formal
semantics. We illustrate these concepts with a simple quantitative CLG
and show how pruning techniques can be used to guide the search for
the highest weighted analysis in such quantitative systems.

This chapter is based upon work previously published in
\citeN{Riezler:96}. 

\section{Introduction and Overview}
\markboth{Chapter 3. Quantitative CLP}{3.1 Introduction and Overview}

Quantitative frameworks have been presented as extensions of both
logic programming and constraint-based grammars. For the area of logic
programmming, a system of quantitative deduction which is sound and
complete with respect to a related fixpoint semantics
was introduced firstly by \citeN{Emden:86}. Like this seminal
approach, most of the subsequent work on quantitative extensions of
logic programming has concentrated on theoretical issues such as 
questions of the expressivity of systems for quantitative logic
programming, or issues of the correctness of the connection of model-theory,
fixpoint-theory, and proof-theory for such systems. However, none on
these approaches seemed to have a specific application in mind.

On the contrary, quantitative extensions of constraint-based grammars
have mainly been motivated by practical considerations. Most
approaches in this area come as numerical extensions of the parsing
strategy of existing constraint-based frameworks.
However, even if for such systems the formal foundation of the
underlying framework may be clear enough, none of these approaches
comes with a well-defined semantics for its quantitative extension.
That is, such quantitative extensions have to be seen as extralogical
extensions of, e.g., the deduction scheme of the underlying CLP
framework, and are not related to the model-theoretic counterpart of
this operational semantics. 

This is clearly an undesirable state of affairs. Rather, in the same
way as CLGs provide a model-theoretic characterization of linguistic
objects coupled with an operational parsing system, one would like to relate a
quantitative deduction system to a quantitative model-theory in a
sound and complete way. The aim of this chapter is to present a sound and
complete system of quantitative CLP which satisfies the following
conditions. It should
\begin{itemize}
\item generally be applicable to CLP over arbitrary constraint
languages,
\item provide a precise, but yet simple formal semantics for
quantitative CLP deduction,
\item from the outset be designed with a specific application in mind,
in our case, with respect to efficient ambiguity resolution in CLGs.
\end{itemize}

The first point means that in quantitative CLP one should not have to
bother about the peculiarities of the constraint languages embedded
into the CLP scheme. Rather, the quantitative
extension should work in the same way for every constraint logic
program irrespective of the embedded constraint language.
For the NLP application, this means that for arbitrary
constraint-based grammars a quantitative extension should be
obtainable from the CLG resulting from an embedding of the grammar
constraint language into a CLP scheme.

The second point addresses the tradeoff between the expressive power of the
quantitative system and the intuitivity and simplicity of its
semantics. That is, since the aim of a formal semantics is to provide
a precise unambiguous way to specify the meaning of all aspects of an operational
system at the design and implementation stage, it is justified only by
its understandability and applicability. Our approach respects these ideas of
simplicity and elegance by using the simple concepts of fuzzy set
algebra as a basis for a formal semantics for quantitative CLP.

The third point, which refers to the
intended application of ambiguity resolution and best-parse search in
CLGs, is realized in quantitative CLP by stating the proof theory of
quantitative CLP in terms of min/max trees, which in turn enables
strategies such as alpha/beta-pruning to be used for efficient
searching for best parses in CLGs.  

Clearly, generalizations of this specific choice of design for quantitative
CLP should be straightforward. However, they will not made explicit in the
following chapters.

This chapter is organized as follows. Sect. \ref{PreviousQCLP}
discusses previous work on quantitative logic programming and
quantitative extensions of constraint-based grammars.

Sect. \ref{Syntax} introduces
the concept of a quantitative definite clause specification, i.e., a
quantitative constraint logic program. 

Sect. \ref{DeclSem}
introduces the declarative semantics of quantitative definite clause
specifications, i.e., a model-theoretic semantics based on concepts of
fuzzy set algebra and a fixpoint semantics obtained by a minimal
models in this model-theory.  

Sect. \ref{OpSem} presents the operational semantics of
quantitative CLP. That is, based on the concepts of quantitative
derivation trees and quantitative proof trees, soundness and
completeness of quantitative deduction in CLP is proven.

Sect. \ref{QCLG} exemplifies these concepts with a quantitative
feature-based CLG, and shows how the search technique of
alpha/beta-pruning can be applied to quantitative CLGs.

\section{Previous Work}\label{PreviousQCLP}
\markright{3.2 Previous Work}

For the area of logic programming, \citeN{Emden:86} presented in a
seminal paper a quantitative deduction scheme and a fixpoint semantics for
sets of numerically annotated Horn clauses. The aim of this paper was
to enable the expression of a continuum of uncertainties between the
usual two truth values in quantitative logic programs.
The semantics of such quantitative logic programs is based upon
concepts of fuzzy set algebra, and crucially deals with the
truth-functional propagation of weights across conventional definite clauses. 
Van Emden's approach initialized research into a now extensively
studied area of quantitative logic programming. For example, annotated
logic programming (\citeN{Sub:87}, \citeN{Kifer:92}) extends the
expressive power of quantitative rule sets by allowing variables and
evaluable function terms as annotations. Furthermore, in annotated
logic programs, annotations can be attached to atoms and their
conjunctions or disjunctions, and such programs are interpreted in
powerful frameworks of lattice-theoretic semantics. 
Depending on different understandings of annotations, 
further extensions of \citeN{Emden:86}'s and \citeN{Sub:87}'s
approaches have been presented. Among those are approaches to
possibilistic logic programming based on subjective necessity values
\cite{Dubois:91}, probabilistic logic programming based on intervals
of subjective probabilistic truth values (see, e.g., \citeN{Ng:92},
\citeN{Ng:93}), or probabilistic deductive databases based on
subjective confidence levels coming as intervals of belief and doubt
(see, e.g., \citeN{Laks:94}, \citeN{Laks:97}).

Quantitative extensions of constraint-based grammars
have mainly been motivated by practical considerations.
For example, \citeN{Dale:92} presented an approach to robust parsing
in PATR systems where according to a subjective value of
necessity/optionality of constraints, constraint violations are
allowed, and so robustness is introduced into the formalism.
\citeN{Kim:94} presented an approach to best-first chart parsing with
PATR grammars. In this approach, atomic values of feature structures are
annotated with subjective weights, and a weight combination scheme is
defined for feature structure unification. The search space in
best-first parsing then is restricted by a treshold below
which completed and predicted feature structures are discarded.
\citeN{Erbach:93a}, \citeN{Erbach:93b}, or \citeN{Erbach:95}
introduced a model of preference for the CUF system, which is
generalizable to the CLP scheme of \citeN{HuS:88}, and which is used, among
others, for tasks such as best-first parsing for ambiguity resolution
and self-monitored generation. In Erbach's model, definite clauses as
a whole are annotated with subjective preference values. Such preference values are
combined in the resolution process by calculating the preference value of a
clause consequent as the product of the preference value of the clause and the
preference values of the antecedent predicates, which are additionally
weighted to add up to $1$.

The aim of our approach is to combine the mathematical exactness of
the logic-programming approaches with the practical applicability of
the quantitative-grammar approaches. We will build our framework of
quantitative CLP on ideas developed in the simple and elegant framework of \citeN{Emden:86}. This means that we restrict our
attention to numerical weights attached to CLP clauses as a whole, and
use the simple concepts of fuzzy set algebra to provide the basis for
an intuitive formal semantics for quantitative CLP. Furthermore, we employ
a min/max scheme for rule application which enables strategies such
as alpha/beta pruning to be used for efficient searching. Clearly, our
approach improves upon van Emden's approach by not being restricted to
Horn clauses or to finite derivations. Moreover, it enables the
application of quantitative search strategies to constraint-based
grammars in a formally well-defined way.

\section{Syntax of Quantitative CLP}
\markright{3.3 Syntax of Quantitative CLP}
\label{Syntax}

Building upon the CLP scheme of \citeN{HuS:88} reported in
Chap. \ref{Foundations}, we can define the syntax of a quantitative
definite clause specification \Pf very quickly. The following definitions are
made with respect to implicit constraint languages \La and \Rl. A
definite clause specification \Xcal{P} in \Rl then can be extended to
a quantitative definite clause specification \Pf in \Rl simply by
adding numerical factors to program clauses.

\begin{de}[\Pf] \label{Pf} A quantitative definite clause specification
\Pf  in \Rl is a finite set of quantitative
  formulae, called quantitative definite clauses, of the form
\[ 
\phi \:\&\: B_1 \:\&\: \ldots \:\&\: B_n \:\mbox{}_f\!\rightarrow
A, 
\]
where $A$, $B_1, \ldots, B_n$ are \Rl-atoms,
$\phi$ is an \La-constraint, $n \geq 0$, $f \in (0,1]$.
We may write a quantitative formula also as
$A \leftarrow_f \phi$ $\&$ $B_1$ $\&$ $\ldots$ $\&$
$B_n$.
\end{de}
These factors (the $f$ in Definition \ref{Pf}) should be thought of as
abstract weights which receive a concrete interpretation in specific
instantiations of \Pf. 

In the following the notation \Rl will be used more generally to
notate relationally extended constraint languages which possibly
include quantitative formulae of the above form.

\section{Declarative Semantics of Quantitative CLP}
\markright{3.4 Declarative Semantics of Quantitative CLP}
\label{DeclSem}

\subsection{Fuzzy Set Algebra and Model-Theoretic Semantics}

To obtain a formal semantics for \Pf, first we have to introduce an
appropriate quantitative measure into the set-theoretic specification
of \Rl-interpretations. One possibility to obtain quantitative
\Rl-interpretations is to base the set algebra of \Rl-interpretations
on the simple and well-defined concepts of fuzzy set algebra (see
\citeN{Zadeh:65}).

Relying on H\"ohfeld and Smolka's specification of  base equivalent \Rl-interpretations,
i.e., \Rl-interpretations extending the same \La-interpretation, in terms
of the denotations of the relation symbols in these interpretations,
we can ``fuzzify'' such interpretations by regarding the denotations
of their relation symbols as fuzzy subsets of the set of tuples in the
common domain.

Given constraint languages \La and \Rl,  we interpret each n-ary relation symbol
$r \in \R$ as a fuzzy subset of ${\cal D}^n$,
for each \Rl-interpretation \A with domain \Xcal{D}.
That is, we identify the denotation of $r$ under \A with a total function
\[
\mu(\_ \:  ; r^{\cal A}): {\cal D}^n \rightarrow [0,1] , 
\]
which can be
thought of as an abstract membership function. Such membership
functions are generalized characteristic functions, and classical set
membership is coded in this context by characteristic functions taking
only 0 and 1 as values.

Next, we have to give a model-theoretic characterization of
quantitative definite clauses. Clearly, any monotonous mapping could
be used for the model-theoretic specification of the interaction of
weights in quantitative definite clauses and accordingly for the
calculation of weights in the proof-theory of quantitative CLP. For
concreteness, we will instantiate such a mapping to the specific case
of Definition \ref{mod} resembling \citeN{Emden:86}'s mode of rule
application. 
This will allow us to state the proof-theory of
quantitative CLP in terms of min/max trees which in turn enables
strategies such as alpha/beta pruning to be used for efficient
searching. Such a quantitative CLP scheme improves upon
several shortcomings of \citeN{Emden:86}'s system, e.g. our
quantitative CLP scheme clearly is not restricted to ground instances
of Horn theories, and the soundness and completeness results we will
present are not restricted to finite derivations.
However, the choice of the mode of rule application made is not crucial for the substantial
claims of this paper, and generalizations of this particular
combination mode to specific applications should be straightforward,
but are beyond the scope of this thesis.

The following definition of model corresponds to the definition of
model in classical logic when considering only clauses with
$f = 1$ and mappings ${\cal D}^n \rightarrow \{ 0,1 \}$.

\begin{de}[Model] \label{mod}
An \Rl-interpretation \A extending some
  \La-interpretation \I  is a model of
  a quantitative definite clause specification \Pf 
iff for each $\alpha \in \ASS$, for each quantitative formula
$r(\vec{x}) \leftarrow_f \phi$ $\&$ $q_1({\vec{x}}_1)$
$\&$ $\ldots$ $\&$ $q_k({\vec{x}}_k)$ in \Pf holds: 
\begin{center}
If $\alpha \in \denI{\phi}$, then
$\mu(\alpha(\vec{x}); r^{\cal A}) \geq f \times \min \{
\mu (\alpha({\vec{x}}_j); q_j^{\cal A}) | $ $1 \leq j \leq k \}$.
\end{center}
\end{de}
In terms of membership degrees, this definition of model can be
paraphrased as follows: If the antecedent constraint is satisfiable,
then the membership degrees of the denotations of the consequent atom
must not be less than $f$ times the membership degrees of the denotations of the
antecedent atom. A truth-functional view could be obtained by
considering membership degrees as truth degrees of atoms under
variable assignments. From the viewpoint of such a truth-functional
propagation of weights across definite clauses, a clause contributes
to the consequent a truth value which is $f$ times the truth value of
the antecedent.

Note that the notation of an \Rl-interpretation \A will be used
more generally to include interpretations of quantitative
formulae. \Rl-solutions of a quantitative formula are defined as
$\denA{r(\vec{x}) \leftarrow_f \phi \;\&\; q_1({\vec{x}}_1)
\;\&\; \ldots \;\&\; q_k({\vec{x}}_k)} = \{ \alpha \in \ASS |$
If  $\alpha \in \denI{\phi}$, then
$\mu(\alpha(\vec{x}); r^{\cal A}) \geq f \times \min \{
\mu (\alpha({\vec{x}}_j); q_j^{\cal A}) | \; 1 \leq j \leq k \} \}$.

Based on the above definition of model, the concept of logical
consequence can be defined as usual. 

\begin{de}[Logical consequence] \label{logcons}
A quantitative formula
$r(\vec{x}) \leftarrow_f \phi$
is a logical consequence of a
quantitative definite clause specification \Pf iff for each
\Rl-interpretation \A, \A is a model of
\Pf implies that \A is a model of
$\{ r(\vec{x}) \leftarrow_f \phi \}$.
\end{de}
Furthermore, we have that the fact that $r(\vec{x}) \leftarrow_f \phi$
is a logical consequence of \Pf implies that
$r(\vec{x}) \leftarrow_{f'} \phi$ is a logical consequence of
\Pf for every $f' \leq f$.

A {\bf goal} $G$ is defined similar to the non-quantitative case as a
(possibly empty) conjunction of \Rl-atoms and \La-constraints.
We can, without loss of generality, restrict
goals to be of the form $r(\vec{x})$ $\&$ $\phi$, i.e., a (possibly
empty) conjunction of a single relational atom $r(\vec{x})$ and an
\La-constraint $\phi$. This can be done since
for each goal $G = r_1({\vec{x}}_1)$ $\&$ $\ldots$ $\&$
$r_k({\vec{x}_k})$ $\&$ $\phi$ which 
contains more than one relational atom, we
can complete the program with a new clause $C = r({\vec{x}_1}, \ldots,
{\vec{x}_k}) \leftarrow_1 r_1({\vec{x}}_1)$ $\&$ $\ldots$ $\&$
$r_k({\vec{x}_k})$ $\&$ $\phi$, with $G$ as antecedent and
a new predicate, which takes all variables in $G$ as arguments, as consequent.
Submitting the new predicate $r({\vec{x}_1}, \ldots,
{\vec{x}_k})$ as query yields the same results as would be
obtained when querying with the compound goal $G$.

Given some program \Pf and some goal $G$,
a  quantitative {\bf \Pf-answer} $\varphi$ of $G$ is defined as a satisfiable
\La-constraint $\varphi$ s.t.\
$\varphi$ $\mbox{}_f\!\rightarrow G$ is a logical consequence of
\Pf. A quantitative formula
$\varphi$ $\mbox{}_f\!\rightarrow r(\vec{x})$ $\&$ $\phi$ is defined to be
a logical consequence of \Pf iff every model of \Pf is a
model of $\{ \varphi$ $\mbox{}_f\!\rightarrow r(\vec{x})$ $\&$
$\phi \}$.
An \Rl-interpretation \A is a model of 
$\{ \varphi$ $\mbox{}_f\!\rightarrow r(\vec{x})$ $\&$ $\phi \}$ iff 
$\denA{\varphi} \subseteq \denA{\phi}$ and \A is a
model of $\{ r(\vec{x}) \leftarrow_f \varphi \}$.

Next we have to associate a complete lattice of interpretations with
quantitative definite clause specifications.

Adopting Zadeh's definitions for set operations, we can define a
partial ordering on the set of base equivalent
\Rl-interpretations.
This is done by defining set operations on these interpretations with
reference to set operations on the denotations of
relation symbols in these interpretations. 
We get for all base equivalent \Rl-interpretations ${\cal A}, \; {\cal
  A'}$:
\begin{itemize}
\item $\A \subseteq \A'$ iff 
for each n-ary relation symbol $r \in \R$,
for each $\alpha \in \ASS$,
for each $\vec{x} \in {\sf VAR}^n$:
$\mu(\alpha(\vec{x}); r^{\cal A}) \leq
\mu(\alpha(\vec{x}); r^{\cal A'})$,
\item $\A = \bigcup X$ iff 
for each n-ary relation symbol $r \in \R$,
for each $\alpha \in \ASS$,
for each $\vec{x} \in {\sf VAR}^n$:
$\mu(\alpha(\vec{x}); r^{\cal A}) = 
\sup \{ \mu(\alpha(\vec{x}); r^{\cal A'}) | \; \A' \in X \}$,
\item $\A = \bigcap X$ iff 
for each n-ary relation symbol $r \in \R$,
for each $\alpha \in \ASS$,
for each $\vec{x} \in {\sf VAR}^n$:
$\mu(\alpha(\vec{x}); r^{\cal A}) = 
\inf \{ \mu(\alpha(\vec{x}); r^{\cal A'}) | \; \A' \in X \}$.
\end{itemize}
Note that we define furthermore 
$\sup \; \emptyset = 0$, $\inf \; \emptyset = 1$.
Clearly, the set of all base equivalent \Rl-interpretations is a
complete lattice under the partial ordering of set inclusion. The
supremum is given by the union, and the infimum by the intersection,
for any set of base-equivalent \Rl-interpretations. The top element is
the \Rl-interpretation $\mathcal{A}^\top$ such that for each $r \in
\R$, for each $\vec{u} \in \mathcal{D}^{\mathsf{Ar}(r)}$:
$\mu(\vec{u}; r^{\mathcal{A}^\top})=1$, and the bottom element is the
\Rl-interpretation $\mathcal{A}^\bot$ such that for each $r \in
\R$, for each $\vec{u} \in \mathcal{D}^{\mathsf{Ar}(r)}$:
$\mu(\vec{u}; r^{\mathcal{A}^\bot})=0$.

\subsection{Minimal Model Semantics}

Based upon the definition of a complete lattice of \Rl-interpretations of a
quantitative definite clause specification \Pf, we can state the
following equations, which link the declarative and operational
semantics of \Pf. These equations define the notion of a \Pf-chain,
which will be crucial for the construction of minimal models for
\Pf. Similar to the non-quantitative case, these equations are based on the
respective definition of model, and take for the quantitative case the
following form.

\begin{de} \label{equations}
Let \Pf be a quantitative definite clause specification in \Rl, \I
be an \La-interpretation. Then the countably infinite sequence $\left<
  {\cal A}_0, {\cal A}_1, {\cal A}_2, \ldots \right>$ of
\Rl-interpretations extending \I is a \Pf-chain iff for each n-ary
relation symbol $r \in \R$, for each $\alpha \in \ASS$, for each
$\vec{x} \in {\sf VAR}^n$:
\begin{description}
\item[]$\mu (\alpha(\vec{x}); r^{{\cal A}_0}) := 0$,
\item[]$\mu (\alpha(\vec{x}); r^{{\cal A}_{i+1}}) :=
\max \{ f \times \min \{
\mu (\alpha({\vec{x}}_j); q_j^{{\cal A}_i}) | $
$1 \leq j \leq n \}$ 
 $|$  there is a variant 
$r(\vec{x}) \leftarrow_f \phi$ $\&$ $q_1({\vec{x}}_1)$
$\&$ $\ldots$ $\&$ $q_n({\vec{x}}_n)$ of a clause in \Pf 
and $\alpha \in \denAi{\phi}{i} \}$.
\end{description}
\end{de}

Before turning to the construction of minimal models, we have to prove
the following useful lemma (see \citeN{Emden:86}, Lemmata 2.10',
2.11'). Lemma \ref{supremum} assures that for each tuple of objects in
the denotation of a relation symbol under a minimal model, there is a
corresponding finite step in the \Pf-chain which introduces these
objects into the minimal model denotation.

\newtheorem{lm}[po]{Lemma}
\begin{lm} \label{supremum}
For each \Pf, for each \Pf-chain
$\left<{\cal A}_0, {\cal A}_1, {\cal A}_2, \ldots \right>$,
for each k-ary relation symbol $r \in \R$, for each  $\alpha \in
\ASS$, for each $\vec{x} \in {\sf VAR}^k$,
there exists some $n \in \Nats$ s.t.\ $\mu(\alpha(\vec{x});
 r^{\bigcup_{i \geq 0} {\cal A}_i}) = \mu(\alpha(\vec{x}); r^{{\cal A}_n})$.
\end{lm}

\begin{proof}
We have to show that the supremum
$v = \sup \{ \mu(\alpha(\vec{x}); r^{{\cal A}_i}) | \; i \geq 0 \} $ can be attained
for some $n \in \Nats$.

\begin{description}
\item[$v=0$:] For $v = 0$, we have $n = 0$.

\item[$v > 0$:] For $v > 0$, we have to show that for any real $\epsilon$,
  $0 < \epsilon <v$, the set $\{ \mu(\alpha(\vec{x}); r^{{\cal A}_i})|\; i \geq
  0 \; and \; \mu(\alpha(\vec{x}); r^{{\cal A}_i}) \geq \epsilon \}$ is
  finite.

Let $F$ be the finite set of real numbers of factors of clauses
  in \Pf, $m$ be the greatest element in $F$ s.t.\ $m < 1$ and let $q$ be
  the smallest integer s.t.\ $m^q < \epsilon$.\\
Then, since each real number
$\mu(\alpha(\vec{x}); r^{{\cal A}_i})$ is
a product of a sequence of elements of $F$, the number of different
products $\geq \epsilon$ is not greater than $|F|^q$, the permutation
of $|F|$ different things taken $q$ at a time with repetitions, and thus finite.\\
Hence, the supremum is the maximum attained for some $n \in \Nats$.
\qed
\end{description}
\renewcommand{\qed}{}
\end{proof}

Now we can obtain minimal model properties for quantitative definite clause
specifications similar to those for the non-quantitative programs of
\citeN{HuS:88}. Based on the constructive definition of a \Pf-chain of
\Rl-interpretations extending an \La-interpretation \I, an
\Rl-interpretation \A is obtainable as the \Rl-interpretation which is
both a model of \Pf and minimal with respect to the lattice of base
equivalent \Rl-interpretations extending \I.
Theorem \ref{definite} states that we can construct a
minimal model \A of \Pf for each quantitative definite clause
specification \Pf in the extension of an arbitrary constraint language
\La and for each \La-interpretation. This means that---due to
the definiteness of \Pf---we can restrict our attention to a minimal model semantics of \Pf.

\newtheorem{te}[po]{Theorem}
\begin{te}[Definiteness] \label{definite}
For each \La-interpretation \I, for each quantitative definite clause
specification \Pf in \Rl, 
for each \Pf-chain 
$\left<{\cal A}_0, {\cal A}_1, {\cal A}_2, \ldots \right>$ of
\Rl-interpretations extending some \La-interpretation \I:

\begin{description}
\item[(i)] ${\cal A}_0 \subseteq {\cal A}_1 \subseteq \ldots$,
\item[(ii)] the union $\A := \bigcup_{i \geq 0} {\cal A}_i$ is a model
  of \Pf extending \I,
\item[(iii)] \A is the minimal model of \Pf extending \I.
\end{description}
\end{te}

\begin{proof}
{\bf (i)} We have to show that 
${\cal A}_i \subseteq {\cal A}_{i+1}$.
We prove by induction on $i$ showing for each constraint language \La,
for each quantitative definite clause specification \Pf in \Rl, for
each \La-interpretation \I, for each \Pf-chain 
$\left<{\cal A}_0, {\cal A}_1, {\cal A}_2, \ldots \right>$ of
\Rl-interpretations extending some \La-interpretation \I,
for each n-ary relation symbol $r \in \R$, for each  $\alpha \in
\ASS$, for each $\vec{x} \in{\sf VAR}^n$, for each $i \in \Nats$:
$\mu(\alpha(\vec{x}); r^{{\cal A}_i}) \leq \mu(\alpha(\vec{x});
r^{{\cal A}_{i+1}})$.

\begin{description}
\item[\rm Base:] $\mu(\alpha(\vec{x}); r^{{\cal A}_0}) = 0 \leq
  \mu(\alpha(\vec{x}); r^{{\cal A}_1})$.

\item[\rm Hypothesis:] Suppose $\mu(\alpha(\vec{x});r^{{\cal A}_{n-1}}) \leq
  \mu(\alpha(\vec{x});r^{{\cal A}_{n}})$.

\item[\rm Step:] $\mu(\alpha(\vec{x});r^{{\cal A}_{n}}) = v > 0$

\begin{description}
\then there exists a variant $r(\vec{x}) \leftarrow_f \phi$ $\&$
$q_1({\vec{x}}_1)$ $\&$ $\ldots$ $\&$ $q_k({\vec{x}}_k)$ of a clause
in \Pf s.t.\  $v = f \times \min
\{ \mu(\alpha({\vec{x}_1});{q_1}^{{\cal A}_{n-1}}), \ldots,
\mu(\alpha({\vec{x}_k});{q_k}^{{\cal A}_{n-1}}) \}$ and $\alpha\in
\denAi{\phi}{n-1}$, by Definition \ref{equations}

\then $\mu(\alpha({\vec{x}_1});{q_1}^{{\cal A}_{n}}) \geq
\mu(\alpha({\vec{x}_1});{q_1}^{{\cal A}_{n-1}}), 
\ldots,
\mu(\alpha({\vec{x}_k});{q_k}^{{\cal A}_{n}}) \geq
\mu(\alpha({\vec{x}_k});{q_k}^{{\cal A}_{n-1}})$ and
$\alpha\in \denAi{\phi}{n}$, by the hypothesis

\then $\mu(\alpha({\vec{x}});r^{{\cal A}_{n+1}}) \geq v$, by
definition of $\mu(\alpha({\vec{x}});r^{{\cal A}_{i+1}})$
\end{description}

\then $\mu(\alpha({\vec{x}});r^{{\cal A}_{n}}) \leq 
\mu(\alpha({\vec{x}});r^{{\cal A}_{n+1}})$.

\item[\rm For] $v=0$ it follows immediately that
$\mu(\alpha({\vec{x}});r^{{\cal A}_{n}}) \leq 
\mu(\alpha({\vec{x}});r^{{\cal A}_{n+1}})$.

\item[] Claim (i) follows by arithmetic induction.
\end{description}
{\bf (ii)} We have to show that $\A := \bigcup_{i \geq 0} {\cal A}_i$ is a
model of \Pf extending \I. We prove that for each clause
$r(\vec{x}) \leftarrow_f \phi$ $\&$
$q_1({\vec{x}}_1)$ $\&$ $\ldots$ $\&$ $q_k({\vec{x}}_k)$ in \Pf,
for each $\alpha \in \ASS$:
If $\alpha \in \denA{\phi}$, then
$\mu(\alpha(\vec{x}); r^{\cal A}) \geq f \times \min \{
\mu(\alpha({\vec{x}_j}); {q_j}^{\cal A}) | \; 1 \leq j \leq k \}$.

\begin{description}
\item[] Note that since every ${\cal A}_i$ is an \Rl-interpretation
  extending \I, \A is an \Rl-interpretation extending \I.

\item[] Now let $r(\vec{x}) \leftarrow_f \phi$ $\&$
$q_1({\vec{x}}_1)$ $\&$ $\ldots$ $\&$ $q_k({\vec{x}}_k)$ be a clause
in \Pf s.t.\ for some $\alpha \in \ASS$:
$\alpha \in \denA{\phi}$ and $\mu(\alpha({\vec{x}_i});
{q_i}^{\cal A}) = \min \{ \mu(\alpha({\vec{x}_j}); {q_j}^{\cal A}) | \;
1 \leq j \leq k \} = v$.

\item[]Then there exists some $n \in \Nats$ s.t.\ $v =
\mu(\alpha({\vec{x}_i}); {q_i}^{{\cal A}_n}) =
\min \{ \mu(\alpha({\vec{x}_j}); {q_j}^{{\cal A}_n}) | $
$ 1 \leq j \leq k \} $,
by Lemma \ref{supremum} and
since for all $j$ s.t. $1 \leq j \leq k: 
\mu(\alpha({\vec{x}_j}); {q_j}^{{\cal A}}) =
\sup \{ \mu(\alpha({\vec{x}_j}); {q_j}^{{\cal A}_i}) | \; i \geq 0 \}$

\begin{description}
\then $\mu(\alpha({\vec{x}}); {r}^{{\cal A}_{n+1}}) \geq f \times v$,
by Definition \ref{equations}

\then $\mu(\alpha({\vec{x}}); {r}^{{\cal A}}) \geq 
\mu(\alpha({\vec{x}}); {r}^{{\cal A}_{n+1}})$, since 
$\mu(\alpha({\vec{x}}); {r}^{{\cal A}}) = \sup \{ \mu(\alpha({\vec{x}}); {r}^{{\cal A}_i}) | \; i \geq 0 \}$
\end{description}

\item{} $\Longrightarrow \mu(\alpha({\vec{x}}); {r}^{{\cal A}}) \geq f \times \min
\{ \mu(\alpha({\vec{x}_j}); {q_j}^{\cal A}) | \;1 \leq j \leq k \}$.

\item[] This completes the proof for claim (ii).
\end{description}
{\bf (iii)} We have to show that \A is the minimal model of \Pf extending
\I. We prove for every base
equivalent model ${\cal B}$ of \Pf: ${\cal A}_i \subseteq {\cal B}$,
which gives $\A \subseteq {\cal B}$, by induction on $i$
showing for each constraint language \La,
for each quantitative definite clause specification \Pf in \Rl, for
each \La-interpretation \I, for each \Pf-chain 
$\left<{\cal A}_0, {\cal A}_1, {\cal A}_2, \ldots \right>$ of
\Rl-interpretations extending some \La-interpretation \I,
for each n-ary relation symbol $r \in \R$, for each  $\alpha \in
\ASS$, for each $\vec{x} \in{\sf VAR}^n$, for each $i \in \Nats$:
$\mu(\alpha(\vec{x}); r^{{\cal A}_i}) \leq \mu(\alpha(\vec{x});r^{\cal
  B})$.

\begin{description}
\item[\rm Base:] $\mu(\alpha(\vec{x}); r^{{\cal A}_0}) = 0 \leq
\mu(\alpha(\vec{x}); r^{{\cal B}})$.

\item[\rm Hypothesis:] Suppose $\mu(\alpha(\vec{x}); r^{{\cal A}_{n-1}})
  \leq \mu(\alpha(\vec{x}); r^{{\cal B}})$.

\item[\rm Step:] $\mu(\alpha(\vec{x});r^{{\cal A}_{n}}) = v > 0$

\begin{description}
\then there exists a variant $r(\vec{x}) \leftarrow_f \phi$ $\&$
$q_1({\vec{x}}_1)$ $\&$ $\ldots$ $\&$ $q_k({\vec{x}}_k)$ of a clause
in \Pf s.t.\ $v = f \times \min \{ \mu(\alpha({\vec{x}_1});{q_1}^{{\cal
    A}_{n-1}}), \ldots, \mu(\alpha({\vec{x}_k});{q_k}^{{\cal
    A}_{n-1}}) \}$ 
and $\alpha \in
\denAi{\phi}{n-1}$, by Definition \ref{equations}

\then $\mu(\alpha({\vec{x}_1});{q_1}^{{\cal B}}) \geq
\mu(\alpha({\vec{x}_1});{q_1}^{{\cal A}_{n-1}}), 
\ldots,
\mu(\alpha({\vec{x}_k});{q_k}^{{\cal B}}) \geq
\mu(\alpha({\vec{x}_k});{q_k}^{{\cal A}_{n-1}})$ and 
$\alpha \in \denX{\phi}{{\cal B}}$, by the hypothesis

\then $\mu(\alpha({\vec{x}});r^{{\cal B}}) \geq v$,
since ${\cal B}$ is a model of \Pf

\end{description}
\then $\mu(\alpha({\vec{x}});r^{{\cal A}_{n}}) \leq 
\mu(\alpha({\vec{x}});r^{{\cal B}})$.

\item[\rm For] $ v=0$ it follows immediately that
$\mu(\alpha({\vec{x}});r^{{\cal A}_{n}}) \leq 
\mu(\alpha({\vec{x}});r^{{\cal B}})$.

\item[] Claim (iii) follows by arithmetic induction.
\qed
\end{description}
\renewcommand{\qed}{}
\end{proof}

The following proposition allows us to link the declarative description of the
desired output from \Pf and a goal, i.e., a quantitative \Pf-answer, to
the minimal model semantics of \Pf. That is, Proposition
\ref{logicalcons} shows that quantitative \Pf-answers are completely
characterized by minimal models of \Pf. Similar to the non-quantitative
case, this is done for the quantitative case by connecting the concept
of logical consequence with the concept of minimal model. 

\begin{po} \label{logicalcons}
Let \Pf be a quantitative definite clause specification in 
  \Rl, $\varphi$ be an \La-constraint and $G$ be a goal.
Then $\varphi$ $\mbox{}_v\!\rightarrow G$ is a
  logical consequence of \Pf iff every minimal model \A of \Pf
is a model of $\{ \varphi$ $\mbox{}_v\!\rightarrow G \}$.
\end{po}

\begin{proof}
\begin{description}
\item[\rm If:] For each minimal model \A of \Pf:
\Xcal{A} is a model of $\{ \varphi$ $\mbox{}_v\!\rightarrow G \}$

\begin{description}
\then for every model \Xcal{B} of \Pf base equivalent to some minimal
model \A of \Pf: \Xcal{B}
is a model of $\{ \varphi$ $\mbox{}_v\!\rightarrow G \}$,
since $\Xcal{A} \subseteq \Xcal{B}$ by Theorem \ref{definite}, (iii)

\then $\varphi$ $\mbox{}_v\!\rightarrow G$
is a logical consequence of \Pf.
\end{description}

\item[\rm Only if:] $\varphi$ $\mbox{}_v\!\rightarrow G$
is a logical consequence of \Pf

\begin{description}
\then every model of \Pf is a model of $\{ \varphi$
$\mbox{}_v\!\rightarrow G \}$, by Definition \ref{logcons}

\then \Xcal{A} is a model of $\{ \varphi$ $\mbox{}_v\!\rightarrow G
\}$.
\qed
\end{description}
\end{description}
\renewcommand{\qed}{} 
\end{proof}

The following example illustrates the basic concepts of the
declarative semantics of quantitative definite clause specifications.
The program of Fig. \ref{QCLPprogram} is a quantitative
version of the program of Fig. \ref{CLPprogram}. The factors attached
to clauses \texttt{2} and \texttt{3} express a preference of the
\La-constraint $X=a$ over the \La-constraint $X=b$ in the definition of
the predicate \texttt{p}. Predicate \texttt{q} is defined uniquely in
clause \texttt{1} and gets assigned the factor $1$.

\begin{figure}[htbp]
\begin{center}
\begin{tabular}{l}
\texttt{1} $\texttt{q}(X) \leftarrow_1 \texttt{p}(X).$ \\
\texttt{2} $\texttt{p}(X) \leftarrow_{.7} X=a.$\\
\texttt{3} $\texttt{p}(X) \leftarrow_{.5} X=b.$
\end{tabular}
\end{center}

\caption{Quantitative constraint logic program}
\label{QCLPprogram}
\end{figure}

The construction of a minimal model for the program of
Fig. \ref{QCLPprogram} is shown in Fig.  \ref{QCLPmm}. For a
variable assignment $\alpha \in \denI{X=a}$,
the membership value of $.7$ of the object $\left<\alpha(X)\right>$ in
the denotation of the predicate \texttt{p} (resp. \texttt{q}) under
the minimal model \A is obtained in step 1 (resp. step 2) of the
\Pf-chain construction. For a variable assignment $\alpha \in \denI{X=b}$,
a membership degree of $.5$ is obtained in similar manner.

\begin{figure}[htbp]
  \begin{center}
    \begin{tabular}{l}
      $\alpha \in \denI{X=a}$: \\
      $\mu(\left< \alpha(X) \right> ; \texttt{p}^{\mathcal{A}_0}) = 0$, \\
      $\mu(\left< \alpha(X) \right> ; \texttt{p}^{\mathcal{A}_1}) = 
      \max \{ .7 \times \min \emptyset \} = .7$, \\
      $\mu(\left< \alpha(X) \right> ; \texttt{p}^{\mathcal{A}_2}) = 
      \max \{ .7 \times \min \emptyset \} = .7$, \\
      $\vdots$  \\
      $\mu(\left< \alpha(X) \right> ; \texttt{p}^{\bigcup_{i \geq 0} 
        \mathcal{A}_i }) =  
      \sup \{ 0,.7,.7,\ldots \} = .7,$ \\
      \vspace{3ex} \\
      $\mu(\left< \alpha(X) \right> ; \texttt{q}^{\mathcal{A}_0}) = 0$, \\
      $\mu(\left< \alpha(X) \right> ; \texttt{q}^{\mathcal{A}_1}) = 0$, \\
      $\mu(\left< \alpha(X) \right> ; \texttt{q}^{\mathcal{A}_2}) = 
      \max \{ 1 \times \min \{ .7 \} \} = .7$,\\
      $\vdots$ \\
      $\mu(\left< \alpha(X) \right> ; \texttt{q}^{\bigcup_{i \geq 0}
        \mathcal{A}_i }) =
      \sup \{ 0,0,.7,\ldots \} = .7 $. \\
      \vspace{5ex} \\
      $\alpha \in \denI{X=b}$: \\
      $\mu(\left< \alpha(X) \right> ; \texttt{p}^{\mathcal{A}_0}) = 0$, \\
      $\mu(\left< \alpha(X) \right> ; \texttt{p}^{\mathcal{A}_1}) = 
      \max \{ .5 \times \min \emptyset \} = .5$, \\
      $\mu(\left< \alpha(X) \right> ; \texttt{p}^{\mathcal{A}_2}) = 
      \max \{ .5 \times \min \emptyset \} = .5$, \\
      $\vdots$ \\
      $\mu(\left< \alpha(X) \right> ; \texttt{p}^{\bigcup_{i \geq 0}
        \mathcal{A}_i }) = \sup \{ 0,.5,.5,\ldots \} = .5,$ \\
      \vspace{3ex} \\
      $\mu(\left< \alpha(X) \right> ; \texttt{q}^{\mathcal{A}_0}) = 0$, \\
      $\mu(\left< \alpha(X) \right> ; \texttt{q}^{\mathcal{A}_1}) = 0$, \\
      $\mu(\left< \alpha(X) \right> ; \texttt{q}^{\mathcal{A}_2}) = 
      \max \{ 1 \times \min \{ .5 \} \} = .5$,\\
      $\vdots$ \\
      $\mu(\left< \alpha(X) \right> ; \texttt{q}^{\bigcup_{i \geq 0}
        \mathcal{A}_i }) =
      \sup \{ 0,0,.5,\ldots \} = .5 $, 
    \end{tabular}
  \end{center}
  \caption{\Pf-chain for quantitative constraint logic program}
  \label{QCLPmm}
\end{figure}
Clearly, $\A =\bigcup_{i \geq 0} \mathcal{A}_i$ is a minimal model of
the quantitative program of Fig. \ref{QCLPprogram}.

\section{Operational Semantics for Quantitative CLP}
\markright{3.5 Operational Semantics for Quantitative CLP}
\label{OpSem}

\subsection{Min/Max Trees and Quantitative Proof Trees}

The proof procedure for quantitative CLP can be stated conveniently as a search of a tree,
corresponding to the search of an SLD-and/or
tree in conventional logic programming or to the search of a derivation tree as
defined in Chap. \ref{Foundations} for CLP. 
The structure of such a tree exactly mirrors the construction of a
minimal model and thus may be defined as a min/max tree. That is,
according to the minimal model construction, which is based on the
operations $\min$ and $\max$, a min/max tree combines the standard left-right
selection and depth-first search with a min/max calculation of
node-values. A relation node of a derivation tree corresponds in the
quantitative case to a max-node, and a constraint node to a
min-node. In contrast to derivation trees, in min/max trees the unique
successor of a constraint node is split up into several successor
nodes, one for each relational atom in the goal. This is necessary to
calculate a minimum of node values at a min-node.

In the following we will assume implicit constraint languages \La and \Rl and
a given quantitative definite clause specification \Pf in \Rl. Furthermore, \Xsf{V} will
denote the finite set of variables in the query and the $\mathsf{V}$-solutions of
a constraint $\phi$ in an interpretation \I are defined as
$\denIV{\phi} := \{ \alpha|_\mathsf{V} |\; \alpha \in \denI{\phi} \}$ and
$\alpha|_\mathsf{V}$ is the restriction of $\alpha$ to \Xsf{V}.

\begin{de}[Min/max tree] A min/max tree determined by a query $G_1$ and
a quantitative definite clause specification \Pf has to satisfy the
following conditions:
\begin{itemize}
\item Each max-node is labeled by a goal.
The value of each nonterminal max-node is the maximum of the values of
its successors.
\item Each min-node is labeled by a clause from \Pf and a goal.
The value of each nonterminal min-node is $f \times m$, where $f$ is
the factor of the clause and $m$ is the minimum of the values of its
successors.
\item The successors of every max-node are all min-nodes s.t.\ for
  every clause $C$ with \gr-resolvent $G'$ obtained by $C$ from goal
  $G$ in a max-node, there is a min-node successor labeled by $C$
  and $G'$.
\item The successors of every min-node are all max-nodes s.t.\ for
  every \Rl-atom $r(\vec{x})$ in goal
$G \: \& \: \phi \: \&\: \phi'$ in a min-node with \cs-resolvent 
$G \: \& \: \phi''$, there is a max-node successor labeled by
$r(\vec{x}) \: \& \: \phi''$.
\item The root node is a max-node labeled by $G_1$.
\item A success node is a terminal max-node labeled by a satisfiable
  \La-constraint. The value of a success node is 1.
\item A failure node is a terminal max-node which is not a success
  node. The value of a failure node is 0.
\end{itemize}
\end{de}

Similar to the non-quantitative case, a proof tree in the quantitative
case is a subtree of a derivation tree. However, in a
quantitative proof tree, each min-node takes all of the successors of
the min-node of the min/max tree as its successors. Furthermore, to
check the consistency of the constraint solving results in the
min-node successors, an additional \cs-step has to be applied to the
conjunction of all success nodes of a quantitative proof tree. This
step yields a satisfiable \La-constraint, called answer constraint, if
the conjunction of the \La-constraints in the success nodes is
satisfiable. 

\begin{de}[Quantitative proof tree]
A quantitative proof tree for a goal $G_1$ from quantitative definite
clause specification \Pf  is a subtree of a min/max
supertree determined by $G_1$ and \Pf and defined as follows:
\begin{itemize}
\item The root node of the proof tree is the root node of the
  supertree.
\item A max-node of the proof tree is a max-node of the supertree and
  takes {\em one} of the successors of the supertree max-node as its
  successor.
\item A min-node of the proof tree is a min-node of the supertree and
  takes {\em all} of the successors of the supertree max-node as its
  successors.
\item All terminal nodes in the proof tree are success nodes $\phi,\:
  \phi', \ldots$
 s.t.\ $\phi \;\&\; \phi'\;\&\;\ldots \cs \varphi$
and $\varphi$ is a satisfiable \La-constraint, called answer
constraint.
\item Values are assigned to proof tree nodes in the same way as to
  min/max tree nodes.
\end{itemize}
\end{de}

\subsection{Soundness and Completeness}

To prove soundness and completeness of the generalized SLD-resolution proof
procedure defined via min/max trees and quantitative proof trees, some
further concepts have to be introduced.  

Note that the definitions of renaming, $\rho$-variant, and variant
carry over to the quantitative case without changes. Clearly, we have
the property that a constraint language \Rl containing quantitative
definite clauses is closed under renaming if the
underlying constraint language \La is closed under
renaming. Furthermore, for each such generalized constraint language
\Rl which is closed under renaming, and for each
\Rl-interpretation \A, we have that \A is a model of an \Rl-constraint iff \A
is a model of each of its variants.

Next, we have to redefine a complexity measure for goal
reduction for the quantitative case. This measure is crucial in
proving termination of goal reduction and works by keying steps of the
minimal model construction to steps of the goal reduction process.

\begin{itemize}
\item The complexity of a variable assignment $\alpha$ for an atom
$r(\vec{x})$ in the minimal model \A s.t.\
$\mu(\alpha(\vec{x}); r^{\cal A}) > 0$ 
is defined as 
\[
comp(\alpha, r(\vec{x}), \A)
:= \min \{ i | \; \mu(\alpha(\vec{x}); r^{\cal A}) =
\mu(\alpha(\vec{x}); r^{{\cal A}_i})\}
;
\]
\item The complexity of $\alpha$ for goal
$G = r_1(\vec{x}_1)$ $\&$ $\ldots$
$\&$ $r_k(\vec{x}_k)$ $\&$ $\phi$
in \A s.t.\ $\alpha \in \denA{\phi}$ and 
$\mu(\alpha(\vec{x}_i); {r_i}^{\cal A}) > 0$
for all $i: 1 \leq i \leq k$
is  defined as 
\[
comp(\alpha,G,\A) :=
\{ comp(\alpha,r_i(\vec{x}_i),\A)|\;
1 \leq i \leq k \}
\]
where $\{ \ldots \}$ is a multiset.
\item The $\mathsf{V}$-complexity of $\alpha$ for goal $G=r_1(\vec{x}_1)$ $\&$
$\ldots$ $\&$ $r_k(\vec{x}_k)$ $\&$ $\phi$ in \A s.t.\
$\alpha \in \denAV{\phi}$ and 
$\mu(\alpha(\vec{x}_i); {r_i}^{\cal A}) > 0$
for all $i: 1 \leq i \leq k$
is defined as 

\begin{center}
\begin{tabular}{ll}
$comp_\mathsf{V}(\alpha,G,\A) := \min \{ comp(\beta,G,
\A)$
& 
$| \; \beta \in \denA{\phi}, \; \mu(\beta(\vec{x}_i); {r_i}^\mathcal{
  A}) > 0 $
\\
& 
 $\textrm{ for all } i: 1 \leq i \leq
k \textrm{ and } \alpha = \beta |_\mathsf{V} \}
$.
\end{tabular}
\end{center}

The minimum is taken with respect to a total ordering on
multisets s.t.\ $M \leq M'$ iff $\forall x \in M \setminus M',
\exists x' \in M'\setminus M$ s.t.\ $x < x'$.
\end{itemize}

The following proofs show that the quantitative proof procedure  is
sound and complete with respect to the above stated semantic concepts. Again,
there is a close similarity to the corresponding statements for the
non-quantitative case of \citeN{HuS:88}. 

\begin{te}[Soundness] For each quantitative definite clause
  specification \Pf, for each goal $G$, for each \La-constraint
  $\varphi$:
If there is a quantitative proof tree for $G$ from \Pf with answer constraint
$\varphi$ and root value $v$, then
$\varphi$ $\mbox{}_v\!\rightarrow G$ is a logical consequence of \Pf.
\end{te}

\begin{proof}
The result is proven by induction on the depth $d$
of the quantitative proof tree, where one unit of depth is from max-node to
max-node.

\begin{description}
\item[\rm Base:] We know that quantitative proof trees
  of depth $d=0$ have to take the form of a single max-node labeled by a satisfiable
  \La-constraint $\psi$ with root value 1.
Then $\psi$ $\mbox{}_1\!\rightarrow \psi$ is a logical
  consequence of \Pf.

\item[\rm Hypothesis:] Suppose the result holds for quantitative proof trees of depth $d <
  n$.

\item[\rm Step:] Let $G_0 = r(\vec{x}) \: \& \: \phi$
be a goal labeling a quantitative proof tree of depth $d = n$ with answer
constraint $\psi$ and root value $h$,\\
 let $G_0' =
q_1({\vec{x}}_1)$ $\&$ $\ldots$ $\&$ $q_k({\vec{x}}_k)$ $\&$
$\phi$ $\&$ $\phi'$ be a goal labeling the min-node
obtained from $G_0$ via \gr using the variant
$C' = r(\vec{x}) \leftarrow_f \phi'$ $\&$
$q_1({\vec{x}}_1)$ $\&$ $\ldots$ $\&$ $q_k({\vec{x}}_k)$ of a clause
$C$ in \Pf, \\
and let $G_1 = q_1({\vec{x}}_1)$ $\&$ $\phi'', \ldots,
G_k = q_k({\vec{x}}_k)$ $\&$ $\phi''$ be goals labeling max-nodes 
obtained from $G_0'$ via \cs.

Then each goal $G_1, \ldots, G_k$ labels a quantitative proof tree of depth $d <
n$ with respective answer constraint
$\psi_1, \ldots, \psi_k$ and root value
$g_1, \ldots, g_k$ s.t.\ $h = f \times \min \{g_1, \ldots, g_k \}$ and
for each model \A of \Pf: $\denA{\psi} = \denA{\psi_1 \:\&\: \ldots \:\&\:
  \psi_k}$, by definition of min/max tree

\begin{description}
\then $\psi_1$ $\mbox{}_{g_1}\!\rightarrow G_1, \ldots, 
\psi_k$ $\mbox{}_{g_k}\!\rightarrow G_k$ are logical consequences of
\Pf, by the hypothesis

\then for each model \A of \Pf, for each $\alpha \in \ASS$:
$\denA{\psi} \subseteq \denA{\phi''}$ and if
$\alpha \in \denA{\psi}$, then $\mu(\alpha(\vec{x}_1); {q_1}^{\cal A})
\geq g_1, \ldots, \mu(\alpha(\vec{x}_k); {q_k}^{\cal A})
\geq g_k$, by definition of logical consequence

\then  for each model \A of \Pf, for each $\alpha \in \ASS$:
$\denA{\psi} \subseteq \denA{\phi'}$ and if
$\alpha \in \denA{\psi}$, then $\mu(\alpha(\vec{x}); {r}^{\cal A})
\geq f \times \min \{ \mu(\alpha(\vec{x}_1); {q_1}^{\cal A}), \ldots,
\mu(\alpha(\vec{x}_k); {q_k}^{\cal A}) \}$, since each model \A of \Pf is a model
of $C'$ iff \A is a model of $C$

\then for each model \A of \Pf, for each $\alpha \in \ASS$:
$\denA{\psi} \subseteq \denA{\phi}$ and if
$\alpha \in \denA{\psi}$, then $\mu(\alpha(\vec{x}); {r}^{\cal A})
\geq h$
\end{description}

\then $\psi$ $\mbox{}_{h}\!\rightarrow r(\vec{x})$ $\&$
$\phi$ is a logical consequence of \Pf.
 
\item[] The result follows by arithmetic induction.
\qed
\end{description}
\renewcommand{\qed}{}
\end{proof}

\begin{te}[Completeness] Let \Pf be a quantitative definite clause
  specification in \Rl, \La be closed under renaming, \A be a minimal model of \Pf,
 $G$ be a goal of the form $r(\vec{x})$ $\&$
  $\phi$, $\alpha \in \denAV{\phi}$ and
$\mu(\beta(\vec{x}); r^{\cal A}) = v$ s.t.\ $v > 0$ and
$\alpha = \beta|_\mathsf{V}$.
Then there exists a quantitative proof tree for $G$ from \Pf with answer constraint
$\varphi$ and root value $v$ and $\alpha \in \denAV{\varphi}$.
\end{te}

\begin{proof}
The result is proven by induction on $c =
comp_\mathsf{V}(\alpha, G, \A)$.

\begin{description}
\item[\rm Base:] We know that goals with complexity $c=\emptyset$ have
  to take the form of a satisfiable \La-constraint $\chi$.
Then there exists a quantitative proof tree for $\chi$ from \Pf consisting of a
single max-node labeled with $\chi$ and root value 1.

\item[\rm Hypothesis:] Suppose the result holds for goals with
  complexity $c < N$.

\item[\rm Step:] Let $G_0 = q(\vec{x})$ $\&$ $\psi$, $\alpha' \in
  \denAV{\psi}$, $\alpha'' \in \denA{\psi}$,
$\alpha' = \alpha''|_\mathsf{V}$, $comp_\mathsf{V}(\alpha',G_0,\A) $
$=$ 
\\
$ comp(\alpha'',G_0,\A) = N$, $comp(\alpha'',q(\vec{x}),\A) := i$,
$\mu(\alpha''(\vec{x}); q^{\cal A}) = h$
and $h>0$.

\begin{description}
\item[] First we observe, that
$\mu(\alpha''(\vec{x}); q^{{\cal A}_i}) = h$,
since $comp(\alpha'',q(\vec{x}),\A) := i$

\begin{description}
\then there exists a variant
$q(\vec{x}) \leftarrow_f \psi'$ $\&$
$q_1({\vec{x}}_1)$ $\&$ $\ldots$ $\&$ $q_k({\vec{x}}_k)$ s.t.\ 
$h= f \times \min \{ \mu(\alpha({\vec{x}_1});
{q_1}^{{\cal A}_{i-1}}),$ $\ldots,$ 
$\mu(\alpha({\vec{x}_k});{q_k}^{{\cal A}_{i-1}}) \}$
and $\alpha'' \in \denAi{\psi'}{i-1}$
and $(\Xsf{V} \cup \Xsf{V}(\psi)) \cap
\Xsf{V}(\psi'$ $\&$ $q_1({\vec{x}}_1)$ $\&$ $\ldots$ $\&$
$q_k({\vec{x}}_k)) \subseteq \Xsf{V}(q(\vec{x}))$,
by Definition \ref{equations} and renaming closure of \Rl, finite
\Xsf{V} and infinitely many variables in \Xsf{VAR}

\then $G_0 \stackrel{r,c}{\longrightarrow} G_0'$ s.t.\ $G_0' =
q_1({\vec{x}}_1)$ $\&$ $\ldots$ $\&$ $q_k({\vec{x}}_k)$ $\&$
$\psi''$
and $\denAV{\psi''} = \denAV{\psi \:\&\: \psi'}$, by definition
of the inference rules.
\end{description}

\item[] Next, $\alpha' \in \denAV{\psi''}$, since
$\alpha'' \in \denA{\psi}$, $\alpha'' \in \denAi{\psi'}{i-1}
\subseteq \denA{\psi'}$, 
$\alpha'' \in \denA{\psi \:\&\: \psi'}$, 
\\
$\denAV{\psi \:\&\: \psi'} = \denAV{\psi''}$
and $\alpha' = \alpha''|_\mathsf{V}$.

\item[] Finally, $comp_\mathsf{V}(\alpha', G_0', \A) < N$, 
since $comp_\mathsf{V}(\alpha', G_0', \A)$
$\leq$ $comp(\alpha'', G_0', \A)$
$<$ $\{ i\}$
$=$ $\{ comp(\alpha'', q(\vec{x}), \A) \}$
$=$ $comp(\alpha'',G_0, \A)$
$=$ $comp_\mathsf{V}(\alpha',G_0,\A)$
$=$ $N$.

\item[] Now we can obtain goals $G_1 = q_1(\vec{x}_1)$ $\&$
  $\psi'',\ldots, G_k$ $
= q_k(\vec{x}_k)$ 
$\&$ $\psi''$ from $G_0'$
s.t.\ $\alpha' \in \denAV{\psi''}$, 
$\mu(\alpha''({\vec{x}_1}); {q_1}^{{\cal A}}) = g_1 > 0,
\ldots,$ 
$\mu(\alpha''({\vec{x}_k}); {q_k}^{{\cal A}}) $
$= g_k > 0$,
$\alpha' = \alpha''|_\mathsf{V}$ and
$comp_\mathsf{V}(\alpha',G_1,\A) < N, $\ldots 
$, comp_\mathsf{V}(\alpha', G_k, \A) < N$

\begin{description}
\then for each goal $G_1, \ldots, G_k$, there exists a quantitative proof tree from
\Pf with respective answer constraint $\chi_1, \ldots, \chi_k$ and
respective root value 
\\
$g_1' = g_1, \ldots, g_k' = g_k$ and
$\alpha' \in \denAV{\chi_1 \:\&\: \ldots \:\&\: \chi_k} = \denAV{\chi}$,
by the hypothesis
\end{description}
\end{description}

\then there exists a quantitative proof tree for $G_0$ from \Pf with answer
constraint $\chi$ and root value $h' = f \times \min \{ g_1', \ldots,
g_k' \} = f \times \min \{g_1, \ldots, g_k \} = h$ and $\alpha' \in
\denAV{\chi}$.

\item[] The result follows by arithmetic induction.
\qed
\end{description}
\renewcommand{\qed}{}
\end{proof}

\begin{figure}[htbp]
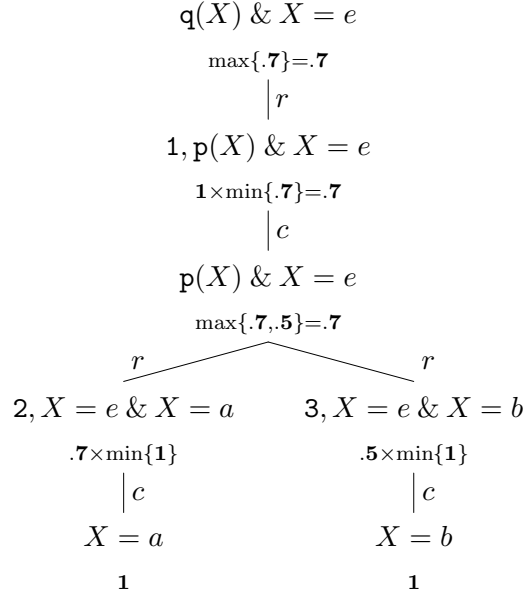

\begin{center}
\setlength{\GapWidth}{70pt}

\begin{bundle}{$\begin{array}{c}
      \texttt{q}(X) \:\&\: X=e\\
       \mathbf{\scriptstyle \max \{ .7 \} = .7}
        \end{array}$}
  \chunk[$\quad r$]{
    \begin{bundle}{$\begin{array}{c}
          \texttt{1}, \texttt{p}(X) \:\&\: X=e \\
          \mathbf{\scriptstyle 1 \times \min \{ .7 \} = .7}
          \end{array}$}
      \chunk[$\quad c$]{
        \begin{bundle}{$\begin{array}{c}
              \texttt{p}(X) \:\&\: X=e\\
              \mathbf{\scriptstyle \max \{ .7,.5 \} = .7 }
              \end{array}$}
          \chunk[$\quad r$]{
            \begin{bundle}{$\begin{array}{c}
                  \texttt{2}, X=e \:\&\: X=a \\
                  \mathbf{\scriptstyle .7 \times \min \{ 1 \} }
                  \end{array}$}
              \chunk[$\quad c$]{$\begin{array}{c}
                  X=a\\
                  \mathbf{\scriptstyle 1}
                  \end{array}$}
            \end{bundle}
            }
          \chunk[$\quad r$]{
            \begin{bundle}{$\begin{array}{c}
                  \texttt{3}, X=e \:\&\: X=b\\
                  \mathbf{\scriptstyle .5 \times \min \{ 1 \}}
                  \end{array}$}
              \chunk[$\quad c$]{$\begin{array}{c}
                  X=b\\
                  \mathbf{\scriptstyle 1}
                  \end{array}$}
            \end{bundle}
            }
        \end{bundle}
        }
    \end{bundle}
    }
\end{bundle}

\end{center}
\caption{Min/max tree for quantitative constraint logic program}
\label{QCLPminmaxtree}
\end{figure}

Returning to our toy example, the proof procedure for quantitative
definite clause specifications can be illustrated as follows. A
min/max derivation tree for the query $\texttt{q}(X) \:\&\: X=e$ and
the program of Fig. \ref{QCLPprogram} is given in
Fig. \ref{QCLPminmaxtree}.

\begin{figure}[htbp]
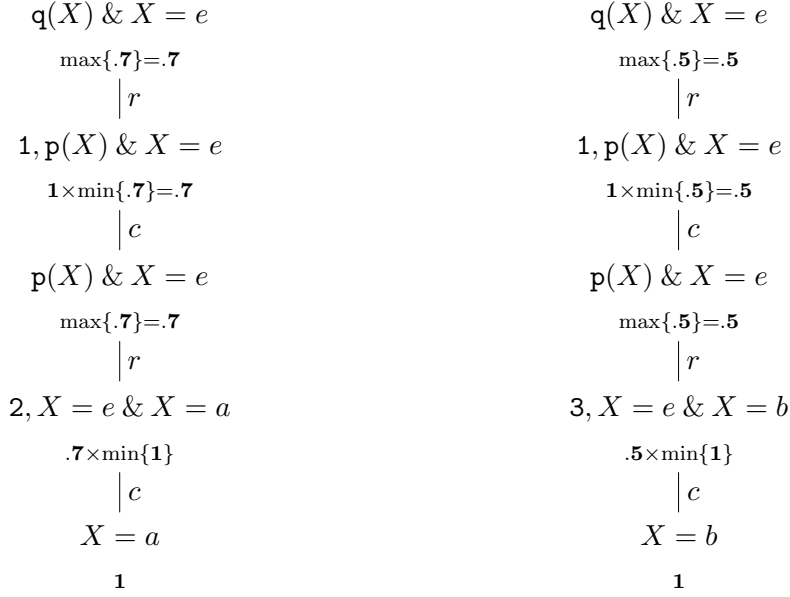

\begin{center}
\begin{tabular}{p{200pt}p{100pt}}

\begin{bundle}{$\begin{array}{c}
      \texttt{q}(X) \:\&\: X=e\\
      \mathbf{\scriptstyle \max \{ .7 \} = .7}
    \end{array}$}
  \chunk[$\quad r$]{
    \begin{bundle}{$\begin{array}{c}
          \texttt{1}, \texttt{p}(X) \:\&\: X=e \\
          \mathbf{\scriptstyle 1 \times \min \{ .7 \} = .7}
        \end{array}$}
      \chunk[$\quad c$]{
        \begin{bundle}{$\begin{array}{c}
              \texttt{p}(X) \:\&\: X=e\\
              \mathbf{\scriptstyle \max \{ .7 \} = .7 }
            \end{array}$}
          \chunk[$\quad r$]{
            \begin{bundle}{$\begin{array}{c}
                  \texttt{2}, X=e \:\&\: X=a \\
                  \mathbf{\scriptstyle .7 \times \min \{ 1 \} }
                \end{array}$}
              \chunk[$\quad c$]{$\begin{array}{c}
                  X=a\\
                  \mathbf{\scriptstyle 1}
                \end{array}$}
            \end{bundle}
            }
        \end{bundle}
        }
    \end{bundle}
    }
\end{bundle}

&

\begin{bundle}{$\begin{array}{c}
      \texttt{q}(X) \:\&\: X=e\\
      \mathbf{\scriptstyle \max \{ .5 \} = .5}
        \end{array}$}
  \chunk[$\quad r$]{
    \begin{bundle}{$\begin{array}{c}
          \texttt{1}, \texttt{p}(X) \:\&\: X=e \\
          \mathbf{\scriptstyle 1 \times \min \{ .5 \} = .5}
          \end{array}$}
      \chunk[$\quad c$]{
        \begin{bundle}{$\begin{array}{c}
              \texttt{p}(X) \:\&\: X=e\\
              \mathbf{\scriptstyle \max \{ .5 \} = .5 }
              \end{array}$}
          \chunk[$\quad r$]{
            \begin{bundle}{$\begin{array}{c}
                  \texttt{3}, X=e \:\&\: X=b\\
                  \mathbf{\scriptstyle .5 \times \min \{ 1 \}}
                  \end{array}$}
              \chunk[$\quad c$]{$\begin{array}{c}
                  X=b\\
                  \mathbf{\scriptstyle 1}
                  \end{array}$}
            \end{bundle}
            }
        \end{bundle}
        }
    \end{bundle}
    }
\end{bundle}

\end{tabular}
\end{center}
\caption{Quantitative proof trees for quantitative constraint logic program} 
\label{QCLPprooftrees} 
\end{figure}

This tree contains two success nodes, $X=a$ and $X=b$, from which two
distinct quantitative proof trees can be obtained (see Fig. \ref{QCLPprooftrees}).

Soundness of quantitative CLP tells us that corresponding to the quantitative proof
tree with answer constraint $X=a$ (resp. $X=b$) and root value $.7$
(resp. $.5$), we know that the quantitative formula
$X=a\; \mbox{}_{.7}\!\rightarrow \texttt{q}(X) \:\&\: X=e$ 
(resp. $X=b \;\mbox{}_{.5}\!\rightarrow \texttt{q}(X) \:\&\: X=e$ ) is a
logical consequence of the program of Fig. \ref{QCLPprogram}. This can
easily be verified from the minimal model constructed in
Fig. \ref{QCLPmm}.

Completeness says that for an object $\left< \alpha(X) \right>$
assigned by $\alpha \in \denI{X=e}$ with membership degree
$\mu(\left< \alpha(X) \right>; \texttt{q}^\mathcal{A}) = .7$
to the denotation of \texttt{q} under the minimal model \A, we have
a corresponding proof tree with answer constraint $X=a$ and root value $.7$
and $\alpha \in \denI{X=a}$. 
Similarly, for an object  $\left< \alpha'(X) \right>$ with 
$\alpha' \in \denI{X=e}$ and 
$\mu(\left< \alpha'(X) \right>; \texttt{q}^\mathcal{A}) = .5$,
we have a proof tree with answer constraint $X=b$ and root value $.5$
and $\alpha' \in \denI{X=b}$.

\section{Parsing and Searching in Quantitative CLGs}
\markright{3.6 Parsing and Searching in Quantitative CLGs}
\label{QCLG}

The quantitative CLP scheme presented in the last chapter allows for a definition
of the parsing problem (and similarly of the
generation problem) for quantitative CLGs in the following way:
Given a program $\Xcal{P}_F$ (encoding some quantitative CLG) and a query $G$
(encoding some input sentence), we ask if we can infer a \Pf-answer
$\varphi$ of $G$ (encoding a parse of the input sentence)
at a value $\upsilon$ (encoding the weight of the parse) proving
$\varphi \;\mbox{}_{\upsilon}\!\rightarrow G$ to be a logical
consequence of \Pf.
That is, according to the soundness and completeness results presented above, the
operational concept of a quantitative proof tree has a declarative
counterpart in the form of a quantitative \Pf-answer. 
Truth-functionally, a quantitative \Pf-answer tells us that the answer
constraint $\varphi$ contributes a truth-value $\upsilon$ to the goal
$G$ in every model of \Pf. In terms of membership values, this means
that a \Pf-answer to a query $G=r(\vec{x}) \;\&\; \phi$
at value $\upsilon$ is a satisfiable \La-constraint $\varphi$ such
that for each model \A of \Pf holds:
If $\varphi$ is satisfiable, then $\phi$ is satisfiable and all
objects assigned to $\vec{x}$ by a solution of $\varphi$ are in the
denotation of $r(\vec{x})$ at a membership value of at least
$\upsilon$.

\subsection{Quantitative Feature-Based CLGs}

Returning to the simple linguistic CLG of Fig. \ref{CLGprogram}, the
formal scheme described above can be illustrated as follows.

The quantitative constraint logic program \Pf of Fig. \ref{QCLGprogram} is
obtained from the program of Fig. \ref{CLGprogram} simply by adding
numerical factors to the program clauses.

\begin{figure}[htbp]
\begin{description}
\item[{\tt 1}  $\texttt{phrase}(X) \leftarrow_{f_1}$] $ X = (phrase \:\wedge\:
\textsc{CAT}:s \:\wedge\: \textsc{DTR1:CAT}:n \:\wedge\:
\textsc{DTR2:CAT}:v \:\wedge\: \textsc{DTR1:AGR}:Y \:\wedge\:
\textsc{DTR2:AGR}:Y \:\wedge\: \textsc{DTR1}:Z_1
\:\wedge\: \textsc{DTR2}:Z_2) \:\&\: \texttt{sign}(Z_1) \:\&\: \texttt{sign}(Z_2)$.
\item[{\tt 2} $\texttt{phrase}(X) \leftarrow_{f_2}$] $ X = (phrase \:\wedge\:
\textsc{CAT}:np \:\wedge\:\textsc{DTR1:CAT}:n \:\wedge\: \textsc{DTR2:CAT}:n \:\wedge\: \textsc{DTR1}:Z_1
\:\wedge\: \textsc{DTR2}:Z_2) \:\&\: \texttt{sign}(Z_1) \:\&\: \texttt{sign}(Z_2)$.
\item[{\tt 3}  $ \texttt{word}(X) \leftarrow_{f_3}$] $ X = (word \:\wedge\:
  \textsc{CAT}:n \:\wedge\: \textsc{PHON}:Clinton \:\wedge\:
  \textsc{AGR}:sg)$.
\item[{\tt 4} $\texttt{word}(X) \leftarrow_{f_4}$] $ X = (word \:\wedge\:
  \textsc{CAT}:v \:\wedge\: \textsc{PHON}:talks \:\wedge\:
  \textsc{AGR}:sg)$.
\item[{\tt 5} $\texttt{word}(X) \leftarrow_{f_5}$] $ X = (word \:\wedge\:
  \textsc{CAT}:n \:\wedge\: \textsc{PHON}:talks \:\wedge\:
  \textsc{AGR}:pl)$.
\item[{\tt 6} $\texttt{sign}(X) \leftarrow_{f_6}$] $\texttt{phrase}(X)$.
\item[{\tt 7} $\texttt{sign}(X) \leftarrow_{f_7}$] $\texttt{word}(X)$.
\end{description}
\caption{Quantitative feature-based constraint logic grammar}
\label{QCLGprogram}
\end{figure}

Given the quantitative CLG of Fig. \ref{QCLGprogram} and a goal $G$ of
the form
\[
X = (sign \: \wedge \: \textsc{DTR1: PHON}:Clinton \:\wedge\:
\textsc{DTR2: PHON}:talks) \:\&\: \texttt{sign}(X),
\]
again encoding the input sentence \emph{Clinton talks},
we can infer two different proof trees for $G$, each with a specific
answer constraint, encoding a parse, and a specific root value, 
encoding the preference value of the parse. Again, we will depict only
success branches and consider the constraint solver as a black box.
The two derivations are shown in Figs. \ref{QCLG1st} and
\ref{QCLG2nd}.

\begin{figure}[htbp]
  \begin{center}
    \footnotesize{
      \setlength{\GapWidth}{70pt}
      
      \begin{bundle}
        {$\begin{array}{c}
            X = (sign \: \wedge \: \textsc{DTR1: PHON}:Clinton
            \:\wedge\: \textsc{DTR2: PHON}:talks) 
            \:\&\: \texttt{sign}(X)\\
            \mathbf{f_6 \times f_1 \times min \{ f_7 \times f_3, f_7 \times f_4
              \} }
          \end{array}$}
        
        \chunk[$\quad r$]
        {\begin{bundle}
            {$\begin{array}{c}
                {\tt 6,} \; X = (sign \: \wedge \: \textsc{DTR1:
                  PHON}: Clinton
                \:\wedge\: \textsc{DTR2: PHON}:talks)
                \:\&\: \texttt{phrase}(X)\\
                \mathbf{f_6 \times f_1 \times min \{ f_7 \times f_3, f_7 \times f_4
                  \} }
              \end{array}$}
            
            \chunk[$\quad c$]
            {\begin{bundle}
                {$\begin{array}{c}
                    X = (sign \: \wedge \: \textsc{DTR1: PHON}:Clinton
                    \:\wedge\: \textsc{DTR2: PHON}:talks) 
                    \:\&\: \texttt{phrase}(X)\\
                    \mathbf{f_1 \times min \{ f_7 \times f_3, f_7 \times f_4 \} }
                  \end{array}$}
                
                \chunk[$\quad r$]
                {\begin{bundle}
                    {$\begin{array}{c}
                        {\tt 1,} \; 
                        X = (sign \: \wedge \: \textsc{DTR1: PHON}:Clinton
                        \:\wedge\: \textsc{DTR2: PHON}:talks) \\
                        \:\&\:
                        X = (phrase \:\wedge\: \textsc{CAT}:s \:\wedge\:
                        \textsc{DTR1: CAT}:n \:\wedge\: \textsc{DTR2: CAT}:v 
                        \:\wedge\: \textsc{DTR1:AGR}:Y \\
                        \:\wedge\: \textsc{DTR2:AGR}:Y 
                        \:\wedge\: \textsc{DTR1}:Z_1 \:\wedge\: \textsc{DTR2}:Z_2)
                        \:\&\: \texttt{sign}(Z_1)
                        \:\&\: \texttt{sign}(Z_2)\\
                        \mathbf{f_1 \times min \{ f_7 \times f_3, f_7 \times f_4 \} }
                      \end{array}$}

                    \chunk[$\quad c$]
                    {
                      \begin{bundle}
                        {$ \begin{array}{c}
                            \ast[X = (phrase \:\wedge\:  \textsc{CAT}:s \:\wedge\:
                            \textsc{DTR1}:word  \:\wedge\: \textsc{DTR1: CAT}:n \\
                            \:\wedge\: \textsc{DTR1: PHON}:Clinton
                            \:\wedge\: \textsc{DTR1:AGR}:Y
                            \:\wedge\: \textsc{DTR2}:word\: 
                            \wedge\:\textsc{DTR2: CAT}:v \\
                            \:\wedge\: \textsc{DTR2: PHON}:talks 
                            \:\wedge\: \textsc{DTR2:AGR}:Y
                            \:\wedge\: \textsc{DTR1}:Z_1\:\wedge\: \textsc{DTR2}:Z_2)]\\
                            \:\&\: \texttt{sign}(Z_1)\\
                            \mathbf{f_7 \times f_3}
                          \end{array}
                          $}
                        
                        \chunk[$\quad r$]
                        {\begin{bundle}
                            {$\begin{array}{c}
                                {\tt 7,} \; \ast[X = (phrase \:\wedge\:  \textsc{CAT}:s \:\wedge\:
                                \textsc{DTR1}:word  \:\wedge\: \textsc{DTR1: CAT}:n \\
                                \:\wedge\: \textsc{DTR1: PHON}:Clinton
                                \:\wedge\: \textsc{DTR1:AGR}:Y
                                \:\wedge\: \textsc{DTR2}:word\: 
                                \wedge\:\textsc{DTR2: CAT}:v \\
                                \:\wedge\: \textsc{DTR2: PHON}:talks 
                                \:\wedge\: \textsc{DTR2:AGR}:Y
                                \:\wedge\: \textsc{DTR1}:Z_1 \:\wedge\: \textsc{DTR2}:Z_2)]\\
                                \:\&\: \texttt{word}(Z_1)\\
                                \mathbf{f_7 \times f_3}
                              \end{array}$}
                            
                            \chunk[$\quad c$]
                            {\begin{bundle}
                                {$\begin{array}{c}
                                    \ast[X = (phrase \:\wedge\:  \textsc{CAT}:s \:\wedge\:
                                    \textsc{DTR1}:word  \:\wedge\: \textsc{DTR1: CAT}:n \\
                                    \:\wedge\: \textsc{DTR1: PHON}:Clinton
                                    \:\wedge\: \textsc{DTR1:AGR}:Y
                                    \:\wedge\: \textsc{DTR2}:word\: 
                                    \wedge\:\textsc{DTR2: CAT}:v \\
                                    \:\wedge\: \textsc{DTR2: PHON}:talks 
                                    \:\wedge\: \textsc{DTR2:AGR}:Y
                                    \:\wedge\: \textsc{DTR1}:Z_1 \:\wedge\: \textsc{DTR2}:Z_2)]\\
                                    \:\&\: \texttt{word}(Z_1)\\
                                    \mathbf{f_3}
                                  \end{array}$}
                                
                                \chunk[$\quad r$]
                                {\begin{bundle}
                                    {$\begin{array}{c}
                                        {\tt 3,} \; \ast[X = (phrase \:\wedge\:  \textsc{CAT}:s \:\wedge\:
                                        \textsc{DTR1}:word  \:\wedge\: \textsc{DTR1: CAT}:n \\
                                        \:\wedge\: \textsc{DTR1: PHON}:Clinton
                                        \:\wedge\: \textsc{DTR1:AGR}:Y
                                        \:\wedge\: \textsc{DTR2}:word\: 
                                        \wedge\:\textsc{DTR2: CAT}:v \\
                                        \:\wedge\: \textsc{DTR2: PHON}:talks 
                                        \:\wedge\: \textsc{DTR2:AGR}:Y
                                        \:\wedge\: \textsc{DTR1}:Z_1 \:\wedge\: \textsc{DTR2}:Z_2)] \\
                                        \:\&\:
                                        Z_1= (word \:\wedge\: \textsc{CAT}:n \:\wedge\:
                                        \textsc{PHON}:Clinton \:\wedge\: \textsc{AGR}:sg)\\
                                        \mathbf{f_3}
                                      \end{array}$}
                                    
                                    \chunk[$\quad c$]
                                    {$\begin{array}{c}
                                        \star[X = (phrase \:\wedge\:  \textsc{CAT}:s \:\wedge\:
                                        \textsc{DTR1}:word  \:\wedge\: \textsc{DTR1: CAT}:n \\
                                        \:\wedge\: \textsc{DTR1: PHON}:Clinton
                                        \:\wedge\: \textsc{DTR1:AGR}:Y
                                        \:\wedge\: \textsc{DTR1:AGR}:sg \\
                                        \:\wedge\: \textsc{DTR2}:word\: 
                                        \wedge\:\textsc{DTR2: CAT}:v 
                                        \:\wedge\: \textsc{DTR2: PHON}:talks \\
                                        \:\wedge\: \textsc{DTR2:AGR}:Y
                                        \:\wedge\: \textsc{DTR2:AGR}:sg )]\\
                                        \mathbf{1}
                                      \end{array}$}
                                  \end{bundle}
                                  }
                              \end{bundle}
                              }
                          \end{bundle}
                          }
                      \end{bundle}
                      }

                    \chunk[$\quad c$]
                    {
                      \begin{bundle}
                        {$\begin{array}{c}
                            \ast[\ldots] \:\&\: \texttt{sign}(Z_2)\\
                            \mathbf{f_7 \times f_4}
                          \end{array}$}
                        
                        \chunk[$\quad r$]
                        {\begin{bundle}
                            {$\begin{array}{c}
                                {\tt 7,} \; \ast[\ldots] \:\&\: \texttt{word}(Z_2)\\
                                \mathbf{f_7 \times f_4}
                              \end{array}$}
                            
                            \chunk[$\quad c$]
                            {\begin{bundle}
                                {$\begin{array}{c}
                                    \ast[\ldots] \:\&\: \texttt{word}(Z_2)\\
                                    \mathbf{f_4}
                                  \end{array}$}
                                
                                \chunk[$\quad r$]
                                {\begin{bundle}
                                    {$\begin{array}{c}
                                        {\tt 4,} \; \ast[\ldots] \\
                                        \:\&\: Z_2 = (word  \:\wedge\: \textsc{CAT}:v \\
                                        \:\wedge\: \textsc{PHON}:talks \:\wedge\: \textsc{AGR}:sg)\\
                                        \mathbf{f_4}
                                      \end{array}$}
                                    
                                    \chunk[$\quad c$]
                                    {$\begin{array}{c}
                                        \star[\ldots]\\
                                        \mathbf{1}
                                      \end{array}$}
                                    
                                  \end{bundle}
                                  }
                              \end{bundle}
                              }
                          \end{bundle}
                          }
                      \end{bundle}
                      }
                  \end{bundle}
                  }
              \end{bundle}
              }
          \end{bundle}
          }
      \end{bundle}
      }
  \end{center}
  
  \caption{Quantitative derivation of $[Clinton_N \; talks_V]_S$}
\label{QCLG1st}
\end{figure}

\begin{figure}[htbp]
  \begin{center}
    \footnotesize{

      \begin{bundle}
        {$\begin{array}{c}
            X = (sign \: \wedge \: \textsc{DTR1: PHON}:Clinton
            \:\wedge\: \textsc{DTR2: PHON}:talks) \:\&\: \texttt{sign}(X)\\
            \mathbf{f_6 \times f_2 \times min \{ f_7 \times f_3, f_7
              \times f_5 \} }
          \end{array}$}
        
        \chunk[$\quad r$]
        {\begin{bundle}
            {$\begin{array}{c}
                {\tt 6,} \; X = (sign \: \wedge \: \textsc{DTR1:
                  PHON}: Clinton
                \:\wedge\: \textsc{DTR2: PHON}:talks) \:\&\: \texttt{phrase}(X)\\
                \mathbf{f_6 \times f_2 \times min \{ f_7 \times f_3, f_7
                  \times f_5 \} }
              \end{array}$}
            
            \chunk[$\quad c$]
            {\begin{bundle}
                {$\begin{array}{c}
                    X = (sign \: \wedge \: \textsc{DTR1: PHON}:Clinton
                    \:\wedge\: \textsc{DTR2: PHON}:talks) \:\&\: \texttt{phrase}(X)\\
                    \mathbf{f_2 \times min \{ f_7 \times f_3, f_7 \times f_5 \} }
                  \end{array}$}
                
                \chunk[$\quad r$]
                {\begin{bundle}
                    {$\begin{array}{c}
                        {\tt 2,} \; 
                        X = (sign \: \wedge \: \textsc{DTR1: PHON}:Clinton
                        \:\wedge\: \textsc{DTR2: PHON}:talks) \\
                        \:\&\:
                        X = (phrase \:\wedge\: \textsc{CAT}:np \:\wedge\:
                        \textsc{DTR1: CAT}:n \:\wedge\: \textsc{DTR2: CAT}:n \\
                        \:\wedge\: \textsc{DTR1}:Z_1 \:\wedge\: \textsc{DTR2}:Z_2)\\
                        \:\&\: \texttt{sign}(Z_1)
                        \:\&\: \texttt{sign}(Z_2) \\
                        \mathbf{f_2 \times min \{ f_7 \times f_3, f_7 \times f_5 \} }
                      \end{array}$}

                    \chunk[$\quad c$]
                    {
                      \begin{bundle}
                        {$ \begin{array}{c}
                            \ast[X = (phrase \:\wedge\:  \textsc{CAT}:np \:\wedge\:
                            \textsc{DTR1}:word  \:\wedge\: \textsc{DTR1: CAT}:n \\
                            \:\wedge\: \textsc{DTR1: PHON}:Clinton
                            \:\wedge\: \textsc{DTR2}:word\: 
                            \wedge\:\textsc{DTR2: CAT}:n \\
                            \:\wedge\: \textsc{DTR2: PHON}:talks 
                            \:\wedge\: \textsc{DTR1}:Z_1 \:\wedge\: \textsc{DTR2}:Z_2)]\\
                            \:\&\: \texttt{sign}(Z_1)\\
                            \mathbf{f_7 \times f_3}
                          \end{array}
                          $}
                        
                        \chunk[$\quad r$]
                        {\begin{bundle}
                            {$\begin{array}{c}
                                {\tt 7,} \; \ast[X = (phrase \:\wedge\:  \textsc{CAT}:np \:\wedge\:
                                \textsc{DTR1}:word  \:\wedge\: \textsc{DTR1: CAT}:n \\
                                \:\wedge\: \textsc{DTR1: PHON}:Clinton
                                \:\wedge\: \textsc{DTR2}:word\: 
                                \wedge\:\textsc{DTR2: CAT}:n \\
                                \:\wedge\: \textsc{DTR2: PHON}:talks 
                                \:\wedge\: \textsc{DTR1}:Z_1 \:\wedge\: \textsc{DTR2}:Z_2)] \\
                                \:\&\: \texttt{word}(Z_1)\\
                                \mathbf{f_7 \times f_3}
                              \end{array}$}
                            
                            \chunk[$\quad c$]
                            {\begin{bundle}
                                {$\begin{array}{c}
                                    \ast[ X = (phrase \:\wedge\:  \textsc{CAT}:np \:\wedge\:
                                    \textsc{DTR1}:word  \:\wedge\: \textsc{DTR1: CAT}:n \\
                                    \:\wedge\: \textsc{DTR1: PHON}:Clinton
                                    \:\wedge\: \textsc{DTR2}:word\: 
                                    \wedge\:\textsc{DTR2: CAT}:n \\
                                    \:\wedge\: \textsc{DTR2: PHON}:talks 
                                    \:\wedge\: \textsc{DTR1}:Z_1 \:\wedge\: \textsc{DTR2}:Z_2)] \\
                                    \:\&\: \texttt{word}(Z_1)\\
                                    \mathbf{f_3}
                                  \end{array}$}
                                
                                \chunk[$\quad r$]
                                {\begin{bundle}
                                    {$\begin{array}{c}
                                        {\tt 3,} \; \ast[ X = (phrase \:\wedge\:  \textsc{CAT}:np \:\wedge\:
                                        \textsc{DTR1}:word  \:\wedge\: \textsc{DTR1: CAT}:n \\
                                        \:\wedge\: \textsc{DTR1: PHON}:Clinton
                                        \:\wedge\: \textsc{DTR2}:word 
                                        \:\wedge\:\textsc{DTR2: CAT}:n \\
                                        \:\wedge\: \textsc{DTR2: PHON}:talks 
                                        \:\wedge\: \textsc{DTR1}:Z_1 \:\wedge\: \textsc{DTR2}:Z_2)]\\
                                        \:\&\:
                                        Z_1= (word \:\wedge\: \textsc{CAT}:n \:\wedge\:
                                        \textsc{PHON}:Clinton)\\
                                        \mathbf{f_3}
                                      \end{array}$}
                                    
                                    \chunk[$\quad c$]
                                    {$\begin{array}{c}
                                        \star[X = (phrase \:\wedge\:  \textsc{CAT}:np \:\wedge\:
                                        \textsc{DTR1}:word
                                        \:\wedge\: \textsc{DTR1: CAT}:n \\
                                        \:\wedge\: \textsc{DTR1: PHON}:Clinton 
                                        \:\wedge\: \textsc{DTR1: AGR}:sg
                                        \:\wedge\: \textsc{DTR2}:word \\
                                        \:\wedge\: \textsc{DTR2: CAT}:n
                                        \:\wedge\: \textsc{DTR2: PHON}:talks)] \\
                                        \mathbf{1}
                                      \end{array}$}
                                  \end{bundle}
                                  }
                              \end{bundle}
                              }
                          \end{bundle}
                          }
                      \end{bundle}
                      }

                    \chunk[$\quad c$]
                    {
                      \begin{bundle}
                        {$\begin{array}{c}
                            \ast[\ldots] \:\&\: \texttt{sign}(Z_2)\\
                            \mathbf{f_7 \times f_5}
                          \end{array}$}
                        
                        \chunk[$\quad r$]
                        {\begin{bundle}
                            {$\begin{array}{c}
                                {\tt 7,} \; \ast[\ldots] \:\&\: \texttt{word}(Z_2)\\
                                \mathbf{f_7 \times f_5}
                              \end{array}$}
                            
                            \chunk[$\quad c$]
                            {\begin{bundle}
                                {$\begin{array}{c}
                                    \ast[\ldots] \:\&\: \texttt{word}(Z_2)\\
                                    \mathbf{f_5}
                                  \end{array}$}
                                
                                \chunk[$\quad r$]
                                {\begin{bundle}
                                    {$\begin{array}{c}
                                        {\tt 5,} \; \ast[\ldots] \\
                                        \:\&\: Z_2 = (word  \:\wedge\: \textsc{CAT}:n \\
                                        \:\wedge\: \textsc{PHON}:talks 
                                        \:\wedge\: \textsc{AGR}:pl)\\
                                        \mathbf{f_5}
                                      \end{array}$}
                                    
                                    \chunk[$\quad c$]
                                    {$\begin{array}{c}
                                        \dagger[X = (phrase \:\wedge\:  \textsc{CAT}:np \\
                                        \:\wedge\: \textsc{DTR1}:word  
                                        \:\wedge\: \textsc{DTR1: CAT}:n \\
                                        \:\wedge\: \textsc{DTR1: PHON}:Clinton \\
                                        \:\wedge\: \textsc{DTR2}:word 
                                        \: \wedge\:\textsc{DTR2: CAT}:n \\
                                        \:\wedge\: \textsc{DTR2: PHON}:talks \\
                                        \:\wedge\: \textsc{DTR2: AGR}:pl)] \\
                                        \mathbf{1}
                                      \end{array}
                                      $}
                                    
                                  \end{bundle}
                                  }
                              \end{bundle}
                              }
                          \end{bundle}
                          }
                      \end{bundle}
                      }
                  \end{bundle}
                  }
              \end{bundle}
              }
          \end{bundle}
          }
      \end{bundle}
      }
  \end{center}
  
  \caption{Quantitative derivation of $[Clinton_N \: talks_N]_{NP}$ }
  \label{QCLG2nd}
\end{figure}

The answer constraint $\phi$ of the first derivation is obtained by
constraint solving of the terminal constraints of the first proof
tree. We get 
\begin{center}
$\star[\ldots] \:\&\: \star[\ldots] \cs
X = (phrase
\:\wedge\:  \textsc{CAT}:s  \:\wedge\:
\textsc{DTR1}:word  
\:\wedge\: \textsc{DTR1: CAT}:n  \:\wedge\:
\textsc{DTR1: PHON}:Clinton 
\:\wedge\: \textsc{DTR1: AGR}:Y
\:\wedge\: \textsc{DTR1: AGR}:sg
 \:\wedge\: \textsc{DTR2}:word
\:\wedge\: \textsc{DTR2: CAT}:v 
 \:\wedge\: \textsc{DTR2: PHON}:talks 
\:\wedge\: \textsc{DTR2: AGR}:Y
\:\wedge\: \textsc{DTR2: AGR}:sg )
$
\end{center}
yielding the reading
 $[Clinton_N \; talks_V  ] _S$ with weight 
\[ \upsilon = f_6 \times f_1 \times
min \{ f_7 \times f_3, f_7 \times f_4 \}.
\]

The answer constraint $\psi$ of the second derivation is 
\begin{center}
$\star[\ldots] \:\&\: \dagger[\ldots] \cs
X = (phrase
\:\wedge\:  \textsc{CAT}:np \:\wedge\: 
\textsc{DTR1}:word  
\:\wedge\: \textsc{DTR1: CAT}:n  
\:\wedge\: \textsc{DTR1: PHON}:Clinton
\:\wedge\: \textsc{DTR1: AGR}:sg
 \:\wedge\: \textsc{DTR2}:word
\:\wedge\: \textsc{DTR2: CAT}:n
 \:\wedge\: \textsc{DTR2: PHON}:talks
\:\wedge\: \textsc{DTR2: AGR}:pl)
$
\end{center}
yielding the reading
 $[Clinton_N \; talks_N  ] _{NP}$ with weight 
\[
\tau = f_6 \times f_2 \times
min \{ f_7 \times f_3, f_7 \times f_5 \}.
\]

Suppose now that we have a subjective weight assignment for the
factors of the quantitative CLG of Fig. \ref{QCLGprogram} where $f_1 >
f_2$ and $f_4 > f_5$. That is, we prefer the rule
$S \rightarrow N \; V$
over the rule 
$NP \rightarrow N \: N$  to describe a phrase.
Furthermore, the terminal rule $V \rightarrow talks$, encoding the
word \emph{talks} as a verb, is preferred over the rule $N \rightarrow
talks$, encoding it as a noun. 
Clearly, we get a preference of the answer constraint $\phi$, encoding
the reading $[Clinton_N \; talks_V  ] _S$, over the answer constraint
$\psi$, encoding the reading $[Clinton_N \; talks_N  ] _{NP}$, with 
$\upsilon > \tau$.

\subsection{Alpha-Beta Searching in Quantitative CLGs}

As proposed by \citeN{Emden:86}, search strategies such as alpha-beta
pruning that are well known in game theory can be used quite directly to define
efficient search strategies for quantitative rule sets. The same technique can be
applied to the proof procedure of quantitative CLP. Alpha-beta pruning
is a technique to speed up the search in min/max trees without
loss of information. For our application, alpha-beta pruning can be
used to efficiently search a min/max derivation tree for the maximal
valued proof tree. The fact that no information is lost in alpha-beta
pruning means in our context that the maximal valued proof tree is
guaranteed to be found. Furthermore, in general, the amount of search
needed to find the best proof for a goal, i.e. the maximal valued
proof tree for a goal from a program, will be reduced remarkably by
controlling the search by the alpha-beta algorithm.

The central concepts of alpha-beta pruning can be summarized as
follows (see \citeN{Nilsson:82}).

Usually some form of depth-first search is employed in alpha-beta
pruning. The search procedure associates with
each max-node (resp.\ min-node) a dynamic alpha-value
(resp.\ beta-value). These values are based on the static values of
terminal nodes and will be backed-up in subsequent search by lookahead
in the tree.  

The search procedure starts with a maximum depth execution of
depth-first search, initializing the alpha and beta
values of the first subtree. During search, alpha and beta values are
computed as follows:
\begin{itemize}
\item The \textbf{alpha value} of a max-node is the maximum of the current
  values of its successors.
\item The \textbf{beta value} of a min-node is the minimum of the current
  values of its successors, multiplied by the factor
  of the clause labeling the min-node.
\end{itemize}
The rules for discontinuing the search are as follows:
\begin{itemize}
\item \textbf{Alpha-cutoff}: Search can be discontinued below any min-node having a beta
  value less than or equal to the alpha value of any of its max-node
  ancestors. The final backed-up value of this min-node can then be set to its beta
value. 
\item \textbf{Beta-cutoff}: Search can be discontinued below any max-node with
  the product of its alpha value and the factor of the rule labeling its min-node
  ancestor being greater than or equal to the beta value of this
  min-node ancestor for all min-node ancestors. The
  final backed-up value of this max-node can then be set to its alpha value.
\end{itemize}
The procedure terminates when all of the successors of the root node
have been given a final backed-up value. The maximal valued proof tree
is then the one taking as single successor of each of its max-nodes
the successor with the maximal final backed-up value. This proof tree
is found efficiently if the original min/max tree can be pruned
by the alpha-beta procedure to a tree consisting of a relatively small number of nodes.

Let us illustrate these concepts with a simple example. A sample artificial
program is given in Fig. \ref{alphabetaprogram}.

\begin{figure}[htbp]
\begin{center}
\begin{tabular}{l}
\texttt{1} $\texttt{p}(X)\leftarrow_{.7} \texttt{r}(X) \:\&\: \texttt{s}(X).$ \\
\texttt{2} $\texttt{r}(X) \leftarrow_{.8} X=a.$ \\
\texttt{3} $\texttt{s}(X) \leftarrow_{.9} X=a.$ \\
\texttt{4} $\texttt{s}(X) \leftarrow_{.2} \texttt{r}(X).$ \\
\texttt{5} $\texttt{p}(X) \leftarrow_{.7} \texttt{t}(X) \:\&\:
\texttt{r}(X) \:\&\: \texttt{s}(X).$\\
\texttt{6} $\texttt{t}(X) \leftarrow_{.1} X=a.$ \\
\end{tabular}
\end{center}
\caption{Quantitative constraint logic program}
\label{alphabetaprogram}
\end{figure}

\begin{figure}[htbp]
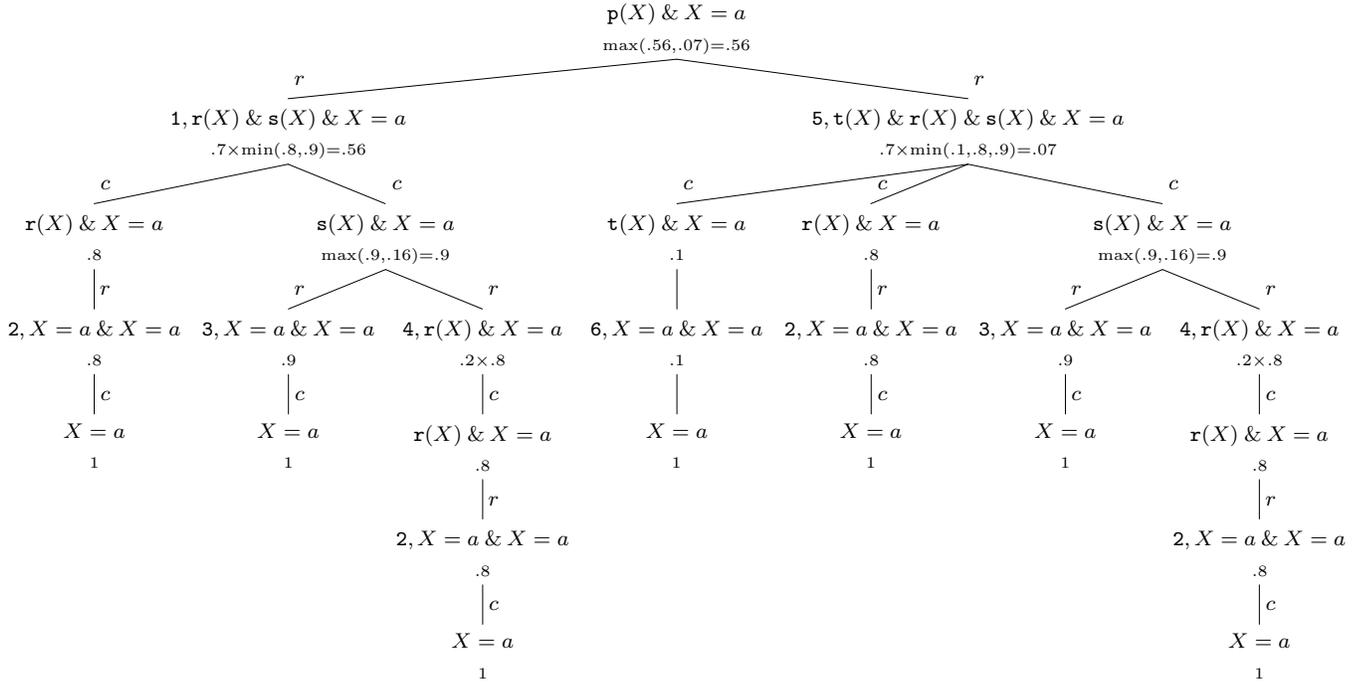


{\scriptsize

\setlength{\GapWidth}{40pt}
\begin{bundle}{$\begin{array}{c}
      \texttt{p}(X) \:\&\: X=a \\
      {\scriptstyle \max(.56,.07)=.56}
    \end{array}$}
  \chunk[$\quad r$]{
    \begin{bundle}{$\begin{array}{c}
          \texttt{1}, \texttt{r}(X) \:\&\: \texttt{s}(X) \:\&\: X=a \\
          {\scriptstyle .7 \times \min (.8,.9) = .56}
        \end{array}$}
      \chunk[$\quad c$]{
        \begin{bundle}{$\begin{array}{c}
              \texttt{r}(X) \:\&\: X=a \\
              {\scriptstyle .8}
            \end{array}$}
          \chunk[$\quad r$]{
            \begin{bundle}{$\begin{array}{c} 
                  \texttt{2}, X=a \:\&\: X=a \\ 
                  {\scriptstyle .8}
                \end{array}$}
              \chunk[$\quad c$]{$\begin{array}{c}
                  X=a \\
                  {\scriptstyle 1}
                \end{array}$}
            \end{bundle}}
        \end{bundle}}
      \chunk[$\quad c$]{
        \begin{bundle}{$\begin{array}{c}
              \texttt{s}(X) \:\&\: X=a \\
              {\scriptstyle \max(.9,.16)=.9}
            \end{array}$}
          \chunk[$\quad r$]{
            \begin{bundle}{$\begin{array}{c} 
                  \texttt{3}, X=a \:\&\: X=a \\ 
                  {\scriptstyle .9}
                \end{array}$}
              \chunk[$\quad c$] {$\begin{array}{c} 
                  X=a \\ 
                  {\scriptstyle 1}
                \end{array}$}
            \end{bundle}}
          \chunk[$\quad r$]{
            \begin{bundle}{$\begin{array}{c} 
                  \texttt{4}, \texttt{r}(X)\:\&\: X=a \\ 
                  {\scriptstyle .2 \times .8}
                \end{array}$}
              \chunk[$\quad c$]{
                \begin{bundle}{$\begin{array}{c} 
                      \texttt{r}(X)\:\&\: X=a \\ 
                      {\scriptstyle .8}
                    \end{array}$}
                  \chunk[$\quad r$]{
                    \begin{bundle}{$\begin{array}{c} 
                          \texttt{2}, X=a \:\&\: X=a \\ 
                          {\scriptstyle .8}
                        \end{array}$}
                      \chunk[$\quad c$]{$\begin{array}{c} 
                          X=a \\ 
                          {\scriptstyle 1}
                        \end{array}$}
                    \end{bundle}}
                \end{bundle}}
            \end{bundle}}
        \end{bundle}}
    \end{bundle}}
  \chunk[$\quad r$]{
    \begin{bundle}{$\begin{array}{c} 
          \texttt{5}, \texttt{t}(X) \:\&\:
          \texttt{r}(X) \:\&\: \texttt{s}(X) \:\&\: X=a \\
          {\scriptstyle .7 \times \min
            (.1,.8,.9) = .07}
        \end{array}$}
      \chunk[$\quad c$]{
        \begin{bundle}{$\begin{array}{c} 
              \texttt{t}(X)\:\&\: X=a \\ 
              {\scriptstyle .1}
            \end{array}$}
          \chunk{
            \begin{bundle}{$\begin{array}{c}  
                  \texttt{6}, X=a \:\&\: X=a \\ 
                  {\scriptstyle .1}
                \end{array}$}
              \chunk{$\begin{array}{c} 
                  X=a \\ 
                  {\scriptstyle 1}
                \end{array}$}
            \end{bundle}}
        \end{bundle}}
      \chunk[$\quad c$]{
        \begin{bundle}{$\begin{array}{c}
              \texttt{r}(X) \:\&\: X=a \\
              {\scriptstyle .8}
            \end{array}$}
          \chunk[$\quad r$]{
            \begin{bundle}{$\begin{array}{c} 
                  \texttt{2}, X=a \:\&\: X=a \\ 
                  {\scriptstyle .8}
                \end{array}$}
              \chunk[$\quad c$]{$\begin{array}{c}
                  X=a \\
                  {\scriptstyle 1}
                \end{array}$}
            \end{bundle}}
        \end{bundle}}
      \chunk[$\quad c$]{
        \begin{bundle}{$\begin{array}{c}
              \texttt{s}(X) \:\&\: X=a \\
              {\scriptstyle \max(.9,.16)=.9}
            \end{array}$}
          \chunk[$\quad r$]{
            \begin{bundle}{$\begin{array}{c} 
                  \texttt{3}, X=a \:\&\: X=a \\ 
                  {\scriptstyle .9}
                \end{array}$}
              \chunk[$\quad c$] {$\begin{array}{c} 
                  X=a \\ 
                  {\scriptstyle 1}
                \end{array}$}
            \end{bundle}}
          \chunk[$\quad r$]{
            \begin{bundle}{$\begin{array}{c} 
                  \texttt{4}, \texttt{r}(X)\:\&\: X=a \\ 
                  {\scriptstyle .2 \times .8}
                \end{array}$}
              \chunk[$\quad c$]{
                \begin{bundle}{$\begin{array}{c} 
                      \texttt{r}(X)\:\&\: X=a \\ 
                      {\scriptstyle .8}
                    \end{array}$}
                  \chunk[$\quad r$]{
                    \begin{bundle}{$\begin{array}{c} 
                          \texttt{2}, X=a \:\&\: X=a \\ 
                          {\scriptstyle .8}
                        \end{array}$}
                      \chunk[$\quad c$]{$\begin{array}{c} 
                          X=a \\ 
                          {\scriptstyle 1}
                        \end{array}$}
                    \end{bundle}}
                \end{bundle}}
            \end{bundle}}
        \end{bundle}}
    \end{bundle}}
\end{bundle}

}

\caption{Complete search of a quantitative derivation tree}
\label{completesearch}
\end{figure}

The complete min/max derivation tree for the query $\texttt{p}(X)
\:\&\: X=a$ to the program of Fig. \ref{alphabetaprogram} is given in
Fig. \ref{completesearch}.

\begin{figure}[htbp]
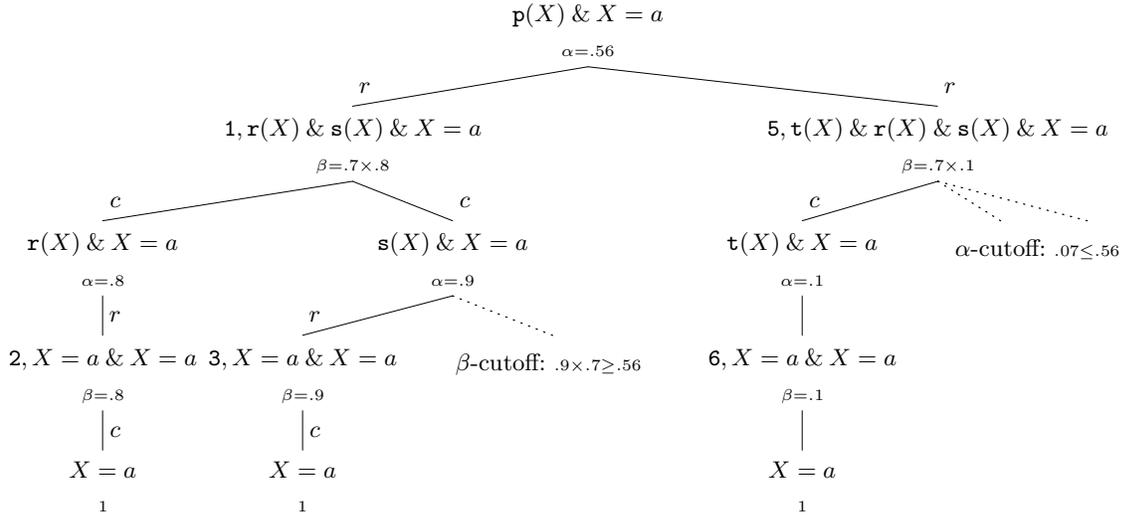

\begin{center}
{\footnotesize

\setlength{\GapWidth}{40pt}
\begin{bundle}{$\begin{array}{c}
      \texttt{p}(X) \:\&\: X=a \\
      {\scriptstyle \alpha=.56}
    \end{array}$}
  \chunk[$\quad r$]{
    \begin{bundle}{$\begin{array}{c}
          \texttt{1}, \texttt{r}(X) \:\&\: \texttt{s}(X) \:\&\: X=a \\
          {\scriptstyle \beta  = .7 \times .8}
        \end{array}$}
      \chunk[$\quad c$]{
        \begin{bundle}{$\begin{array}{c}
              \texttt{r}(X) \:\&\: X=a \\
              {\scriptstyle \alpha = .8}
            \end{array}$}
          \chunk[$\quad r$]{
            \begin{bundle}{$\begin{array}{c} 
                  \texttt{2}, X=a \:\&\: X=a \\ 
                  {\scriptstyle \beta = .8}
                \end{array}$}
              \chunk[$\quad c$]{$\begin{array}{c}
                  X=a \\
                  {\scriptstyle 1}
                \end{array}$}
            \end{bundle}}
        \end{bundle}}
      \chunk[$\quad c$]{
        \begin{bundle}{$\begin{array}{c}
              \texttt{s}(X) \:\&\: X=a \\
              {\scriptstyle \alpha =.9}
            \end{array}$}
          \drawwith{\drawwith{\dottedline{3}}\drawline}
          \chunk[$\quad r$]{
            \begin{bundle}{$\begin{array}{c} 
                  \texttt{3}, X=a \:\&\: X=a \\ 
                  {\scriptstyle \beta = .9}
                \end{array}$}
              \chunk[$\quad c$] {$\begin{array}{c} 
                  X=a \\ 
                  {\scriptstyle 1}
                \end{array}$}
            \end{bundle}}
          \chunk{$\beta$-cutoff: ${\scriptstyle .9 \times .7 \geq .56
              } $ }
        \end{bundle}}
    \end{bundle}}
  \chunk[$\quad r$]{
    \begin{bundle}{$\begin{array}{c} 
          \texttt{5}, \texttt{t}(X) \:\&\:
          \texttt{r}(X) \:\&\: \texttt{s}(X) \:\&\: X=a\\
          {\scriptstyle \beta = .7 \times .1}
        \end{array}$}
 \drawwith{\drawwith{\dottedline{3}}\drawline}
      \chunk[$\quad c$]{
        \begin{bundle}{$\begin{array}{c} 
              \texttt{t}(X)\:\&\: X=a \\ 
              {\scriptstyle \alpha = .1}
            \end{array}$}
          \chunk{
            \begin{bundle}{$\begin{array}{c}  
                  \texttt{6}, X=a \:\&\: X=a \\ 
                  {\scriptstyle \beta =.1}
                \end{array}$}
              \chunk{$\begin{array}{c} 
                  X=a \\ 
                  {\scriptstyle 1}
                \end{array}$}
            \end{bundle}}
        \end{bundle}}
      \setlength{\GapWidth}{3pt}
      \chunk{$\alpha$-cutoff:}
      \chunk{$\scriptstyle .07  \leq .56$}
\end{bundle}}
\end{bundle}

}
\end{center}
\caption{Alpha-beta search of quantitative derivation tree}
\label{alphabetasearch}
\end{figure}

The concept of alpha-beta pruning can be illustrated with this example
as follows (see Fig. \ref{alphabetasearch}). The alpha value $\alpha =
.9$ of the max-node $\texttt{s}(X) \:\&\: X=a$ times the factor $.7$
of the min-node ancestor is greater than the beta value $\beta = .56$
of this min-node. Since we know that this alpha value cannot be
decreased by further evaluation of the subtrees of this max-node, and
since we are interested in the minimum of the values of the successors
of this min-node, we can cut off the search below this max-node
without a risk of losing information relevant to the final maximal
valued proof tree. This cutoff is indicated by the dotted line below
this max-node in Fig. \ref{alphabetasearch}. 
In a similar way, search below the min-node 
$\texttt{5}, \texttt{t}(X) \:\&\: \texttt{r}(X) \:\&\: \texttt{s}(X)
\:\&\: X=a$ 
can be discontinued because the non-decreasing beta value $\beta = .07$ 
of this node is already smaller than the alpha value $\alpha = .56$  of its max-node
ancestor. The pruning of the two subtrees of this min-node again is
indicated by dotted lines in Fig. \ref{alphabetasearch}. Again, there
is no risk of information loss in this  pruning step. 

Clearly, in each application of the alpha-beta procedure, the number
of nodes to be generated and evaluated is minimal when the number of
cutoffs is maximal.  The best case occurs when the maximal valued proof
tree is reached first in the depth-first search. In the worst case, no
gain in search efficiency is obtained at all, i.e., all nodes of the
min/max tree have to be generated.
In either case, the maximal valued proof tree is guaranteed to be left unpruned.
Risking loss of relevant information, the alpha-beta procedure can be
improved by setting an initial alpha value for the root note which
allows to cut off search branches with root value lower than this initial
value.
For a thorough analysis of the properties of alpha-beta pruning the
reader is referred to \citeN{Knuth:75}.

Furthermore, it should be noted that a strict application of
alpha-beta pruning is possible only for quantitative CLP based on
min/max trees. Suppose for example that the minimum operator is replaced by a
product operator throughout the declarative as well as operational semantics of
quantitative CLP. This replacement could be motivated by the aim to consider the
contribution of all instead only one antecedent atom to the weight of
the consequent. To efficiently search for the maximal valued proof
tree in such a setting, a version of alpha-beta pruning employing
only alpha-cutoffs has to be used. In this setting, additional beta-cutoffs can improve the
search efficiency for finding a \emph{good} proof tree, but possibly
cut off parts of the \emph{best} proof tree, i.e., here attention has
to be paid to the risk of losing information relevant to the maximal
valued proof tree.

\section{Summary and Discussion}
\markright{3.7 Summary and Discussion}

In this chapter we presented a formal framework for quantitative
CLP. In this framework CLP clauses were attached by arbitrary
numerical weights. Such weighted clauses were interpreted in a
model-theory based on concepts of fuzzy set algebra. The quantitative
system was shown to be sound and complete with respect to a fixpoint
semantics based on minimal models in this model-theory. We illustrated
the concepts of quantitative CLP by a simple quantitative
feature-based CLG. Furthermore, we showed how to adapt the search
algorithm of alpha-beta pruning to searching efficiently for the
highest weighted proof tree in a quantitative CLP system.

The advantage of quantitative CLP clearly is the freedom it offers to the
grammar writer or implementer to specify arbitrary weights in a formally
clear and efficient programming framework. Such
weights could be specified, e.g., as subjective preference values 
(\citeNP{Erbach:93b}, \citeNP{Kim:94}, \citeNP{Dale:92}), subjective
values expressing graded grammaticality \cite{Erbach:93a},
or subjective probabilities \cite{Erbach:95}. Calculation with
arbitrary such weights can be interpreted in a unique well-defined formal
framework. Furthermore, generalizations of the formal system to particular
applications which require particular calculation schemes are easily
obtainable. For example, if we want to model
probabilistic context-free grammars \cite{Booth:73} in quantatitive
CLP, we simply must attach subjective probability measures to a
context-free program according to the conditions of \citeN{Booth:73},
and replace the minimum operator by a product operator in all relevant
definitions of the declarative and operational semantics of
quantitative CLP. Unfortunately, such changes in the weight calculation 
model prevent a direct application of alpha-beta pruning for
efficient disambiguation. For example, in the case of probabilistic
context-free CLP, only a restricted version of alpha-beta pruning
using exclusively alpha-cutoffs is applicable. 
Alternatively, one could use a form of
best-first pruning for the search task, i.e., a search rule which
selects the highest weighted clause at each derivation step. Clearly,
this approach does not guarantee that the highest weighted proof tree
is found, but offers only an approximate heuristic search procedure.

However, regardless of the specific choice of weights, a proper
specification of a multitude of weights can be very complex and is
always user-dependent. In several applications, one would like to
trade in the flexibility of subjective weight assignment for automatic
and reusable methods for estimating weights from empirical data. One
solution to this problem is to use automatic methods for statistical
inference to induce values of probabilistic parameters from empirical data. 

In the next chapter, we present an framework of probabilistic CLP
which addresses the problem of finding a proper probability
distribution over the set of proof trees of a constraint logic program
and of using statistical estimation methods to infer parameters
from empirical data. Clearly, even if it would be possible to specify
a model-theoretic semantics for such a system, it is superfluous to do so in
the context of automatic statistical inference. Rather, the interest
is here in the stochastic semantics of CLP provided by the
probabilistic and statistical methods used.

\chapter{Probabilistic CLP: Probabilistic
Modeling and Statistical Inference from Incomplete Data} 
\label{P}

In this chapter we present a probabilistic model for CLP and a novel
method for statistical inference of the parameter values of such a
model from incomplete training data. We show mononoticity
and convergence of the new algorithm to the desired maximum likelihood
estimates. Furthermore, we show the usefulness of the
statistical approach by a small-scale experiment on
estimating feature-based CLGs. We present a novel algorithm to infer the
properties of such parametric probability models from incomplete data
and discuss different approaches for approximate computation for the
inference task. Moreover, we discuss the possibilities of using the
structure of the probabilistic model to guide the search in finding
the most probable proof tree in probabilistic CLP and present as
heuristic search method for this task.

This chapter is based upon work previously published in
\citeN{Riezler:97}, \citeN{RiezlerKon:98}, \citeN{Riezler:98}, and \citeN{Johnson:99}.

\section{Introduction and Overview}
\markboth{Chapter 4. Probabilistic CLP}{4.1 Introduction and Overview}

In the previous chapter we presented a formal semantics for a system of
quantitative CLP. This formal semantics and the connected quantitative
inference system were crucially based upon open parameters for
subjective weights. 
Most approaches to probabilistic logic programming interpret such weights as subjective probabilities, and concentrate on inference systems
and formal semantics for programming systems with
user-defined probabilities attached to the formulae of the language. The aim of such
approaches is the development of sound and complete logic programming
systems where the handling of weights is restricted to accord to the
laws of probability theory. That is, these approaches aim to connect
logical inference with probabilistic inference.

In this chapter, we present a completely different approach to
probabilistic CLP. In this approach, subjective assignment of probabilities is
replaced by automatic and reusable methods for estimating
empirical probabilities from data.
The central aims of this approach are the specification of a
probability distribution over the proof trees given by a program, and the use of statistical methods to
infer the values of the probabilistic parameters from empirical data. 
That is, in this setting, the weight of a CLP proof tree is determined
directly by a probability distribution over proof trees rather than by
quantitative calculation scheme referring to weighted clauses.
The parameters of the probability distribution are determined by
statistical inference from empirical data rather than by an assignment
of subjective weights to clauses. Furthermore, the specific properties
of the parametric probability model can be inferred by statistical
methods. 
That means, in this chapter we do not only turn from quantitative to
probabilistic inference but, what is more, to statistical inference.
In such a setting, the connection of probability theory, semantic
fixpoint theory and logical inference theory is not of interest since
the specification of probabilistic parameters is done by automatic
statistical methods and not manipulable by the user. Rather, we are
interested in the stochastic semantics defined by the methods of
probabilistic modeling and statistical inference.

The statistical problem we consider here is the problem of
statistical parameter estimation. We assume that the
statistical properties of a given sample of observations $O=O_1,
\ldots, O_n$ can be described by a parametric family of probability
distributions. That is, the probability distribution that generated
the data is assumed to be completely known except for the values of a vector
$\theta$ of parameters.
We then ask how the unknown value of
$\theta$ can be estimated from the observation sequence 
$O$, i.e., a statistical inference is made about the values of the
parameters defining that family.
Recent interest in statistical approaches to NLP
can be attributed to the fact that solutions to such statistical
problems can lead quite directly to effective, but
conceptually simple and mathematically clear solutions to various
problems in NLP. In the context of structural ambiguity resolution in
NLP systems, this connection is as follows: Given a probabilistic
grammar depending on parameter vector $\theta$ and given a training
corpus $O$, a solution $\hat\theta$ to the parameter estimation
problem will adapt the model parameters to best account for the input
corpus. This tuning of the grammar to a particular natural language
corpus is a necessary prerequisite for probabilistic
disambiguation. That is, when the plausibility of a parse is
connected with its probability, the assumption that the correct parse of a
sentence is its most probable parse can be made with some
justification if the underlying probabilistic grammar is based on
parameter values  $\hat\theta$ estimated from large data sets of
natural language. 

The aim of this chapter is to solve open problems in statistical 
inference and probabilistic modeling of constraint-based grammars. Following
\citeN{Abney:96}, we choose the parametric family of log-linear probability
distributions to model such grammars. The great advantage of
log-linear models is their generality and flexibility. Log-linear
models allow to describe arbitrary context dependencies in the data by choosing a few
salient properties of the data as the defining properties of the
model. In contrast to most approaches to probabilistic grammars, with
log-linear models we are not restricted to build our models on
production rules or other configurational properties of the data.
Rather, we have the virtue of employing essentially arbitrary
properties in our models. For example, heuristics on preferences of
grammatical functions or on attachment preferences as used in \citeN{Srinivas:95}, or 
the preferences in lexical relations as used in 
\citeN{Alshawi:94} can be integrated into a log-linear model very
easily. However, the step from simple rule-based probability models to
general log-linear models requires also a more
general and more complex estimation algorithm. The estimation
algorithm for log-linear models used by \citeN{Abney:96} is the
iterative scaling method of \citeN{DDL:97}. This algorithm allows to
recast the optimization of weights of preference functions
as done by \citeN{Srinivas:95} or \citeN{Alshawi:94} as
estimation of parameters associated with the properties of a
log-linear model. However, there is a drawback: In contrast to
rule-based models where efficient estimation algorithms from
incomplete, i.e., unannotated data exist, the iterative scaling
estimation method of \citeN{DDL:97} applies only to complete, i.e.,
fully annotated training data. Unfortunately, the need
to rely on large samples of complete data is impractical. For parsing
applications, complete data means several person-years of
hand-annotating large corpora with specialized grammatical analyses.
This task is always labor-intensive, error-prone, and restricted to a
specific grammar framework, a specific language, and a specific
language domain. 

Thus, the first open problem to solve is fo find automatic and reusable techniques for
parameter estimation of probabilistic constraint-based grammars from
incomplete data. We will present a general estimation algorithm for
log-linear models from incomplete data which can be seen as an
extension of the iterative scaling method of \citeN{DDL:97}. We prove
monotonicity and convergence of the new algorithm to (local) maxima of
the incomplete-data log-likelihood function, and show how automatic
property selection can be done from incomplete data.

A further open problem is the empirical evaluation of the performance
of probabilistic constraint-based grammars in terms of finding
human-determined correct parses. We present an
experiment with a log-linear model employing a few hundred general
properties encoding grammatical functions, attachment preferences,
branching behaviour, parallelism, and other general properties of
constraint-based parses. The experiment was conducted on a small scale
but clearly shows the usefulness of general properties in order to get
good results in a linguistic evaluation.

Clearly, for larger scales, problems arise concerning the tractability
of the estimation formulae. We discuss the applicability of several
approximation methods to our problem of statistical inference from incomplete data,
including Newton's method, Monte Carlo methods, or methods for
approximating expectations via pseudo-likelihood approaches.

A further open problem is the efficient search for most probable
parses, i.e, best-parse search,  in parsing systems based on
probabilistic constraint-based grammars. Instead of listing all
possible parses and selecting the most probable one, one would like to
use the structure of the probabilistic model to guide the search for the most probable
analysis. Most popular approaches use the search technique of the
Viterbi algorithm (\citeN{Viterbi:67}, \citeN{Forney:73})  to solve
this problem, but there is as yet no solution for probabilistic constraint-based
grammars. We show that standard methods for best-parse search are only of
limited use for probabilistic models involving
context-dependencies, and make the move to approximate heuristic methods.

To summarize, our approach satisfies the following requirements. It

\begin{itemize}
\item is generally applicable to probability models involving context-dependencies,
  and especially to a probabilistic model for CLP over arbitrary constraint languages,
\item provides automatic and reusable techniques for statistical
  inference from incomplete data for such probability models, and
\item is accompanied with search techniques for finding most probable
  analyses in probabilistic CLP.
\end{itemize}

This chapter is organized as follows.
Is Sect. \ref{PreviousPCLP} we discuss related previous approaches to
statistical inference for probabilistic constraint-based grammars.

In Sect. \ref{EM} we introduce the basic
concepts of maximum likelihood estimation from incomplete data via the
EM algorithm. 

Sect. \ref{EMExample} discusses the problem of applying a
popular instance of this algorithm, namely Baum's maximization
technique for stochastic context-free models, to parameter estimation
for probabilistic CLP. 

Sect. \ref{LLM} and \ref{SI} present in detail a solution to this
problem by introducing a log-linear probability model for CLP coupled
with an incomplete-data inference algorithm for such
models. This section includes a detailed proof of monotonicity and
convergence of the inference algorithm.

Sect. \ref{Experiment} presents an empirical evaluation of the
applicability of general log-linear
distributions to probabilistic constraint-based grammars in a small-scale experiment on estimating a log-linear
model on constraint-based parses.

Sect. \ref{Approximation} discusses computation issues such as the
use of Monte Carlo methods, Newton's numerical method, and other
approximation techniques in the context of this inference process. 

Sect. \ref{PCLG} discusses the applicability of standard parsing and
search methods to context-dependent constraint-based models, and
presents a heuristic method for searching for best parses in CLGs.

\section{Previous Work}
\markright{4.2 Previous Work}
\label{PreviousPCLP}

An approach to define estimators for probabilistic constraint-based
grammars which has been applied to nearly all constraint-based
formalisms is a renormalized extension of the estimator for stochastic
regular \cite{Baum:70} or context-free grammars \cite{Baker:79}
to constraint-based models. Examples for this approach are, e.g., 
stochastic unification-based grammars
\cite{Briscoe:92,BriscoeCarroll:93}, stochastic constraint logic
programming \cite{Eisele:94}, stochastic head-driven phrase structure grammar
\cite{Brew:95},  stochastic logic programming \cite{Miyata:96},
stochastic categorial grammars \cite{Osborne:97} or data-oriented
approaches to lexical-functional grammar \cite{Bod:98}. 
Since the estimation technique for context-free models is based on the
assumption of mutual independence of the model's derivation steps, but
context-dependent constraints on derivations are inherent to
constraint-based grammars, a loss in probability mass due to failure
derivations is caused in these approaches. However, the
necessary renormalization of the probability distribution on
derivations with respect to consistent derivations causes a general
deviance of the resulting estimates from the desired maximum
likelihood estimates. This was shown firstly by \citeN{Abney:96} for
estimation of constraint-based models from complete data. We will make
a similar argument for incomplete data in the following.
Optimization-theoretically these approaches can be described as
maximization procedures for pseudo-likelihood
functions for context-free models where the probability
distribution on context-free derivations is restricted to consistent
derivations in the constraint-based sense. Maximum pseudo-likelihood
estimators for context-free models certainly are sensible, e.g., if
the aim is to constrain an inherently context-free language to include
only linguistically plausible derivations as is done by introducing bracketing
constraints on context-free derivations by \citeN{Pereira:92}.
However, it is questionable if is the best way to
model constraint-based grammars probabilistically by context-free
models which respect constraints only indirectly to discard derivations. 
The move to log-linear models as is done in our approach clearly has
several advantages. Since there is linguistically no reason
to base probabilistic grammars on rule-properties, we can now exploit the
flexibility of log-linear distributions and model the
context-dependencies in the data directly. Furthermore, since the new
family of parametric probability models requires new estimation
techniques, we can again take consistent maximum likelihood estimators
as the optimization procedures of our choice.

Other approaches to probabilistic constraint-based models have been
presented which define custom-built statistical inference procedures
for specialized parsing models including a limited amount of
context-dependency. For example, the model presented by
\citeN{Goodman:98} conditions on a finite set of categorial features
beyond the nonterminal of each node which makes it possible to
explicitly unfold the dependencies in the parsing model. This allows
for the use of standard dynamic programming techniques for
computation. In the approaches of \citeN{Magerman:94}
and \citeN{Rat:98} general statistical inference methods, namely
decision trees and maximum-entropy methods, are used to infer weights
associated to the actions of specialized parsing models including
limited context-dependency. However, it
is difficult to generalize these models to arbitrary log-linear models
on constraint-based grammars, concerning both the choice of properties
and the issue of efficient computation. Clearly, a careful choice of
properties and dependencies makes it possible to tune specialized models to maximum accuracy and
efficiency, which does not hold for the general case\footnote{For
  example, as noted by \citeN{Goodman:98}, the computational complexity of his dynamic
  programming algorithm for probabilistic feature-grammars is
  exponential in the general case.} The aim of our approach is to
address problems concerning estimation, property design, or
approximation methods for general log-linear models and show these
general ideas to be applicable in practice.

\section{Maximum Likelihood Estimation from Incomplete Data via the EM Algorithm}
\markright{4.3 MLE from Incomplete Data via the EM Algorithm}
\label{EM}

A constant companion during the course of this chapter will be the
statistical estimation technique of the Expectation-Maximization (EM)
algorithm. The fact that both Baum's estimation technique, which is
shown not to be applicable to probabilistic CLP in Sect. \ref{Baum},
and the incomplete-data estimation algorithm for log-linear models we
present in Sects. \ref{LLM}-\ref{Approximation}, can be seen as instances of
the EM algorithm, justifies a closer look at this estimation scheme.

\subsection{General Theory of the EM Algorithm}
\label{GeneralTheory}

The EM algorithm has been introduced by \citeN{Dempster:77},
although central parts of the general theory can be found earlier
in special applications, e.g., in \citeN{Baum:70}. Various applications
and extensions of the algorithm are discussed in \citeN{Little:87}
and, more recently, in \citeN{McLachlan:97}.

The EM method is a technique for maximum likelihood estimation (MLE) from
incomplete data. For a parametric family of probability distributions
depending on parameter vector $\theta$ and a given sample of training
data from this parametric family, MLE
defines the estimate $\hat\theta$  of $\theta$ as a value of $\theta$
which maximizes the likelihood of the training sample. MLE from observed
complete data is particularily easy for many statistical problems,
thanks to the nice form of the complete-data (log-)likelihood
function. The problem the EM algorithm especially addresses is the
case where the observed data are incomplete. That is, we observe only
a function of complete data, which themselves are unobserved. Because of
this indirect, hidden character of the complete data, MLE from incomplete data is
difficult.

In the following, an incomplete-data estimation setting is assumed to
consist of

\begin{itemize}
\item a sample space $\mathcal{Y}$ of observed, incomplete data,
\item a sample space $\mathcal{X}$ of unobserved, complete data,
\item a many-to-one function $Y:\mathcal{X} \rightarrow \mathcal{Y}$ s.t. $Y(x) =y$ is the unique observation corresponding to the complete datum
$x$, and its inverse $X:\mathcal{Y} \rightarrow 2^\mathcal{X}$
s.t. $X(y) = \{ x | Y(x) = y \}$ is the countably infinite set of complete data
corresponding to the observation $y$,
\item a complete-data specification $p_\theta(x)$ with parameters
  $\theta \in \Theta$,
\item an incomplete data specification $g_\theta(y)$ which is related
  to the complete-data specification by marginalization as
\[ g_\theta(y) = \sum_{x \in X(y)} p_\theta(x) .\]
\end{itemize}
Let $y_1, y_2, \ldots, y_N$ be a random sample from $\mathcal{Y}$,
i.e., values of independently and identically distributed (i.i.d.)
random variables on  $\mathcal{Y}$. 
Let $p[f] = \sum_{\omega \in \Omega} p(\omega) f(\omega)$
denote the expectation of a function $f:\Omega \rightarrow \Reals$ with respect
to a probability distribution $p$ on $\Omega$. If $f$ is a
multivariable function $f(\omega', \omega)$, then the expectation of
$f$ with respect to $p(\omega)$ is written $p[f(\omega', \cdot)]$. 
Furthermore, let the empirical probability $\tilde p(y)$ of an
incomplete data type be defined as 
\(
\tilde p : \mathcal{Y} \rightarrow \Reals \textrm{ s.t. } 
\tilde p(y) = N^{-1} \sum_{i = 1}^N \delta_{y_i, y}
\)
where the Kronecker delta 
$\delta_{y_i,y} = 
\left\{ \begin{array}{ll}
1 & \textrm{ if } y_i = y,\\
0 & \textrm{ otherwise. }
\end{array}\right.$ \\
Then the incomplete-data log-likelihood $L$ is defined for a random
sample from $\mathcal{Y}$ as a function of $\theta$ as

\[
L(\theta) = \ln \prod_{y \in \mathcal{Y}} g_\theta (y) ^{\tilde p
  (y)} 
 = \sum_{y \in \mathcal{Y}} \tilde p(y) \ln g_\theta (y) 
 = \tilde p [ \ln g_\theta ].
\]
The EM algorithm is directed at finding a value $\hat\theta$ of
$\theta \in \Theta$
that maximizes $L$ as a function of $\theta$ for a given random sample
from $\mathcal{Y}$, i.e.,  

\[ \hat\theta = \underset{\theta \in \Theta}{\arg\max\;} L(\theta) 
\textrm{ where } L(\theta) =  \tilde p [ \ln g_\theta ] = \tilde p
[ \ln \sum_{x \in X(\cdot)} p_\theta (x) ].
\] 
The summation inside this logarithm can make MLE from incomplete data
difficult even when complete-data MLE is easy.

The old idea formalized in the EM algorithm can be stated informally as follows:
1. Replace unobserved data values by expected values, 2. perform MLE
from the expected complete data, 3. recompute the unobserved data expectations
using the new parameter estimates, 4. reestimate parameters using the
new expectations,  5. iterate until convergence of the likelihood
function.

The trick of the EM algorithm thus is to solve the incomplete-data
estimation problem for $\ln g_\theta(y)$ indirectly by proceeding iteratively
in terms of complete-data estimation for $\ln p_\theta(x)$. 
Since the $x$ are not observable, $\ln p_\theta(x)$ is replaced by its
conditional expectation given the observed data $y$ and the current
fit of the parameter values $\theta^{(t)}$. That is, complete-data
log-likelihood values are constructed from a conditional expectation given the
observed data of the incomplete data problem and the current value of
the unknown parameters (E-step). From the thus manufactured
complete-data, maximization is simpler and often exists in
closed form (M-step). Starting from suitable
initial parameter values, the E- and M-steps are iterated until
convergence of the incomplete-data log-likelihood $L$.

More formally, let $k_\theta(x|y) = p_\theta(x) / g_\theta(y)$ 
be the conditional probability of $x$ given $y$ and $\theta$, then 
\begin{eqnarray*}
L(\theta') & = & \sum_{y \in \mathcal{Y}} \tilde p (y) \ln g_{\theta'}
(y) \\
& = & \sum_{y \in \mathcal{Y}} \tilde p (y) k_\theta [  \ln g_{\theta'}
(y)] \\
& = & \sum_{y \in \mathcal{Y}} \tilde p (y) \sum_{x \in X(y)}
k_\theta (x|y) \ln \frac{p_{\theta'}(y)}{k_{\theta'}(x|y)} \\
& = & \sum_{y \in \mathcal{Y}} \tilde p (y) \sum_{x \in X(y)}
k_\theta (x|y) \ln p_{\theta'}(y) 
- \sum_{y \in \mathcal{Y}} \tilde p (y) \sum_{x \in X(y)}
k_\theta (x|y) \ln k_{\theta'}(x|y) \\
& = & \tilde p [ k_\theta [ \ln p_{\theta'} ] ] - \tilde p [
k_\theta [ \ln k_{\theta'} ] ] \\
& = & Q(\theta' ; \theta) - H(\theta' ; \theta).
\end{eqnarray*}
$Q(\theta';\theta)$, the conditional expectation of the complete-data
log-likelihood function $\ln p_{\theta'}(x)$ given $y$ and $\theta$,
then is used as an auxiliary function to construct an EM algorithm via a mapping $M: \Theta
\rightarrow \Theta$, where each iteration is defined by
$\theta^{(t+1)} = M(\theta^{(t)})$ as follows:

\begin{quote}
E-step: Compute $Q(\theta; \theta^{(t)}) = \tilde p [ k_{\theta^{(t)}}
  [ \ln p_{\theta} ] ] $\\
M-step: Choose $\theta^{(t+1)}$ to be a value of $\theta \in \Theta$
which maximizes $Q(\theta; \theta^{(t)})$.
\end{quote}
That is, $M$ is a point-to-set map $M(\theta^{(t)}) =
\underset {\theta\in \Theta}{\arg\max\;} Q(\theta; \theta^{(t)})$. This use of $Q$ as an auxiliary function in the EM algorithm can be justified by the fact that an iterative maximization of $Q$ guarantees that the incomplete-data log-likelihood function $L$ is non-decreasing on each iteration of an EM algorithm. This can easily be shown with the inequality
\[
L(M(\theta)) - L(\theta) = ( Q(M(\theta);\theta) - Q(\theta;\theta))
+ (H(\theta;\theta) - H(M(\theta);\theta)) \geq 0, \textrm{ for all }
\theta \in \Theta,
\]
which follows from the positivity of the difference both in the $Q$ functions (by definition of $M$) and in the $H$ functions (by Jensen's inequality (see \citeN{CoverThomas:91})). That is, we have the following proposition, due to
\citeN{Dempster:77}.

\begin{po}[\shortciteN{Dempster:77}, Theorem 1]
  \label{IncreasingLikelihood} 
\mbox{}\\
For each EM algorithm, $L(M(\theta)) \geq L(\theta)$, 
for all $\theta
\in \Theta$.
\end{po}

Although $Q$ is globally maximized in each M-step, the term $H$ may
hinder a straight global maximization of $L$. As a general result for
EM algorithms, \citeN{Wu:83} shows that under continuity and differentiability
conditions on $L$ and $Q$, a sequence of EM iterates $\{
L(\theta^{(t)}) \}$ bounded from above converges monotonically to a
critical point of $L$.

\begin{po}[\citeN{Wu:83}, Theorem 2] For continuous $Q$, 
  all limit points of any instance $\{ \theta^{(t)} \}$ of an EM
  algorithm are critical points of $L$, and for continuous and
  differentiable $L$, a sequence $\{L(\theta^{(t)}) \}$ bounded from
  above converges monotonically to $L^\ast = L(\theta^\ast)$ for some
  critical point $\theta^\ast$ of $L$.
\end{po}

To summarize, the popularity of the EM algorithm is due to its easy
computation because it relies only on complete-data computations:
the E-step involves complete-data conditional expectations, and the M-step
requires MLE from these completed data. Even if the algorithm may
converge slowly, it conservatively increases the likelihood function
at each iteration and in almost all cases converges to a local
maximum of $L$.  If a sequence of EM iterates is stuck at some critical point which is
not a local or global maximum of $L$, e.g., a saddle point or even a
local minimum, a small random perturbation will help it to diverge from
this critical point. If $L$ has several critical points, the
convergence properties of an EM sequence will be extremely dependent on the choice of
the starting value of the sequence of iterates.

\subsection{Partial M-Steps: The GEM Algorithm}
\label{PartialM}

As discussed in the last section, one main feature of the EM algorithm
is to provide a simplified M-step where MLE from complete data rather
than from incomplete data is performed. In some cases, even this
maximization is complicated and does not exist in closed form. An EM
algorithm involving such a complicated M-step would be computationally
unattractive. For such cases, \citeN{Dempster:77} defined a so-called
generalized EM (GEM) algorithm where the M-step is only partially
computed, i.e, each M-step only increases the
$Q$ function rather than globally maximizing it.

That is, for a GEM algorithm, $\theta^{(t+1)}$ is chosen s.t.
\[ Q(\theta^{(t+1)} ; \theta^{(t)}) \geq Q(\theta^{(t)};
\theta^{(t)}). \]

As shown by \citeN{Dempster:77}, this condition suffices for
increasing the incomplete-data log-likelihood at each interation, i.e., Proposition
\ref{IncreasingLikelihood} also holds for each GEM algorithm.
However, appropriate convergence of a GEM algorithm does not follow directly
without further specification on the process on increasing the $Q$
function. For each instance of a GEM algorithm, one can either show
the general convergence conditions for a GEM algorithm as given by
\citeN{Wu:83} to hold, or directly prove convergence
of the specific GEM instance in question. The latter approach is
pursued in Sect. \ref{IM} where we explicitly show convergence for a
GEM algorithm for log-linear models.

\subsection{Partial E-steps and Maximum Pseudo-Likelihood Estimation}
\label{PartialE}

For many cases, a partial computation of the E-step is also
useful. These are especially cases where the sample space
$\mathcal{X}$ is too large to be summed over explicitly in the
expectations to be calculated in the E-steps. The idea here is to
replace the intractable probability function with respect to which the expectation
is taken by a probability function which is more tractable. This change in
probability functions results in a corresponding change of the
likelihood function to a pseudo-likelihood function which is now
defined with respect to the new tractable distribution.
Thus from a general optimization-theoretic point of view EM with partial
E-steps is an example of maximum pseudo-likelihood estimation.

A theoretical justification for maximum pseudo-likelihood estimation in the
context of EM is given in
\citeN{Neal:93} or \citeN{Csiszar:84}. In terms of \citeN{Neal:93},
the EM algorithm can be seen as maximizing a joint function
$\mathcal{F}$ of the parameters and of the distributions over the
unobserved data. Using an arbitrary distribution $q$ over the
unobserved variables, $\mathcal{F}$  can be obtained as a lower bound
on the incomplete-data log-likelihood function $L$ as follows.

\begin{eqnarray*}
L(\theta)  & = & \tilde p [ \ln \sum_{x \in X(y)} p_\theta (x) ] \\
& = & \tilde p [ \ln \sum_{x \in X(\cdot)} q(x) \frac{p_\theta (x)}{q(x)}
] \\
& \geq & \tilde p [ \sum_{x \in X(\cdot)} q(x) \ln \frac{p_\theta
  (x)}{q(x)} ] \textrm{, by Jensen's inequality} \\ 
& = & \tilde p [ q [ \ln p_\theta ] ] - \tilde p [ q [\ln q ] ] \\
& = & \mathcal{F}(q,\theta).
\end{eqnarray*}
Provided that values of $x$ are seen as physical states and the energy
of  a state is $- \ln p_\theta(x)$, the function
$\mathcal{F}(q,\theta)$ can be seen as analogous to the negative of
the ``free energy'' of statistical physics, i.e., the expected energy
under $q$ minus the entropy of $q$. 
The EM algorithm can be interpreted in this framework as alternating
between maximizing $\mathcal{F}$ as a function of $q$ and
$\theta$.
The E-step maximizes $\mathcal{F}$ with respect to $q$ and
holds $\theta$ fixed;
the M-step maximizes $\mathcal{F}$ with respect
to $\theta$ for fixed $q$.

\begin{quote}
E-step: Set $q^{(t+1)}$ to $\underset{q}{\arg\max\;}
\mathcal{F}(q,\theta^{(t)})$. \\
M-step: Set $\theta^{(t+1)}$ to $\underset{\theta}{\arg\max\;}
\mathcal{F}(q^{(t+1)}, \theta)$.
\end{quote}
\citeN{Neal:93} show that at a true joint maximization, these
iterations are equivalent to the classical EM iterations defined in
Sect. \ref{GeneralTheory}. That is, the maximum in the E-step is obtained by
taking $q^{(t+1)}(x) = k_{\theta^{(t)}}(x|y)$, and at this point we
have the equality $\mathcal{F}(q^{(t+1)}, \theta^{(t)}) =
L(\theta^{(t)})$.
The maximum in the M-step is obtained by maximizing the term in
$\mathcal{F}$ depending on $\theta$, which is in this case 
$\tilde p [ k_{\theta^{(t)}} [ \ln p_\theta ] ]
= Q(\theta;\theta^{(t)})$.
Since each such E-step guarantees that $\mathcal{F} = L$, and since we
maximize $Q(\theta;\theta^{(t)})$ in each M-step, we are guaranteed
not to decrease $L$ at each combined EM step. 

In a partial E-step, $q^{(t+1)}$ is set to a tractable approximation of
$k_{\theta^{(t)}}(x|y)$, which yields the inequality
$\mathcal{F}(q^{(t+1)}, \theta^{(t)}) \leq  L(\theta^{(t)})$. In the
corresponding M-step, the term in $\mathcal{F}$ depending on $\theta$ is
maximized. Together, these combined EM steps guarantee not to decrease
the lower bound $\mathcal{F}$ on the incomplete-data log-likelihood
$L$  at each iteration.
Thus, for partial E-steps, monotonicity and convergence of the
resulting algorithm have to be shown in terms of the pseudo-likelihood
function $\mathcal{F}$ which bounds the true likelihood function $L$
from below.

\section{An EM Example: Baum's Maximization Technique}
\markright{4.4 Baum's Maximization Technique}
\label{EMExample}

\subsection{Basic Concepts}

A special instance of the EM algorithm for MLE of hidden Markov
models, i.e., stochastic regular grammars, from incomplete data was
presented in \citeN{Baum:70} and \citeN{Baum:72}. The form of this algorithm using
dynamic programming techniques for efficient computation is well-known
as the ``forward-backward algorithm'' (see \citeN{Rabiner:89}).
Most popular approaches to parameter estimation for probabilistic
grammars are based upon this technique. \citeN{Baker:79} generalized
this algorithm to the so-called ``inside-outside algorithm'', which efficiently
estimates the parameters of stochastic context-free grammars (see also
\citeN{Booth:73}, \citeN{Lari:90} and \citeN{Jelinek:90}). This algorithm can
successfully be applied also to other stochastic grammars which assume
independence of their derivation units of each other. Such models are,
e.g, stochastic dependency grammars \cite{Carroll:92} or stochastic
lexicalised tree-adjoining grammars \cite{Resnik:92,Schabes:92}. In
the following, we will refer to the basic version of this algorithm as
Baum's maximization technique.

In the following, we will give a quick review of the
basic concepts of Baum's maximization technique. The probabilistic
models the algorithm is applied to can be abstracted by
stochastic derivation models which define a derivation process as a
stochastic process as follows:
Make a stochastic choice at each derivation step and assume the stochastic
choices to be independent of each other; calculate the
probability of a derivation as the joint probability of the
independent stochastic choices made, and the probability of an input 
as the sum of the probabilities of its derivations.

More formally, let $\pi = ( \pi_{ij}) \in \Pi$
be the parameter vector of the probabilistic processing model where
$\pi_{ij} \geq 0$ and $\sum_j \pi_{ij} = 1$.
The variable $i$ ranges over the types of choices that the stochastic process
makes, and the variable $j$ ranges over the
alternatives to choose from when a choice of type $i$ is made.
Furthermore, let $y$ denote an input of the
probabilistic processing model, i.e., an observation sequence, and let
$x$ denote an output of the model, i.e., an analysis, and let $Y(x) = y$ be the unique observation
corresponding to analysis $x$ and $X(y) = \{ x | Y(x) = y \}$ be the
set of analyses of observation $y$. Finally, let $\nu_{ij}(x)$ be the
number of selections of alternative $j$ for a choice of type $i$ in analysis
$x$. The probability of an analysis is the joint probability of the
stochastic choices made in producing it. Since these stochastic
choices are assumed to be independent of each other, the probability
of an analysis is calculated as the product of the probabilities of
the stochastic choices made in producing it:

 \[ p_\pi(x) = \prod_{ij} \pi_{ij}^{\nu_{ij}(x)} .\]
The probability of an observation is the sum of the probabilities
of its analyses:
\[ g_\pi(y) = \sum_{x \in X(y)} p_\pi(x) .\]
For a given random sample of observations, the purpose of Baum's
maximization technique is to find maximum likelihood parameter values
for the incomplete-data likelihood function $L$ where 

\[ L(\pi) = \prod_{y \in \mathcal{Y}}  g_\pi(y)^{\tilde p(y)}. \]
The EM mapping $M$ is instantiated here to a particularily simple
case. Let $k_\pi(x|y) = p_\pi(x)/g_\pi(y)$, then

\[ M(\pi_{ij}) = \frac{\tilde p [N_{ij}]}{\tilde p [\sum_l N_{il} ]} =
\frac{\tilde p [ k_\pi [\nu_{ij} ] ]}
{\tilde p [  \sum_l k_\pi [\nu_{il} ] ] } .\]
Intuitively, the estimated value of parameter $\pi_{ij}$ is obtained
by prorating $N_{ij}$, the expected number of times choice $ij$ is made during
the derivation, by $\sum_l N_{il}$, the expected total number of times
a choice of type $i$ is made during the derivation, for all
observations $y$.
\citeN{Baum:70} showed that this algorithm is hill-climbing, i.e.,
$L(M(\pi)) \geq L(\pi)$ for all $\pi \in \Pi$,
and that the incomplete-data likelihood $L$ eventually converges to a
critical point, i.e., to a local maximum.

\subsection{Baum's Maximization Technique and Context-Dependence in CLP}
\label{Baum}

The intuitive appeal and the efficient computability of Baum's
maximization technique has led to a multiplicity of applications of this
technique to various grammar frameworks. Recently, an attempt to apply this
technique to a probabilistic version of the constraint-based formalism
CUF, which is an instance of the CLP scheme of \citeN{HuS:88}, has
been presented by \citeN{Eisele:94}. 
As recognized by \citeN{Eisele:94}, there is a
context-dependence problem associated with applying this technique to
such constraint-based systems. In CLP terms, the problem is that
incompatible variable bindings can lead to failure derivations, which
cause a loss of probability mass in the estimated probability distribution over
derivations. A similar problem appears in every constraint-based
system which constrains derivations by restrictions dependent
of the context of the derivation.
Approaches embedding Baum's maximization technique into estimation
procedures for context-sensitive constraint-based systems have been
presented, e.g., by \citeN{Briscoe:92}, \citeN{BriscoeCarroll:93},
\citeN{Brew:95}, \citeN{Miyata:96}, \citeN{Osborne:97} or
\citeN{Bod:98}. 
From an optimization-theoretic point of view, all such constraint-based
approaches contradict the inherent assumptions of Baum's maximization
technique which require that the derivation steps are mutually
independent and thus the set of licensed derivations is
unconstrained.

This problem of context-dependence is discussed in detail in \citeN{Abney:97} in
connection with the so-called Empirical Relative Frequency (ERF) estimation
method, which can be seen a complete-data version of Baum's estimation
technique. He shows that applying this method to context-sensitive
stochastic attribute-value grammars does not generally yield
maximum-likelihood estimates. 

In the following, this general argument shall be illustrated with a
simple CLP example.
Let us apply the stochastic derivation model of
Sect. \ref{EMExample} to a simple context-sensitive constraint logic
program (see Fig. \ref{PCLPprogram}). The stochastic choices of the
abstract model correspond to application probabilities of definite
clauses in the generalized SLD-resolution procedure;
the alternatives to choose from when an atom is selected in goal
reduction are the different clauses defining the selected atom. To
indicate a probabilistic parameter $\pi_{ij}$, each
clause will be annotated by a number \texttt{ij}.

\begin{figure}[htbp]
\begin{center}
\begin{tabular}{l}
\texttt{11} $\texttt{s}(Z)\leftarrow \texttt{p}(Z) \:\&\: \texttt{q}(Z).$ \\
\texttt{21} $\texttt{p}(Z) \leftarrow Z=a.$ \\
\texttt{22} $\texttt{p}(Z) \leftarrow Z=b.$ \\
\texttt{31} $\texttt{q}(Z) \leftarrow Z=a.$ \\
\texttt{32} $\texttt{q}(Z) \leftarrow Z=b.$ 
\end{tabular}
\end{center}
\caption{Constraint logic program}
\label{PCLPprogram}
\end{figure}

The relational atom $\texttt{s}(Z)$ is defined uniquely in clause \texttt{11}.
The atoms $\texttt{p}(Z)$ and $\texttt{q}(Z)$ each are defined in two different ways,
which for the sake of the example are considered to be incompatible.
This incompatibility together with the variable sharing in the body of
clause \texttt{11} introduces context-dependence into the program.
For a selection of atom $\texttt{p}(Z)$ one can choose between clauses
\texttt{21} and \texttt{22} in a goal reduction step,
whereas  for a choice of atom $\texttt{q}(Z)$ the alternatives to choose from are clauses
\texttt{31} and \texttt{32}.

Suppose we have a training corpus of three queries, consisting of
two tokens of query $y_1: \texttt{s}(Z)\:\&\: Z=a$ and one token of query
$y_2: \texttt{s}(Z)\:\&\: Z=b$.  Each query gets a unique proof tree
from the program of Fig. \ref{PCLPprogram}, i.e., a query of type
$y_1$ gets a proof tree of type $x_1$, and a query
of type $y_2$ gets one of type $x_2$ (see Fig. 
\ref{prooftrees}). Note that in the proof trees of
Fig. \ref{prooftrees} goal reduction and constraint solving are
applied in one step.

\begin{figure}[htbp]
\begin{center} 
\begin{tabular}{p{75pt}c}

$x_1:$

&

\begin{bundle}{$\texttt{s}(Z)\:\&\: Z=a$}
  \chunk[$\qquad r,c$]
  {
    \begin{bundle}{\texttt{11}, $\texttt{p}(Z) \:\&\: \texttt{q}(Z) \:\&\: Z=a$}
      \chunk[$\qquad r,c$]
      {
        \begin{bundle}{\texttt{21}, $\texttt{q}(Z) \:\&\: Z=a$}
          \chunk[$\qquad r,c$]
          {\texttt{31}, $Z=a$}
        \end{bundle}
        }
    \end{bundle}
    }
\end{bundle}

\end{tabular}

\end{center}

\vspace{2ex}

\begin{center} 
  \begin{tabular}{p{75pt}c}
    
    $x_2:$ 
    
    & 
    
    \begin{bundle}{$\texttt{s}(Z)\:\&\: Z=b$}
      \chunk[$\qquad r,c$]
      {
        \begin{bundle}{\texttt{11}, $\texttt{p}(Z) \:\&\: \texttt{q}(Z) \:\&\: Z=b$}
          \chunk[$\qquad r,c$]
          {
            \begin{bundle}{\texttt{22}, $\texttt{q}(Z) \:\&\: Z=b$}
              \chunk[$\qquad r,c$]
              {\texttt{32}, $Z=b$}
            \end{bundle}
            }
        \end{bundle}
        }
    \end{bundle}
    
  \end{tabular}
  
\end{center}

\caption{Proof trees from constraint logic program}
\label{prooftrees}
\end{figure}

For parameter estimation according to Baum's method, we must
calculate conditional probabilities $k(x|y)$ for $x \in X(y)$. These
probabilities will be 1 in each case, since there is a unique proof tree for each query.
Thus for the calculation of
$\tilde p[N_{ij}] = \tilde p [ k_\pi [\nu_{ij} ] ]$,
the expected number of occurences of clauses in proof trees, we simply
have to count and can ignore the respective probabilities of the proof
trees. As in an application of the complete-data ERF method, for this
case Baum's algorithm will give unique parameter estimates 
$\hat \pi_{ij} = \frac{\tilde p [N_{ij}]}{\tilde p [\sum_l N_{il} ]}$
in one step  (see Table \ref{estimation}).

\begin{table}
\begin{center}
\begin{tabular}{c c c c|c|c c|c c|}
$y \in \mathcal{Y}$ &
$x \in X(y)$ &
$\tilde p(y)$ &
$k(x|y)$  &
$N_{11}$ &
$N_{21}$ &
$N_{22}$ &
$N_{31}$ &
$N_{32}$ \\
\hline
$ y_1$ & $x_1$ & 2/3 &1 & $1 \cdot1$ & $1 \cdot1$ & $1 \cdot0$  & $1 \cdot1$ &  $1 \cdot 0$ \\
$y_2$ & $x_2$ & 1/3 & 1 & $1 \cdot1$ & $1 \cdot0$ & $1 \cdot1$ & $1 \cdot0$  & $1 \cdot1$ \\
\hline
 & & & $\tilde p[N_{ij}]=$ & 3/3 & 2/3 & 1/3 & 2/3 & 1/3 \\
 & & & $\tilde p[ \sum_l N_{il}]=$ & 3/3 & 3/3 & 3/3 & 3/3 & 3/3\\
\hline
 & & & $\hat\pi_{ij}=$ & 1 & 2/3 & 1/3 & 2/3 & 1/3
\end{tabular}
\end{center}
\caption{Estimation using Baum's maximization technique}
\label{estimation}
\end{table}

If we consider the calculation of the probability distribution
over the proof trees of such a probabilistic CLP model, we see that we
cannot simply calculate a product for each proof tree. Instead, we
have to introduce a normalization constant in order to ensure the sum
over the sample space of proof trees to be 1. For the program of
Fig. \ref{PCLPprogram}, this partition function is taken as the sum of the unnormalized
probabilities of the proof trees under the estimated model: $p_{\hat\pi}(x_1) + p_{\hat\pi}(x_2) =  (1 \cdot 2/3 \cdot 2/3) + (1 \cdot 1/3
\cdot 1/3) = 4/9 + 1/9 = 5/9$. The normalized 
probability distribution over proof trees then is: 
$p'_{\hat\pi}(x_1) = (4/9) / (5/9) = 4/5, \; p'_{\hat\pi}(x_2) =
(1/9)/(5/9) =  1/5$. 
The likelihood $L'$ of our training corpus under the normalized
distribution is: $L'= (4/5)^2 \cdot 1/5 = .128$. 
However, note that there is no analytical solution to the
problem of finding parameter values $\pi'$ for the clauses of the program of
Fig. \ref{PCLPprogram} which define $p'$ as a probabilistic context-free model
on the proof trees of Fig. \ref{prooftrees}. Rather, what has
happened here is that we implicitly moved to another family of
probability distributions by introducing the normalization constant
into $p'$. This new family of probability distributions obviously no
longer requires the parameter values to sum up to 1 for identical
left-hand sides of rules, but introduces a normalization constant
instead in order to guarantee the function to be a probability
function. We will acknowledge this family of probability distributions
as log-linear distributions in the next section. Clearly, we can
easily find parameters of a log-linear model which assigns a higher likelihood to this sample. We could take for example a parameterization $\pi''$ which assigns
$\pi''_{\mathtt{21}} = 2$ and $\pi''_{\mathtt{ij}} = 1$ forall
$\mathtt{ij} \not = \mathtt{21}$. This yields a normalized probability
distribution over the proof trees with $p''_{\pi''}(x_1) = 2/3$,
$p''_{\pi''}(x_2) = 1/3$ and likelihood $L'' = (2/3)^2 \cdot 1/3 =
.148.$ The fact that $L'' > L'$ clearly contradicts the assumption that the parameter
estimates $\hat \pi$ given by applying Baum's estimation technique to a
normalized context-free probability model yield the desired maximum
likelihood values.

\section{A Log-Linear Probability Model for CLP}
\markright{4.5 A Log-Linear Probability Model for CLP}
\label{LLM}

As shown in the last section, we cannot simply apply a stochastic
context-free derivation model to CLP but have to go to more expressive
probability models. In fact, we implicitly already have made this move in
the above example by introducing a partition function into the
probabilistic context-free model. We will show in the following that
acknowledging this model as a log-linear model not only opens the
possibility to find new consistent maximum likelihood estimators but also enables a more
flexible parameterization of the probability models. 

\subsection{Motivation}

Log-linear models are widely used in probabilistic modelling but come
with different names in different applications. The name
log-linear is standardly used in contingency table analysis (see,
e.g, \citeN{Knoke:80}). The model itself originated under the name
of the Gibbs- or Boltzmann-distribution in statistical physics as a
flexible probability model of equilibrium states of physical
systems. \citeN{Jaynes:57} interpreted such equilibrium models in a
more abstract framework and coined the name maximum-entropy
model. Log-linear models have been applied successfully in the area of image
processing, where they are known under the name of random fields (see
\citeN{Geman:84}). These special log-linear models are closely related
to other probabilistic network models such as Boltzmann machines (see
\citeN{Ackley:85}) or Bayesian networks 
(see \citeN{Frey:98}). 
Log-linear models have been used with effort also in various NLP
applications. To name only a few, these applications include
probabilistic grammar models \cite{Mark:92,Abney:97},
word spellings \cite{DDL:97},
machine translation \cite{Berger:96},
language modelling \cite{Rosenfeld:96},
prepositional phrase attachment \cite{Rat:94},
part-of-speech tagging \cite{Rat:96},
history-based parsing \cite{RatP:97},
lexical correlations \cite{LaffertyACL:97}
text segmentation \cite{LaffertyEMNLP:97}, 
and text classification \cite{Nigam:99}.

The popularity of log-linear models is clearly due to the great
expressive power they provide with very simple means. That is,
log-linear models can be seen as an \emph{exponential family} of probability
distributions where the probability of a datum is simply defined as being
proportional to the product of weights assigned to selected properties
of the datum. Let $(\pi_i)$ be
a vector of weights and $\nu_i(\omega)$ the number of times property
$i$ appears in datum $\omega$, for all $i=1, \ldots, n$, then

\[
p(\omega) \propto \prod_{i=1}^n \pi_i^{\nu_i(\omega)}.
\]
A log-linear form is obtained from this simply by replacing proportionality by a constant $C = Z^{-1}$ and parameters $\pi_i$ by log-parameters
$\lambda_i=\ln \: \pi_i$, for all $i=1, \ldots, n$, i.e., taking the logarithm of this probability function yields a linear combination of parameters and properties and a constant.
\begin{eqnarray*}
p(\omega) & = & C \prod\nolimits_{i=1}^n \pi_i^{\nu_i(\omega)} \\
 & = & Z^{-1} \prod\nolimits_{i=1}^n \pi_i^{\nu_i(\omega)} \\
 & = & Z^{-1} \prod\nolimits_{i=1}^n e^{\lambda_i \nu_i(\omega)} \\
 & = & Z^{-1} e^{\sum_{i=1}^n \lambda_i \nu_i(\omega)} .
\end{eqnarray*}
A more general form of log-linear models is obtained by including a fixed initial or
reference distribution $p_0$ into the model such that 
$p(\omega) = Z^{-1} e^{\sum_{i=1}^n \lambda_i \nu_i(\omega)} p_0(\omega)$ and 
$Z=\sum_\omega e^{\sum_i \lambda_i \nu_i(\omega)} p_0(\omega)$.

Clearly, the main advantage of log-linear models is their great
flexibility, which includes the normalized models used in
Sect. \ref{Baum} and even probabilistic context-free models as
special cases (the normalization constant has value 1 in this case). However, considering CLP, with log-linear models we are free to
select as properties arbitrary features of proof trees rather than
being restricted to clauses only. For example, we could take
subtrees of proof trees as properties. This possibility to combine
arbitrary clauses to properties allows us to model arbitrary
context-dependencies in proof trees. Clearly, linguistically there is
no particular reason for assuming rules or clauses as the best
properties to use in a probabilistic grammar. As we will see in Sect.
\ref{Experiment}, more abstract properties referring to grammatical
functions, attachment preferences, or other general features of
constraint-based parses can be employed successfully to probabilistic
CLGs. Furthermore, the log-parameters 
corresponding to these properties are not required to constitute a
probablity distribution over clauses defining the same predicate,
i.e., the parameters do not have to sum to 1 for clauses defining the
same predicate. That is, log-linear models allow us to define a
probability distribution over proof trees directly rather than
indirectly as a joint probability of clause applications as in the
context-free models above.

Let us illustrate this with the simple CLP example of
Sect. \ref{Baum}. A training corpus consisting of two tokens of
query $y_1: \texttt{s}(Z)\:\&\: Z=a$ and one token of query 
$y_2: \texttt{s}(Z)\:\&\: Z=b$ together with the corresponding proof
trees generated by the program of Fig. \ref{PCLPprogram} is depicted in
Fig. \ref{sample}. Note that for ease of readability, we will omit in
the following figures the labelings of nodes and edges of proof trees.

\begin{figure}[htbp]
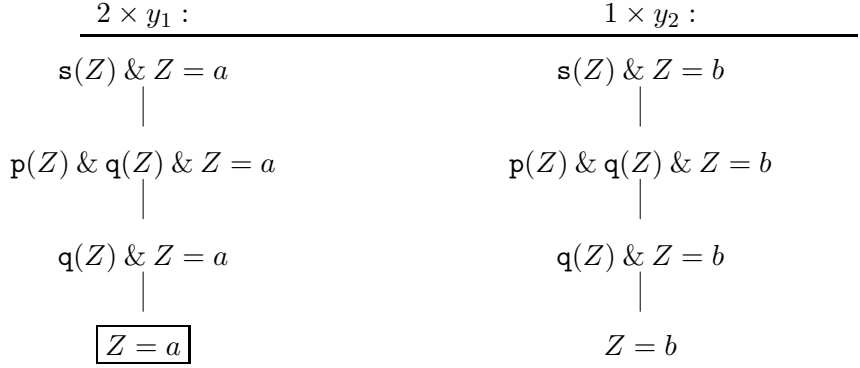

\begin{center} 
\begin{tabular}{p{180pt}p{90pt}}

$2 \times y_1:$ & $1 \times y_2:$ \\ \hline

\begin{bundle}{$\texttt{s}(Z)\:\&\: Z=a$}
  \chunk
  {
    \begin{bundle}{$\texttt{p}(Z) \:\&\: \texttt{q}(Z) \:\&\: Z=a$}
      \chunk
      {
        \begin{bundle}{$\texttt{q}(Z) \:\&\: Z=a$}
          \chunk
          {\fbox{$Z=a$}}
        \end{bundle}
        }
    \end{bundle}
    }
\end{bundle}

&

\begin{bundle}{$\texttt{s}(Z)\:\&\: Z=b$}
  \chunk
  {
    \begin{bundle}{$\texttt{p}(Z) \:\&\: \texttt{q}(Z) \:\&\: Z=b$}
      \chunk
      {
        \begin{bundle}{$\texttt{q}(Z) \:\&\: Z=b$}
          \chunk
          {$Z=b$}
        \end{bundle}
        }
    \end{bundle}
    }
\end{bundle}

\end{tabular}

\end{center}

\caption{Queries and proof trees for constraint logic program}
\label{sample}
\end{figure}

To capture the statistics of the training sample of 
Fig. \ref{sample}, it is sufficient to define a single property which
is able to differentiate between the proof tree types. Such a property
could be, for example, the terminal node \fbox{$Z=a$} of proof tree
$x_1$. Setting the value of the corresponding parameter of this single-parameter
model to $\ln 2$ will yield the desired probability distribution
$p(x_1) = 2/3, \; p(x_2) = 1/3$ with incomplete-data log-likelihood $L=.148$. 

Another way to understand log-linear models is as \emph{maximum-entropy
models}. From this viewpoint we do statistical inference and, believing
that entropy is the unique consistent measure of the amount of
uncertainty represented by a probability distribution, we obey the
following principle:  
\begin{quote}
In making inferences on the basis of partial information we must use that
probability distribution which has maximum entropy subject to
whatever is known. This is the only unbiased assignment we can make;
to use any other would amount to arbitrary assumption of information
which by hypothesis we do not have. \cite{Jaynes:57}
\end{quote}
More formally, suppose a random variable $X$ can take on values $x_k,
k=1, \ldots, m$ and we want to estimate the corresponding
probabilities $p_k, k=1, \ldots, m$. All we have are expectations of
functions $f_i(X), i=1,\dots, n$. Let these expectations be defined
with respect to a given empirical distribution $\tilde p_k,
k=1,\ldots, m$ on complete data $x_k, k=1,\ldots,m$ s.t. $\sum_{k=1}^m
p_k f_i(x_k) = \sum_{k=1}^m \tilde p_k f_i(x_k), i=1,\ldots,n$. Then
the maximum-entropy principle can be stated as follows. 

\begin{quote}
Maximize the entropy $H(p) =
- \sum_{k=1}^m p_k \: \ln\: p_k$ subject to the constraints \\
$\sum_{k=1}^m p_k f_i(x_k) = \sum_{k=1}^m \tilde
p_k f_i(x_k),  i=1,\ldots,n$ and $\sum_{k=1}^m
p_k = 1$. 
\end{quote}
For all $p_k, k=1, \ldots, m$ which solve the above problem, we get
the following parametric solution:

\[
p_k = \frac{\textstyle e^{\sum_{i=1}^n \lambda_i f_i(x_k)}}
{\textstyle \sum_{k=1}^m e^{\sum_{i=1}^n \lambda_i f_i(x_k)}}
\]
Following \citeN{Jaynes:57}, this result can be derived directly from
a constrained optimization argument where the parameters are viewed as
Lagrange multipliers. That is, by applying the standard technique of Lagrange multipliers (see, e.g., \citeN{ThomasFinney:96}) to the constrained optimization problem stated in
the maximum-entropy principle, the above parametric probability model
can be derived by solving this constrained optimization problem with
respect to the probabilities $p_k$. Let $\Lambda$ denote the
Lagrangian defined by 
\begin{eqnarray*}
\Lambda(p,\lambda)& = &
\sum_{k=1}^m (p_k \: \ln\: p_k) 
- (\lambda_0 + 1) (\sum_{k=1}^m p_k -1) \\
& & - \lambda_1 (\sum_{k=1}^m p_k f_1(x_k) + 
\sum_{k=1}^m \tilde p_k f_1(x_k)) \\
& & \vdots \\
& & - \lambda_n (\sum_{k=1}^m p_k f_n(x_k) +
\sum_{k=1}^m \tilde p_k f_n(x_k)).
\end{eqnarray*}
Then the first partial derivative of $\Lambda$ with respect to the
$p_k$ is 
\[ \frac{\partial}{\partial p_k} \Lambda = 
(\ln\: p_k + 1) - (\lambda_0 +1) - \lambda_1 f_1(x_k) - \cdots -
\lambda_n f_n(x_k). \]
Now set 
\[ \frac{\partial}{\partial p_k} \Lambda = 0 , \]
then
\[ p_k = e^{\lambda_0 + \sum_{i=1}^n \lambda_i f_i(x_k)}.\]
Since the sum of all probabilities $p_k$ has to be 1, we have 
\[ 1 = \sum_{k=1}^m p_k =
e^{\lambda_0} \sum_{k=1}^m e^{\sum_{i=1}^n \lambda_i f_i(x_k)}.\]
If we define a partition function $Z$ as
\[ Z = \sum_{k=1}^m e^{\sum_{i=1}^n \lambda_i f_i(x_k)} ,\]
then \[ \lambda_0 = \ln\: Z^{-1} \]
and the maximum-entropy distribution is 
\[ p_k = Z^{-1} e^{\sum_{i=1}^n \lambda_i f_i(x_k)} = 
\frac{e^{\sum_{i=1}^n \lambda_i f_i(x_k)}}
{\sum_{k=1}^m e^{\sum_{i=1}^n \lambda_i f_i(x_k)}}. \]

To sum up, the parametric form of maximum-entropy probability models
can be derived by solving a constrained optimization problem with
respect to the probabilities $p_k, k=1, \ldots, m$. The remaining problem,
namely solving this constrained maximum-entropy problem with
respect to the parameters $\lambda_i, i=1, \ldots, n$, can be shown to
be equivalent to solving a maximum likelihood problem for log-linear
models. This duality can be stated as follows (see
\citeN{Berger:96}). The complete-data log-likelihood $L_c$ of a random
sample from a log-linear model $p_\lambda$ on $X$, with empirical
probability $\tilde p(x_k)$ of the
values $x_k, k=1,\ldots,m$ is defined as

\[
L_c(\lambda) = \ln \prod_{k=1}^m p_\lambda(x_k)^{\tilde
p (x_k)}
=
\sum_{k=1}^m \tilde p(x_k) \ln p_\lambda(x_k)
\]
This function is equivalent to the Lagrangian $\Lambda$ instantiated
to the parametric model $p_\lambda$:

\begin{eqnarray*}
\Lambda(p_\lambda,\lambda) 
& = & \sum_{k=1}^m Z_\lambda^{-1} e^{\lambda \cdot f(x_k)} 
\ln ( Z_\lambda^{-1} e^{\lambda \cdot f(x_k)} ) \\
& & - \sum_{i=1}^n \lambda_i \sum_{k=1}^m Z_\lambda^{-1} e^{\lambda
\cdot f(x_k)} f_i(x_k) 
\\ 
& & + \sum_{i=1}^n \lambda_i \sum_{k=1}^m \tilde p(x_k) f_i(x_k) \\
& = & -\ln Z_\lambda + p_\lambda[\lambda \cdot f] - p_\lambda[\lambda \cdot f]
+ \tilde p[\lambda \cdot f] \\
& = & - \ln Z_\lambda + \tilde p[\lambda \cdot f]. \\
& & \\
L_c(\lambda) & = & 
\ln \prod_{k=1}^m p_\lambda(x_k) ^{\tilde p (x_k)} \\
& = &  \sum_{k=1}^m \tilde p(x_k) \ln (Z_\lambda^{-1}
e^{\lambda \cdot f(x_k)}) \\ 
& = &  - \ln Z_\lambda + \tilde p [\lambda \cdot f].
\end{eqnarray*}
Thus, the values $\lambda^\ast$ that solve the constrained
maximum-entropy problem with respect to the parameters $\lambda_i,
i=1, \ldots, n$ are equivalently a solution to the complete-data
maximum likelihood problem for the log-linear model $p_\lambda$.

The more general model which includes an initial or reference
distribution $p_0$ is
derived in a similar way as the unique parametric probability
distribution $p$ that minimizes the Kullback Leibler (KL) distance
$D(p||p_0)$ between $p$ and a given reference distribution $p_0$, subject
to certain constraints. That is, the generalized log-linear model  

\[
p_\lambda(\omega) = \frac{e^{\lambda \cdot \nu(\omega)}
p_0(\omega)}{\sum_{\omega \in \Omega} e^{\lambda \cdot \nu(\omega})p_0(\omega)}
\]
is the parametric solution to the following constraint optimization problem:

\begin{quote}
Minimize $D(p||p_0) = \sum_{\omega \in \Omega} p(\omega) \ln
\frac{p(\omega)}{p_0(\omega)}$ subject to the constraints \\
$p[f_i] = \tilde p[f_i],
i=1, \ldots,n$ and $\sum_{\omega \in \Omega} p(\omega)=1$.
\end{quote}
For uniformly distributed $p_0(\omega)$, the KL distance $D(p||p_0)$ is the
negative of the entropy $H(p)$, minus a constant not involving
$\lambda$:

\[
D(p||p_0) = \sum_{\omega \in \Omega} p(\omega) \ln p(\omega) - \ln
p_0(\omega)  = - H(p) - K.
\]
In this case, minimizing the KL distance subject to certain
constraints is equivalent to maximizing the entropy subject to these
constraints. Furthermore, a connection to a maximum likelihood
problem can be established for the KL distance miminization problem
in a similar way as for the maximum entropy problem.

\subsection{The Form of Log-Linear Models}

Log-linear probability distributions define the probability of a datum
simply as proportional to weights assigned to selected properties of
the datum. Formally, the parametric family of such distributions is
defined as follows.

\begin{de} [Log-linear distribution] \label{LogLinearDistribution}
A log-linear probability distribution $p_{\lambda \cdot \nu}$ on a set
  $\Omega$ is defined s.t.\ for all $\omega \in \Omega$:
\[
p_{\lambda \cdot \nu}(\omega) =
{Z_{\lambda \cdot \nu}}^{-1}
e^{\lambda \cdot \nu(\omega)} p_0(\omega),
\]
\begin{description}

\setlength{\parsep}{0ex}
\setlength{\itemsep}{0ex}

\item[]
$Z_{\lambda \cdot \nu} = \sum_{\omega \in \Omega} 
e^{\lambda \cdot \nu(\omega)} p_0(\omega)$
is a normalizing constant,
\item[] $\lambda = (\lambda_1, \ldots, \lambda_n)$
is a vector of log-parameters s.t.\ $\lambda \in \Rn$,
\item[] $\mathbf{\chi} = (\chi_1, \ldots, \chi_n)$
is a vector of properties, 
\item[] $\mathbf{\nu} = (\nu_1, \ldots, \nu_n)$
is a vector of property-functions s.t. for each 
$\nu_i:\Omega
\rightarrow \Nats $, 
$\nu_i(\omega)$
is the number of occurences of property $\chi_i$ in $\omega$, 
\item[]$\lambda \cdot \nu (\omega)$
is a weighted property-function s.t. 
$\lambda \cdot \nu (\omega) =
\sum^n_{i=1}\lambda_i \nu_i(\omega)$,
\item[]$p_0$ is a fixed initial distribution.
\end{description}
\end{de}
For the following discussion, it will be convenient to introduce some further
notation. Properties will be referred to for most purposes by vectors
$\nu$ of property functions rather than by explicit vectors
$\chi$ of properties. Slightly abusing terminology, we will call properties both
$\chi$ and $\nu$.
 
As in Definition
\ref{LogLinearDistribution}, a log-linear probability distribution 
depending on property vector $\nu$ and parameter vector $\lambda$ will
be written in subscript notation as $p_{\lambda \cdot \nu}$. In case
the property vector is fixed and clear from the context, the model
(resp. the normalization constant) will be written $p_\lambda$
(resp. $Z_\lambda$) to indicate the dependence on the
parameter vector $\lambda$. 

Furthermore, it will be convenient to have a recursive definition of
log-linear models based on weighted property-functions which are extended by
additional properties and corresponding parameters.

\begin{po} For each weighted property-funtion
$\phi(\omega) = \mathbf{\lambda \cdot \nu}(\omega)$, 
$\psi(\omega) = \mathbf{\gamma \cdot \mu}(\omega)$ 
let $(\psi + \phi)(\omega) = \psi(\omega) + \phi(\omega)$
be an extended  property-function.
Then 
\[
p_{\psi + \phi}(\omega) = {Z_{\psi \circ \phi}}^{-1}
e^{\psi(\omega)} p_{\phi}(\omega)
\; \textrm{where} \;
Z_{\psi \circ \phi} = p_\phi [e^\psi].
\]
\end{po}

\begin{proof}
\begin{eqnarray*}
p_{\psi + \phi}(\omega)
& = & 
{Z_{\psi + \phi}}^{-1} e^{\psi + \phi(\omega)}
p_0(\omega) \\
& = & 
(\sum_{\omega \in \Omega} e^{\psi(\omega) + \phi(\omega)}
p_0(\omega) )^{-1} e^{\psi(\omega) + \phi(\omega)} p_0(\omega)\\
& = & 
( \sum_{\omega \in \Omega} e^{\psi(\omega)} e^{\phi(\omega)}
p_0(\omega) Z_{\phi} {Z_{\phi}}^{-1} ) ^{-1} e^{\psi(\omega)}
e^{\phi(\omega)} p_0(\omega)\\
& = & 
{Z_\phi}^{-1} (\sum_{\omega \in \Omega} e^{\psi(\omega)}
p_{\phi}(\omega) )^{-1}
e^{\psi(\omega)} e^{\phi(\omega)} p_0(\omega) \\
& = & 
(\sum_{\omega \in \Omega} e^{\psi(\omega)}  p_{\phi}(\omega)
)^{-1} e^{\psi(\omega)} p_{\phi}(\omega) \\
& = & {Z_{\psi \circ \phi}}^{-1} e^{\psi(\omega)} p_\phi(\omega).
\qed
\end{eqnarray*}
\renewcommand{\qed}{}
\end{proof}
For an extended model with weighted property functions 
$\phi(\omega) = \mathbf{\lambda \cdot \nu}(\omega)$ and
$\psi(\omega) = \mathbf{\gamma \cdot \nu}(\omega)$,
written $p_{\gamma + \lambda}$, we have accordingly 
\begin{eqnarray*}
p_{\gamma + \lambda} (\omega) 
& = & 
Z_{\gamma + \lambda}^{-1}
e^{\gamma \cdot \nu (\omega) + \lambda \cdot \nu (\omega)}
p_0 (\omega) \\
& = & 
(\sum_{\omega \in \Omega}
e^{(\gamma + \lambda) \cdot \nu(\omega)}
p_0(\omega))^{-1}
e^{(\gamma + \lambda) \cdot \nu(\omega)}
p_0(\omega) \\
& = & 
(\sum_{\omega \in \Omega}
e^{\gamma \cdot \nu (\omega)}
p_{\lambda \cdot \nu} (\omega))^{-1}
e^{\gamma \cdot \nu (\omega)}
p_{\lambda \cdot \nu} (\omega)\\
& = & {Z_{\gamma \circ \lambda}}^{-1}
e^{\gamma \cdot \nu(\omega)}
p_{\lambda \cdot \nu} (\omega).
\end{eqnarray*}

\section{Statistical Inference for Log-Linear Models from Incomplete
  Data}
\markright{4.6 Statistical Inference for Log-Linear Models from Incomplete
  Data}
\label{SI}

In the last two sections we argued that a solution to the context-dependence
problem in probabilistic CLP requires probability models which are more
expressive than context-free and proposed log-linear models for this
purpose. The price we have to pay for this gain in expressivity
clearly is a gain in complexity of parameter estimation. Furthermore,
the gain in flexibility due to property selection is an additional
complexity factor which calls for an automatic solution.
Fortunately, \citeN{DDL:97} have presented a
statistical inference algorithm for combined property selection and
parameter estimation for log-linear models. \citeN{Abney:97} has shown the
applicability of this algorithm to stochastic attribute-value
grammars, which can be seen as a special case of context-sensitive CLGs. 

This algorithm, however, applies only to complete data. Unfortunately, the need to
rely on large training samples of complete data is a problem if such
data are difficult to gather. For example, in natural language parsing
applications, complete data means several person-years of
hand-annotating large corpora with detailed analyses of specialized
grammar frameworks. This is always a labor-intensive and error-prone
task, which additionally is restricted to the specific grammar
framework, the specific language, and the specific language domain in
question. Clearly, for such applications automatic and
reusable techniques for statistical inference from incomplete data are
desirable.

In the following, we present a version of the statistical inference
algorithm of \citeN{DDL:97} especially designed for incomplete data
problems. We present a parameter estimation technique
for log-linear models from incomplete data (Sect. \ref{IM}) and a
property selection procedure from incomplete data
(Sect. \ref{PS}). These algorithms are combined into a statistical
inference algorithm for log-linear models from incomplete data (Sect.
\ref{CSI}). Empirical results on experimenting with these algorithms
on a small scale are presented in Sect. \ref{Experiment}.

This section is based on work presented in shortened form in
\citeN{RiezlerKon:98}.

\subsection{Motivation}\label{MotivIM}

\begin{table}
\renewcommand{\arraystretch}{1.5}
\begin{center}
\begin{tabular}{|l|c|c|}\hline
& log-likelihood & auxiliary function \\ \hline
complete data 
& $\frac{\partial L_c(\lambda)}{\partial \lambda_i} =
\tilde p [ \nu_i ] - p_\lambda [ \nu_i ] $ 
& $\frac{\partial A_c(\gamma; \lambda)}{\partial \gamma_i} =
\tilde p [ \nu_i ] - p_\lambda [ \nu_i e^{\gamma_i \nu_\#} ] $ \\ 
incomplete data 
& $ \frac{\partial L(\lambda)}{\partial \lambda_i} = 
\tilde p [ k_\lambda [ \nu_i ] - p_\lambda [ \nu_i ] ] $
& $\frac{\partial Q(\lambda; \lambda')}{\partial \lambda_i} =
\tilde p [ k_{\lambda'} [ \nu_i ] - p_\lambda [ \nu_i ] ]$ \\
\hline
\end{tabular}
\label{maximization-equations}
\caption{Partial derivatives of objective functions for MLE of
  log-linear models}
\end{center}
\end{table}

Why is incomplete-data estimation for log-linear models difficult? The
answer is because complete-data estimation for such models is
difficult, too. Let us have a look at the first partial derivatives of
some objective functions which are considered in MLE of log-linear
models from complete and incomplete data (see Table \ref{maximization-equations}).
The system of equations to be solved at
the points where the first partial derivatives of the complete data
log-likelihood function $L_c$ are zero, i.e., at the critical points of
$L_c$, is 
\[ 
\sum_{x \in \mathcal{X}} Z_\lambda^{-1} e^{\lambda \cdot \nu(x)}
\nu_i(x) 
= \sum_{x \in \mathcal{X}} \tilde p(x) \nu_i(x) \textrm{ for all } i=1,
\ldots, n.
\] 
Clearly, because of the dependence of both $Z_\lambda$ and $e^{\lambda
  \cdot \nu(x)}$ on $\lambda$ this system of equations cannot be
  solved coordinate-wise in $\lambda_i$. 
This problem is even more severe for the case of incomplete-data
  estimation. The incomplete-data log-likelihood $L$ has its critical
  points at the solution of the following system of equations in $\lambda_i$:
\[ 
\sum_{y \in \mathcal{Y}} \tilde p(y) \sum_{x \in \mathcal{X}}
  Z_\lambda^{-1} e^{\lambda \cdot \nu(x)} \nu_i(x) 
= \sum_{y \in \mathcal{Y}} \tilde p(y) \sum_{x \in X(y)} k_\lambda(x|y) \nu_i(x) \textrm{ for all } i=1,
\ldots, n.
\] 
Here additionally a dependence of $ k_\lambda(x|y) $ on $\lambda$ has
to be respected.
However, an application of the standard EM theory to incomplete-data
estimation of log-linear models only partially solves the problem. The
equations to be solved to find the critical points of the auxiliary
function $Q(\lambda;\lambda')$ for a log-linear model depending on
$\lambda$ are
\[ 
\sum_{y \in \mathcal{Y}} \tilde p(y) \sum_{x \in \mathcal{X}}
  Z_\lambda^{-1} e^{\lambda \cdot \nu(x)} \nu_i(x) 
= \sum_{y \in \mathcal{Y}} \tilde p(y) \sum_{x \in X(y)} k_{\lambda'}(x|y) \nu_i(x) \textrm{ for all } i=1,
\ldots, n.
\] 
Here $k_{\lambda'}(x|y)$ depends on $\lambda'$ instead of $\lambda$.
However, the dependency of $Z_\lambda$ and $e^{\lambda
  \cdot \nu(x)}$ on $\lambda$ still remains a problem.

Solutions for the system of equations can be found, e.g., by applying
general-purpose numerical optimization methods (see \citeN{Fletcher:87}) to
the problem in question. For the smooth and strictly concave
complete-data log-likelihood $L_c$, e.g., a conjugate gradient
approach could be used. However, optimization methods specifically
tailord to the problem of MLE from complete data for log-linear models
have been presented by \citeN{Darroch:72} and \citeN{DDL:97}. The
``improved iterative scaling'' algorithm of \citeN{DDL:97} itself is
an extension of the ``generalized iterative scaling'' algorithm of
\citeN{Darroch:72}. In the first algorithm properties are required to
sum up to a constant independent of the complete data,
 i.e., $\nu_\# = \sum_{i = 1}^n \nu_i(x) = K$ for
all $x \in \mathcal{X}$, whereas in the latter algorithm $\nu_\#$ is
allowed to vary as a function of $x$. This property of ``generalized
iterative scaling'' is claimed to improve the convergence rate by
increasing the step size taken toward the maximum at each iteration.
Both iterative scaling algorithms iteratively maximize an auxiliary
function $A_c(\gamma;\lambda)$ which is defined as a lower bound on
the difference $L_c(\gamma + \lambda) - L_c(\lambda)$ in complete-data
log-likelihood when going from a basic model $p_\lambda$ to an
extended model $p_{\gamma + \lambda}$. The function $A_c(\gamma;\lambda)$ is
maximized as a function of $\gamma$ for fixed $\lambda$ which makes it
possible to solve the following equation coordinate-wise in
$\gamma_i, i = 1, \ldots, n$:
\[
 \sum_{x \in \mathcal{X}} p_\lambda(x) \nu_i(x) e^{\gamma_i \nu_\#(x)} =
\sum_{x \in \mathcal{X}} \tilde p(x) \nu_i(x) \textrm{ for all } i = 1, \ldots, n.
\]
A closed form solution for $\gamma_i$ is given for constant $\nu_\#$;
otherwise simple numerical methods such as Newton's method can be used
to solve for the $\gamma_i$. 
It is shown in \citeN{DDL:97} and \citeN{Darroch:72} that iteratively replacing
$\lambda^{(t+1)}$ by $\lambda^{(t)} + \gamma^{(t)}$ conservatively
increases $L_c$ and such a sequence of likelihood values eventually converges
to the the global maximum of the strictly
concave function $L_c$. 

For the case of incomplete-data estimation things are more
complicated. Since the incomplete-data log-likelihood function $L$ is
not strictly concave, general-purpose numerical methods such as
conjugate gradient cannot be applied. However, such methods can be
applied to the auxiliary function $Q$ as defined by a standard EM algorithm for
log-linear models. Alternatively, iterative scaling methods can be
used to perform maximization of the auxiliary function $Q$ of the EM algorithm. Both
approaches result in a doubly iterative algorithm where an iterative
algorithm for the M-step is interweaved in the iterative EM algorithm.
Clearly, this is computationally burdensome and should be avoided.

The aim of this chapter is exactly to avoid such doubly iterative
algorithms. The idea of our approach is to interleave the auxiliary
functions $Q$ of the EM algorithm and $A_c$ of iterative scaling in
order to define a singly-iterative incomplete-data estimation
algorithm using a new combined auxiliary function. Similar to the case
of iterative scaling for complete data, the new auxiliary function
will be defined as a lower bound on the improvement in log-likelihood. This
allows for an intuitive and elegant proof of convergence of the new
algorithm.  Our proofs are completely self-contained and do not rely
on the convergence of alternating minimization procedures for
maximum-entropy models as presented by \citeN{Csiszar:75} or
\citeN{Csiszar:89} or on the regularity conditions for generalized EM
algorithms as presented by \citeN{Wu:83} or \citeN{Meng:93}. The
relation of our algorithm to generalized EM estimation and
maximum-entropy estimation is discussed in Sects. \ref{GEM}
and \ref{ME}.

\subsection{Parameter Estimation}\label{IM}

\subsubsection{General Theory}

Let us start with a problem definition. Applying the incomplete-data
framework defined in Sect. \ref{GeneralTheory} to a log-linear
probability model for CLP, we can assume the following to be given:

\begin{itemize}
\item observed, incomplete data $y \in \mathcal{Y}$,
  corresponding to a finite sample of queries for a
  constraint logic program \Po, 
\item unobserved, complete data $x \in \mathcal{X}$, corresponding to
  the countably infinite sample of proof trees for queries
$\mathcal{Y}$ 
  from \Po,
\item a many-to-one function $Y:\mathcal{X} \rightarrow
\mathcal{Y}$ s.t. $Y(x) = y$ corresponds to the unique query labeling proof tree 
  $x$, and its inverse  $X:\mathcal{Y} \rightarrow 2^\mathcal{X}$
  s.t. $X(y) = \{x |\; Y(x) = y \}$
  is the countably infinite set of proof trees for query $y$ from \Po,
\item a complete-data specification $p_\lambda (x)$, which is a
  log-linear distribution on $\mathcal{X}$ 
  with given initial distribution $p_0$,
  fixed property vector $\chi$ and
  property-functions vector $\nu$
  and depending on parameter vector $\lambda$, 
\item an incomplete-data specification $g_\lambda (y)$, which is
  related to the complete-data specification by
  \[ g_\lambda(y) = \sum_{x \in X(y)} p_\lambda(x) .\]
\end{itemize}
The problem of
maximum-likelihood estimation for log-linear models from incomplete
data can then be stated as follows. 

\begin{quote}
  Given a fixed sample from $\mathcal{Y}$ and a set $\Lambda =
  \{ \lambda |\; p_{\lambda}(x)$ is a log-linear distribution on
  $\mathcal{X}$ with fixed $p_0$, fixed $\nu$ and
  $\lambda \in \Rn \}$,
  we want to find a maximum likelihood estimate
  $\lambda^\ast$ of $\lambda$ s.t.
  $\lambda^\ast = \underset{\lambda \in \Lambda}{\arg\max\;}
  L(\lambda) = \ln \prod_{y \in \mathcal{Y}} g_\lambda(y)^{\tilde p(y)}$.
\end{quote}
For the rest of this section we will refer to a given vector $\nu$ of
property functions. Furthermore, we
assume that for each property function $\nu_i$ some proof tree $x \in
\mathcal{X}$ with $\nu_i(x) > 0$ exists,
and require $p_\lambda$ to be strictly positive on $\mathcal{X}$,
i.e., $p_\lambda(x) > 0$ for all $x \in \mathcal{X}$. These conditions
guarantee that $p_\lambda(x) > 0$ for all $x \in \mathcal{X}$ and for
all $\lambda \in \Lambda$ which is a desirable property in the
following discussion.

Similar to the case of iterative scaling for complete-data estimation,
we define an auxiliary function $A(\gamma , \lambda)$ as a
conservative estimate of the difference $L(\gamma + \lambda) -
L(\lambda)$ in log-likelihood. The lower bound for the incomplete-data
case can be derived from the complete-data case, in essence, by
replacing an expectation of complete, but unobserved data by a
conditional expectation given the observed data and the current fit of
the parameter values.
Clearly, this is the same trick that is used in the EM algorithm, but
applied in the context of a different auxiliary function. From the
lower-bounding property of
the auxiliary function it can immediately be seen that each
maximization step of $A(\gamma , \lambda)$ as a function of $\gamma$
will increase or hold constant the improvement $L(\gamma + \lambda) -
L(\lambda)$. This is a first
important property of a MLE algorithm. Furthermore, our approach
to view the incomplete-data auxiliary function directly as a lower
bound on the improvement in incomplete-data log-likelihood enables an
intuitive and elegant proof of convergence. 

Let the conditional probability of complete data $x$ given incomplete
data $y$ and parameter values $\lambda$ be defined as
\[
k_\lambda (x|y) = p_\lambda (x)/g_\lambda (y) = 
\frac{e^{\lambda \cdot \nu(x)} p_0(x)}{\sum_{x \in X(y)} e^{\lambda
\cdot \nu(x)} p_0(x)}.
\]
Then a two-place auxiliary function $A$ can be defined as follows.

\begin{de} \label{A} Let $\lambda \in \Lambda$, $\gamma \in \Rn$, $\nu_\#(x) =
\sum_{i=1}^n \nu_i (x)$, $\bar\nu_i (x) = \nu_i(x) / \nu_\#(x)$.
Then 
\[
A(\gamma , \lambda) = \tilde p[
 1 + k_{\lambda} [ \gamma \cdot \nu ] -
p_{\lambda} [ \sum^n_{i=1} {\bar{\nu}}_i e^{\gamma_i \nu_{\#} }]].
\]
\end{de}

The particular form of the auxiliary function $A$ and the connection
of $A$ and $L$ is discussed in detail in Lemmata \ref{A<L-L}, 
\ref{A0=0}, and \ref{dA=dL} below. 
Let us first have a look at the extreme value properties of $A$, which
are crucial for the iterative maximization of $A$.

By considering the first and second derivatives of $A$, we see that
$A$ can be maximized directly and uniquely. This can be explained as
follows. Suppose the parameters $\gamma \in \Rn$ to be a convex set; the
Hessian matrix of $A$ is a diagonal matrix filled only with negative
elements 
\[
\frac{\partial^2 A(\gamma;\lambda)}{\partial \gamma_i \partial
  \gamma_j} =
\frac{\partial}{\partial \gamma_j} ( \frac{\partial A(\gamma;\lambda)}{\partial
  \gamma_i}) =
\left\{ 
\begin{array}{cc}
< 0 & \textrm{ if } i = j \\
0 & \textrm{ else}
\end{array}
\right.
\]
and thus negative definite. Unique maximization follows from this
since a function whose Hessian is negative definite throughout a
convex set is strictly concave, and a strictly concave function
attains a maximum at most one point of a convex set, and thus a critical
point is necessarily a maximum (see \citeN{Horn:85}).

\begin{po} \label{maxA} 
For each $\lambda \in \Lambda$, $\gamma \in \Rn$:
$A(\gamma , \lambda)$ takes its maximum as a function of
  $\gamma$ at the unique point $\hat\gamma$ satisfying for each
  $\hat\gamma_i, i=1, \ldots, n$:
\[
\tilde p [  k_{\lambda} [ \nu_i ] ]
= \tilde p [ p_{\lambda} [ \nu_i
    e^{\hat\gamma_i \nu_{\#} }]] .
\]
\end{po}

\begin{proof}
\begin{eqnarray*}
\frac{\partial}{\partial \gamma_i}
A(\gamma , \lambda)
& = & 
\frac{\partial}{\partial \gamma_i}
\tilde p [ 
 1 + k_{\lambda} [ \gamma \cdot \nu ] -
p_{\lambda} [ \sum^n_{j=1} {\bar{\nu}}_j e^{\gamma_j \nu_{\#} }] 
] \\
& = & 
\tilde p [ 
 \frac{\partial}{\partial \gamma_i}
\sum^n_{j=1}
( \frac{1}{n} + k_{\lambda} [ \gamma_j \cdot \nu_j ] -
p_{\lambda} [  {\bar{\nu}}_j e^{\gamma_j \nu_{\#} }] 
) ]\\
& = &  
\tilde p [ 
 \sum_{j \not= i} 
( \frac{\partial}{\partial \gamma_i}
( \frac{1}{n} + k_{\lambda} [ \gamma_j \cdot \nu_j ] -
p_{\lambda} [  {\bar{\nu}}_j e^{\gamma_j \nu_{\#} }] 
) ) \\
& &
+ \frac{\partial}{\partial \gamma_i}
( \frac{1}{n} + k_{\lambda} [ \gamma_i \cdot \nu_i ] -
p_{\lambda} [  {\bar{\nu}}_i e^{\gamma_i \nu_{\#} }] 
) ] \\
& = & 
\tilde p [ 
 k_{\lambda} [ \nu_i ] 
- \sum_{x \in \mathcal{X}} 
(  p_{\lambda} (x) \bar\nu_i (x) e^{\gamma_i \nu_{\#}(x) } \nu_{\#}(x)
) ] \\
& = & 
\tilde p [ 
 k_{\lambda} [ \nu_i ] 
- \sum_{x \in \mathcal{X}} 
( p_{\lambda} (x) \nu_i (x) e^{\gamma_i \nu_{\#}(x)}
) ] \\
& = & 
\tilde p [ 
 k_{\lambda} [ \nu_i ] 
- p_{\lambda} [ \nu_i e^{\gamma_i \nu_{\#}} ] ].\\
& & \\
\frac{\partial^2}{{\partial \gamma_i}^2} A(\gamma , \lambda)
& = & 
\frac{\partial}{\partial \gamma_i}
\tilde p [ 
 k_{\lambda} [ \nu_i ] 
- p_{\lambda} [ \nu_i e^{\gamma_i \nu_{\#}} ] ] \\
& = & 
- \tilde p [
\frac{\partial}{\partial \gamma_i} p_{\lambda} [ \nu_i e^{\gamma_i \nu_{\#}}] 
] \\
& = & 
- \tilde p [ 
 \sum_{x \in \mathcal{X}} 
( p_{\lambda}(x) \nu_i(x) e^{\gamma_i
    \nu_{\#}(x)} \nu_{\#}(x) ) ]\\
& = & 
- \tilde p [ 
p_{\lambda} [ \nu_i \nu_\# e^{\gamma_i \nu_\#} ] ] \\
& < & 0.
\qed
\end{eqnarray*}
\renewcommand{\qed}{}
\end{proof}

From the auxiliary function $A$ an iterative algorithm for maximizing
$L$ is constructed. For want of a name, we will call this
algorithm the ``Iterative Maximization (IM)'' algorithm. At each step
of the IM algorithm, a log-linear model based on parameter vector
$\lambda$
is extended to a model based on parameter vector
$\lambda + \hat\gamma$, where  
$\hat\gamma$ is an estimation of the parameter vector that maximizes
the improvement in $L$ when moving away in the parameter space from
$\lambda$.
This  increment $\hat\gamma$ is estimated by maximizing the auxiliary
function $A(\gamma , \lambda)$ as a function of $\gamma$ and, by
Proposition \ref{maxA}, determined for each $i = 1, \ldots, n$
uniquely as the solution $\hat\gamma_i$ to the equation 
$\tilde p [  k_{\lambda} [ \nu_i ] ]
= \tilde p [ p_{\lambda} [ \nu_i
    e^{\hat\gamma_i \nu_{\#} }]] .$ If $\nu_\# = \sum_{i=1}^n \nu_i(x) =
    K$ sums to a constant independent of $x \in \mathcal{X}$, there
    exists a closed form solution for the $\hat \gamma_i$:
\[
\hat \gamma_i = \frac{1}{K} \ln \frac{\tilde p [ k_\lambda [ \nu_i ] ]
  } {p_\lambda [ \nu_i ] } \textrm{ for all } i = 1, \ldots, n.
\]
For $\nu_\#$ varying as a function of $x$ Newton's method can be
applied to find an approximate solution (see Sect. \ref{Approximation}). The IM
algorithm in its general form is defined as follows:

\begin{de}[Iterative maximization] \label{IterativeMaximization}
Let $\mathcal{M}:\Lambda
  \rightarrow \Lambda$ be a mapping defined by
\[
\mathcal{M}(\lambda) = \hat \gamma + \lambda
\textrm{ with }
\hat \gamma = \underset{\gamma \in \Rn}{\arg\max\;}
A(\gamma , \lambda).
\]
\textrm{ Then each step of the IM algorithm is
  defined by }
\[
\lambda^{(k+1)} = \mathcal{M}(\lambda^{(k)}).
\]
\end{de}

In order to show the monotonicity and convergence properties of the IM
algorithm, we first must prove some provisional results.
Lemma \ref{A<L-L} shows that the auxiliary function
$A(\gamma , \lambda)$ is a lower bound on the incomplete-data log-likelihood
difference $L(\gamma + \lambda) - L(\lambda)$. In the first inequality
we apply Jensen's inequality to the
natural logarithm of an expectation. We get a simplified form similar
to the log-likelihood difference for complete data, modulo an empirical
distribution over complete data being replaced by the conditional
distribution $k_\lambda(x|y)$. This form is simplified further by
omitting the logarithm, using the inequality $\ln x \leq
x-1$. Furthermore, a random variable $\nu_\#$ on $\mathcal{X}$ is
introduced in order to define a probability distribution $\bar\nu_i$
on $\mathcal{X}$. Applying Jensen's inequality to an expectation with
respect to $\bar\nu_i$ in the power of $e$, we arrive at a final
simplified form, defining the auxiliary function $A$.

\newtheorem{lemma}[po]{Lemma}
\begin{lemma} \label{A<L-L}
$A(\gamma , \lambda) \leq L(\gamma + \lambda) - L(\lambda)$.
\end{lemma}

\begin{proof}
\begin{eqnarray*}
L(\gamma + \lambda) - L(\lambda)
& = & \sum_{y \in \mathcal{Y}} \tilde p(y) \ln g_{\gamma + \lambda}(y)
- \sum_{y \in \mathcal{Y}} \tilde p \ln g_\lambda(y)
\\
& = &
\tilde p [ 
\ln \frac{g_{\gamma +\lambda}(\cdot)}{g_{\lambda}(\cdot)}
] \\
& = & 
\tilde p [ 
\ln \frac{1}{g_{\lambda}(\cdot)} 
\sum_{x \in X(\cdot)} (
p_{\gamma + \lambda}(x) 
\frac{p_{\lambda}(x)}{p_{\lambda}(x)}
)
] \\
& = & 
\tilde p [ 
\ln \sum_{x \in X(\cdot)} (
\frac{p_{\lambda}(x)}{g_{\lambda}(\cdot)}
\frac{p_{\gamma + \lambda}(x)}{p_{\lambda}(x)}
)
] \\
& \geq & 
\tilde p [ 
\sum_{x \in X(\cdot)} (
\frac{p_{\lambda}(x)}{g_{\lambda}(\cdot)}
\ln \frac{p_{\gamma + \lambda}(x)}{p_{\lambda}(x)}
)
] 
\textrm{ by Jensen's inequality} \\
& = & 
\tilde p [ 
\sum_{x \in X(\cdot)} (
\frac{p_{\lambda}(x)}{g_{\lambda}(\cdot)} (
\ln p_{\gamma + \lambda}(x) 
- \ln p_{\lambda}(x)
)
)
]\\
& = & 
\tilde p [ 
\sum_{x \in X(\cdot)} (
\frac{p_{\lambda}(x)}{g_{\lambda}(\cdot)} (
\ln Z_{\gamma \circ \lambda}^{-1}
+ \ln e^{\gamma \cdot \nu(x)} 
+ \ln p_{\lambda}(x)
- \ln p_{\lambda}(x)
))] \\
& = & 
\tilde p [ 
k_{\lambda} [ \gamma \cdot \nu ]
- \ln p_{\lambda} [ e^{\gamma \cdot \nu} ]
]\\
& \geq & 
\tilde p [ 
k_{\lambda} [ \gamma \cdot \nu ]
+1 
-  p_{\lambda} [ e^{\gamma \cdot \nu} ]
] \quad \textrm{since } \ln x \leq x -1 \\
& = & 
\tilde p [ 
k_{\lambda} [ \gamma \cdot \nu ]
+1 
- \sum_{x \in \mathcal{X}} (
p_{\lambda} (x) e^{ \sum^n_{i=1} \gamma_i \nu_i(x)
  \frac{\nu_\#(x)}{\nu_\#(x)}}
)
]\\
& = & 
\tilde p [ 
k_{\lambda} [ \gamma \cdot \nu ]
+1 
- \sum_{x \in \mathcal{X}} (
p_{\lambda} (x) e^{ \sum^n_{i=1} \gamma_i \bar\nu_i(x) \nu_\#(x)}
)
]\\
& \geq &
\tilde p [ 
k_{\lambda} [ \gamma \cdot \nu ]
+1 
- \sum_{x \in \mathcal{X}} (
p_{\lambda} (x) \sum^n_{i=1} \bar\nu_i(x) e^{  \gamma_i \nu_\#(x)}
)
]
\textrm{ by Jensen's inequality}
\\
& = & 
\tilde p [ 
k_{\lambda} [ \gamma \cdot \nu ]
+1 
- p_{\lambda} [ \sum^n_{i=1} \bar\nu_ie^{  \gamma_i \nu_\#}
]
] \\
& = & 
A(\gamma , \lambda).
\qed
\end{eqnarray*}
\renewcommand{\qed}{}
\end{proof}
Lemma \ref{A0=0} shows that there is no estimated improvement in
log-likelihood at the origin.

\begin{lemma} \label{A0=0}
$A(0 , \lambda) = 0$. 
\end{lemma}

\begin{proof}
\[
A(0 , \lambda) 
= \tilde p [ 
k_{\lambda} [ 0 \cdot \nu ]
+ 1 
- \sum_{x \in \mathcal{X}} p_{\lambda}(x) \sum^n_{i=1} \bar\nu_i (x)
e^0
]
=0.
\qed
\]
\renewcommand{\qed}{}
\end{proof}
Lemma \ref{dA=dL} shows that the critical points of $A$ and $L$ as
functions of $\gamma$ for fixed $\lambda$ are the same.

\begin{lemma} \label{dA=dL}
$ \left. \frac{d}{dt} \right|_{t=0} A(t\gamma , \lambda)
= \left. \frac{d}{dt} \right|_{t=0} L(t\gamma + \lambda)$.
\end{lemma}

\begin{proof}
\begin{eqnarray*}
\frac{d}{dt} A(t\gamma , \lambda) 
& = & \frac{d}{dt} 
\tilde p [ 
k_{\lambda} [ t\gamma \cdot \nu ]
+1 
- \sum_{x \in \mathcal{X}} (
p_{\lambda} (x) \sum^n_{i=1} \bar\nu_i(x) e^{  t\gamma_i \nu_\#(x)}
)
] \\
& = & 
\tilde p [ 
k_{\lambda} [ \gamma \cdot \nu ]
- \sum_{x \in \mathcal{X}} (
p_{\lambda} (x) \sum^n_{i=1} \frac{\nu_i(x)}{\nu_\#(x)} e^{  t\gamma_i
  \nu_\#(x)} \gamma_i \nu_\#(x)
)
] \\
& = &
\tilde p [ 
k_{\lambda} [ \gamma \cdot \nu ]
- \sum_{x \in \mathcal{X}} (
p_{\lambda} (x) 
\sum^n_{i=1} \nu_i (x) \gamma_i e^{t \gamma_i \nu_\# (x)} 
)]. \\
& & \\
\left. \frac{d}{dt} \right|_{t=0} A(t\gamma , \lambda)
& = & 
\tilde p [ 
k_{\lambda} [ \gamma \cdot \nu ] 
- \sum_{x \in \mathcal{X}} ( 
p_{\lambda} (x) 
\sum^n_{i=1} \nu_i (x) \gamma_i e^0 
)] \\
& = &  
\tilde p [ 
k_{\lambda} [ \gamma \cdot \nu ] 
- p_{\lambda} [ \gamma \cdot \nu ]
]. \\
& & \\
\frac{d}{dt} L(t\gamma + \lambda) 
& = & 
\tilde p [ 
\frac{d}{dt} \ln \sum_{x \in X)(\cdot)} p_{t\gamma + \lambda}(x)
] \\
& = & 
\tilde p [ 
( \sum_{x \in X)(\cdot)} p_{t\gamma + \lambda}(x) )^{-1}
\frac{d}{dt}
\sum_{x \in X)(\cdot)}
e^{t\gamma \cdot \nu(x)} 
p_{\lambda}(x)
Z_{t\gamma \circ \lambda}^{-1}
]\\
& = & 
\tilde p [ 
( \sum_{x \in X)(\cdot)} p_{t\gamma + \lambda}(x) )^{-1}
\sum_{x \in X)(\cdot)}
p_{\lambda}(x)  
( 
- e^{t\gamma \cdot \nu(x)} Z_{t\gamma \circ \lambda}^{-2} \\
& &
\sum_{x \in \mathcal{X}} e^{t\gamma \cdot \nu(x)} \gamma \cdot \nu(x) 
p_{\lambda}(x) 
+
Z_{t\gamma \circ \lambda}^{-1} e^{t\gamma \cdot \nu(x)} \gamma \cdot
\nu(x)
)]
\\
& = & 
\tilde p [ 
- \sum_{x \in X)(\cdot)} p_{t\gamma + \lambda}(x)
p_{t\gamma + \lambda} [ \gamma \cdot \nu ]
( \sum_{x \in X)(\cdot)} p_{t\gamma + \lambda}(x))^{-1} \\
& & 
+ \sum_{x \in X)(\cdot)} p_{t\gamma + \lambda} [ \gamma \cdot \nu
]
( \sum_{x \in X)(\cdot)} p_{t\gamma + \lambda}(x))^{-1} 
] \\
& = & 
\tilde p [ 
-  p_{t\gamma + \lambda} [ \gamma \cdot \nu
] 
+
k_{t\gamma + \lambda} [\gamma \cdot \nu ]
]. \\
& & \\
\left. \frac{d}{dt} \right|_{t=0} L(t\gamma + \lambda)
& = & \tilde p [ 
k_{\lambda} [\gamma \cdot \nu ]
- p_{\lambda} [ \gamma \cdot \nu
] 
].
\qed
\end{eqnarray*}
\renewcommand{\qed}{}
\end{proof}

One central result of this section is stated in Theorem
\ref{IncreasingIMLikelihood}. It  shows the monotonicity of the IM
algorithm, i.e., the incomplete-data log-likelihood
$L$ is increasing on each iteration of the IM algorithm
except at fixed points of $\mathcal{M}$ or equivalently at critical
points of $L$. 

\newtheorem{theorem}[po]{Theorem}
\begin{theorem}[Monotonicity]\label{IncreasingIMLikelihood}
For all $\lambda \in \Lambda$:
$L(\mathcal{M}(\lambda) )\geq L(\lambda)$ with equality iff $\lambda$
is a fixed point of $\mathcal{M}$ or equivalently is a critical point of
$L$.
\end{theorem}

\begin{proof}
\begin{eqnarray*}
L(\mathcal{M}(\lambda)) - L(\lambda) 
& \geq & 
A(\mathcal{M}(\lambda)) \quad \textrm{by Lemma \ref{A<L-L}} \\
& \geq &
0 \quad \textrm{by Lemma \ref{A0=0} and definition of $\mathcal{M}$}.
\end{eqnarray*}
The equality $L(\mathcal{M}(\lambda)) = L(\lambda)$  holds iff
$\lambda$ is a fixed point of $\mathcal{M}$,
i.e., $\mathcal{M}(\lambda) = \hat\gamma + \lambda$ with
$\hat\gamma = 0$.
Furthermore, $\lambda$ is a fixed point of $\mathcal{M}$ iff 
$\hat\gamma = \underset{\gamma \in \Rn}{\arg\max\;} A(\gamma ,
\lambda) = 0$, \\
$\iff \textrm{for all } \gamma \in \Rn : \hat t = \underset{t \in
\Reals}{\arg\max\;} A(t\gamma , \lambda) = 0$, \\
$\iff \textrm{for all } \gamma \in \Rn: \left. \frac{d}{dt} \right|_{t=0} A(t
\gamma , \lambda) = 0$, \\
$\iff \textrm{for all }\gamma \in \Rn: \left. \frac{d}{dt} \right|_{t=0}
L(t\gamma + \lambda) = 0 $, by Lemma \ref{dA=dL} \\
$\iff  \lambda \textrm{ is a critical point of $L$}.$
\end{proof}
Corollary \ref{MaximumLikelihoodEstimates} implies that a maximum
likelihood estimate is a fixed point of the mapping $\mathcal{M}$.

\newtheorem{corollary}[po]{Corollary}
\begin{corollary} \label{MaximumLikelihoodEstimates}
Let $\lambda^\ast = \underset{\lambda \in \Lambda}{\arg\max\;} L(\lambda)$.
Then $\lambda^\ast$ is a fixed point of $\mathcal{M}$.
\end{corollary} 
Theorem \ref{convergence} discusses the convergence properties of the
IM algorithm. In constrast to the improved iterative scaling
algorithm, we cannot show convergence to a global maximum of a
strictly concave objective function. Rather we can show convergence of
a sequence of IM iterates to a critical point of the non-concave
incomplete-data log-likelihood function $L$. The central property to
show is that all limit points of a sequence of IM iterates are critical points of $L$.

\begin{theorem}[Convergence]\label{convergence}
Let $\{ \lambda^{(k)} \}$ be a sequence in $\Lambda$
determined by the IM Algorithm.
Then all limit points of $\{ \lambda^{(k)} \}$ are fixed points of
$\mathcal{M}$ or equivalently are critical points of $L$.
\end{theorem}

\begin{proof}
Let $\{ \lambda^{(k_n)} \} $ be a subsequence of
$\{ \lambda^{(k)} \}$ converging to $\bar\lambda$.
Then for all $\gamma \in
\Rn$:
\begin{eqnarray*}
A(\gamma , \lambda^{(k_n)} ) 
& \leq &
A(\hat\gamma^{(k_n)} , \lambda^{(k_n)} ) \quad \textrm{by definition
  of $\mathcal{M}$} \\
& \leq &
L(\hat\gamma^{(k_n)} + \lambda^{(k_n)}) - L(\lambda^{(k_n)}) \quad
\textrm{by Lemma \ref{A<L-L}} \\
& = &  
L( \lambda^{(k_n + 1)}) - L(\lambda^{(k_n)}) \quad \textrm{by
  definition of IM} \\
& \leq &
L( \lambda^{(k_{n+1})}) - L(\lambda^{(k_n)}) \quad \textrm{by
  monotonicity of $L(\lambda^{(k)})$,}
\end{eqnarray*}
and in the limit as $n \rightarrow \infty$, for continuous $A$ and $L$:
$A(\gamma , \bar\lambda) \leq L(\bar\lambda) -  L(\bar\lambda) = 0$.
Thus $\gamma = 0$ is a maximum of $A(\gamma , \bar\lambda)$,
using Lemma \ref{A0=0},
and $\bar\lambda$ is a fixed point of $\mathcal{M}$.
Furthermore, $\left. \frac{d}{dt} \right|_{t=0} A(t\gamma ,
\bar\lambda) = \left. \frac{d}{dt} \right|_{t=0} L(t\gamma +
\bar\lambda) = 0$,
using Lemma \ref{dA=dL},
and $\bar\lambda$ is a critical point of
$L$. 
\end{proof}
From this and Theorem \ref{IncreasingIMLikelihood} it follows
immediately that each sequence of likelihood values for which an
upper bound exists monotonically converges to a critical point of $L$.

\begin{corollary}
Let $\{ L (\lambda^{(k)} \}$ be a sequence of likelihood values
bounded from above. Then $\{ L (\lambda^{(k)} \}$ converges
monotonically to a value
$L^\ast = L(\lambda^\ast)$ for some critical point
$\lambda^\ast$ of $L$. 
\end{corollary}

Thus, the general properties of the IM algorithm are as follows: The
IM algorithm conservatively increases the incomplete-data
log-likelihood function $L$. Furthermore, it converges monotonically
to a critical point of $L$, which in almost all cases is a local
maximum. And it shows a chaotic behaviour in that for functions $L$
with several extreme values, convergence will be extremely sensitive
to the starting value of a sequence of iterates. 

\subsubsection{Relation to Generalized EM Estimation}
\label{GEM}

As discussed in Sect. \ref{MotivIM}, a direct application of the standard EM theory to
log-linear models is complicated, since complete-data MLE is
complicated for log-linear models. That is, a direct application of
the EM algorithm to log-linear models always is doubly iterative,
because the M-step itself involves some kind of iterative scaling
procedure. Examples using iterative M-steps in MLE of log-linear
models for partially classified contingency tables are given in
\citeN{Little:87}.  

Iterative M-steps can be avoided by going to partial M-steps, i.e., to
GEM algorithms, as shown in Sect. \ref{PartialM}.
In a GEM algorithm, the auxiliary function $Q$ is increased in each
M-step rather than maximized. That means, if the improved iterative
scaling algorithm is used in the M-step, a single maximization step on
the auxiliary function of this algorithm suffices to increase the
objective function of this algorithm. 
\citeN{DDL:97} use the auxiliary function 
$A_c(\gamma , \lambda) = 1 + \tilde p [\gamma \cdot \nu] 
- p_\lambda [ \sum_{i=1}^n \bar \nu_i e^{\gamma_i \nu_\#} ]$ 
for the objective complete-data log-likelihood function
$L_c(\lambda)$ = $\ln \prod_{x \in \mathcal{X}} p_\lambda(x)^{\tilde p(x)}$.
An incorporation of this complete-data MLE algorithm into a GEM
setting yields the following procedure: First, for a given sample from
$\mathcal{Y}$, the auxiliary function $Q$ for the
incomplete-data log-likelihood
$L= \ln \prod_{y \in \mathcal{Y}} g_\lambda(y)^{\tilde p(y)}$
is computed as prescribed by the E-step of the EM theory. Next,
$\lambda^{t+1}$ is set to increase $Q$. That is, we perform only a
partial M-step. This task can be fulfilled by
tuning the complete-data auxiliary function $A_c$ of \citeN{DDL:97} to a
new auxiliary function $\hat A$ for the manufactured objective
function $Q$, and by performing a one-step maximization of the
complete-data auxiliary function $ A_c$.

\begin{description}
\item[] E-step: Compute $Q(\lambda; \lambda^{(t)}) =
\tilde p [ k_{\lambda^{(t)} } [ \ln p_\lambda ] ]$ for a
log-linear model $p_\lambda$. 

\item[] M-step: Choose $\lambda^{(t+1)}$ s.t.
$Q(\lambda^{(t+1)} ; \lambda^{(t)}) \geq
Q(\lambda^{(t)}; \lambda^{(t)})$, 

i.e., $\lambda^{(t+1)} = \gamma^{(t)} + \lambda^{(t)}$
with $\gamma^{(t)} = \underset{\gamma \in \Reals}{\arg\max\;}
\hat A(\gamma , \lambda^{(t)})$,\\
and $\hat A(\gamma , \lambda^{(t)}) =
\tilde p[ 1 + k_{\lambda^{(t)}} [\gamma \cdot \nu] -
p_{\lambda^{(t)}} [\sum_{i=1}^n \bar \nu_i e^{\gamma_i \nu_\#} ]]$.
\end{description}
Note that the auxiliary function $\hat A$ which is constructed by applying 
the complete-data auxiliary function $A_c$ to the manufactured complete-data
log-likelihood $Q$  is identical to our auxiliary function $A$ as
specified in Definition \ref{A}.
From the theory of the improved iterative scaling
algorithm we can deduce that $Q$ is increased at each M-step of the above
procedure. Given this, the theory of the GEM algorithm tells us that
the incomplete-data log-likelihood $L$ also is increased at each GEM
step of the above procedure.
However, convergence of this combined procedure has yet to be studied. An
intuitive and elegant way to do this is by considering the auxiliary
function $A$ as a lower bound not only on the  manufactured
complete-data log-likelihood $Q$ but also directly on the
incomplete-data log-likelihood $L$, and prove convergence directly
from the relation of $A$ to $L$. This is the approach we took in the last section.

\subsubsection{Relation to Maximum-Entropy Estimation}
\label{ME}

The improved iterative scaling algorithm can be
seen also from the perspective of maximum-entropy estimation. 
\citeN{DDL:97} and \citeN{Berger:96} show a duality between maximum
likelihood and maximum entropy problems, which can be stated as follows.

\begin{quote}
The probability distribution $p^\ast$ with maximum entropy subject to
constraints $p[f_i] = \tilde p[f_i], i=1,\ldots,n$ from a distribution
$\tilde p(x)$ over complete data $\mathcal{X}$ is the model in
the parametric family of log-linear models $p_\lambda$ that maximizes
the likelihood of the training sample $\mathcal{X}$ distributed
according to $\tilde p(x)$.
\end{quote}

Clearly, due to the lack of a distribution $\tilde p(x)$ over complete
data $\mathcal{X}$, a similar result cannot hold for the
incomplete-data case. Rather, in each M-step we get a maximum of a
manufactured complete-data likelihood  $Q(\lambda; \lambda') =
\tilde p [ k_{\lambda'} [ \ln p_{\lambda} ] ]$ which
corresponds to a maximum-entropy solution subject to constraints from
the conditional distribution $k_{\lambda'} (x|y)$. If the
M-steps are partial themselves , i.e., if we use a GEM setting, then we
get the following ``increasing-entropy'' theorem:

\begin{quote}
The probability distribution $p^\ast$ that increases the entropy
$H(p)$ for any probability distribution
$p$ subject to the constraints
$p[f_i] = k_{\lambda'}[f_i], i=1, \ldots, n$
from a conditional distribution $k_{\lambda'} (x|y)$
is the model in the parametric family of log-linear probability
distributions $p_{\lambda}$ 
with $Q(\lambda;\lambda') \geq Q(\lambda;\lambda)$.
\end{quote}

\subsection{Property Selection}
\label{PS}

For the task of parameter estimation discussed in the last section, we
assumed a vector of properties to be given. Clearly,
exhaustive sets of properties can grow unmanageably large and must be
curtailed. An appropriate quality measure on properties can then be
used to define an algorithm for automatic property selection. 

More generally, property selection can be seen from the viewpoint of
model induction. That means, selecting prominent properties out of a
set of possible properties can be seen as incrementally inducing a
model that captures only the salient statistical qualities of the
training data. Such induced models disallow overfitting the
training data, which would be the case with models with one unique property per
training element. Instead, compact models allow generalizations to
new data and temper the overtraining problem.

Different approaches to model induction have been
presented. For example, \citeN{Stolcke:94} have given a Bayesian
approach to inducing the structure of hidden Markov models. This
approach starts with a hidden Markov model that directly encodes the data, and
proceeds by incrementally generalizing by merging states according to
a Bayesian posterior probability measure. This measure trades off the
likelihood of the data, which prefers overfitting models, against a
prior probability, which prefers simpler models. Maximization of the
posterior probability, i.e., the product of the prior and the
likelihood, determines which states to merge and when to stop generalizing. 

The property selection approach presented by \citeN{DDL:97} and
\citeN{Berger:96} proceeds from the opposite direction. Starting from
a uniform distribution over the data, which is encoded by a model with
no properties at all, properties are incrementally added to the model according to a
likelihood measure. A naive form of this measure is the improvement
in complete-data log-likelihood when extending a model by a single
candidate property $c$ with corresponding log-parameter
$\alpha$. Unfortunately, when a new parameter is added to the
parameter vector of the model, the optimal values can change for
all parameters. Thus the calculation of the likelihood-improvement due
to adding a single property requires MLE for all parameters.
Clearly, this is infeasible for models with large parameter
spaces. \citeN{DDL:97} and \citeN{Berger:96} propose an
approximate solution where the complete-data log-likelihood function is maximized
directly as a function of a single parameter $\alpha$. That is, the
improvement due to adding a single candidate is approximated by adjusting only the
parameter of this candidate and holding all other parameters
fixed. This yields a greedy algorithm which makes it practical to evaluate a large
number of candidates at each stage of the combined inference algorithm.

Let us turn now to property selection for log-linear CLP
models. For the sake of concreteness, let properties of proof trees
be specified as connected, non-overlapping subtrees of proof trees as follows:
A property of a proof tree is a connected subgraph of a
proof tree, where each node of such a subtree has either zero
descendants or the same number of descendants as the corresponding
node of the supertree, and the node sets of every two subtrees in the
set of properties must not intersect.

Suppose furthermore that properties can be incrementally constructed
by selecting from an initial set of goals and from subtrees built by
performing a resolution step at a terminal node of a subtree already
in the model.

Clearly, an exhaustive set of such properties must be pruned
according to some quality measure. What could be an appropriate
quality measure for the case of incomplete data? For a MLE framework,
the approach of \citeN{DDL:97} and \citeN{Berger:96} offers itself.
Unfortunately, we cannot apply the approximate solution of maximizing
the likelihood as a function of a single parameter $\alpha$, since the
incomplete-data log-likelihood $L$ is not concave in the parameters.
However, we can express a conservative estimate of the
likelihood-gain by instantiating the auxiliary function $A$ of
Definition \ref{A} to the extension of a model $p_{\lambda \cdot \nu}$
by a single  property $c$ with parameter $\alpha$. 
\begin{eqnarray*}
A(\alpha , \lambda) 
& = & 
\tilde p [ 1+ k_\lambda [\alpha_i c_i ] 
- p_\lambda [\sum_{i=1}^n \bar c_i e^{\alpha_i c_\#} ] ] \\
& = & 
\tilde p [ 1+ k_\lambda [\alpha c] 
- p_\lambda [e^{\alpha c} ] ] \\
& & \textrm{since } \alpha_i = \alpha, c_i(x) = c(x), c_\# (x) = c(x),
\bar c_i(x) = 1.
\end{eqnarray*}
From this, we can define an estimated likelihood-gain
$G_c(\alpha , \lambda)$ for a candidate $c$ as follows.

\begin{de}
Let $\lambda \cdot \nu(x)$ be a weighted property function,
$c$ be a candidate property,
and $\alpha \in \Reals$ the log-parameter corresponding
to $c$.
Then the estimated gain $G_c(\alpha , \lambda)$
of adding candidate property $c$ with parameter value $\alpha$ to the
log-linear model $p_{\lambda \cdot \nu}$ is defined s.t. 
\begin{center}
$ G_c(\alpha , \lambda) 
= \tilde p [ 
1 + 
k_{\lambda \cdot \nu} [\alpha c] - 
p_{\lambda \cdot \nu} [e^{\alpha c}] ]$.
\end{center}
\end{de}
Clearly, this estimated likelihood-gain $G_c(\alpha , \lambda)$
 is a lower bound on the true likelihood-gain
$L(\alpha + \lambda) - L(\lambda)$
for a parameter $\alpha$ corresponding to a property $c$.
$G_c(\alpha , \lambda)$ also is strictly concave in the parameters and
can be maximized directly and uniquely.

\begin{po} $G_c(\alpha ,  \lambda)$ takes its maximum as a
function of $\alpha$ at the unique point $\hat \alpha$ 
satisfying
\[
\tilde p [
k_{\lambda \cdot \nu} [c] ]
= 
\tilde p [
p_{\lambda \cdot \nu} [ c \: e^{\hat\alpha c} ] ].
\]
\end{po}

\begin{proof}
\[\frac{\partial}{\partial \alpha} G_c(\alpha , \lambda)
= \tilde p [ 
k_{\lambda\cdot\nu} [ c] - 
p_{\lambda\cdot\nu} [c \: e^{ \alpha c}] ] ,
\]

\[\frac{\partial^2}{\partial \alpha^2} 
 G_c(\alpha , \lambda) 
= - \tilde p [ p_{\lambda\cdot\nu} [c^2 e^{\alpha c}] ] < 0.
\qed
\]
\renewcommand{\qed}{}
\end{proof}

Property selection then will incorporate that property out of the set
of candidates that gives the greatest improvement to the model at the property's
best adjusted parameter value. Since we are interested only in
relative, not absolute gains, a single, non-iterative maximization of
the estimated gain will suffice to choose from the candidates. This
yields a greedy algorithm for approximate property selection defined
as follows.

\begin{de}[Property selection] \label{PropertySelection}
Let $C$ be a set of candidate properties,
$c \in C$ be a candidate property with log-parameter
$\alpha \in \Reals$,
and $G_c (\lambda) = \max\limits_\alpha G_c (\alpha , \lambda)$
the maximal estimated gain that property $c$ can give to model
$p_{\lambda \cdot \nu}$.
Then $c$ is selected in a property selection step for model
$p_{\lambda \cdot \nu}$ if $c = \underset{c' \in C}{\arg\max\;} G_{c'}(\lambda)$.
\end{de} 
A reasonable stopping criterion for property selection is to employ
cross-validation techniques. That is, the training corpus from
$\mathcal{Y}$ has to be divided into a training portion and a
held-out portion. Each candidate property is subjected to maximization of the
likelihood for both the training portion and the held-out portion. If the
likelihood is increasing for the training portion, but no longer for the held-out
portion, the property is discarded. The idea is that at such a point
overfitting is indicated for a set of properties that too tighly fits
the training portion (and its noise) but no longer provides a good
statistical model for both the training and held-out portion of the
training corpus. A similar approach of cross-validation can be used to
provide a stopping criterion in parameter estimation.

\subsection{Combined Statistical Inference}
\label{CSI}

The IM procedure for parameter estimation (Definition
\ref{IterativeMaximization}) and the procedure for property selection
(Definition \ref{PropertySelection}) can be combined into a statistical
inference algorithm for log-linear models from incomplete data as
shown in Table \ref{AlgCSI}.
The initial model of the Combined Statistical Inference
algorithm is assumed to be chosen according to the respective
application. For example, $p_0$ can be chosen as uniform distribution
for finite $\mathcal{X}$, or as the estimate resulting from an
applicaton of Baum's maximization technique to CLP (see
Sect. \ref{Baum}) for infinite $\mathcal{X}$.  After each
property-selection step $t$, a good starting point for parameter
estimation is a $p_0$ based upon parameter value $\hat\alpha +
\lambda ^{(t)}$, where $\hat\alpha $ is the parameter value of the selected
property $\hat c$ that maximizes the gain $G_{\hat c}(\alpha ,
\lambda^{(t)})$. Note that $\mathcal{X}$ is defined as the disjoint
union of the complete data corresponding to the incomplete data in the
random sample, i.e., $\mathcal{X} := \sum_{y \in \mathcal{Y} | \tilde
  p(y) > 0} X(y)$.

\begin{table}[htbp]
\begin{center}
\fbox{\begin{minipage}{14cm}

\begin{description}
\item[{Input}] Initial model $p_0$, incomplete-data sample from
  $\mathcal{Y}$.

\item[{Output}] Log-linear model $p^\ast$ on complete-data sample
$\mathcal{X} = \sum_{y \in \mathcal{Y} | \tilde
  p(y) > 0} X(y)$ with selected property function vector $\nu^\ast$ and 
log-parameter vector $\lambda^\ast = \underset{\lambda \in
\Lambda}{\arg\max\;} L(\lambda) $ 
where $\Lambda = \{ \lambda | \; p_\lambda$ 
is a log-linear model on $\mathcal{X}$ 
based on $p_0$, $\nu^\ast$ and $\lambda \in \Rn \} $.

\item[Procedure] \mbox{}

\begin{enumerate}
\item $p^{(0)} := p_0$ with $C^{(0)} := \emptyset$,

\item Property selection: For each candidate property $c \in C^{(t)}$,
compute the gain 
$G_c(\lambda^{(t)}) := \max\limits_{\alpha \in \Reals}
G_c(\alpha , \lambda^{(t)})$, 
and select the property $\hat c := \underset{c \in C^{(t)}}{\arg\max\;}
G_c(\lambda^{(t)})$.

\item Parameter estimation: Compute a maximum likelihood parameter
  value 
$\hat \lambda := \underset{\lambda\in\Lambda}{\arg\max\;} L(\lambda)$ 
where $\Lambda = \{ \lambda | \; p_\lambda (x) $
is a log-linear distribution on $\mathcal{X}$
with initial model $p_0$,
property function vector $\hat \nu := (\nu_1^{(t)}, \nu_2^{(t)},
  \ldots, \nu_n^{(t)}, \hat c)$, 
and $\lambda \in \Reals^{n+1} \} $.

\item Until the model converges, set \\
$p^{(t+1)} := p_{\hat\lambda \cdot \hat\nu}$, \\
$t := t+1$, \\
go to $2$.

\end{enumerate}
\end{description}

\end{minipage}}
\caption{Algorithm (Combined Statistical Inference)}
\label{AlgCSI}
\end{center}
\end{table}

\begin{figure}[htbp]
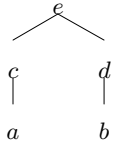

\begin{center}

{\footnotesize
\setlength{\GapWidth}{30pt}
\setlength{\GapDepth}{10pt}

\begin{bundle}{$e$}
  \chunk{
    \begin{bundle}{$c$}
      \chunk{$a$}
    \end{bundle}
    }
  \chunk{
    \begin{bundle}{$d$}
      \chunk{$b$}
    \end{bundle}
    }
\end{bundle}
}

\end{center}
\caption{Type hierarchy}
\label{PCLPtypes}
\end{figure}

Let us illustrate this procedure with a simple CLP example. Suppose our
sample program is the same as in Fig. \ref{PCLPprogram} but with
\La-constraints taken from a language of hierarchical types. The
ordering on the types is defined by the operation of set inclusion on
the denotations of the types and depicted graphically in
Fig. \ref{PCLPtypes}.

\begin{figure}[htbp]
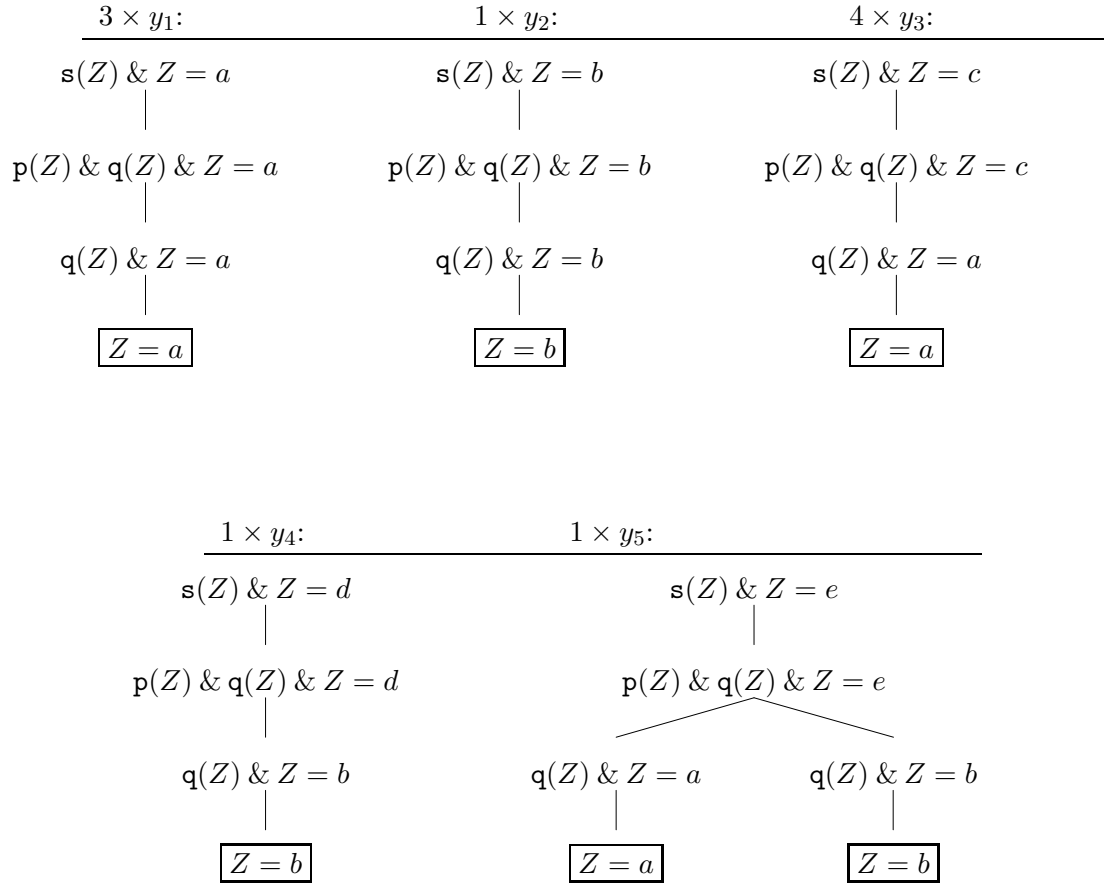


\begin{center} 
\begin{tabular}{p{130pt}p{130pt}p{90pt}}

$3 \times y_1$: & $1 \times y_2$: & $4 \times y_3$: \\ \hline

\begin{bundle}{$\texttt{s}(Z)\:\&\: Z=a$}
  \chunk
  {
    \begin{bundle}{$\texttt{p}(Z) \:\&\: \texttt{q}(Z) \:\&\: Z=a$}
      \chunk
      {
        \begin{bundle}{$\texttt{q}(Z) \:\&\: Z=a$}
          \chunk
          {\fbox{$Z=a$}}
        \end{bundle}
        }
    \end{bundle}
    }
\end{bundle}

& 

\begin{bundle}{$\texttt{s}(Z)\:\&\: Z=b$}
  \chunk
  {
    \begin{bundle}{$\texttt{p}(Z) \:\&\: \texttt{q}(Z) \:\&\: Z=b$}
      \chunk
      {
        \begin{bundle}{$\texttt{q}(Z) \:\&\: Z=b$}
          \chunk
          {\fbox{$Z=b$}}
        \end{bundle}
        }
    \end{bundle}
    }
\end{bundle}

&

\begin{bundle}{$\texttt{s}(Z)\:\&\: Z=c$}
  \chunk
  {
    \begin{bundle}{$\texttt{p}(Z) \:\&\: \texttt{q}(Z) \:\&\: Z=c$}
      \chunk
      {
        \begin{bundle}{$\texttt{q}(Z) \:\&\: Z=a$}
          \chunk
          {\fbox{$Z=a$}}
        \end{bundle}
        }
    \end{bundle}
    }
\end{bundle}

\end{tabular}
\end{center}

\vspace{5ex}

\begin{center} 
  \begin{tabular}{p{120pt}p{150pt}}
    
    $1 \times y_4$: & $1 \times y_5$: \\ \hline

    \begin{bundle}{$\texttt{s}(Z)\:\&\: Z=d$}
      \chunk
{
  \begin{bundle}{$\texttt{p}(Z) \:\&\: \texttt{q}(Z) \:\&\: Z=d$}
    \chunk
    {
      \begin{bundle}{$\texttt{q}(Z) \:\&\: Z=b$}
        \chunk
        {\fbox{$Z=b$}}
      \end{bundle}
      }
  \end{bundle}
  }
\end{bundle}

&

\setlength{\GapWidth}{70pt}
\begin{bundle}{$\texttt{s}(Z)\:\&\: Z=e$}
  \chunk
  {
    \begin{bundle}{$\texttt{p}(Z) \:\&\: \texttt{q}(Z) \:\&\: Z=e$}

      \chunk
      {
        \begin{bundle}{$\texttt{q}(Z) \:\&\: Z=a$}
          \chunk
          {\fbox{$Z=a$}}
        \end{bundle}
        }

      \chunk
      {
        \begin{bundle}{$\texttt{q}(Z) \:\&\: Z=b$}
          \chunk
          {\fbox{$Z=b$}}
        \end{bundle}
        }
      
    \end{bundle}
    }
\end{bundle}

\end{tabular}

\end{center}

\caption{Queries and proof trees for constraint logic program}
\label{QueriesProoftrees}
\end{figure}

Furthermore, suppose we have a training corpus of ten queries,
consisting of three tokens of query $y_1: \: \texttt{s}(Z)\:\&\: Z=a$,
four tokens of $y_3: \: \texttt{s}(Z)\:\&\: Z=c$, and one token each of query
$y_2:\: \texttt{s}(Z)\:\&\: Z=b,
\; y_4: \: \texttt{s}(Z)\:\&\: Z=d$, and
$y_5: \: \texttt{s}(Z)\:\&\: Z=e$.
The corresponding proof trees generated by the program in
Fig. \ref{PCLPprogram} are given in Fig. \ref{QueriesProoftrees}. Note that
queries $y_1$, $y_2$, $y_3$ and $y_4$ are unambiguous, being assigned
a single proof tree, while $y_5$ is ambiguous.

A useful first distinction between the proof trees of
Fig. \ref{QueriesProoftrees} can be obtained by
selecting the two subtrees $\chi_1: \fbox{$Z=a$}$ and $\chi_2:
\fbox{$Z=b$}$ as properties. These properties allow us to cluster the
proof trees in two disjoint sets on the basis of similar statistical
qualities of the proof threes in these sets.
Since in our training corpus seven out of ten queries
come unambiguously with a proof tree including property $\chi_1$, we
would expect the maximum likelihood parameter value corresponding to
property $\chi_1$ to be higher than the parameter value of property
$\chi_2$. However, we cannot simply recreate the proportions of the
training data from the corresponding proof trees as we did in the
unambiguous example of Sect. \ref{LLM}. Here we are confronted with an
incomplete-data problem, which means that we do not know the frequency
of the possible proof trees of query $y_5$. 

Let us apply the IM algorithm to this incomplete-data problem. For the selected
properties $\chi_1$ and $\chi_2$, we have
$\nu_\#(x) = \nu_1(x) + \nu_2(x) = 1$
for all possible proof trees $x$ 
for the sample of Fig. \ref{QueriesProoftrees}. Thus the parameter updates
$\hat\gamma_i$ can be calculated from a particularly simple closed
form $
\hat\gamma_i = \ln \frac{\tilde p [ k_\lambda [\nu_i]]}
{ p_\lambda [\nu_i]}.
$
A sequence of IM iterates is
given in Table \ref{IMestimation}. 
Probabilities of proof trees involving property $\chi_i$ are
denoted by $p_i$. Starting from an initial uniform probability
of $1/6$ for each proof tree, this sequence of likelihood values
converges with an accuracy in the third place after the decimal point
after three iterations and yields probabilities $p_1 \approx .259$ and
$p_2 \approx .074$ for the respective proof trees.

\begin{table}[htbp]
\begin{center}
\begin{tabular}{c|l|l|l|l|c}
Iteration $t$& $\lambda_1^{(t)}$ & $\lambda_2^{(t)}$ & $p_1^{(t)}$ & $p_2^{(t)}$ & $L(\lambda^{(t)})$ \\
\hline

0 & 0 & 0 & $1/6$ & $1/6$ & $-17.224448$ \\ \hline

1 & $\ln 1.5$ & $\ln .5$ & $.25$ & $.08\dot{3}$ & $-15.772486$ \\
\hline

2 & $\ln 1.55$ & $\ln .45$ & $.258\dot{3}$ & $.075$ & $-15.753678$ \\
\hline

3 & $\ln 1.555$ & $\ln .445$ & $.2591\dot{6}$ & $.0741\dot{6}$ &
$-15.753481$ 
\end{tabular}
\end{center}
\caption{Estimation using the IM algorithm}
\label{IMestimation}
\end{table}

\section{An Experiment}
\markright{4.7 An Experiment}
\label{Experiment}

In this section we present an empirical evaluation of the
applicability of log-linear probability models and iterative
scaling techniques to constraint-based grammars. We present
a computationally tractable maximum pseudo-likelihood estimation
procedure for log-linear models and apply it to estimating a probabilistic constraint-based
grammar from a small corpus of LFG analyses provided by Xerox
PARC. The log-linear models employ a small set of about 200 properties
to induce a probability distribution on 3000 parses where on average
each sentence is ambiguous in 10 parses. The empirical evaluation
shows that the correct parse from the set of all parses is found about
59 \% of the time. 

This section is based on joint work described in \citeN{Johnson:99}.

\subsection{Incomplete-Data Estimation as Maximum Pseudo-Likelihood Estimation
  for Complete Data}

As we saw in Sect. \ref{SI}, the equations to be solved in statistical
inference of log-linear models involve the computation of
expectations of property-functions $\nu_i(x)$ with respect to
$p_\lambda(x)$. Clearly it is possible to find constraint-based
grammars where the sample space $\mathcal{X}$ of parses to be summed
over in these expectations is unmanageably large or even infinite. 

One possibility to sensibly reduce the summation space is to employ the
definition of the sample space $\mathcal{X}:= \sum_{y \in \mathcal{Y}|
  \tilde p(y) > 0} X(y)$ used in incomplete-data estimation as a
reduction factor in complete-data estimation. That is, we approximate
expectations with respect to the distribution $p_\lambda(\cdot)$ on
$\mathcal{X}$ by considering only such parses $x \in \mathcal{X}$
whose terminal yield $y = Y(x)$ is seen in the training corpus.
Furthermore, the distribution $g_\lambda(y)$ on terminal yields is
replaced by the empirical distribution $\tilde p(y)$:
\begin{eqnarray*}
p_\lambda[\nu_i] & = & \sum_{x \in \mathcal{X}} p_\lambda(x) \nu_i(x)
\\
& = & \sum_{y \in \mathcal{Y}} \sum_{x \in X(y)} p_\lambda(x) \nu_i(x)
  \\
& = & \sum_{y \in \mathcal{Y}} g_\lambda(y) \sum_{x \in X(y)}
  k_\lambda(x|y) \nu_i(x) \\
& \approx & \sum_{y \in \mathcal{Y}} \tilde p(y) \sum_{x \in X(y)}
  k_\lambda(x|y) \nu_i(x).
\end{eqnarray*}
Clearly, for most cases the approximate expectation is easier to calculate since the
space $\sum_{y \in \mathcal{Y}| \tilde p(y) > 0}
X(y)$ is smaller than the original full space $\mathcal{X}$. 

The equations to be solved in complete-data estimation for log-linear
models are then
\[ 
\sum_{y \in \mathcal{Y}} \tilde p(y) \sum_{x \in X(y)}
k_\lambda(x|y) \nu_i(x) 
= \sum_{x \in \mathcal{X}} \tilde p(x) \nu_i(x) \textrm{ for all } i=1,
\ldots, n.
\] 
These equations are solutions to the maximization problem of
another criterion, namely a complete-data log-pseudo-likelihood function
$PL_c$ which is defined with respect to
the conditional probability of parses given the yields observed in the
training corpus.
\[
PL_c(\lambda) = \ln \prod_{x \in \mathcal{X}, y \in \mathcal{Y}}
k_\lambda(x|y)^{\tilde p(x,y)}
\]

In the actual implementation described in \citeN{Johnson:99} a
slightly different function involving a regularization term promoting
small values of $\lambda$ onto the objective function was maximized.
The maximization equations were solved 
using a conjugate-gradient approach adapted from \citeN{Press:92}. A
similar approach to maximum pseudo-likelihood estimation for
log-linear models from complete data but in the context of an
iterative scaling approach can be found in \citeN{Berger:96}.

\subsection{Property Design for Feature-Based CLGs}

One central aim of our experiment was to take advantage of the high
flexibility of log-linear models and evaluate the usefulness of this
issue in hard terms of empirical performance. 

The properties employed in our models
clearly deviate from the rule or production properties employed in
most other probabilistic grammars by encoding as property-functions general linguistic
principles as proposed by \citeN{Alshawi:94}, \citeN{Srinivas:95} or
\citeN{Hobbs:95}. The definition of properties
of LFG parses refers to both the c(onstituent)- and
f(eature)-structures of the parses. Examples for the properties employed
in our model are

\begin{itemize}
\item properties counting the number of adjuncts, arguments and segments in an analysis,
\item properties corresponding to grammatical functions used in LFG,
  including SUBJ, OBJ, OBJ2, COMP, XCOMP, ADJUNCT, etc.
\item properties measuring the complexity of the phrase being
  attached to, thus indicating both high and low attachment, 
\item properties indicating non-right-branching of nonterminal nodes,
\item properties indicating non-parallel coordinate structures,
\item properties for atomic attribute-value pairs in feature
  structures,
\item properties for particular syntactic structures such as date-NPs,
\item standard rule-properties.
\end{itemize}

The number of properties defined for each of the two
corpora we worked with was about 200 including about 50
rule-properties respectively.

We would also have liked to have included 
properties corresponding to lexical-semantic head-head relations, but
found the small size of our training corpora to be an obstacle in estimating the associated
parameters accurately.

\subsection{Empirical Evaluation}

The two corpora provided to us by Xeroc PARC contain appointment
planning dialogs (Verbmobil corpus, henceforth VM-corpus),
and a documentation of Xerox printers (Homecentre corpus, henceforth
HC-corpus). The basic properties of the corpora
are summarized in Table \ref{corpora}. The corpora consist of a packed
representation of the c- and f-structures of parses produced for the sentences by a LFG grammar.
The LFG parses have been produced automatically by the XLE system (see
\citeN{MaxKap:89}) but corrected manually in addition. Furthermore, it
is indicated for each sentence which of its parses is the
linguistically correct one. The ambiguity of the sentences in the
corpus is 10 parses on average. 

\begin{table}[htbp]
\renewcommand{\arraystretch}{1.3}
\begin{center}
\begin{tabular}{|l|c|c|}\hline
& VM-corpus & HC-corpus \\ \hline
number of sentences
& 540 
& 980 \\
number of ambiguous sentences
& 314
& 481 \\
number of parses of ambiguous sentences
& 3245
& 3169 \\
\hline
\end{tabular}
\caption{Properties of the corpora used for the estimation experiment}
\label{corpora}
\end{center}
\end{table}

In order to cope with the small size of the corpora a 10-way
cross-validation framework has been used for estimation and
evaluation. That is, the sentences of each corpus were assigned randomly
into 10 approximately equal-sized subcorpora. In each run, 9 of the
subcorpora served as training corpus, and one subcorpus as test
corpus. The evaluation scores presented in Tables \ref{VMevaluation}
and \ref{HCevaluation} are
sums over the the evaluation scores gathered by using each subcorpus
in turn as test corpus and training on the 9 remaining subcorpora.

We used two evaluation measures on the test corpus. The first measure
$C_\mathrm{test}(\lambda)$ gives the accuracy of disambiguation based
on most probable parses. That
is, $C_\mathrm{test}(\lambda)$ counts the percentage of sentences in
the test corpus whose most probable parse according to a model
$p_\lambda$ \emph{is} the manually determined correct parse. If a sentence has $k$
most probable parses and one of these parses is the correct one, this
sentence gets score $1/k$. 
The second evaluation measure is $- PL_\mathrm{test}(\lambda)$, the
negative log-pseudo-likelihood for the correct parses of the test
corpus given their yields. This metric measures how much of the
probability mass the model puts onto the correct analyses. 

In the empirical evaluation, the maximum pseudo-likelihood estimator
is compared against a baseline estimator which treats all parses as
equally likely. Furthermore, another objective function is considered:
The function $C_\mathcal{\tilde X}(\lambda)$ is the number of times
the highest weighted parse under $\lambda$ is the manually determined
correct parse in the training corpus $\mathcal{\tilde X}$. This
function directly encodes the criterion which is used in the
linguistic evaluation. 
However, $C_\mathcal{\tilde X}(\lambda)$ is a highly discontinuous function in $\lambda$ and
hard to maximize. Experiments using a simulated annealing optimization
procedure \cite{Press:92} for this objective function showed that the
computational difficulty of this procedure grows and the quality of
the solutions degrades rapidly with the number of properties employed
in the model.

The results of the empirical evaluation are shown in Tables
\ref{VMevaluation} and \ref{HCevaluation}. The maximum
pseudo-likelihood estimator performed superior to both the simulated
annealing estimator and the uniform baseline estimator on both
corpora. The simulated annealing procedure typically scores better
than the maximum pseudo-likelihood approach if the number of
properties is very small. However, the pseudo-likelihood approach
outperforms simulated annealing already for a property-size of 200 as
used in our experiment. Furthermore it should be noted that the
absolute numbers of 59 \% accuracy on the disambiguation task have to
be assessed relative to a number of on average 10 parses per sentence.

\begin{table}[htbp]
\renewcommand{\arraystretch}{1.3}
\begin{center}
\begin{tabular}{|l|c|c|}\hline
& $C_\mathrm{test}$ for VM-corpus  & $- PL_\mathrm{test}$ for
VM-corpus \\ \hline
uniform baseline estimator
& 9.7 \% & 533 \\
simulated annealing estimator
& 53.7 \%  & 469 \\
maximum pseudo-likelihood estimator
& 58.7 \% & 396 \\
\hline
\end{tabular}
\caption{Empirical evaluation of estimators on  $C_\mathrm{test}$ (accuracy of
  disambiguation with most probable parse) and $-PL_\mathrm{test}$
  (negative log-pseudo-likelihood of correct parses in test corpus) on
  VM-corpus}
\label{VMevaluation}
\end{center}
\end{table}

\begin{table}[htbp]
\renewcommand{\arraystretch}{1.3}
\begin{center}
\begin{tabular}{|l|c|c|}\hline
& $C_\mathrm{test}$ for HC-corpus & $- PL_\mathrm{test}$ for HC-corpus \\ \hline
uniform baseline estimator
& 15.2 \% & 655 \\
simulated annealing estimator
& 53.2\%  & 604 \\
maximum pseudo-likelihood estimator
& 58.8 \%  & 583 \\
\hline
\end{tabular}
\caption{Empirical evaluation of estimators on HC-corpus}
\label{HCevaluation}
\end{center}
\end{table}

\section{Approximation Methods}
\markright{4.8 Approximation Methods}
\label{Approximation}

With the algorithms and proofs of the preceding sections in hand, it
seems that statistical inference of log-linear models from incomplete data
reduces to solving simple equations and computing expectations of
simple functions. However, depending on the size of the sample spaces
over which these expectations must be taken and depending on the complexity of the
parameter- and property-space, these equations can become intractable
both analytically and numerically. In order to give a self-contained
recipe for statistical inference of log-linear models from incomplete
data, we will discuss the possibilities of applying various
approximation methods to achieve both analytical and computational
tractability in complex applications. 

\subsection{Enforcing a Closed-Form Solution}
\label{ClosedForm}

As mentioned above, if the property-functions sum to a constant independent of $x$, i.e., if
\[
\nu_\#(x) = \sum^n_{i=1} \nu_i(x) = K \textrm{ for all } x \in
\mathcal{X}, 
\]
then the maximum $\hat\gamma$ of the auxiliary function $A$ used in
parameter estimation is given in closed form.

For a given vector of property-functions $\nu$ with $\nu_\#(x)=K$, the
IM algorithm can be stated as shown in Table \ref{AlgPECF}. Note that
the complete-data sample $\mathcal{X}$ is computed as $\mathcal{X} =
\sum_{y \in \mathcal{Y}| \tilde p(y) > 0} X(y)$. 

\begin{table}[htbp]
\begin{center}
\fbox{\begin{minipage}{14cm}

\begin{description}
\item[Input] Initial model $p_0$, 
  property-functions vector $\nu$, 
 incomplete-data sample from $\mathcal{Y}$.

\item[Output] MLE model $p_{\lambda^\ast}$ on $\mathcal{X} =
  \sum_{y \in \mathcal{Y}| \tilde p(y) > 0} X(y)$.

\item[Procedure] \mbox{}

\begin{tabbing}
Until \= convergence do \\
\> Compute $p_\lambda, \; k_\lambda$, based on $\lambda = (\lambda_1,
\ldots, \lambda_n)$, \\
\> For \= $i$ from $1$ to $n$ do \\
\>\> $\gamma_i := 
\frac{1}{K} \ln
\frac{\sum_{y \in \mathcal{Y}} \tilde p(y) 
\sum_{x \in X(y)} 
k_{\lambda} (x|y) \nu_i(x)}
{ \sum_{y \in \mathcal{Y}} \sum_{x \in \mathcal{X}} p_{\lambda} (x) \nu_i (x)} $, \\
\>\> $\lambda_i := \lambda_i + \gamma_i$, \\
Return $\lambda^\ast = (\lambda_1, \ldots, \lambda_n)$.
\end{tabbing}
\end{description}

\end{minipage}}
\caption{Algorithm (Iterative Maximization, Closed-Form)}
\label{AlgPECF}
\end{center}
\end{table}

In this case, the IM algorithm can be seen as an
incomplete-data version of the generalized iterative scaling algorithm of \citeN{Darroch:72}. 

If the constancy-condition is not fulfilled, it can be
enforced by introducing a ``correction'' property-function $\nu_l$ as
follows:

\begin{quote}
Choose $K = \max_{x \in \mathcal{X}} \: \nu_\#(x)$ and
$\nu_l(x) = K - \nu_\#(x)$ for all $x \in \mathcal{X}$,\\
then $\sum^l_{i=1} \nu_i(x) = K$ for all $x \in \mathcal{X}$.
\end{quote}
Unfortunately, defining a correction property can be expensive, e.g.,
in case a property selection procedure is used in statistical inference, a
correction property has to be defined after each property selection
step.

Correction properties can be avoided by letting $\nu_\#$ vary over $x
\in \mathcal{X}$. This approach is also claimed to improve the
convergence rate of iterative scaling methods by increasing the step
size taken toward the maximum at each iteration.

\subsection{Numerical Approximation via Newton's Method}
\label{Newton}

If $\nu_\#(x)$ does not add up to a constant for all $x \in \mathcal{X}$,
the solutions to the maximization equations in parameter estimation
and property selection cannot, in general, be determined in closed
form. Fortunately, numerical methods such as Newton's method can be
used to efficiently compute approximate solutions to these equations.

Newton's method approximates the solution $\alpha$ of an
equation $f(\alpha) = 0$ by using a sequence of linearizations of
$f$. At each step, the intersection of the tangent to
$f$ at $\alpha_t$ with the $\alpha$-axis is taken, yielding an
improved estimate $\alpha_{t+1}$. The iteration formulae to approach
the solution up to a desired accuracy are defined as follows.

\begin{center}
$\alpha_{t+1} = \alpha_t - \frac{f(\alpha_t)}
{f'(\alpha_t)}$ where $f'(\alpha_t)$ is the derivative of $f$ at
$\alpha_t$.
\end{center}

This method directly suits our application when we replace
$f(\alpha)$ by the first derivative of the auxiliary function $A$, $\frac{\partial}{\partial \gamma_i} A(\gamma ,
\lambda)$, in case of parameter estimation, and by the first derivative
of the approximate gain $G_c$, $\frac{\partial}{\partial \alpha}
G_c(\alpha , \lambda)$, in case of property selection. Newton's method
usually converges rapidly for such functions.

To efficiently compute the functions in the Newton formulae, we can
use a cashing technique similar to the one used in \citeN{Abney:97}
and apply it to our incomplete-data problem. First, we have to define tables of
total probabilities as follows. 

\begin{itemize}
\item $S_{i,v} = \sum_{x \in \mathcal{X}}  p_\lambda(x)
\delta_{\nu_i(x),v} $
is the expexted number of times property function $\nu_i$ takes value $v$,
\item $T_{i,y}=\sum_{x \in X(y)} k_\lambda(x|y)
\nu_i(x)$
is the conditionally expected number of times property $\chi_i$ occurs,
\item $U_{i,m}=\sum_{x \in \mathcal{X} | \; 
  \nu_\# ( x) = m} p_\lambda(x) \nu_i(x)$ 
is the expected number of times property $\chi_i$ occurs when there
is a total number of $m$ property instances.
\end{itemize}
Corresponding to these expectations, we define the following counting variables:

\begin{itemize}
\item $s_r(\alpha,i) = \sum_v S_{i,v} e^{\alpha v} v^r$, 
\item $t_r(\alpha,i) = \sum_y T_{i,y}\alpha^r $, 
\item $u_r(\alpha, i) = \sum_m U_{i,m} e^{\alpha m} m^r$.
\end{itemize}
The Newton formulae for property selection can then be filled with
these expected counts as follows:

\begin{eqnarray*}
\alpha_{t+1}
& = & 
\alpha_t +
\frac{ \frac{\partial}{\partial \alpha_t}
G_c(\alpha_t , \lambda) }
{ \frac{\partial^2}{\partial \alpha_t^2}
G_c(\alpha_t , \lambda) }
\\
& = & 
\alpha_t +
\frac{\tilde p [ k_\lambda [c] - N p_\lambda
[ c \: e^{\alpha_t c} ] }
{N p_\lambda [c^2 e^{\alpha_t c} ]}
\\
 &  = & 
\alpha_t +
\frac{t_0(\alpha_t,c) - N s_1(\alpha_t,c)}
{ N s_2(\alpha_t,c)}.
\end{eqnarray*}

The tables of total probabilities defined above also allow us to express
the gain $G_c(\hat\alpha , \lambda)$ of adding property $c$
with best parameter value $\hat\alpha$ to model $p_\lambda$ in terms
of expected counts:

\begin{eqnarray*}
G_c(\hat\alpha , \lambda) 
& = & 
N + \tilde p [ k_\lambda [\hat\alpha c] - N p_\lambda
[e^{\hat\alpha c} ] \\
& = & 
N + t_1(\hat\alpha,c) - N s_0(\hat\alpha, c).
\end{eqnarray*}

For the task of parameter estimation, similar Newton formulae can be
obtained from the expected counts:

\begin{eqnarray*}
\alpha_{t+1}
& = & 
\alpha_t +
\frac{ \frac{\partial}{\partial \alpha_t}
A(\gamma , \lambda) }
{ \frac{\partial^2}{\partial \alpha_t^2}
A(\gamma , \lambda) }
\\
& = & 
\alpha_t +
\frac{\tilde p [ k_\lambda [\nu_i] - N p_\lambda [\nu_i e^{\alpha_t
  \nu_\#}]}
{N p_\lambda [\nu_i \nu_\# e^{\alpha_t \nu_\#} ]}
\\
 &  = & 
\alpha_t +
\frac{ t_0(\alpha_t,i) - N u_0(\alpha_t, i)}
{ N u_1(\alpha_t, i)}.
\end{eqnarray*}

For a random sample from $\mathcal{Y}$ of size $N$, an algorithm for
approximate parameter estimation can be defined from the above Newton
formulae as shown in Table \ref{AlgPENE}.

\begin{table}[htbp]
\begin{center}
\fbox{\begin{minipage}{14cm}

\begin{description}
\item[Input] Initial model $p_0$,
  property-functions vector $\nu$, 
 incomplete-data sample from $\mathcal{Y}$.

\item[Output] MLE model $p_{\lambda^\ast}$ on $\mathcal{X} =
  \sum_{y \in \mathcal{Y}| \tilde p(y) > 0} X(y)$.

\item[Procedure] \mbox{}

\begin{tabbing}
Until \= convergence do \\
\> Compute tables $T$, $U$, based on $\lambda = (\lambda_1, \ldots,
\lambda _n)$,\\
\> For \= $i$ from $1$ to $n$ do \\
\>\> $\alpha := 0$,\\
\>\> Until \= $\alpha$ is accurate enough do \\
\>\>\> $u_0:=0,\: u_1:=0,\: t_0:=0$,\\
\>\>\> For $m$ \= from $0$ to $m_{\max}$ do \\
\>\>\>\> $a:= U_{i,m} e^{\alpha m}$,\\
\>\>\>\> $u_0:=u_0+a$, \\
\>\>\>\> $u_1 := u_1 + am$,\\
\>\>\> For $y \in \mathcal{Y}$ where $\tilde p(y) > 0$ do \\
\>\>\>\> $b:= T_{i,y}$,\\
\>\>\>\> $t_0 := t_0 + b$, \\
\>\>\> $\alpha := \alpha + \frac{t_0-Nu_0}{Nu_1}$,\\
\>\> $\lambda_i := \lambda_i + \alpha$, \\
Return $\lambda^\ast = (\lambda_1, \ldots, \lambda_n)$.
\end{tabbing}

\end{description}

\end{minipage}}
\caption{Algorithm (Iterative Maximization, Newton-Estimate)} 
\label{AlgPENE}
\end{center}
\end{table}

Similarly, an algorithm for approximate property selection can be
given as in Table \ref{AlgPSNE}. 

\begin{table}[htbp]
\begin{center}
\fbox{\begin{minipage}{14cm}

\begin{description}
\item[Input] Model $p_\lambda$, set of candidate properties $C$,
 incomplete-data sample from $\mathcal{Y}$.

\item[Output] Selected property $c^\ast$ with maximal parameter value
  $\alpha^\ast$. 

\item[Procedure] \mbox{}

\begin{tabbing} 
Compute tables $S$, $T$, based on $\lambda$,\\
$G^\ast := 0, \; c^\ast := \emptyset$, $\alpha^\ast := 0$,\\
For \= all candidates $c \in C$ do \\
\> $\alpha := 0$, \\
\> Until \= $\alpha$ is accurate enough do \\
\>\> $s_0 :=0,\: s_1:=0,\: s_2:=0,\: t_0:=0, \: t_1:=0$,\\
\>\> For $v$ \= from $0$ to $v_{\max}$ do \\
\>\>\> $a:= S_{c,v} e^{\alpha v}$, \\
\>\>\> $s_0 := s_0 + a$, \\
\>\>\> $s_1 := s_1 + av$, \\
\>\>\> $s_2 := s_2 + av^2$,\\
\>\> For $y \in \mathcal{Y}$ where $\tilde p(y) > 0$ do \\
\>\>\> $b:=T_{i,y}$,\\
\>\>\> $t_0:=t_0+b$,\\
\>\>\> $t_1:=t_1+b\alpha$,\\
\>\> $\alpha := \alpha + \frac{t_0-Ns_1}{Ns_2}$,\\
\> $G:= N+t_1-Ns_0$,\\
\> If $G >G^\ast$, then $G^\ast := G, \: c^\ast := c, \: \alpha^\ast :=
\alpha$. \\
Return $c^\ast, \: \alpha^\ast$.
\end{tabbing}

\end{description}

\end{minipage}}
\caption{Algorithm (Property Selection, Newton-Estimate)}
\label{AlgPSNE}
\end{center}
\end{table}

\subsection{Approximating Expectations via Monte Carlo Methods}
\label{MonteCarlo}

Independent of whether the solutions of the maximization
equations exist in closed form, a further problem arises in
connection with large or infinite sample spaces. That is, if the sample space
$\mathcal{X}$ is too large to be summed over in the calculation of the
expectations in the maximization equations, methods must be used to
approximate these expectations.  

One possibility is to use Monte Carlo Methods. Following
\citeN{Abney:97}, we use the Metropolis-Hastings method and show how
it can be applied to our incomplete-data problem. 

The strategy behind this method is to generate a random sample from a
target distribution $p$ by choosing a nominating matrix $p'$ from
which sampling is easy, and performing a Bernoulli trial with parameter
$\alpha$ to determine whether to accept or reject the nominated
sample point. That means, this method converts a sampler for $p'$
into a sampler for $p$ via an evaluation matrix $\alpha$.
For our application, we can take as nominating matrix for each query
$y \in \mathcal{Y}$ a stochastic context-free CLP model
$p_\pi(x)$ on $X(y)$  as defined in Sect. \ref{Baum}. From this
stochastic derivation model sampling is easy and can be converted by a
standard evaluation matrix to sampling from the desired log-linear
distribution $p_\lambda(x)$ on $X(y)$. 

Following standard textbooks such as \citeN{Fishman:96}, an
application of the Metropolis-Hastings algorithm to our problem is as
shown in Table \ref{MH}.

\begin{table}
\begin{center}
\fbox{\begin{minipage}{14cm}

\begin{description}

\item[Input] Initial state $x_0 \in X(y)$, \\
Nominating matrix $p' = p_\pi(x)$ on $X(y)$,\\
Log-linear distribution $p = p_\lambda(x)$ on $X(y)$,\\
Evaluation matrix $\alpha_{x,z} = 
\left\{ \begin{array}{ll}
1 & \textrm{ if } p(x) p'(z) \leq p(z) p'(x) \\
\frac{p(z)p'(x)}{ p(x)p'(z)} & \textrm{ if } 
p(x) p'(z) > p(z) p'(x)  
\end{array} 
\right.$, \\ 
Terminal number of steps $k$.

\item[Output] Random sample $X_0, \ldots, X_k$
from $p_\lambda$ on $X(y)$.

\item[Procedure] \mbox{} 
\begin{tabbing}
$X_0 := x_0$, \\
$i := 1$ ,\\
While \=  $i \leq k$ \\
\> $x := X_{i-1}$, \\
\> Randomly generate $z$ from $p'$, \\
\> If \= $z = X_{i-1}$ , then $X_i := X_{i-1}$, \\
\> \> Else evaluate $\alpha_{x,z}$, \\
\> \> Randomly generate $u$ from uniform distribution on $[0,1]$, \\
\> \> If \= $u \leq \alpha_{x,z}$ , then $X_i := z$ , \\
\> \> \> Else $X_i := X_{i-1}$, \\
\> $i:= i + 1$, \\
Return $X_0, \ldots, X_k$.
\end{tabbing}

\end{description}

\end{minipage}}
\caption{Algorithm (Metropolis-Hastings Sampling)}
\label{MH}
\end{center}
\end{table}

Note that the evaluation matrix $\alpha_{x,z}$ reduces to a
particularly simple form for our application which does not require the
computation of normalization constants $Z_\lambda$. That is, by taking
the initial model $p_0$ of the log-linear CLP model $p_\lambda$ to be
of the form of a stochastic CLP model $p_\pi$, and by assuming 
independence of the nominated sample points, we get the following form
of $\alpha_{x,z}$:

\begin{eqnarray*}
\alpha_{x,z} = \min \left( 1, \frac{p(z)p'(x)}{p(x)p'(z)} \right)
\textrm{ where }
\frac{p(z)p'(x)}{p(x)p'(z)} 
& = & 
\frac{p_\lambda(z) p_\pi(x)}{p_\lambda(x) p_\pi(z)}\\
& = & 
\frac{Z_\lambda^{-1} e^{\lambda \cdot \nu(z)} p_\pi(z) p_\pi(x)}
{Z_\lambda^{-1} e^{\lambda \cdot \nu(x)} p_\pi(x) p_\pi(z)} \\
& = & \frac{e^{\lambda \cdot \nu(z)}}
{e^{\lambda \cdot \nu (x)}}\\
& = & e^{(\lambda \cdot \nu(z) -\lambda \cdot \nu(x))}.
\end{eqnarray*}

It can be shown for this sampling method that the  distribution of the
i.i.d. random variables  $X_i$ converges in distribution to the target distribution
$p_\lambda$ as $i\rightarrow \infty$:
\[
\lim\limits_{i \rightarrow \infty} P(X_i = x) = p_\lambda(x)
\textrm{ for all } x \in X(y).
\]
Furthermore, a proper random sample  from a probability distribution
$p$ enables to estimate expectations of functions $f$ 
with respect to  $p$ directly from the sample points $X_i$. That is,
the estimated expectation converges to the true expectation with
probability $1$:
\[
\lim\limits_{K \rightarrow \infty} \frac{1}{K} \sum_{i=1}^K f(X_i) =
\sum_x f(x) p(x) \textrm{ with probability 1}.
\]
Applying the Metropolis-Hastings algorithm to a log-linear model for
CLP yields for each $y \in \mathcal{Y}$ where $\tilde p(y) > 0$ a random sample
$\tilde X(y)$ from $p_\lambda$ on $X(y)$. Such samples can be combined
into a sample $\tilde\mathcal{X} =
\sum_{y \in \mathcal{Y}| \tilde p(y) > 0} \tilde
X(y)$ from $p_\lambda$ on $\mathcal{X}$.
From these random samples the desired estimates of expectations
of functions with respect to $p_\lambda$ can be computed. 

Note that we can use the same random sample for each iteration of
Newton's method to estimate the gain for each candidate property
simultaneously. After adding the selected property to the model, again
a single random sample from the extended model can be used to estimate
the MLE values for each parameter in parallel.
Suppose we have a random sample from $\mathcal{Y}$ of size $N$,
a complete data sample $\tilde X(y)$ of size $M_y$
for $y$, and combined complete data sample $\tilde \mathcal{X} =
\sum_{y \in \mathcal{Y}| \tilde p(y) > 0} \tilde X(y)$
of size $L$.
Then we can define tables similar to the tables of total
probabilities used in Sect. \ref{Newton} as follows.

\begin{itemize}
\item $S_{i,v} = \sum_{\tilde x \in \tilde \mathcal{X}} 
\delta_{\nu_i(\tilde x), v }$
is the number of times property function $\nu_i$ takes value $v$ in
combined random sample $\tilde \mathcal{X}$,
\item $T_{i,y}=\sum_{\tilde x \in \tilde X(y)}
\nu_i(\tilde x)$
is the number of times property $\chi_i$ occurs in random sample
$\tilde X(y)$, 
\item $U_{i,m}=\sum_{\tilde x \in \tilde \mathcal{X} | \;
  \nu_\# (\tilde x) = m} \nu_i(\tilde x)$ 
is the number of times property $\chi_i$ occurs in combined random
sample $\tilde \mathcal{X}$ when there is a total number of $m$
property instances for each sample point. 
\end{itemize}
For the expectations involved in the closed-form updates in parameter
estimation of Sect. \ref{ClosedForm}, the following counting variables
will be convenient:

\begin{itemize}
\item $s(i) = \sum_v S_{i,v} v$,
\item $t(i) = \sum_y T_{i,y} M_y^{-1}$.
\end{itemize}
The closed-form parameter update $\hat\gamma$ can then be approximated
by random sampling as follows.

\[
\hat\gamma_i \approx \frac{1}{K} \ln
\frac{t(i)}
{\frac{N}{L} s(i)}.
\]
For the expectations involved in the Newton formulae, the counting
variables are the same as those of Sect. \ref{Newton}, except for

\begin{itemize}
\item 
$ t_r(\alpha,i) = \sum_y T_{i,y}\alpha^r M_y^{-1} $.
\end{itemize}
The Newton update used in property selection is approximated by
random sampling as follows.

\[
\alpha_{t+1}
\approx
\alpha_t +
\frac{t_0(\alpha_t,c) - \frac{N}{L} s_1(\alpha_t,c)}
{ \frac{N}{L} s_2(\alpha_t,c)}.
\]
The gain is approximated as

\[
G_c(\hat\alpha , \lambda) \approx  N + t_1(\hat\alpha,c) -
\frac{N}{L} s_0(\hat\alpha, c).
\]
Similar random sampling estimates can be obtained for the Newton
update used in parameter estimation:

\[
\alpha_{t+1}
\approx
\alpha_t +
\frac{t_0(\alpha_t,i) - \frac{N}{L} u_0(\alpha_t, i)}
{ \frac{N}{L} u_1(\alpha_t, i)}.
\]

\subsection{Approximating Expectations via Maximum Pseudo-Likelihood Estimation }

As stated above, Monte Carlo methods offer the theoretical
assurance that the approximation of an expectation converges to the
true expectation in the limit. This means that one can get arbitrarily
close to the true value of the expectation with increasing sample
size. However, convergence can be very slow, i.e., the sample size
necessary for an appropriate approximation may be very large. This is
especially the case if the distributions of the nominating model $p_\pi$ and the
target model $p_\lambda$ are far apart. This may be the case if 
probabilistic context-free grammars are used as nominating model for a
log-linear model on constraint-based grammars. Besides the
compensation for sampling errors, many samples may have to be generated to
guarantee a reliable estimate of the desired expectations. Together,
these problems can make Monte Carlo approximations infeasible in practice.

An alternative to Monte Carlo methods is to approximate expectations
in a maximum pseudo-likelihood estimation framework. In Sect. \ref{PartialE} we
introduced partial E-steps in the EM algorithm as in instance of
maximum pseudo-likelihood estimation. The idea was there to
replace an intractable probability function with respect to which an expectation
is taken by a probability function which is more tractable. One
possibility to achieve such tractable expectations is to use sparse
expectations: Instead of replacing the intractable sample space by a
Monte-Carlo sample and counting from this,
the original sample space is restricted to an appropriate finite
subset over which the expectation is calculated.

The general form of such sparse approximations is as follows (cf.
\citeN{Neal:93}).
Let $S(y)$ be a finite subset of the set $X(y)$ of complete data
corresponding to an incomplete datum $y \in \mathcal{Y}$.
Then a sparse conditional distribution $s_{\lambda}(x|y)$ on complete
data $x$ given incomplete data $y$ and the current value of the
parameters $\lambda$ can be defined s.t.
\[
s_{\lambda^{(t)}}(x|y) = 
\left\{ \begin{array}{ll}
 0 & \textrm{ if } x \not\in S^{(t)}(y), \\
 \frac{p_{\lambda^{(t)}}(x)}
{\sum_{x \in S^{(t)}(y)} p_{\lambda^{(t)}}(x)} & \textrm{ if } 
x \in S^{(t)}(y).
\end{array} 
\right.
\]
That is, for a given subset $S^{(t)}(y)$ of the sample space $X(y)$
defined at time $t$, the sparse probability distribution
$s_{\lambda^{(t)}}(x|y)$ is defined as the normalized probability distribution
that assigns a positive probability only to the elements in $S^{(t)}(y)$.
The calculation of expectations $\sum_{x \in S^{(t)}(y)}
s_{\lambda^{(t)}} (x|y) f(x)$ of functions $f(x)$ with respect to 
$s_{\lambda^{(t)}}(x|y)$ then only takes time proportional to the size of
$S^{(t)}(y)$ at time $t$. 

Various heuristics
can be used for a flexible definition of $S^{(t)}(y)$. 
A sensible approach is to define $S^{(t)}(y)$ as the $N$ most
probable $x \in X(y)$, and recalculate this set at each step $t$, and
frequently perform a full iteration with $S^{(t)}(y) = X(y)$ for all $y \in
\mathcal{Y}$ with $\tilde p(y) > 0$. For $N=1$, this approach yields the well-known
\emph{Viterbi-approximation} of the EM algorithm. Here each $y$ is assumed
to come with a unique $x \in X(y)$ at time $t$. Given algorithms for
efficiently searching for the most probable proof tree $x$ for a given
query $y$, a Viterbi-approximation can be
defined for parameter estimation of a probabilistic CLP model. A
recursive use of this algorithm also enables an
\emph{N-best-approximation}. 

A linguistically motivated definition of $S^{(t)}(y)$ as the trees $x
\in X(y)$ of a context-free grammar which correspond to a bracketing
structure annotated to the sample of training sentences has been
presented by \citeN{Pereira:92}. Since the bracketing does not change
during the estimaton process, $S^{(t)}(y)$ is constant for all $t$. 
Clearly, such \emph{bracketing
  constraints} yield on the
one hand better linguistic results in terms of a constituent structures
of trees consistent with hand-annotated bracketings. On the other
hand, the restriction of the sample space to the $x \in \mathcal{X}$
which correspond to the bracketing structure of the sample from
$\mathcal{Y}$ also reduce the computational load of the
estimation process.

A general form of the IM algorithm using sparse approximations
$s_\lambda(x|y)$ is given in Table \ref{AlgSparse}.  
  
\begin{table}[htbp]
\begin{center}
\fbox{\begin{minipage}{14cm}

\begin{description}
\item[Input] Initial model $p_0$, 
 initial set $S_0(y)$,
  property-functions vector $\nu$, 
 incomplete-data sample from $\mathcal{Y}$.

\item[Output] Approximated MLE model $p_{\lambda^\ast}$ on
  $\mathcal{X} = \sum_{y \in \mathcal{Y} | \tilde p(y) > 0} X(y)$.

\item[Procedure] \mbox{}

\begin{tabbing}
Until \= convergence do \\
\> Compute $S(y), \; p_\lambda, \; s_\lambda$, based on $\lambda =
(\lambda_1, \ldots, \lambda_n)$, \\
\> For \= $i$ from $1$ to $n$ do \\
\>\> $\gamma_i := 
\frac{1}{K} \ln
\frac{\tilde p [ 
\sum_{x \in S(y)} s_{\lambda} (x|y) \nu_i(x) ]  }
{ \sum_{x \in \mathcal{X}} p_{\lambda} (x) \nu_i (x)} $, \\
\>\> $\lambda_i := \lambda_i + \gamma_i$, \\
Return $\lambda^\ast = (\lambda_1, \ldots, \lambda_n)$.
\end{tabbing}
\end{description}

\end{minipage}}
\caption{Algorithm (Sparse Iterative Maximization, Closed-Form)} 
\label{AlgSparse}
\end{center}
\end{table}

A theoretical justification of such approaches can be given 
in terms of partial expectations in the context of the EM algorithm. In
Sect. \ref{PartialE}, we saw that the incomplete-data log-likelihood
$L(\lambda) = \tilde p[\ln g_\lambda(y)]$ for a given random sample from
$\mathcal{Y}$ is lower bounded by a pseudo-likelihood function
$\mathcal{F}(q,\lambda)$ which is a joint function of the
parameters and of the distributions over the unobserved data. 
The function $q$ can be set to a tractable
sparse approximation $s_{\lambda^{(t)}}$ of $k_{\lambda^{(t)}}$.
Thus a sparse distribution $s_{\lambda^{(t)}}$ yields a lower bound
$\mathcal{F}(s_{\lambda^{(t)}},\lambda^{(t)}) \leq L(\lambda^{(t)})$
in the E-step, which is maximized as a function of $\lambda$ in the
M-step. 
As shown by \citeN{Neal:93} or \citeN{Csiszar:84}, even if some
iterations may decrease $L$, we are guaranteed that the
pseudo-likelihood $\mathcal{F}$ which bounds $L$ from below is
increased or held constant with every iteration. The interpretation of
the IM algorithm as an instance of a GEM algorithm given in Sect.
\ref{GEM} thus justifies a replacement of $k_{\lambda^{(t)}}$ by a
sparse approximation  $s_{\lambda^{(t)}}$ also for an IM algorithm.

However, it has to be kept in mind that for such partial E-steps
monotonicity and convergence of the estimation algorithm has to be
proven in terms of the lower bound $\mathcal{F}$ on $L$. Clearly,
convergence can be shown easily for approaches with constant
$S^{(t)}(y)$ for all $t$ induced, e.g., by fixed bracketing constraints, but is hard to
verify for approaches which let $S^{(t)}(y)$ vary as a function of $t$
such as Viterbi-approximations.
More subtle versions of pseudo-likelihood approaches to EM include variational
approximation methods, where a parameterized approximating
distribution $q$ is used and the parameters are varied to minimize the
Kullback Leibler distance between $q$ and $k_\lambda$. 
Minimizing this distance clearly results in a minimization of the
distance between the pseudo-likelihood function $\mathcal{F}$ and the
true likelihood function $L$.
\begin{eqnarray*}
L(\lambda) - \mathcal{F}(q,\lambda) 
& = & 
\tilde p [ \ln g_\lambda (\cdot) ] - \tilde p [ \sum_{x \in X(\cdot)}q(x) \ln
\frac{p_\lambda(x)}{q(x)} ]\\
& = & \tilde p [ \ln g_\lambda(\cdot) - \sum_{x \in X(\cdot)} q(x)( \ln
k_\lambda(x|\cdot) + \ln g_\lambda(\cdot) - \ln q(x)   )  ] \\
& = & \tilde p [ \sum_{x \in X(\cdot)} q(x) \ln
\frac{q(x)}{k_\lambda(x|\cdot)} ]   \\
& = & \tilde p [ D(q||k_\lambda) ]
\end{eqnarray*}
The parametric models used, e.g., in the context of
large-scale neural networks, are models assuming complete independence of the
variables of the network (mean field approximation, see
\citeN{Parisi:88}) or approximated models
probabilistic dependencies of the original model (structured
variational approximation, see \citeN{Saul:96}). Possible applications
of variational approximation to estimating probabilistic CLGs could
follow these lines. A discussion of such approaches yet is beyond the scope of this thesis.

\section{Parsing and Searching}
\markright{4.9 Parsing and Searching}
\label{PCLG}

In the foregoing chapters we discussed the mathematical and
algorithmic details of statistical inference of log-linear models from
incomplete data, and experimented with these techniques on a 
small set of real-world data of parses of a constraint-based grammar. On this
small scale it was possible to do ambiguity resolution by explicitly
listing all parses according to the induced probability distribution
and picking the most probable one as the correct one. However, for
applications on a larger scale an important question is how the
structure of the probability model on parses can be used to guide the
search for the most probable parse efficiently without having to list
all parses explicitly. Thus the question is whether the search
techniques standardly used for probabilistic grammars can be
re-applied to the log-linear CLP and CLG models.

We begin our discussion in Sect. \ref{Earley} with an application of
the tabular parsing method of Earley deduction \cite{Pereira:83} to
CLGs. The table of pending derivations defined in this method will lay
the ground for probabilistic search methods for finding most probable
parses. In Sect. \ref{Viterbi} we show that the 
probabilistic search method of the Viterbi algorithm
(\citeN{Viterbi:67}, \citeN{Forney:73}) standardly used  in
context-free tabular processing models finds the most probable parse
of a probabilistic CLG model only under certain restrictions. Since
such restrictions may trade off against the search complexity, methods for sensibly relaxing the restrictions are
desirable. A heuristic search algorithm resulting from such a
relaxation is discussed in Sect. \ref{Heuristics}.

\subsection{Earley Deduction for Feature-Based CLGs} 
\label{Earley}

Earley deduction has been introduced by \citeN{Pereira:83} as a
generalization of Earley's efficient context-free parsing algorithm
(\citeN{Earley:70}, \citeN{Aho:72}) to a tabular parsing algorithm for definite
clause grammars. In contrast to backtracking methods, in tabular
parsing methods a table, or chart, of pending subderivations is
built up during derivation. In Earley deduction, subderivations
correspond to definite clauses derived from the grammar axioms and a
query. Storing such derivation states for future use as items in a
chart may avoid the redundancy of backtracking methods which leads in
the worst case to an exponential search complexity. Instead, this
dynamic-programming technique of storing solutions to subproblems may reduce the search complexity to be polynomial in input length. 

The very basic concepts of an application of Earley deduction to CLP
can be given as follows.
Earley deduction works on two sets of definite clauses, the set of
program clauses $\mathcal{P}$ and the set of derived clauses
constituting the chart $\mathcal{C}$. An active item of a context-free
Earley parser corresponds here to a definite clause with at least one
relational atom on its righthandside, i.e., to a non-unit clause. Passive items
correspond to clauses whose righthandsides consist only of an
\La-constraint, i.e., to unit-clauses. 
A selection function determines for each non-unit clause its selected
\Rl-atom. We adopt here the standard Prolog selection rule where the
first atom on the righthandside of a clause is selected in each step.
The input to the algorithm consists of a set of program clauses \Po
and a query $G$. The  content of the chart $\mathcal{C}$ initially
consists of $G$ and is continually added to by an exhaustive
application of the following two inference rules\footnote{Prediction
  is called ``instantiation'' in \citeN{Pereira:83} and completion
  corresponds to their ``reso-lution''. In context-free Earley parsing standardly a
  distinction between ``predictor'', ``scanner'' and ``completer''
  operations is made. The first operation corresponds to
  prediction and the latter two operations are subsumed by the
  completion operation of the Earley decuction framework defined below.}
(the rules are to be read as ``If there are clauses $c_1$
and $c_2$ and the conditions on these clauses are satisfied, then add
clause $c_3$ to the chart.'').

\begin{quote}
\textbf{Prediction:} \\
\mbox{} \\
$c_1 = (H_1 \leftarrow B_1) \: \in \mathcal{C}$ \\
$c_2 = (H_2 \leftarrow B_2) \: \in \Po$ \\
\put(0,0){\line(1,0){110}} \\
$c_3 = (S \leftarrow B_2' \cup \phi) \: \in \mathcal{C}$\\
\mbox{}\\
where $c_1$ is non-unit, $c_2$ is unit or non-unit, $S$ is the
selected atom in $B_1$, $\phi$ is the \La-constraint in $B_1$, and
 there exists a variant
$c_2' = (S \leftarrow B_2')$
of $c_2$ 
s.t. $\Xsf{V}(c_1) \cap \Xsf{V}(B_2')\subseteq \Xsf{V}(S)$ 
and the \La-constraint $\phi'$ of $c_3$ is satisfiable.
\end{quote}

\newpage

\begin{quote}
\textbf{Completion:} \\
\mbox{}\\
$c_1 = (H_1 \leftarrow B_1) \: \in \mathcal{C}$ \\
$c_2 = (H_2 \leftarrow B_2) \: \in \mathcal{C}$\\
\put(0,0){\line(1,0){140}} \\
$c_3 = (H_1 \leftarrow ( B_1 \setminus S) \cup B_2') \: \in
\mathcal{C}$ \\
\mbox{} \\
where $c_1$ is non-unit, $c_2$ is unit, $S$ is the
selected atom in $B_1$, and
 there exists a variant $c_2' = (S \leftarrow B_2')$
of $c_2$ 
s.t. $\Xsf{V}(c_1) \cap \Xsf{V}(B_2')\subseteq \Xsf{V}(S)$
and the \La-constraint $\phi'$ of $c_3$ is satisfiable.
\end{quote}

These rules can be rationalized as follows: The prediction rule
proposes for the selected atom of a clause $c_1$ a possible variant of a
program clause $c_2$ using which an
$\stackrel{r,c}{\longrightarrow}$-step, i.e., a combined
goal-reduction and constraint-solving step, can be performed. 
For a unit clause $c_2$, the completion rule then performs a combined
$\stackrel{r,c}{\longrightarrow}$-step
on the lefthandside atom of $c_2$ and substitutes this selected atom
in clause $c_1$ by the resulting righthandside \La-constraint. 
Both rules collect the \La-constraints of the antecedent clauses
and take care of successful constraint solving and prevent accidental
variable sharing in the consequent clause.

Clearly, this combination of top-down prediction and bottom-up
completion defines a search rule which can reduce the parsing complexity
in comparison to backtracking methods.
However, to make these inference rules a workable algorithm, several
issues concerning the effective applicability of Earley deduction to
different purposes have to be addressed. Since these topics are not of direct
relevance for our problem, we refer the reader to the extensive
literature on this subject (see, e.g., \citeN{Pereira:87},
\citeN{Doerre:93a}, \citeN{Doerre:95}). 

Let us illustrate the basic concepts of Earley deduction with a
simple feature-based CLG. In the following example we will make use of
a standard technique for string position indexing, e.g., the indexed clause
\[
\texttt{sign}(X,0,1) \leftarrow X=\phi.
\]
abbreviates an actual CLP clause
\[
\texttt{sign}(X,Y,Z) \leftarrow X=\phi \:\&\: Y=0 \:\&\: Z=1.
\]
where the constants 0 and 1 denote the start and end position of the span
of the predicate in the input string. The string position can
be read off for unit clauses from the lefthandside atom, but for non-unit
clauses from the first string position argument of the head atom and
the first string position argument of the leftmost atom in the body.
Note that string position indexing is not mentioned in the definition of
the inference rules for Earley deduction. In fact, this indexing is not
necessary for Earley deduction to work. Rather, it is an effective
way to reduce the number of unsuccesful rule applications in an
implementation, and will also make our example more transparent.

Let us return for illustration to the simple example of
Fig. \ref{QCLGprogram}. An indexed variant of this program is given in
Fig. \ref{PCLGprogram}. 

\vspace{3ex}

\begin{figure}[htbp]
\begin{description}
\item[{\tt 1}  $\texttt{phrase}(X,S_0,S) \leftarrow$] $ X = (phrase \:\wedge\:
\textsc{CAT}:s \:\wedge\: \textsc{DTR1:CAT}:n \:\wedge\:
\textsc{DTR2:CAT}:v \:\wedge\: \textsc{DTR1:AGR}:Y \:\wedge\:
\textsc{DTR2:AGR}:Y \:\wedge\: \textsc{DTR1}:Z_1
\:\wedge\: \textsc{DTR2}:Z_2) \:\&\: \texttt{sign}(Z_1,S_0,S_1) \:\&\: \texttt{sign}(Z_2,S_1,S)$.
\item[{\tt 2} $\texttt{phrase}(X,S_0,S) \leftarrow$] $ X = (phrase \:\wedge\:
\textsc{CAT}:np \:\wedge\:\textsc{DTR1:CAT}:n \:\wedge\: \textsc{DTR2:CAT}:n \:\wedge\: \textsc{DTR1}:Z_1
\:\wedge\: \textsc{DTR2}:Z_2) \:\&\: \texttt{sign}(Z_1,S_0,S_1) \:\&\: \texttt{sign}(Z_2,S_1,S)$.
\item[{\tt 3}  $ \texttt{word}(X,0,1) \leftarrow$] $ X = (word \:\wedge\:
  \textsc{CAT}:n \:\wedge\: \textsc{PHON}:Clinton \:\wedge\:
  \textsc{AGR}:sg)$.
\item[{\tt 4} $\texttt{word}(X,1,2) \leftarrow$] $ X = (word \:\wedge\:
  \textsc{CAT}:v \:\wedge\: \textsc{PHON}:talks \:\wedge\:
  \textsc{AGR}:sg)$.
\item[{\tt 5} $\texttt{word}(X,1,2) \leftarrow$] $ X = (word \:\wedge\:
  \textsc{CAT}:n \:\wedge\: \textsc{PHON}:talks \:\wedge\:
  \textsc{AGR}:pl)$.
\item[{\tt 6} $\texttt{sign}(X,S_0,S) \leftarrow$] $\texttt{phrase}(X,S_0,S)$.
\item[{\tt 7} $\texttt{sign}(X,S_0,S) \leftarrow$] $\texttt{word}(X,S_0,S)$.
\end{description}
\caption{Indexed feature-based constraint logic grammar}
\label{PCLGprogram}
\end{figure}

An application of Earley deduction to parsing the query 
\[
\texttt{sign}(X,0,2) \:\&\: X = (sign \: \wedge \:
\textsc{DTR1: PHON}:Clinton \:\wedge\: \textsc{DTR2: PHON}:talks).
\]
denoting the input sentence 
\[
\mbox{}_0 Clinton_1 \; talks_2.
\]
is given in Figs. \ref{earley} and \ref{earleycont}.
\begin{figure}
{\footnotesize

\begin{tabular}{lp{12cm}l}
\texttt{9} & 
$\leftarrow \texttt{sign}(X,0,2) \:\&\: X = (sign \: \wedge \:
\textsc{DTR1: PHON}:Clinton \:\wedge\: \textsc{DTR2: PHON}:talks).$
& (I) \\

\texttt{10} &
$\texttt{sign}(X,0,2) \leftarrow \texttt{phrase}(X,0,2) \:\&\: X = (sign \:
\wedge \: \textsc{DTR1: PHON}:Clinton \:\wedge\: \textsc{DTR2:
  PHON}:talks).$
& (P \texttt{9,6})\\

\texttt{11} &
$\texttt{phrase}(X,0,2) \leftarrow 
\texttt{sign}(Z_1,0,S_1) \:\&\: \texttt{sign}(Z_2,S_1,2) 
\:\&\: X = (phrase \:\wedge\:  \textsc{CAT}:s 
\:\wedge\: \textsc{DTR1}:word  
\:\wedge\: \textsc{DTR1: CAT}:n 
\:\wedge\: \textsc{DTR1: PHON}:Clinton 
\:\wedge\: \textsc{DTR1: AGR}:Y
 \:\wedge\: \textsc{DTR2}:word
\: \wedge\:\textsc{DTR2: CAT}:v 
 \:\wedge\: \textsc{DTR2: PHON}:talks 
\:\wedge\: \textsc{DTR2: AGR}:Y
\:\wedge\: \textsc{DTR1}:Z_1 \:\wedge\: \textsc{DTR2}:Z_2).$
& (P \texttt{10,1})\\

\texttt{12} &
$\texttt{sign}(Z_1,0,S_1) \leftarrow \texttt{word}(Z_1,0,S_1) \:\&\: 
X = (phrase \:\wedge\:  \textsc{CAT}:s 
\:\wedge\: \textsc{DTR1}:word 
\:\wedge\: \textsc{DTR1: CAT}:n 
\:\wedge\: \textsc{DTR1: PHON}:Clinton 
\:\wedge\: \textsc{DTR1: AGR}:Y
 \:\wedge\: \textsc{DTR2}:word
\: \wedge\:\textsc{DTR2: CAT}:v 
 \:\wedge\: \textsc{DTR2: PHON}:talks 
\:\wedge\: \textsc{DTR2: AGR}:Y
\:\wedge\: \textsc{DTR1}:Z_1 \:\wedge\: \textsc{DTR2}:Z_2).$
& (P \texttt{11,7}) \\

\texttt{13} &
$\texttt{word}(Z_1,0,1) \leftarrow X = (phrase 
\:\wedge\: \textsc{CAT}:s 
\:\wedge\:  \textsc{DTR1}:word  
\:\wedge\: \textsc{DTR1: CAT}:n 
\:\wedge\: \textsc{DTR1: PHON}:Clinton 
\:\wedge\: \textsc{DTR1: AGR}:Y
\:\wedge\: \textsc{DTR1: AGR}:sg 
 \:\wedge\: \textsc{DTR2}:word 
\: \wedge\:\textsc{DTR2: CAT}:v 
 \:\wedge\: \textsc{DTR2: PHON}:talks 
\:\wedge\: \textsc{DTR2: AGR}:Y
\:\wedge\: \textsc{DTR2: AGR}:sg 
\:\wedge\: \textsc{DTR1}:Z_1 \:\wedge\: \textsc{DTR2}:Z_2).$
& (P \texttt{12,3})\\

\texttt{14} &
$\texttt{phrase}(X,0,2) \leftarrow \texttt{sign}(Z_1,0,S_1)
\:\&\: \texttt{sign}(Z_2,S_1,2)
\:\&\: X = (phrase \:\wedge\:  \textsc{CAT}:np 
\:\wedge\: \textsc{DTR1}:word  
\:\wedge\: \textsc{DTR1: CAT}:n 
\:\wedge\: \textsc{DTR1: PHON}:Clinton 
 \:\wedge\: \textsc{DTR2}:word
\: \wedge\:\textsc{DTR2: CAT}:n 
 \:\wedge\: \textsc{DTR2: PHON}:talks 
\:\wedge\: \textsc{DTR1}:Z_1 \:\wedge\: \textsc{DTR2}:Z_2)$.
& (P \texttt{10,2})\\

\texttt{15} &
$\texttt{sign}(Z_1,0,S_1) \leftarrow \texttt{word}(Z_1,0,S_1) 
\:\&\: X = (phrase \:\wedge\:  \textsc{CAT}:np 
\:\wedge\: \textsc{DTR1}:word  
\:\wedge\: \textsc{DTR1: CAT}:n 
\:\wedge\: \textsc{DTR1: PHON}:Clinton 
 \:\wedge\: \textsc{DTR2}:word
\: \wedge\:\textsc{DTR2: CAT}:n 
 \:\wedge\: \textsc{DTR2: PHON}:talks 
\:\wedge\: \textsc{DTR1}:Z_1 \:\wedge\: \textsc{DTR2}:Z_2)$.
& (P \texttt{14,7})\\

\texttt{16} &
$\texttt{word}(Z_1,0,1) \leftarrow 
X = (phrase \:\wedge\:  \textsc{CAT}:np 
\:\wedge\: \textsc{DTR1}:word  
\:\wedge\: \textsc{DTR1: CAT}:n 
\:\wedge\: \textsc{DTR1: PHON}:Clinton 
\:\wedge\: \textsc{DTR1: AGR}:sg 
 \:\wedge\: \textsc{DTR2}:word 
\: \wedge\:\textsc{DTR2: CAT}:n 
 \:\wedge\: \textsc{DTR2: PHON}:talks 
\:\wedge\: \textsc{DTR1}:Z_1 \:\wedge\: \textsc{DTR2}:Z_2)$.
& (P \texttt{15,3})\\

\texttt{17} &
$\texttt{sign}(Z_1,0,1) \leftarrow X = (phrase 
\:\wedge\: \textsc{CAT}:s 
\:\wedge\:  \textsc{DTR1}:word  
\:\wedge\: \textsc{DTR1: CAT}:n 
\:\wedge\: \textsc{DTR1: PHON}:Clinton 
\:\wedge\: \textsc{DTR1: AGR}:Y
\:\wedge\: \textsc{DTR1: AGR}:sg 
 \:\wedge\: \textsc{DTR2}:word 
\: \wedge\:\textsc{DTR2: CAT}:v 
 \:\wedge\: \textsc{DTR2: PHON}:talks 
\:\wedge\: \textsc{DTR2: AGR}:Y
\:\wedge\: \textsc{DTR2: AGR}:sg 
\:\wedge\: \textsc{DTR1}:Z_1 \:\wedge\: \textsc{DTR2}:Z_2).$
& (C \texttt{12,13})\\

\texttt{18} &
$\texttt{phrase}(X,0,2) \leftarrow \texttt{sign}(Z_2,1,2) 
\:\&\: X = (phrase 
\:\wedge\: \textsc{CAT}:s 
\:\wedge\:  \textsc{DTR1}:word  
\:\wedge\: \textsc{DTR1: CAT}:n 
\:\wedge\: \textsc{DTR1: PHON}:Clinton 
\:\wedge\: \textsc{DTR1: AGR}:Y
\:\wedge\: \textsc{DTR1: AGR}:sg 
 \:\wedge\: \textsc{DTR2}:word 
\: \wedge\:\textsc{DTR2: CAT}:v 
 \:\wedge\: \textsc{DTR2: PHON}:talks 
\:\wedge\: \textsc{DTR2: AGR}:Y
\:\wedge\: \textsc{DTR2: AGR}:sg 
\:\wedge\: \textsc{DTR1}:Z_1 \:\wedge\: \textsc{DTR2}:Z_2).$
& (C \texttt{11,17})\\

\texttt{19} & 
$\texttt{sign}(Z_1,0,1) \leftarrow X = (phrase 
\:\wedge\:  \textsc{CAT}:np 
\:\wedge\: \textsc{DTR1}:word  
\:\wedge\: \textsc{DTR1: CAT}:n 
\:\wedge\: \textsc{DTR1: PHON}:Clinton 
\:\wedge\: \textsc{DTR1: AGR}:sg 
 \:\wedge\: \textsc{DTR2}:word 
\: \wedge\:\textsc{DTR2: CAT}:n 
 \:\wedge\: \textsc{DTR2: PHON}:talks 
\:\wedge\: \textsc{DTR1}:Z_1 \:\wedge\: \textsc{DTR2}:Z_2)$.
& (C \texttt{15,16})\\

\texttt{20} &
$\texttt{phrase}(X,0,2) \leftarrow \texttt{sign}(Z_2,1,2) 
\:\&\: X = (phrase 
\:\wedge\:  \textsc{CAT}:np 
\:\wedge\: \textsc{DTR1}:word  
\:\wedge\: \textsc{DTR1: CAT}:n 
\:\wedge\: \textsc{DTR1: PHON}:Clinton 
\:\wedge\: \textsc{DTR1: AGR}:sg 
 \:\wedge\: \textsc{DTR2}:word 
\: \wedge\:\textsc{DTR2: CAT}:n 
 \:\wedge\: \textsc{DTR2: PHON}:talks 
\:\wedge\: \textsc{DTR1}:Z_1 \:\wedge\: \textsc{DTR2}:Z_2)$.
& (C \texttt{14,19})

\end{tabular}
}

\caption{Earley deduction chart}
\label{earley}
\end{figure}

\begin{figure}
{\footnotesize

\begin{tabular}{lp{12cm}l}

\texttt{21} &
$\texttt{sign}(Z_2,1,2) \leftarrow \texttt{word}(Z_2,1,2) 
\:\&\: X = (phrase  
\:\wedge\: \textsc{CAT}:s  
\:\wedge\:  \textsc{DTR1}:word   
\:\wedge\: \textsc{DTR1: CAT}:n  
\:\wedge\: \textsc{DTR1: PHON}:Clinton  
\:\wedge\: \textsc{DTR1: AGR}:Y  
\:\wedge\: \textsc{DTR1: AGR}:sg  
 \:\wedge\: \textsc{DTR2}:word  
\: \wedge\:\textsc{DTR2: CAT}:v  
 \:\wedge\: \textsc{DTR2: PHON}:talks  
\:\wedge\: \textsc{DTR2: AGR}:Y 
\:\wedge\: \textsc{DTR2: AGR}:sg  
\:\wedge\: \textsc{DTR1}:Z_1 \:\wedge\: \textsc{DTR2}:Z_2).$ 
& (P \texttt{18,7})\\

\texttt{22} &
$\texttt{word}(Z_2,1,2) \leftarrow  
X = (phrase 
\:\wedge\: \textsc{CAT}:s 
\:\wedge\:  \textsc{DTR1}:word  
\:\wedge\: \textsc{DTR1: CAT}:n 
\:\wedge\: \textsc{DTR1: PHON}:Clinton 
\:\wedge\: \textsc{DTR1: AGR}:Y
\:\wedge\: \textsc{DTR1: AGR}:sg 
 \:\wedge\: \textsc{DTR2}:word 
\: \wedge\:\textsc{DTR2: CAT}:v 
 \:\wedge\: \textsc{DTR2: PHON}:talks 
\:\wedge\: \textsc{DTR2: AGR}:Y
\:\wedge\: \textsc{DTR2: AGR}:sg 
\:\wedge\: \textsc{DTR1}:Z_1 \:\wedge\: \textsc{DTR2}:Z_2).$
& (P \texttt{21,4})\\

\texttt{23} &
$\texttt{sign}(Z_2,1,2) \leftarrow \texttt{word}(Z_2,1,2) 
\:\&\: X = (phrase 
\:\wedge\:  \textsc{CAT}:np 
\:\wedge\: \textsc{DTR1}:word  
\:\wedge\: \textsc{DTR1: CAT}:n 
\:\wedge\: \textsc{DTR1: PHON}:Clinton 
\:\wedge\: \textsc{DTR1: AGR}:sg 
 \:\wedge\: \textsc{DTR2}:word 
\: \wedge\:\textsc{DTR2: CAT}:n
 \:\wedge\: \textsc{DTR2: PHON}:talks 
\:\wedge\: \textsc{DTR1}:Z_1 \:\wedge\: \textsc{DTR2}:Z_2).$
& (P \texttt{20,7})\\

\texttt{24} &
$\texttt{word}(Z_2,1,2) \leftarrow  
X = (phrase  
\:\wedge\:  \textsc{CAT}:np 
\:\wedge\: \textsc{DTR1}:word  
\:\wedge\: \textsc{DTR1: CAT}:n 
\:\wedge\: \textsc{DTR1: PHON}:Clinton 
\:\wedge\: \textsc{DTR1: AGR}:sg 
 \:\wedge\: \textsc{DTR2}:word 
\: \wedge\:\textsc{DTR2: CAT}:n
 \:\wedge\: \textsc{DTR2: PHON}:talks 
\:\wedge\: \textsc{DTR2: AGR}:pl
\:\wedge\: \textsc{DTR1}:Z_1 \:\wedge\: \textsc{DTR2}:Z_2).$
& (P \texttt{23,5})\\

\texttt{25} &
$\texttt{sign}(Z_2,1,2) \leftarrow
X = (phrase  
\:\wedge\: \textsc{CAT}:s  
\:\wedge\:  \textsc{DTR1}:word   
\:\wedge\: \textsc{DTR1: CAT}:n  
\:\wedge\: \textsc{DTR1: PHON}:Clinton  
\:\wedge\: \textsc{DTR1: AGR}:Y  
\:\wedge\: \textsc{DTR1: AGR}:sg  
 \:\wedge\: \textsc{DTR2}:word  
\: \wedge\:\textsc{DTR2: CAT}:v  
 \:\wedge\: \textsc{DTR2: PHON}:talks  
\:\wedge\: \textsc{DTR2: AGR}:Y 
\:\wedge\: \textsc{DTR2: AGR}:sg  
\:\wedge\: \textsc{DTR1}:Z_1 \:\wedge\: \textsc{DTR2}:Z_2).$ 
& (C \texttt{21,22}) \\

\texttt{26} &
$\texttt{phrase}(X,0,2) \leftarrow
X = (phrase  
\:\wedge\: \textsc{CAT}:s  
\:\wedge\:  \textsc{DTR1}:word   
\:\wedge\: \textsc{DTR1: CAT}:n  
\:\wedge\: \textsc{DTR1: PHON}:Clinton  
\:\wedge\: \textsc{DTR1: AGR}:Y  
\:\wedge\: \textsc{DTR1: AGR}:sg  
 \:\wedge\: \textsc{DTR2}:word  
\: \wedge\:\textsc{DTR2: CAT}:v  
 \:\wedge\: \textsc{DTR2: PHON}:talks  
\:\wedge\: \textsc{DTR2: AGR}:Y 
\:\wedge\: \textsc{DTR2: AGR}:sg  
\:\wedge\: \textsc{DTR1}:Z_1 \:\wedge\: \textsc{DTR2}:Z_2).$  
& (C \texttt{18,25})\\

\texttt{27} &
$\texttt{sign}(X,0,2) \leftarrow
X = (phrase  
\:\wedge\: \textsc{CAT}:s  
\:\wedge\:  \textsc{DTR1}:word   
\:\wedge\: \textsc{DTR1: CAT}:n  
\:\wedge\: \textsc{DTR1: PHON}:Clinton  
\:\wedge\: \textsc{DTR1: AGR}:Y  
\:\wedge\: \textsc{DTR1: AGR}:sg  
 \:\wedge\: \textsc{DTR2}:word  
\: \wedge\:\textsc{DTR2: CAT}:v  
 \:\wedge\: \textsc{DTR2: PHON}:talks  
\:\wedge\: \textsc{DTR2: AGR}:Y 
\:\wedge\: \textsc{DTR2: AGR}:sg  
\:\wedge\: \textsc{DTR1}:Z_1 \:\wedge\: \textsc{DTR2}:Z_2).$   
& (C \texttt{10,26}) \\

\texttt{28} &
$\texttt{sign}(Z_2,1,2) \leftarrow
X = (phrase  
\:\wedge\:  \textsc{CAT}:np 
\:\wedge\: \textsc{DTR1}:word  
\:\wedge\: \textsc{DTR1: CAT}:n 
\:\wedge\: \textsc{DTR1: PHON}:Clinton 
\:\wedge\: \textsc{DTR1: AGR}:sg 
 \:\wedge\: \textsc{DTR2}:word 
\: \wedge\:\textsc{DTR2: CAT}:n
 \:\wedge\: \textsc{DTR2: PHON}:talks 
\:\wedge\: \textsc{DTR2: AGR}:pl
\:\wedge\: \textsc{DTR1}:Z_1 \:\wedge\: \textsc{DTR2}:Z_2).$
& (C \texttt{23,24}) \\

\texttt{29} &
$\texttt{phrase}(X,0,2) \leftarrow
X = (phrase  
\:\wedge\:  \textsc{CAT}:np 
\:\wedge\: \textsc{DTR1}:word  
\:\wedge\: \textsc{DTR1: CAT}:n 
\:\wedge\: \textsc{DTR1: PHON}:Clinton 
\:\wedge\: \textsc{DTR1: AGR}:sg 
 \:\wedge\: \textsc{DTR2}:word 
\: \wedge\:\textsc{DTR2: CAT}:n
 \:\wedge\: \textsc{DTR2: PHON}:talks 
\:\wedge\: \textsc{DTR2: AGR}:pl
\:\wedge\: \textsc{DTR1}:Z_1 \:\wedge\: \textsc{DTR2}:Z_2).$ 
& (C \texttt{20,28}) \\

\texttt{30} &
$\texttt{sign}(X,0,2) \leftarrow
X = (phrase  
\:\wedge\:  \textsc{CAT}:np 
\:\wedge\: \textsc{DTR1}:word  
\:\wedge\: \textsc{DTR1: CAT}:n 
\:\wedge\: \textsc{DTR1: PHON}:Clinton 
\:\wedge\: \textsc{DTR1: AGR}:sg 
 \:\wedge\: \textsc{DTR2}:word 
\: \wedge\:\textsc{DTR2: CAT}:n
 \:\wedge\: \textsc{DTR2: PHON}:talks 
\:\wedge\: \textsc{DTR2: AGR}:pl
\:\wedge\: \textsc{DTR1}:Z_1 \:\wedge\: \textsc{DTR2}:Z_2).$  
& (C \texttt{10,29}) \\

\end{tabular}
}

\caption{Earley deduction chart, cont.}
\label{earleycont}
\end{figure}

\begin{figure}
\begin{center}
\mbox{\psfig{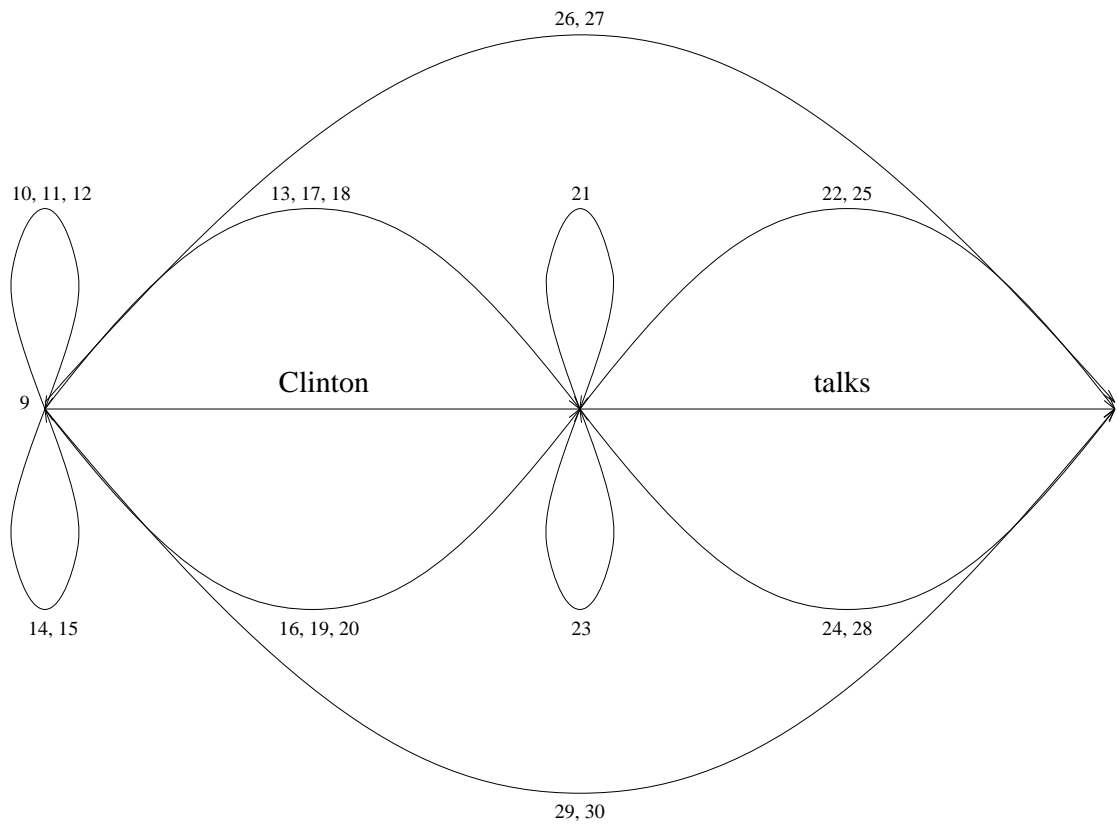}}
\end{center}
\caption{Chart graph}
\label{graph}
\end{figure}

A convenient way to illustrate graphically the relation of derived
items in a chart to partial parses of an input sentence is by a chart
graph. An chart graph for the sequence of derived
clauses of Figs. \ref{earley} and \ref{earleycont} is given in
Fig. \ref{graph}. 

This graph associates the numbers of derived clauses with directed
edges which connect input string position nodes.
Edges which point from a node to the node itself are
attached with the numbers of clauses derived by prediction with
non-unit clauses. In the language of context-free Earley parsing, such
edges represent predictions on non-terminal symbols. Scanning of
terminal symbols is represented by edges connecting a node with the
next node on its right. Such edges are attached with the numbers of
clauses derived by prediction with unit clauses. Completion of
non-terminal symbols is represented by edges connecting a node with a
node possibly further on its right. Such nodes are attached with the numbers
of clauses derived by the completion rule. 

There are two parses of the above input sentence hidden in the Earley
deduction chart of Figs. \ref{earley} and \ref{earleycont}.
In the chart graph of Fig. \ref{graph}, these two parses are
represented by the upper and lower half of the symmetric graph.
From each of the two final
completed clauses, \texttt{27} and \texttt{30}, a proof tree
representing a parse can be reconstructed using the algorithm of
Def. \ref{ConstructTree}.
This algorithm defines the construction of partial proof trees from
completed clauses, and when applied recursively, permits the construction of proof
trees from a given Earley deduction chart.

\begin{de}
\label{ConstructTree}
Let $c_k$ be a completed clause derived from clauses $c_i$ and
$c_j$, let $t_i$ and $t_j$ be the unique partial proof trees
corresponding to $c_i$ and $c_j$, 
and define for each predicted clause $(E \leftarrow F)$ a partial
proof tree
{\footnotesize
$\begin{array}{c}
E \\
| \\
F
\end{array}$
}.
Then the partial proof tree
$ t(t_i, t_j)$ corresponding to $c_k$ is
constructed s.t. \\

$t(t_i, t_j) =
\left\{ 
\begin{array}{ll}

{\scriptstyle
\begin{array}{c}
t_i \\
\oplus \\
t_j
\end{array}
} 

& 

\textrm{ if both } c_i, c_j
\textrm{ are completed clauses,} \\
 
& \\

{\scriptstyle
\begin{array}{c}
t_i \\
\otimes\\
t_j
\end{array}
}

&

\textrm{ if one or both } c_i, c_j
\textrm{ are predicted clauses, }

\end{array} 
\right.
$

and
{\footnotesize
$\begin{array}{ccc}
 & & \vdots \\
 & & A\\
t_1& & | \\
\oplus & = & B \cup C \\
t_2 & & | \\
 & & D \\
 & & \vdots
\end{array} $},
{\footnotesize
$\begin{array}{ccc}
 & & \vdots \\  
 & & A \\ 
t_1& & | \\ 
\otimes & = & B \\ 
t_2 & & | \\ 
 & & B \setminus C \cup D\\ 
 & & \vdots 
\end{array} $}, 
if
{\footnotesize
$\begin{array}{ccc} 
& & \vdots \\ 
t_1 & = & A \\ 
& & | \\ 
& & B 
\end{array}$}, 
{\footnotesize
$\begin{array}{ccc} 
& & C \\ 
& & | \\ 
t_2 & = & D \\ 
& & \vdots 
\end{array}$}.
\end{de}

\begin{figure}[htbp]
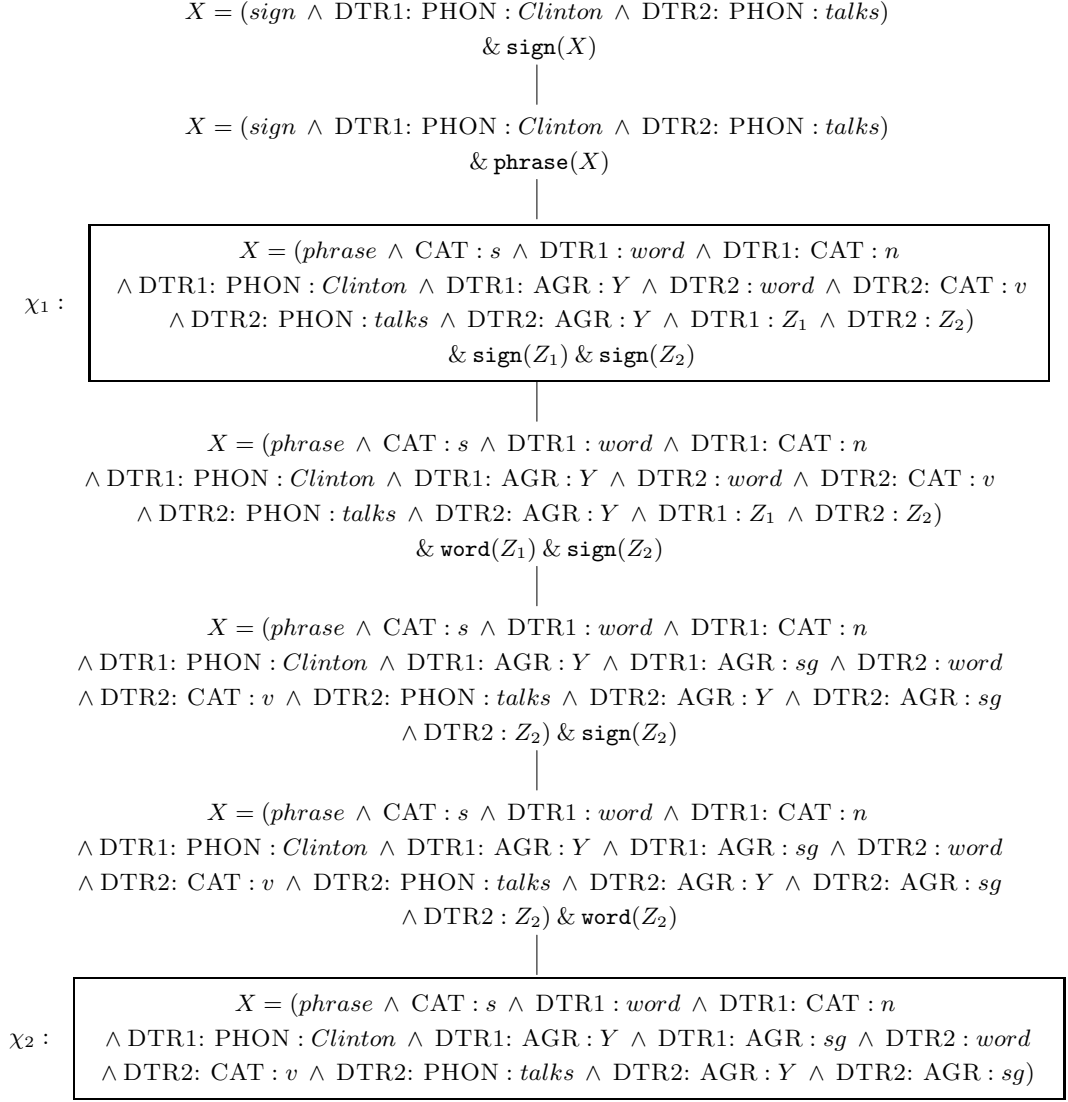

{\footnotesize
\begin{center}
\setlength{\GapWidth}{50pt}

\begin{bundle}
  {$\begin{array}{c} 
      X = (sign \: \wedge \: \textsc{DTR1: PHON}:Clinton 
      \:\wedge\: \textsc{DTR2: PHON}:talks) \\
      \:\&\: \texttt{sign}(X)  
    \end{array}$} 
  
  \chunk
  {\begin{bundle}
      {$\begin{array}{c} 
          X = (sign \: \wedge \: \textsc{DTR1: PHON}: Clinton 
          \:\wedge\: \textsc{DTR2: PHON}:talks) \\
          \:\&\: \texttt{phrase}(X) 
        \end{array}$} 
      
      \chunk
      {\begin{bundle}
          {$\chi_1: \quad$
            \fbox{
              $\begin{array}{c}
                X = (phrase \:\wedge\:  \textsc{CAT}:s \:\wedge\: 
                \textsc{DTR1}:word  \:\wedge\: \textsc{DTR1: CAT}:n \\ 
                \:\wedge\: \textsc{DTR1: PHON}:Clinton 
                \:\wedge\: \textsc{DTR1: AGR}:Y
                \:\wedge\: \textsc{DTR2}:word\:  
                \wedge\:\textsc{DTR2: CAT}:v \\ 
                \:\wedge\: \textsc{DTR2: PHON}:talks 
                \:\wedge\: \textsc{DTR2: AGR}:Y
                \:\wedge\: \textsc{DTR1}:Z_1 \:\wedge\: \textsc{DTR2}:Z_2) \\
                \:\&\: \texttt{sign}(Z_1)
                \:\&\: \texttt{sign}(Z_2)
              \end{array}$}
            }
          
          \chunk
          {\begin{bundle}
              {$\begin{array}{c}
                  X = (phrase \:\wedge\:  \textsc{CAT}:s \:\wedge\: 
                  \textsc{DTR1}:word  \:\wedge\: \textsc{DTR1: CAT}:n \\ 
                  \:\wedge\: \textsc{DTR1: PHON}:Clinton 
                  \:\wedge\: \textsc{DTR1: AGR}:Y
                  \:\wedge\: \textsc{DTR2}:word\:  
                  \wedge\:\textsc{DTR2: CAT}:v \\ 
                  \:\wedge\: \textsc{DTR2: PHON}:talks 
                  \:\wedge\: \textsc{DTR2: AGR}:Y
                  \:\wedge\: \textsc{DTR1}:Z_1 \:\wedge\: \textsc{DTR2}:Z_2) \\
                  \:\&\: \texttt{word}(Z_1) \:\&\: \texttt{sign}(Z_2)
                \end{array}$}

              \chunk
              {\begin{bundle}
                  {$\begin{array}{c}
                      X = (phrase \:\wedge\:  \textsc{CAT}:s \:\wedge\: 
                      \textsc{DTR1}:word  \:\wedge\: \textsc{DTR1: CAT}:n \\
                      \:\wedge\: \textsc{DTR1: PHON}:Clinton 
                      \:\wedge\: \textsc{DTR1: AGR}:Y
                      \:\wedge\: \textsc{DTR1: AGR}:sg 
                      \:\wedge\: \textsc{DTR2}:word \\
                      \: \wedge\:\textsc{DTR2: CAT}:v 
                      \:\wedge\: \textsc{DTR2: PHON}:talks 
                      \:\wedge\: \textsc{DTR2: AGR}:Y
                      \:\wedge\: \textsc{DTR2: AGR}:sg \\
                      \:\wedge\: \textsc{DTR2}:Z_2) 
                      \:\&\: \texttt{sign}(Z_2) 
                    \end{array}$
                    }

                  \chunk
                  {\begin{bundle}
                      {$\begin{array}{c}
                          X = (phrase \:\wedge\:  \textsc{CAT}:s \:\wedge\: 
                          \textsc{DTR1}:word  \:\wedge\: \textsc{DTR1: CAT}:n \\ 
                          \:\wedge\: \textsc{DTR1: PHON}:Clinton 
                          \:\wedge\: \textsc{DTR1: AGR}:Y
                          \:\wedge\: \textsc{DTR1: AGR}:sg 
                          \:\wedge\: \textsc{DTR2}:word \\
                          \: \wedge\:\textsc{DTR2: CAT}:v 
                          \:\wedge\: \textsc{DTR2: PHON}:talks 
                          \:\wedge\: \textsc{DTR2: AGR}:Y
                          \:\wedge\: \textsc{DTR2: AGR}:sg \\
                          \:\wedge\: \textsc{DTR2}:Z_2) 
                          \:\&\: \texttt{word}(Z_2) 
                        \end{array}$
                        }
                      
                      \chunk
                      {$\chi_2: \quad$
                        \fbox
                        {$\begin{array}{c}
                            X = (phrase \:\wedge\:  \textsc{CAT}:s \:\wedge\: 
                            \textsc{DTR1}:word  \:\wedge\: \textsc{DTR1: CAT}:n \\ 
                            \:\wedge\: \textsc{DTR1: PHON}:Clinton 
                            \:\wedge\: \textsc{DTR1: AGR}:Y
                            \:\wedge\: \textsc{DTR1: AGR}:sg 
                            \:\wedge\: \textsc{DTR2}:word \\
                            \: \wedge\:\textsc{DTR2: CAT}:v 
                            \:\wedge\: \textsc{DTR2: PHON}:talks 
                            \:\wedge\: \textsc{DTR2: AGR}:Y
                            \:\wedge\: \textsc{DTR2: AGR}:sg )
                          \end{array}$ }
                        }
                    \end{bundle}
                    }
                \end{bundle}
                }
            \end{bundle}
            }
        \end{bundle}
        }
    \end{bundle}
    }
\end{bundle}

\end{center}
}
\caption{Proof tree for $[Clinton_N \;talks_V]_S$}
\label{PCLG1st}
\end{figure}

\begin{figure}[htbp]
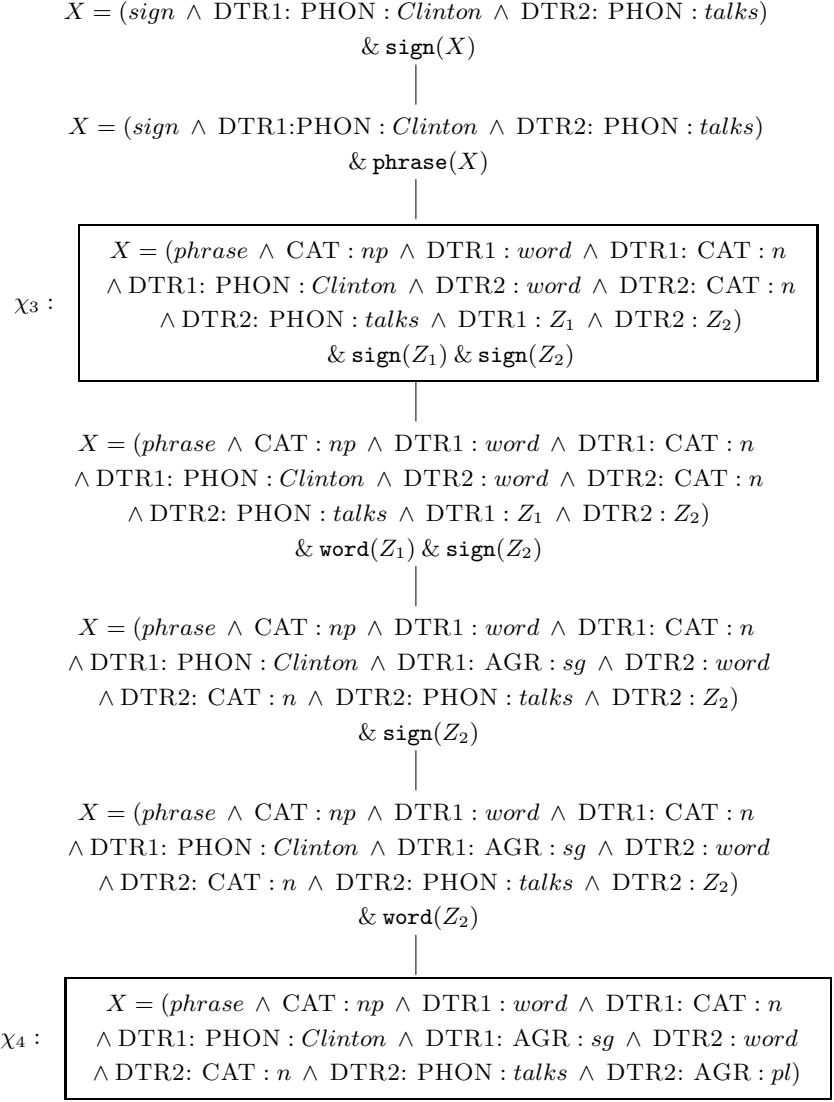

  {\footnotesize
    \begin{center}
      \setlength{\GapWidth}{50pt}
      
      \begin{bundle}
        {$\begin{array}{c} 
            X = (sign \: \wedge \: \textsc{DTR1: PHON}:Clinton 
            \:\wedge\: \textsc{DTR2: PHON}:talks) \\
            \:\&\: \texttt{sign}(X)  
          \end{array}$} 
        
        \chunk
        {\begin{bundle}
            {$\begin{array}{c} 
                X = (sign \: \wedge \: \textsc{DTR1: 
                  PHON}: Clinton 
                \:\wedge\: \textsc{DTR2: PHON}:talks) \\
                \:\&\: \texttt{phrase}(X) 
              \end{array}$} 
            
            \chunk
            {\begin{bundle}
                {$\chi_3: \quad $
                  \fbox
                  {$\begin{array}{c}
                      X = (phrase \:\wedge\:  \textsc{CAT}:np \:\wedge\: 
                      \textsc{DTR1}:word  \:\wedge\: \textsc{DTR1: CAT}:n \\ 
                      \:\wedge\: \textsc{DTR1: PHON}:Clinton 
                      \:\wedge\: \textsc{DTR2}:word\:  
                      \wedge\:\textsc{DTR2: CAT}:n \\ 
                      \:\wedge\: \textsc{DTR2: PHON}:talks 
                      \:\wedge\: \textsc{DTR1}:Z_1 \:\wedge\: \textsc{DTR2}:Z_2) \\
                      \:\&\: \texttt{sign}(Z_1)
                      \:\&\: \texttt{sign}(Z_2)
                    \end{array}$}
                  }
                
                \chunk
                {\begin{bundle}
                    {$\begin{array}{c}
                        X = (phrase \:\wedge\:  \textsc{CAT}:np \:\wedge\: 
                        \textsc{DTR1}:word  \:\wedge\: \textsc{DTR1: CAT}:n \\ 
                        \:\wedge\: \textsc{DTR1: PHON}:Clinton 
                        \:\wedge\: \textsc{DTR2}:word\:  
                        \wedge\:\textsc{DTR2: CAT}:n \\ 
                        \:\wedge\: \textsc{DTR2: PHON}:talks 
                        \:\wedge\: \textsc{DTR1}:Z_1 \:\wedge\: \textsc{DTR2}:Z_2) \\
                        \:\&\: \texttt{word}(Z_1) \:\&\: \texttt{sign}(Z_2)
                      \end{array}$}

                    \chunk
                    {\begin{bundle}
                        {$\begin{array}{c}
                            X = (phrase \:\wedge\:  \textsc{CAT}:np \:\wedge\: 
                            \textsc{DTR1}:word  \:\wedge\: \textsc{DTR1: CAT}:n \\
                            \:\wedge\: \textsc{DTR1: PHON}:Clinton 
                            \:\wedge\: \textsc{DTR1: AGR}:sg 
                            \:\wedge\: \textsc{DTR2}:word \\
                            \: \wedge\:\textsc{DTR2: CAT}:n 
                            \:\wedge\: \textsc{DTR2: PHON}:talks 
                            \:\wedge\: \textsc{DTR2}:Z_2) \\
                            \:\&\: \texttt{sign}(Z_2) 
                          \end{array}$
                          }

                        \chunk
                        {\begin{bundle}
                            {$\begin{array}{c}
                                X = (phrase \:\wedge\:  \textsc{CAT}:np \:\wedge\: 
                                \textsc{DTR1}:word  \:\wedge\: \textsc{DTR1: CAT}:n \\ 
                                \:\wedge\: \textsc{DTR1: PHON}:Clinton 
                                \:\wedge\: \textsc{DTR1: AGR}:sg 
                                \:\wedge\: \textsc{DTR2}:word \\
                                \: \wedge\:\textsc{DTR2: CAT}:n 
                                \:\wedge\: \textsc{DTR2: PHON}:talks 
                                \:\wedge\: \textsc{DTR2}:Z_2) \\
                                \:\&\: \texttt{word}(Z_2) 
                              \end{array}$
                              }
                            
                            \chunk
                            {$\chi_4: \quad$
\fbox
{$\begin{array}{c}
    X = (phrase \:\wedge\:  \textsc{CAT}:np \:\wedge\: 
    \textsc{DTR1}:word  \:\wedge\: \textsc{DTR1: CAT}:n \\ 
    \:\wedge\: \textsc{DTR1: PHON}:Clinton 
    \:\wedge\: \textsc{DTR1: AGR}:sg 
    \:\wedge\: \textsc{DTR2}:word \\
    \: \wedge\:\textsc{DTR2: CAT}:n
    \:\wedge\: \textsc{DTR2: PHON}:talks 
    \:\wedge\: \textsc{DTR2: AGR}:pl)
  \end{array}$ }
}
\end{bundle}
}
\end{bundle}
}
\end{bundle}
}
\end{bundle}
}
\end{bundle}
}
\end{bundle}

\end{center}
}
\caption{Proof tree for $[Clinton_N \; talks_N]_{NP}$}
\label{PCLG2nd}
\end{figure}

The proof tree for completed clause \texttt{27}, corresponding to the parse
$[Clinton_N \;talks_V]_S$, 
is the proof tree of Fig. \ref{CLG1st} and repeated here in
Fig. \ref{PCLG1st}.
The parse $[Clinton_N \; talks_N]_{NP}$ 
is derived via the proof tree of Fig. \ref{CLG2nd}, repeated here in
Fig. \ref{PCLG2nd}, and can be reconstructed from completed clause \texttt{30}.

\subsection{Probabilistic CLGs and the Viterbi Algorithm} 
\label{Viterbi}

Turning to probabilistic CLGs, we see that because CLGs are
simply instances of CLP, all techniques developed for statistical
inference of probabilistic CLP apply to
probabilistic CLGs without modification. The simplest way to
inspect a probability distribution on parses is to list the
respective proof trees and calculate their probabilities from the
subtree-properties and the corresponding parameters. An imaginable
probability model for the proof trees of Figs. \ref{PCLG1st} and
\ref{PCLG2nd} could take as properties the subtrees introduced by the
clauses which are responsible for the two different readings of the input
sentence, namely clauses \texttt{1}, \texttt{4}, \texttt{2}, and
\texttt{5}. The respective properties
$\chi_1, \: \chi_2, \: \chi_3,$ and $\chi_4$
are depicted in Figs. \ref{PCLG1st} and \ref{PCLG2nd} as framed parts of
the proof trees. MLE from a large natural language corpus for the
parameters $\lambda_1,\: \lambda_2, \: \lambda_3$ and 
$\lambda_4$ corresponding to these properties 
would probably return a higher weight for parameters $\lambda_1$ and
$\lambda_2$ than for $\lambda_3$ and $\lambda_4$. Thus this
probability model would tell the proof tree of Fig. \ref{PCLG1st},
corresponding to the parse $[Clinton_N \; talks_V ]_S$, to be more
probable given the input sentence $Clinton \; talks$ than the proof tree of
Fig. \ref{PCLG2nd}, corresponding to the parse $[Clinton_N \; talks_N
]_{NP}$.

However, if we are interested in the most probable parse of a
sentence, listing all possible parses may be too costly in general, even
if the parses just have to be extracted from a chart. Clearly, it
would be nice if we could make use of the structure of the
probabilistic model to guide the search for the most probable
parse. The Viterbi algorithm (\citeN{Viterbi:67}, \citeN{Forney:73}) for finding the most
probable parse implements this idea using a dynamic-programming
approach as follows: 
During derivation, each derivation state must keep
track of the most probable path of derivation states leading towards
it. When the final derivation state is reached, the maximum
probability derivation can be recovered by tracing back the stored
path of most probable derivation states.

Clearly, different specifications of this algorithm depend on the
chosen parsing strategy and on the underlying probability
model. For example, \citeN{Stolcke:93} computes a Viterbi parse for
probabilistic context-free grammars in a framework of probabilistic Earley parsing
as follows: During derivation, each completed item keeps track of the
most probable path of items contributing to it. The rule probabilities are propagated
recursively by associating each predicted item with the probability of
the rewriting rule used in the prediction, and by recording for each
completed item the product of probabilities of the pair of items that
contributes with maximal value to the completion. Storing at each
completion step the item-pair leading to the maximum, finally yields a path of
most probable items from which the most probable derivation can be
retrieved.

Under certain restrictions on the parsing strategy and on the
probabilistic search method, the idea of the Viterbi algorithm is applicable to
Earley deduction for log-linear probabilistic CLGs as follows.

Concerning the parsing strategy, let us strictly adhere to the definition of
Earley deduction given above. That is, we only speak of an ambiguous
derivation of a completed clause if more than one pair of clauses
yields via completion the same clause with the same variable binding.
That is, in this setting a numerical comparison at a completion step is done 
only between clause-pairs contributing via completion to the same
``instantiated'' clause. 

Considering the probabilistic search method,  \citeN{Stolcke:93}'s
model of Viterbi parsing can be reconstructed if we identify the
properties of the log-linear model with program clauses.
If properties are allowed to be subtrees of proof trees, things are
more complictated. In this setting, in order to compare numerically between
alternatives, we have to incrementally build up partial proof trees
and check their properties during derivation. 

First, we have to define a function $w$ to calculate the weight of a partial proof
tree $t_k$ under a log-linear probability model $p_\lambda$.

\begin{de}
Let $\mathcal{C}$ be an Earley deduction chart for query $G$ and
program \Po, let $\mathcal{X}$ be the set of proof trees for $G$ from \Po,
and let $p_\lambda$ be a log-linear distribution on $\mathcal{X}$. Then
the weight $w$ of a partial proof tree $t_k$ constructable for a completed
clause $c_k \in \mathcal{C}$ is defined s.t.
\[
w(t_k) = e^{\lambda \cdot\nu(t_k)}.
\]
\end{de}
Furthermore, a numerical comparison between alternatives leading to
the same completion requires the partial proof trees corresponding to the alternative
completions to include only completely built-up
subtree-properties. This is necessary to avoid the outranking of
highly weighted partial proof trees by lower weighted partial proof
trees at a completion step where the highly weighted
subtree-properties cannot yet be taken under consideration. 
For an appropriate partial ordering on trees based on an operation
$\subseteq$, we can ensure that partial proof trees include only
completely built-up properties as follows.

\begin{quote}
A partial proof tree $t_k$ is complete for a property-vector $\chi =
(\chi_1, \ldots, \chi_n)$ iff for each $i=1, \ldots, n$: 
$\chi_i \subseteq t_k$ or else $\chi_i \cap t_k = \emptyset$.
\end{quote}
The algorithm of Def. \ref{ConstructTree} can be used for a recursive
comparison as follows. Note that we use the definition of variant
given in Chap. \ref{Foundations} for the specification of an
equivalence class of clauses to be compared.

\begin{quote} 
For each equivalence class $[c_k]$ of completed clauses, record the
partial proof tree 
$t^\ast_k = \underset{t_k}{\arg\max\;} w(t_k)$, 
where 
$[c_k] = \{ c \in \mathcal{C} |$
$c$ is a variant of $c_k$,
and there exist clauses $c_i$ and $c_j$ in $\mathcal{C}$ from which $c$ is
derivable  via completion$\}$,
and
$t_k \in \{ t(t^\ast_i, t^\ast_j) |$
$t^\ast_i$ and $t^\ast_j$ are the hightest weighted complete partial proof
trees corresponding to clauses $c_i$ and $c_j\}$.
\end{quote}

Clearly, given the above restrictions, this procedure will yield the
most probable proof tree for a given query to a program.
The possible savings in computational complexity induced by this procedure depend
on the size of the subtree-properties to be worked out during the
search process. That is, small subtrees will permit an efficient pruning at
nearly each completion step whereas subtrees connecting nodes over
long distances may in the worst case yield no gain in efficiency at
all.

\subsection{Heuristic Searching for Most Probable Parses}
\label{Heuristics}

However, the effective applicability of the search procedure stated
above strongly depends the form of the grammars under
consideration. That means, for particular CLGs, it is inefficient to restrict the numerical selection
only to alternative completions which lead to the same clause with
the same variable binding. The storing of variable
bindings in each step of an Earley deduction procedure is necessary to
enable partial proofs to be reused in other partial
proofs. Unfortunately, deriving a new clause with each new variable
binding may introduce overhead which causes in the worst case an
exponential search cost. This can be the case, e.g., for grammars which
encode parses entirely via variable bindings, i.e., via
\La-constraints, and in not via predicates, i.e., \Rl-atoms. The
extreme ends of the spectrum of such examples can be marked, e.g., for
the first case by CLGs resulting from a direct application of the
compilation procedure of \citeN{Goetz:95}. This procedure translates HPSG
descriptions into the \La-constraints of a CLP fragment using a single
\Rl-atom for processing. An example for the second case
are definite clause grammars such as those presented in
\citeN{Pereira:83} which encode each grammar symbol as a distinct CLP
predicate. For cases like the first, it would be more effective if one
could compare alternative completions leading to a variant of a CLP
clause irrespective of the variable bindings. Unfortunately, this
approach to comparing ``uninstantiated'' completed clauses introduces
a context-dependence problem caused by incompatible variable
bindings. That is, we are confronted here with a trade-off between
efficiency and correctness of the search method. 

\begin{figure}
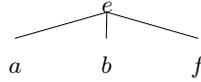

{\footnotesize
\setlength{\GapWidth}{30pt}
\setlength{\GapDepth}{10pt}
\begin{center}

\begin{bundle}{$e$}
  \chunk{$a$}
  \chunk{$b$}
  \chunk{$f$}
\end{bundle}

\end{center}
}
\caption{Type hierarchy}
\label{ftypes}
\end{figure}

Let us illustrate this context-dependence problem with
a simple example. 
For illustration we use the program of Fig. \ref{PCLPprogram}, repeated
here in Fig. \ref{fprogram}, with \La-constraints from a language of
hierarchcal types. The ordering on the types is depicted in Fig. \ref{ftypes}.

\begin{figure}[htbp]
  {\footnotesize
    \begin{center}
      \begin{tabular}{l}
        \texttt{1} $\texttt{s}(Z)\leftarrow \texttt{p}(Z) \:\&\: \texttt{q}(Z).$ \\
        \texttt{2} $\texttt{p}(Z) \leftarrow Z=a.$ \\
        \texttt{3} $\texttt{p}(Z) \leftarrow Z=b.$ \\
        \texttt{4} $\texttt{p}(Z) \leftarrow Z=f.$ \\
        \texttt{5} $\texttt{q}(Z) \leftarrow Z=a.$ \\
        \texttt{6} $\texttt{q}(Z) \leftarrow Z=b.$ 
      \end{tabular}
    \end{center}}
  \caption{Constraint logic program}
  \label{fprogram}
\end{figure}

An Earley deduction chart for the query $\texttt{s}(Z) \:\&\: Z=e$ is
given in Fig. \ref{fearley}.
\begin{figure}
{\footnotesize
\begin{center}
\begin{tabular}{lll}

\texttt{7}
& $\leftarrow \texttt{s}(Z) \:\&\: Z=e$. 
& (I) \\

\texttt{8}
& $\texttt{s}(Z) \leftarrow \texttt{p}(Z) \:\&\: \texttt{q}(Z) \:\&\: Z=e $.
& (P \texttt{7,1}) \\

\texttt{9}
& $\texttt{p}(Z) \leftarrow Z=a$.
& (P \texttt{8,2}) \\

\texttt{10}
& $\texttt{p}(Z) \leftarrow Z=b$.
& (P \texttt{8,3}) \\

\texttt{11}
& $\texttt{p}(Z) \leftarrow Z=f$.
& (P \texttt{8,4})\\

\texttt{12}
& $\texttt{s}(Z) \leftarrow \texttt{q}(Z) \:\&\: Z=a$.
&(C \texttt{8,9})\\

\texttt{13}
& $\texttt{s}(Z) \leftarrow \texttt{q}(Z) \:\&\: Z=b$.
&(C \texttt{8,10})\\

\texttt{14}
& $\texttt{s}(Z) \leftarrow \texttt{q}(Z) \:\&\: Z=f$.
&(C \texttt{8,11})\\

\texttt{15}
& $\texttt{q}(Z) \leftarrow Z=a$.
&(P \texttt{12,5})\\

\texttt{16}
& $\texttt{q}(Z) \leftarrow Z=b$.
&(P \texttt{13,6}) \\

\texttt{17}
& $\texttt{s}(Z) \leftarrow Z=a$.
&(C \texttt{12,15})\\

\texttt{18} 
& $\texttt{s}(Z) \leftarrow Z=b$.
&(C \texttt{13,16})

\end{tabular}
\end{center}
}
\caption{Earley deduction chart}
\label{fearley}
\end{figure}

\begin{figure}[htbp]
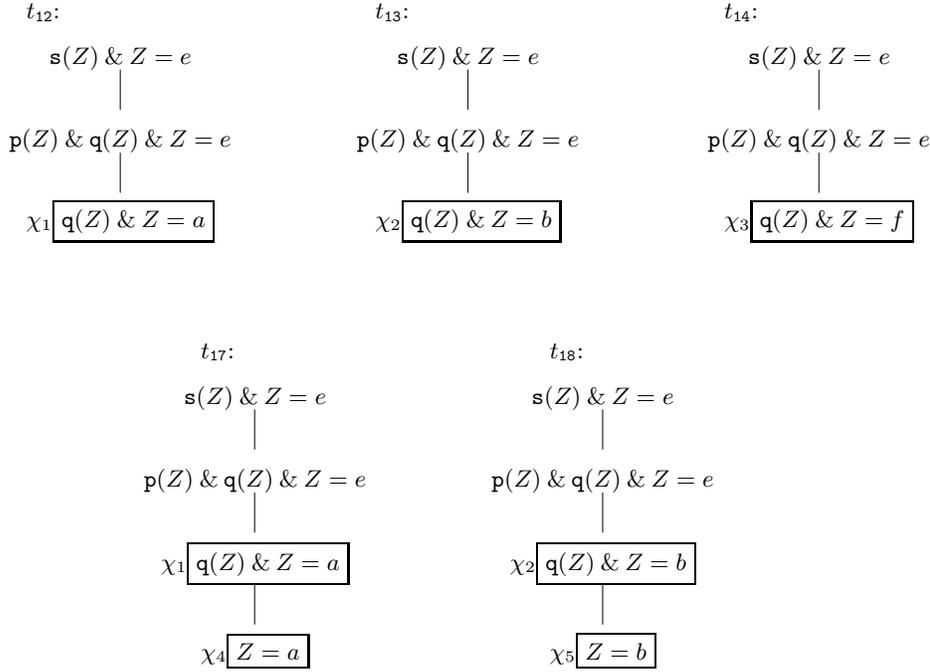

{\footnotesize
\begin{center} 
\begin{tabular}{p{120pt}p{120pt}p{120pt}}

$t_{\mathtt{12}}$: & $t_{\mathtt{13}}$: & $t_{\mathtt{14}}$: \\

\begin{bundle}{$\texttt{s}(Z)\:\&\: Z=e$}
  \chunk
  {
    \begin{bundle}{$\texttt{p}(Z) \:\&\: \texttt{q}(Z) \:\&\: Z=e$}
      \chunk{$\chi_1$ 
        \fbox{$\texttt{q}(Z) \:\&\: Z=a$}}
    \end{bundle}
    }
\end{bundle}

& 

\begin{bundle}{$\texttt{s}(Z)\:\&\: Z=e$}
  \chunk
  {
    \begin{bundle}{$\texttt{p}(Z) \:\&\: \texttt{q}(Z) \:\&\: Z=e$}
      \chunk{$\chi_2$ 
        \fbox{$\texttt{q}(Z) \:\&\: Z=b$}}
    \end{bundle}
    }
\end{bundle}

&

\begin{bundle}{$\texttt{s}(Z)\:\&\: Z=e$}
  \chunk
  {
    \begin{bundle}{$\texttt{p}(Z) \:\&\: \texttt{q}(Z) \:\&\: Z=e$}
      \chunk{$\chi_3$ 
        \fbox{$\texttt{q}(Z) \:\&\: Z=f$}}
    \end{bundle}
    }
\end{bundle}

\end{tabular}
\end{center}
}

\vspace {1ex}

{\footnotesize
  \begin{center} 
    \begin{tabular}{p{120pt}p{120pt}}
      
      $t_{\mathtt{17}}$: & $t_{\mathtt{18}}$: \\

      \begin{bundle}{$\texttt{s}(Z)\:\&\: Z=e$}
        \chunk
        {
          \begin{bundle}{$\texttt{p}(Z) \:\&\: \texttt{q}(Z) \:\&\: Z=e$}
            \chunk
            {
              \begin{bundle}{$\chi_1$
                  \fbox{$\texttt{q}(Z) \:\&\: Z=a$}}
                \chunk
                {$\chi_4$\fbox{$Z=a$}}
              \end{bundle}
              }
          \end{bundle}
          }
      \end{bundle}
      
      &

      \begin{bundle}{$\texttt{s}(Z)\:\&\: Z=e$}
        \chunk
        {
          \begin{bundle}{$\texttt{p}(Z) \:\&\: \texttt{q}(Z) \:\&\: Z=e$}
            \chunk
            {
              \begin{bundle}{$\chi_2$
                  \fbox{$\texttt{q}(Z) \:\&\: Z=b$}}
                \chunk
                {$\chi_5$\fbox{$Z=b$}}
              \end{bundle}
              }
          \end{bundle}
          }
      \end{bundle}

    \end{tabular}
  \end{center}
  }

\caption{Partial proof trees}
\label{fprooftrees}
\end{figure}

Let the properties  $\chi_1$ to $\chi_5$ of a probability distribution
over the proof trees corresponding to the Earley deduction chart of
Fig. \ref{fearley} be defined as the framed subtrees shown in
Fig. \ref{fprooftrees}. Furthermore, let the 
corresponding parameter values be $\lambda_1 = \ln 2, \: \lambda_2 =
\ln 3, \: \lambda_3 = \ln 5, \: \lambda_4 = \ln 5$ and $\lambda_5 =
\ln 3$.  Now let us take a look at how the probability model defined by these
properties and parameters guides the search for the most probable
proof tree of query $\texttt{s}(Z) \:\&\: Z=e$ from the program of
Fig. \ref{fprogram}. The first decision to be made is between the
completed clauses \texttt{12}, \texttt{13} and \texttt{14}, which
differ only by their variable bindings, i.e., by their
\La-constraints. The partial proof trees corresponding to these
clauses, $t_\mathtt{12}$, $t_\mathtt{13}$ and $t_\mathtt{14}$,
are shown in Fig. \ref{fprooftrees}. 
However, $t_\mathtt{14}$, the highest weighted of these partial proof
trees, is not included in any proof tree, $t_\mathtt{17}$ or
$t_\mathtt{18}$, 
corresponding to the final completion steps. That is, in this case
the probabilistic search has not only missed the most probable proof
tree but has led to failure!
Even if we ignore the clauses contributing to this failure, namely
clauses \texttt{4}, \texttt{11} and \texttt{14}, a problem still
remains. In this case, $w(t_\mathtt{13}) > w(t_\mathtt{12})$,
and the weight of proof tree $t_\mathtt{18}$  including the best
partial proof tree $t_\mathtt{13}$ is $w(t_\mathtt{18}) = 9$.
However, the weight of proof tree $t_\mathtt{17}$ including the
partial proof tree $t_\mathtt{12}$, which we just have thrown away, is
$w(t_\mathtt{17}) = 10$
and $w(t_\mathtt{17}) > w(t_\mathtt{18})$. Thus in this case the
probabilistic search method has led us to the lower weighted proof tree.

Clearly, the first of these problems can be solved by constructing the
Earley deduction chart in advance and by checking the terminal
\La-constraint of each partial proof tree corresponding to an
alternative completion against the \La-constraints of the final completion steps in the chart.
This can be accomplished by the the following re-definition of the
equivalence class $[c_k]$ of completed clauses which is subject to a
numerical comparison in a completion step. Note that we refer here to the
definition of variable renaming given in Chap. \ref{Foundations}.

\begin{quote}
Let $\mathcal{C}$ be an Earley deduction chart for query
  $G$ and program \Po, 
$c_k=(A \leftarrow B_1 \:\&\: \ldots
B_n \:\&\: \psi)$
be a clause in $\mathcal{C}$,  
and $c_k' = c_k \setminus \psi$.
Then an equivalence class $[c_k]$ of completed clauses in $\mathcal{C}$
 is defined s.t.
\begin{description}
\item[${[}c_k{]}$] 
$ = \{ c \in \mathcal{C} |$
there exist clauses $c_i$ and $c_j$  in $\mathcal{C}$ from which
clause  $c = (C \leftarrow D_1 \:\&\: \ldots \:\&\:
D_m \:\&\: \phi)$ is derivable via completion,
$c' = c \setminus \phi$
is obtained from $c_k'$ by simultaneously replacing 
each occurence of a variable $X$ in $c_k'$ by a renamed variable
$\rho(X)$ for all variables $X \in
\mathsf{V}(c_k')$ for a renaming $\rho$, 
and there exists a satisfiable \La-constraint $\phi \:\&\: \varphi$
for at least one final completed clause in $\mathcal{C}$ with \La-constraint
$\varphi \}$. 
\end{description}
\end{quote}

However, the latter of the problems stated above is solvable only at the cost of
re-introducing the restriction to compare only ``instantiated'' variants of completed
clauses, i.e., variants of clauses with the same \La-constraints. Each
search algorithm which allows ``uninstantiated'' variants of
completed clauses to be compared, necessarily provides only a
heuristic search procedure in the sense that it does not guarantee
that the most probable proof tree is found. 
According to the definition of $[c_k]$, the equivalence class of
completed clauses which are compared in a completion step, and the
satisfaction of the completeness requirement on partial proof trees,
the algorithm of Table \ref{AlgSearch} defines either an approximate
heuristic or a true best-parse search algorithm. 

An approach to a Viterbi-like heuristic search procedure similar to
ours is used by \citeN{CarrollBriscoe:92} for searching the parse
forest produced by their probabilistic LR parser for unification-based grammars.
There also the parse forest must be built up completely before unpacking 
to ensure that the search algorithm does pursue successful
derivations. 

On the whole, the decision between a possibly more efficient, but only
approximate heuristic procedure and a possibly inefficient, but
optimal Viterbi algorithm has to be made with respect to particular
classes of CLGs in mind. Furthermore, if a heuristic search procedure
is used, an alternative to completing the chart in advance is to using
a backtracking procedure in connection with an incremental computation
of clauses and corresponding best partial proof trees.

\begin{table}[htbp]
\begin{center}

\fbox{

\begin{minipage}{14cm}

\begin{description}

\item[Input] Log-linear model $p_\lambda$ on set $\mathcal{X}$ of proof trees
for goal $G$ from program \Po, weight function $w$, tree-constructor
function $t$, Earley deduction algorithm, choice of equivalence class of
completed clauses.

\item[Output] Best proof tree $t^\ast_k$ for $G$ from \Po.

\item[Procedure] \mbox{} 
\begin{tabbing}
Until \= no clauses can be added \\
\> Compute clauses by Earley deduction algorithm, \\
\> If \=  $c_k$ is a completed clause, \\
\>\> Then $w^\ast := 0$, compute $[c_k]$,\\
\>\> If \=$[c_k] = [c_l]$ for some $l < k$, \\
\>\> \> Then $t^\ast_k := t^\ast_l$, \\
\>\>\> Else \= for each $c \in [c_k]$,\\
\>\>\>\> For \= each $c_i, c_j$ which derive $c$ via completion, \\
\>\>\>\>\> Compute the best proof tree $t_i^\ast$ for $c_i$, \\
\>\>\>\>\> Compute the best proof tree $t_j^\ast$ for $c_j$, \\
\>\>\>\>\> $t_k := t(t^\ast_i, t^\ast_j)$, \\
\>\>\>\>\> If \= $w(t_k) > w^\ast$, \\
\>\>\>\>\>\>Then $w^\ast := w(t_k), \; 
t^\ast_k := t_k$,\\ 
\>\>\>\>\>\>Else delete $c$, \\
Return $t^\ast_k$.
\end{tabbing}

\end{description}

\end{minipage}}

\end{center}
\label{AlgSearch}
\caption{Algorithm (Best-Parse Search)}
\end{table}

\section{Summary and Discussion}
\markright{4.10 Summary and Discussion}

In this chapter we presented a probabilistic model for CLP and a novel
method for statistical inference about the parameters of such models
from incomplete data.
We discussed the problems of previous approaches which applied Baum's
estimation technique for stochastic context-free models to
estimation of stochastic constraint-based models. We showed with a
counterexample that this incomplete-data estimation
method does not generally yield the desired maximum
likelihood values when applied to constraint-based systems. To
overcome the inherent context-dependence problem of such systems, we
introduced a powerful log-linear probability model for
CLP. Furthermore, we presented a new algorithm to infer the parameters of
log-linear models, and also the properties of such parametric models,
from incomplete training data. We showed monotonicity and convergence
of the algorithm to the desired maximum likelihood estimates and
applied it experimentally to estimation of a CLG on a small scale.
Furthermore, we discussed various methods for approximate computation of the formulae
involved in the inference task, and presented methods which use the structure of
the probabilitic model to guide the search for the most probable
analysis. To this end we presented an approximate heuristic search
algorithm based on dynamic programming techniques. Depending on the class of
grammars under consideration, this algorithm can provide a
considerable efficiency gain in searching for the most probable
analysis.

In comparison to the work on quantitative CLP presented above, the
advantages of probabilistic CLP are clearly the possibility to use
automatic techniques for statistical inference for parameter
estimation and property selection. Rather, our
incomplete-data inference algorithm is general enough to be applicable to log-linear
probability distributions in general, and thus is useful in other
incomplete-data settings as well. In this chapter the algorithm has
especially been shown to be useful for probabilistic context-sensitive
NLP models. In contrast to related approaches such as that of
 \citeN{Magerman:94}, \citeN{Rat:98} or \citeN{Goodman:98}, which require
fully annotated corpora for estimation, our statistical inference
algorithm provides general means for automatic and reusable training of
arbitrary probabilistic constraint-based grammars from unannotated
corpora. Furthermore, our approach is the first one since the
introduction of log-linear models into the discussion of probabilistic
parsing by \citeN{Abney:96} which evaluates experimentally the usefulness of general
log-linear models on CLGs.

\chapter{Conclusion}
\label{Conclusion}

In this final chapter we present a short summary of the work of this
thesis. We compare the advantages and shortcomings of the two
presented approaches to quantitative and probabliistic CLP relative to
each other and relative to other approaches. Not surprisingly, the presented work is not definitive but raises several
questions which could not be answered in the course of this
thesis. These questions will be dealt with when we discuss future
continuations of the presented work.

\section{Summary}
\markboth{Chapter 5. Conclusion}{5.1 Summary}

In this thesis, we have presented new mathematical and algorithmic
techniques for quantitative and statistical inference in
constraint-based NLP. We have chosen the general concepts of CLP as the
formal framework to deal with constraint-based NLP, yielding CLGs as
instances of CLP. Aiming at a general solution of the problem of
structural ambiguity in CLGs, we have presented two independent
approaches to weighted CLGs.   

The first approach, called quantitative CLP, is situated in a clear
logical framework, and presents a sound and complete system of
quantitative inference for definite clauses with subjective weights
attached to them. This approach permits to specify weights in
arbitrary ways, e.g., as subjective probabilities, user-defined
preference values, or degrees of grammaticality, and to use search
techniques such as alpha-beta pruning for finding the maximally
weighted proof tree for a given set of queries efficiently.
Related previous work either focussed solely on 
formal semantics of quantitative logic programs without specific
applications in mind, or presented only informal attachments of
weights to grammar components for the aim of weight-based pruning in
natural language parsing. Our approach is the first one to combine
weight-based parsing for constraint-based systems with a rigid formal
semantics for such quantitative inference systems.

The second approach, called probabilistic CLP, addresses the problem
of structural ambiguity resolution by a completely different form of
weighted CLGs. Here a log-linear probability model is presented which
defines a probability distribution over the proof trees of a
constraint logic program on the basis of weights assigned to arbitrary
properties of these trees. The possibility to define arbitrary
features of proof trees as such properties and to estimate appropriate
weights for them permits the probabilistic modeling of arbitrary
context-dependencies. In this thesis we firstly evaluate empirically
the applicability and feasibility of estimation of general log-linear models on CLGs.
In contrast to previous approaches which were restricted to
estimation from annotated data for specialized probabilistic parsing
models we present an algorithm to estimate the
parameters and to induce the properties of log-linear models from
incomplete, unanalyzed data. The new algorithm has the same
computational complexity as related complete-data inference algorithms
for log-linear models. Furthermore, we address the problem of
computational intractability of large summations in the inference task by
discussing various techniques to approximately solve this task and
present an approximate heuristic search algorithm for CLGs.

\section{Future Work}
\markright{5.2 Future Work}

As shown in Sect. \ref{Experiment}, the empirical evaluation of
estimating log-linear models on CLGs showed promising results both for
training and evaluation on a small scale. Clearly, the main task of
future work is a thorough investigation of the performance of the
presented general algorithms on \emph{larger scales} of real-world NLP
applications. In larger experiments issues which were addressed
so far only theoretically shall be evaluated in practice. Such issues are the
empirical evaluation of \emph{property selection}, the evaluation of various
\emph{approximation methods} in parameter estimation, or the empirical
testing of the performance of non-heuristic versus \emph{heuristic search
techniques} in terms of linguistic results. 

New issues which shall be addressed in larger experiments are the use of
\emph{dynamic programming} techniques not only for searching for best parses
but also for efficient calculation of expectations in the estimation
process. Similar to the heuristic Viterbi algorithm presented for best
parse search the application of dynamic programming to computing
expectations will be possible only in a heuristic way. Clearly, the
question to be addressed is how such heuristic estimation procedures
perform in terms of linguistic evaluations.  

Another issue that will become important for larger data sets is the
use of reference distributions as simpler and easier to estimate
\emph{back-off} models. 
A reasonable choice of a reference distribution for our
task is, e.g., a model defining a probability distribution on
lexical-semantic head-head relations such as verb-noun pairs (see, e.g,
\citeN{Rooth:99}). Such a clustering model does not require complex
parsing models or costly annotated corpora, but can be estimated
easily from large corpora of verb-noun pairs. Furthermore, such a class-based model will
also provide a smooth default distribution and thus help to solve the
sparse data problem.

A further task of future work will be the investigation of possible
applications of log-linear models and incomplete-data estimation to NLP
applications different from parsing.


\newpage
\markboth{List of Figures}{List of Figures}
\addcontentsline{toc}{chapter}{List of Figures}
\listoffigures

\newpage
\markboth{List of Tables}{List of Tables}
\addcontentsline{toc}{chapter}{List of Tables}
\listoftables


\bibliographystyle{chicago}

\end{document}